\documentclass[
dvipsnames, table,   
format=sigconf,
anonymous=false,     
review=false,        
authorversion=false, 
screen=true,         
nonacm=true,         
]{acmart}

\usepackage{custom_commands}
\usepackage{custom_aliases}

\copyrightyear{2022} 
\acmYear{2022} 
\setcopyright{acmlicensed}
\acmConference[GECCO '22]{Genetic and Evolutionary Computation Conference}{July 9--13, 2022}{Boston, MA, USA}
\acmBooktitle{Genetic and Evolutionary Computation Conference (GECCO '22), July 9--13, 2022, Boston, MA, USA}
\acmPrice{15.00}
\acmDOI{10.1145/3512290.3528769}
\acmISBN{978-1-4503-9237-2/22/07}

\begin{document}

\title[%
    \alg: Multi-objective Bayesian Optimisation by Density-Ratio Estimation%
]{%
    \alg: Multi-objective Bayesian Optimisation by\\
          Density-Ratio Estimation%
}

\author{George {De Ath}}
\email{g.de.ath@exeter.ac.uk}
\orcid{0000-0003-4909-0257}
\affiliation{%
  \department{Department of Computer Science}
  \institution{University of Exeter}
  \city{Exeter}
  \country{United Kingdom}
}

\author{Tinkle Chugh}
\email{t.chugh@exeter.ac.uk}
\orcid{0000-0001-5123-8148}
\affiliation{%
  \department{Department of Computer Science}
  \institution{University of Exeter}
  \city{Exeter}
  \country{United Kingdom}
}

\author{Alma A. M. Rahat}
\email{a.a.m.rahat@swansea.ac.uk}
\orcid{0000-0002-5023-1371}
\affiliation{%
  \department{Department of Computer Science}
  \institution{Swansea University}
  \city{Swansea}
  \country{United Kingdom}
}

\begin{abstract}
Optimisation problems often have multiple conflicting objectives 
that can be computationally and/or financially expensive. Mono-surrogate
Bayesian optimisation (BO) is a popular model-based approach for optimising
such black-box functions. It combines objective values via scalarisation and
builds a Gaussian process (GP) surrogate of the scalarised values. The location
which maximises a cheap-to-query acquisition function is chosen as the next
location to expensively evaluate. While BO is an effective strategy, the use of
GPs is limiting. Their performance decreases as the problem input dimensionality
increases, and their computational complexity scales cubically with the amount
of data. To address these limitations, we extend previous work on BO by
density-ratio estimation~(BORE) to the multi-objective setting. BORE links the
computation of the probability of improvement acquisition function to that of
probabilistic classification. This enables the use of state-of-the-art
classifiers in a BO-like framework. In this work we present
\alg: multi-objective Bayesian optimisation by density-ratio estimation, and
compare it to BO across a range of synthetic and real-world benchmarks. We find
that \alg performs as well as or better than BO on a wide variety of problems,
and that it outperforms BO on high-dimensional and real-world problems.

\end{abstract}

%



\maketitle

\section{Introduction}
\label{sec:intro}
Optimisation problems, particularly those in real-world settings, often have
conflicting objectives that can be computationally and/or financially expensive
to evaluate. For example, it is often desirable to maximise a robot's speed
and minimise its energy consumption~\citep{liao:robots:2019}, or to maximise
crop yield while minimising the environmental impacts the required agricultural
development~\citep{jiang:agriculture:2020}. These tend to be \emph{black-box}
problems, \ie they have no closed-form or derivative information available. In
order to optimise these expensive black-box functions, a common strategy is to
create surrogate models for each of the objective functions and perform
optimisation on these instead, choosing where to evaluate the expensive
functions next based on locations that are predicted to yield high-quality
solutions.

A plethora of multi-objective surrogate-based approaches have been proposed in
recent years~\citep{rahat:infill:2017,chugh:krvea:2018,belakaria:mobomes:2019,
yang:ehvi:2019,suzuki:pfentropy:2020,tiao:bore:2021}. These are usually
inspired by, or directly use, the framework of single-objective Bayesian
optimisation (BO). In BO, given an initial set of expensively evaluated
solutions, a surrogate model, often a probabilistic model such as a Gaussian
process (GP), is created for either each objective function
(a \emph{multi-surrogate} approach), or a scalarised representation of the
multiple objectives (a \emph{mono-surrogate} approach). In these approaches,
an acquisition function, also known as an \emph{infill criterion}, is optimised
to suggest the next location to expensively evaluate. These attempt to balance
the exploration of locations with high amounts of prediction uncertainty with
the exploitation of locations that have good predicted values. They are also
cheap-to-query and, therefore, transform the expensive optimisation problem
into a sequence of cheaper ones followed by an expensive evaluation. Once a new
candidate solution has been selected and evaluated, the surrogate models are
retrained with the new solution and the process is repeated \emph{ad libitum}.

Mono-surrogate approaches are known to be considerably faster than
multi-surrogate approaches~\citep{chugh:scalarisers:2020}. Using only one
surrogate model offers large computational savings because the model of choice,
GPs, scale cubically in the number of solutions the model is trained 
on~\citep{rasmussen:gpml:2006}, leading to large amounts of computation needed
for larger numbers of solutions. Acquisition functions that have been designed
for BO~\citep[\eg][]{jones:ego:1998,srinivas:ucb:2010,wang:mes:2017,
death:egreedy:2021}, can be used without alteration. Single-objective
acquisition functions are usually cheap to evaluate, unlike those used in
multi-surrogate approaches. These often need to carry out expensive
multidimensional integrations, such as when calculating the expected
hypervolume improvement criterion~\citep{emmerich:ehvi:2005}.

Despite the success of mono-surrogate approaches, they are not without
limitations. The aforementioned cubic scaling of GPs leads to model training
times increasing as the number of solutions grows. Similarly, GPs are known to
suffer greatly from the curse of dimensionality. Specifically, the modelling
ability of nonparametric regression depends exponentially on the problem
dimensionality~\citep{gyorfi:lowerbounds:2002}. Approaches to address these
fundamental problems with GPs usually involve either reducing the number
solutions included in the GP models via \emph{inducing point}
methods~\citep{snelson:sparsegp:2006,titsias:inducing:2009,wilson:kissgp:2015,
salimbeni:OrthogonallyDecoupledVariational:2018,
shi:SparseOrthogonalVariational:2020}, or making assumptions about the
underlying structure of the function of interest to simplify the inference,
such as assuming that it resides on a low-dimensional
manifold~\citep{wang:rembo:2013, kandasamy:hdbo:2015,wang:hdbo:2017,
gardner:hdbo:2017,rolland:hdbo:2018,letham:hdbo:2020}. However, these 
necessarily lead to poorer modelling of target
function~\citep{letham:hdbo:2020}.

An alternative approach to improving the modelling in BO is to use a different
surrogate model, such as a random forest~\citep{hutter:smac:2011} or neural
network~\citep{snoek:scabo:2015}. However, these models do not usually come
with principled ways to compute prediction uncertainty, limiting their
applicability to \emph{Bayesian} optimisation. Recently, \citet{tiao:bore:2021}
proposed to carry out BO via density-ratio estimation (BORE), a generalisation
of the classical tree-structured Parzen estimator~\citep{bergstra:tpe:2011}.
They claimed that maximising the \emph{expected} improvement over the best-seen
function value in a probabilistic regression model, the most popular
formulation of BO, is equivalent to maximising the class-posterior probability
of classifier trained with a proper scoring rule~\citep{gneiting:scoring:2007}.
Unfortunately, this claim is only partially
correct~\citep{song:boreproof:2021}. \citet{garnett:bobook:2022} demonstrates that, in fact, maximising the
class-posterior probability is equivalent to maximising the \emph{probability}
of improvement. Nevertheless, the result is important because it still
facilitates the use of most state-of-the-art classifiers, such as neural
networks or gradient boosted trees~\citep{chen:xgboost:2016}. Naturally, this
leads to a vast reduction in computational complexity as well as an increase in
modelling capability of surrogate models available for use in BO. Therefore,
this allows for higher-dimensional problems to be tackled by using more
suitable models, \eg neural networks, which are well-known for having universal
function approximation guarantees~\citep{hornik:mlp:1989}.

In this work, we extend BORE to the multi-objective setting via the use of
scalarising functions~\citep{hwang:mcd:1979,chugh:scalarisers:2020}, a
mono-surrogate approach. Specifically, we compare \alg, our proposed
multi-objective version of BORE, using two different classification models: a 
fully-connected neural network and gradient boosted machines
(XGBoost~\citep{chen:xgboost:2016}) with the standard BO approach using a GP.
We compare their performance when using three existing scalarisation
approaches, augmented Tchebycheff~\citep{knowles:parego:2006},
hypervolume improvement~\citep{rahat:infill:2017} and
dominance ranking~\citep{rahat:infill:2017}, as well as
\emph{Pareto hypervolume contribution} (PHC), a novel scalarisation method
proposed in this work. Multi-objective BO and \alg are evaluated on two
synthetic benchmark suites, DTLZ~\citep{deb:dtlz:2005} and
WFG~\citep{huband:wfg:2006}, across a range of problem dimensionalities and
numbers of objectives, as well as a recently proposed real-world set of
benchmark problems~\citep{tanabe:rwproblems:2020}. Additionally, we also
investigate the performance of BO and \alg on high-dimensional problems using
the WFG benchmark, and empirically compare their computational costs.

Our main contributions can be summarised as follows:
\begin{enumerate}
    \item We present \alg, a novel mono-surrogate multi-objective 
        classification-based BO approach using scalarisation.
    \item We also present PHC, a new scalariser that directly uses the
        hypervolume contribution of a solution to its Pareto shell. 
    \item We compare \alg and BO, using two probabilistic classifiers
        and several popular scalarisation methods, across a range of synthetic
        and real-world test problems.
    \item Empirically, we show that \alg is equal to or better than BO on a
        wide range of problems, particularly when using PHC, and that \alg is
        considerably better than BO on the evaluated real-world and 
        high-dimensional problems, all while using a fraction of the
        computational resources.
\end{enumerate}

We begin in Section~\ref{sec:bo} by providing an overview of both single- and
multi-objective BO, including a review of GPs, acquisition functions and
scalarisation methods. Section~\ref{sec:bore} introduces BO by density-ratio
estimation and, in Section~\ref{sec:mbore}, we extend it to the
multi-objective setting. An extensive experimental evaluation of BO and  \alg
is carried out in Section~\ref{sec:results}, along with a discussion of the
results. We finish with concluding remarks in Section~\ref{sec:conc}.

\section{Bayesian Optimisation Framework}
\label{sec:bo}
In this section we first describe single-objective Bayesian optimisation,
including its two main components, the Gaussian process surrogate model and
acquisition function. We then go on to show how BO can be extended to the
multi-objective setting via scalarisation, giving examples of the scalarisation
methods used in this work, along with the proposal of a new scalariser.

\subsection{Bayesian Optimisation}
\label{sec:bo:bo}
Bayesian optimisation (BO) was first proposed for single-objective problems by 
\citet{kushner:ego:1964}, and later improved by \citet{mockus:ei:1978} and 
\citet{jones:ego:1998}. Interested readers should refer to
\citep{shahriari:ego:2016}~and~\citep{frazier:tutorial:2018} for recent 
and comprehensive surveys on the topic. BO is a global search strategy that
sequentially samples design space at locations that are likely to contain the
global optimum, while taking into account the prediction of a surrogate model
and its associated uncertainty~\citep{jones:ego:1998}. The single-objective
optimisation problem can be defined as finding a minimum of an unknown
objective function $f : \mX \mapsto \Real$, defined on a compact domain
$\mX \subset \Real^d$:
\begin{equation}
    \label{eq:bo:min}
    \min_{\bx \in \mX} f(\bx).
\end{equation}

BO starts by using a space-filling algorithm, such as Latin hypercube 
sampling~\citep{mckay:lhs:1979}, to generate an initial set of solutions
$\{\bx_i\}_{i=1}^t$, and then expensively evaluates them with the objective
function. These observations form the dataset
$\Data = \{(\bx_i, f_i \triangleq f(\bx_i)) \}_{i=1}^t$ that the
initial surrogate model is trained with. Following model training, and at each
subsequent iteration, the next location to expensively evaluate is chosen as
the location $\xnext$ that maximises an acquisition function $\alpha(\bx)$,
\ie $\xnext = \argmax_{\bx \in \mX} \alpha(\bx)$. The dataset is then augmented
with $\{\xnext, f(\xnext)\}$ and the process is repeated until budget
exhaustion. The global minimum of $f$ is then estimated to be the best-observed
value thus far, $\fstar = \min \{ f_i \}$.

\subsubsection{Gaussian Processes}
\label{sec:bo:bo:gp}
Gaussian processes (GP) are a common choice of surrogate model due to their
strengths in uncertainty quantification and function
approximation~\citep{rasmussen:gpml:2006}. They define a prior distribution
over functions, such that any finite number of drawn function values are
jointly Gaussian, with mean $m(\bx)$ and covariance
$\kappa(\bx, \bx' \given \btheta)$, with hyperparameters $\btheta$. Without
loss of generality, we use a zero-mean prior
$m(\bx) = 0 \, \forall \, \bx \in \mX$; see~\citep{death:gpmean:2020} for
alternatives. Conditioning the GP prior distribution on data consisting of
$t$ sampled locations $\Data = \{(\bx_i, f_i \triangleq f(\bx_i)) \}_{i=1}^t$
leads to a posterior distribution that is also a GP:
\begin{equation}
    \label{eqn:gp:post}
    p ( f(\bx) \given \bx, \Data, \btheta) = \Normal ( \mu(\bx), \sigma^2(\bx) )
\end{equation}
with mean and variance
\begin{align}
    \mu(\bx \given \Data, \btheta)
     & = \bkappa( \bx, \bX ) \bK^{-1} \by
    \label{eqn:gp:pred}                                                  \\
    \sigma^2(\bx \given \Data, \btheta)
     & = \kappa(\bx, \bx) - \bkappa( \bx, \bX)^\top \bK^{-1} \bkappa(\bX, \bx).
    \label{eqn:gp:var}
\end{align}
Here, $\bX \in \Real^{t \times d}$ is a matrix of input locations in each row
and $\by = (f_1, f_2, \dots, f_t)^\top$ is the corresponding vector of
expensive function evaluations. The kernel matrix $\bK \in \Real^{t \times t}$
is $\bK_{ij} = \kappa(\bx_i, \bx_j \given \btheta)$ and
$\bkappa(\bx, \bX) \in \Real^t$ is given by
$[\bkappa(\bx, \bX)]_i = \kappa(\bx, \bx_i \given \btheta)$. In this work we use
a \Matern~$5 / 2$ kernel, as recommended for modelling realistic
functions~\citep{stein:interpolation:1999, snoek:practical:2012}. The kernel's
hyperparameters $\btheta$ are learnt via maximising the log marginal
likelihood~\citep{rasmussen:gpml:2006,death:fullybayes:2021}.

\subsubsection{Acquisition Functions}
\label{sec:bo:bo:acq}
Acquisition functions ${\alpha : \mX \mapsto \Real}$ measure the expected
utility of expensively evaluating $f$ at any location $\bx$. Its maximiser is
chosen as the next location to expensively evaluate. This strategy allows for
an estimate of the global optimum to be generated cheaply via the surrogate
model, rather than by repeatedly querying the expensive objective function.
The probability of improvement~(PI)~\citep{kushner:ego:1964} is one of the
earliest infill criteria. It is the probability that the predicted value at a
location $\bx$ is less than a threshold $\tau$, which, if the posterior
distribution is Gaussian, can be expressed in closed form as
\begin{equation}
    \label{eqn:pi}
    \PI (\bx, \tau) = p( f(\bx) < \tau \given \Data, \btheta) = \Phi(s).
\end{equation}
Here, $\tau$ is usually set to the best solution seen thus far,
$s = (\tau - \mu(\bx)) / \sigma(\bx)$ is the predicted improvement normalised
by its uncertainty, and $\Phi(\cdot)$ is the standard Gaussian cumulative
density function. Its successor, the expected
improvement~(EI)~\citep{mockus:ei:1978,jones:ego:1998}, is one of the most
common acquisition functions and measures the expected positive predicted
improvement over a threshold $\tau$. It is expressible in
closed-form under the same assumptions~\citep{jones:ego:1998}:
\begin{equation}
    \label{eqn:ei}
    \EI (\bx, \tau) = \sigma(\bx) \left[ 
        s \Phi\left( s \right) + \phi\left( s \right)
    \right],
\end{equation}
where $\phi(\cdot)$ is the standard Gaussian probability density function.

In practice, EI is often preferred because PI tends to be
overly-exploitative~\citep{jones:ego:1998,death:egreedy:2021}. However, many
other acquisition functions have been proposed, including optimistic strategies
such as upper confidence bound~\citep{srinivas:ucb:2010},
information-theoretic approaches~\citep{scott:kg:2011,hennig:es:2012,
henrandez-lobato:pes:2014,wang:mes:2017,ru:fitbo:2018}, 
and $\epsilon$-greedy strategies~\citep{bull:convergence:2011,
death:egreedy:2021}.

\subsection{Multi-objective Bayesian Optimisation}
\label{sec:bo:mobo}
Real-world optimisation problems often have multiple conflicting objectives,
all of which must be minimised at the same time~\citep{coello:solvingEAs:2006}.
The multi-objective optimisation problem can be defined as simultaneously
minimising $M \geq 2$ unknown objective functions
\begin{equation}
    \label{eq:mobo:min}
    \min_{\bx \in \mX} \bFF(\bx) = ( f^{1}(\bx), \dots, f^{M}(\bx) ),
\end{equation}
where $f^{i}$ is the $i$-th objective value and $\bFF: \mX \mapsto \Real^M$.
Assuming that problem contains conflicting objectives, then there will not be a
one unique solution. Instead, possibly infinitely many solutions may exist that
trade off between the different objectives. The trade-off between solutions is
often characterised by a \emph{dominance} relationship: $\bx$ is said to
dominate $\xnext$, denoted $\bx \prec \xnext$, iff:
\begin{equation}
    \forall i \in \{1, 2, \dots M\} \left( 
            f^{i}(\bx) \leq f^{i}(\xnext) 
        \right)
    \wedge
    \exists i \, \left( f^{i}(\bx) < f^{i}(\xnext) \right).
\end{equation}
The set containing solutions that optimally trades off between the objectives
is referred to as the Pareto set:
\begin{equation}
    \Pset = \{
        \bx \given \xnext \! \not\prec \bx \, \wedge \,  \bx, \xnext \in \mX
    \},
\end{equation}
where $\xnext \! \not\prec \bx$ indicates that $\xnext$ does not dominate 
$\bx$. Computation of $\Pset$ is infeasible, due to its potentially infinite
size. Thankfully, an approximation is often sufficient, leading to the goal of
generating a good approximation $\Papprox$ to the Pareto set $\Pset$.

\citet{knowles:parego:2006} presented the first mono-surrogate approach for
multi-objective BO. They proposed to \emph{scalarise} the objective function
values $(f^1, f^2, \dots, f^M)$, \ie mapping them to a single value, via the
use of a scalarising function. It uses the randomly-weighted normalised
objective values in an augmented Tchebycheff function, drawing weightings from
a set of predefined weights to scalarise the sets of objective values. These
can then be used in lieu of the objective values, directly within the BO
framework (Section~\ref{sec:bo:bo}). The location that maximises the expected
improvement over the best-seen scalarisation is then chosen to be expensively
evaluated.

\subsubsection{Scalarisation functions}
\label{sec:bo:mobo:sca}
The augmented Tchebycheff scalarisation was the \emph{de facto} choice for
multi-objective BO. However, \citet{rahat:infill:2017} recently proposed
several scalarisation that outperform it. Subsequentially,
\citet{chugh:scalarisers:2020} conducted an exhaustive study of scalarisation
methods for multi-objective BO, comparing those
in~\citep{rahat:infill:2017} to several outside the BO literature. They show
that the choosing the best scalarisation method is far from trivial and is, in
fact, problem dependent. It is also noted that the ease with which a GP may
model a landscape produced by a scalarisation is problem dependent, \ie the
same scalarisation may produce easier or harder landscapes for a GP to model,
relative to other methods.

The main properties required for a scalarisation to be used in the BO framework
is that it should preserve dominance relationships and allow for every Pareto
optimal solution to be reached \citep{hwang:mcd:1979}. In doing so, the
maximisation of a dominance-preserving scalarisation should lead to an
improvement in the Pareto set~\citep{zitzler:indicators:2004}. Although many
methods have been proposed, we focus on the popular augmented Tchebycheff
method~\citep{knowles:parego:2006}, two of the best-performing
in~\citep{rahat:infill:2017}, and a novel scalarisation proposed in this work.

\paragraph{Augmented Tchebycheff (AT)}
First presented for BO by~\citet{knowles:parego:2006}, the augmented
Tchebycheff method combines the Tchebycheff function with a weighted sum,
scaled by a small positive value $\rho$:
\begin{equation}
    \label{eqn:sca:atc}
    \parego(\bx, X) = \max_{i} \left[ w^{i} \bar{f}^{i} \right]
                    + \rho \sum_{i=1}^M w^{i} \bar{f}^{i},
\end{equation}
where $\bar{f}^{i} \in [0, 1]$ are the normalised versions of the observed
objective values. At each BO iteration, the weights
$\bww = (w^{1}, w^{2}, \dots, w^{M})$, where $|\bww| = 1$, are sampled uniformly
from a fixed set of evenly distributed weight vectors on the $M$-dimensional
unit simplex.

\paragraph{Hypervolume Improvement (HypI)}
The hypervolume indicator $H(X, \brr)$ measures the volume of space dominated
by a set of solutions $X = \{\bx_1, \bx_2, \dots, \bx_t\}$ relative to a
reference vector $\brr$~\citep{zitzler:eamoo:1999}. It is a popular choice for
comparing two sets of solutions in because maximising it is equivalent to
locating the optimal Pareto set~\citep{fleischer:pareto:2003}. In order to turn
the hypervolume indicator into a scalariser, it is natural to consider the
\emph{contribution} of each set member, \ie the amount of hypervolume gained by
including a set member. However, only solutions that reside within the Pareto
set of $X$ will have a non-zero contribution, even if solutions dominated by
the Pareto set dominate other solutions, they will all be assigned a value of
zero. This would hinder the progress of BO because it would create a plateau
that lacks spatial information about solution quality.

\citet{rahat:infill:2017} proposed a solution to the inherent
problems of solely using the hypervolume contribution that they named the
hypervolume improvement (HYPI). The members of $X$ are first ranked according
to Pareto shells in which they reside. Let the first Pareto shell
$\Pset^{1}$ be the Pareto set of $X$, \ie $\Pset^{1} = \textsc{nondom}(X)$,
where $\textsc{nondom}(X)$ returns the non-dominated members of $X$. Subsequent
Pareto shells $\Pset^{\lambda}$, with $(\lambda > 1)$, can then be defined as
\begin{equation}
    \Pset^{\lambda} = \textsc{nondom} \left(
            X \setminus \bigcup_{i=1}^{\lambda - 1} \Pset^{i}
    \right).
\end{equation}
Once the members of $X$ have been ranked, the HYPI of a solution $\bx$ is
defined to be the hypervolume of the union of $\bx$ and the first Pareto shell
$\Pset^{\dagger}$ that contains no solutions that dominate it:
\begin{equation}
    \label{eqn:sca:hypi}
    \hypi(\bx, X) = H \left( \{ \bx \} \cup \Pset^{\dagger}, \brr \right).
\end{equation}
Therefore, each solution has a non-zero, dominance-preserving scalarisation,
with values that provide a gradient towards non-dominated space.

\paragraph{Dominance Ranking (DomRank)}
An alternative way to compare multi-objective solutions is to count the number
of solutions that dominate them, with the idea being that we should prefer
solutions that are dominated less~\citep{fonseca:mooga:1993}.
\citet{rahat:infill:2017} use this idea to form the DomRank scalarisation:
\begin{equation}
    \label{eqn:sca:dr}
    \domrank(\bx, X) = 1 - \frac{|\textsc{dom}(\bx, X)|}{ |X| - 1},
\end{equation}
where $\textsc{dom}(\bx, X)$ returns the set of solutions in $X$ that dominate
$\bx$. The authors found that DomRank was performed similarly to other
scalarisations, even though, in the degenerate setting, it is possible for all
members of $X$ to be non-dominated with respect to one another, resulting in
equal scalarisations for all $X$. However, this should be expected to happen
more frequently as the number of objectives $M$ increases because the
likelihood of one solution dominating another is inversely proportional to
$M$~\citep{li:moeas:2015}.

\paragraph{Pareto Hypervolume Contribution (PHC)}
Inspired by the success of HYPI, we present an alternative scalariser that can
directly use the hypervolume contribution of each solution: PHC. Let
$v(\bx, \Pset^{\lambda})$ be a function that calculates the hypervolume
contribution of a solution $\bx$ to the shell $\Pset^{\lambda}$ it resides in:
\begin{equation}
    v(\bx, \Pset^{\lambda}) = 
        H(\Pset^{\lambda}, \brr) - H(\Pset^{\lambda} \setminus \{\bx\}, \brr).
\end{equation}
The PHC of a solution $\bx$ can be then be calculated by taking its hypervolume
contribution and adding the largest contribution from each subsequent shell:
\begin{equation}
    \label{eqn:sca:phc}
    \phc(\bx, X) = v(\bx, \Pset^{\lambda}) + \sum_{i=\lambda+1}^N 
        \max \{ v(\xnext , \Pset^{i}) \given \xnext \in \Pset^{i} \},
\end{equation}
where $N$ is the total number of Pareto shells. If $\bx$ dominates $\xnext$,
then $\xnext$ will be in a subsequent shell to $\bx$ and, by definition,
$\phc(\bx, X) > \phc(\xnext, X)$. Therefore, it preserves dominance
relationships between members of $X$ and ensures monotonicity between shells.

\section{BORE: Bayesian Optimisation by Density-Ratio Estimation}
\label{sec:bore}
BO is a popular and successful strategy for both single- and multi-objective
optimisation of expensive black-box problems. However, GPs, the surrogate 
models of choice in BO, present some notable limitations. Specifically, its
$\bigO(n^3)$ computational complexity in the number $n$ of solutions the GP is
trained with, as well as additional assumptions which, often, do not apply to
the functions that they are trying to model. GPs almost exclusively use
stationary kernels, \ie they only depend on the distance between solutions.
This means that they are incapable of modelling different length-scales of
data in different regions of $\mX$.

Motivated by these shortcomings, \citet{bergstra:tpe:2011} presented a
reformulation of BO by estimating the probability of improvement~\eqref{eqn:pi}
acquisition function by density-ratio estimation. We note that the authors
claimed that the expected improvement was being estimated, but this has since
been proven to be incorrect~\citep{song:boreproof:2021,garnett:bobook:2022}. We
start, following the exposition of~\citep{garnett:bobook:2022}, by considering
two densities that depend on a threshold $\tau = \Phi^{-1}(\gamma)$ that is the
$\gamma$-th quantile of the observed objective values $f$, where
$0 < \gamma \leq 1$ and $\gamma = \Phi(\tau) = p(f \leq \tau)$:
\begin{align}
    \label{eqn:bore:g}
    b(\bx)    & = p( \bx \given f < \tau ) \\
    \label{eqn:bore:l}
    \ell(\bx) & = p( \bx \given f \geq \tau ).
\end{align}
Here, $b$ is the probability density of observed values being less than the
threshold $\tau$, and $\ell$ of observations not being less than $\tau$. Using
Bayes rule, $\ell$ and $b$ can be expressed as being proportional to PI:
\begin{align}
    \label{eqn:bore:gapprox}
    b(\bx)    & \propto p( f < \tau    ) p(\bx) =      \PI(\bx, \tau) p(\bx) \\
    \label{eqn:bore:lapprox}
    \ell(\bx) & \propto p( f \geq \tau ) p(\bx) 
        = \left(1 - \PI(\bx, \tau)\right) p(\bx),
\end{align}
where $p(\bx)$ is a prior density over $\mX$. \citet{bergstra:tpe:2011} use the
ratio of these densities as an acquisition function that monotonically
increases with $\PI(\bx, \tau)$:
\begin{equation}
    \label{eqn:tpe}
    \TPE(\bx, \tau) = \frac{ b(\bx) }{ \ell(\bx) }
        \propto \frac{ \PI(\bx, \tau) } { 1 - \PI(\bx, \tau) },
\end{equation}
where the prior density $p(\bx)$ cancels under the assumption that it has
support over all of $\mX$. Therefore, maximising $\TPE(\bx, \tau)$ is
equivalent to maximising $\PI(\bx, \tau)$.

This transforms the problem training a GP model and maximising
PI~\eqref{eqn:pi}, to one of estimating two densities and maximising their
ratio~\eqref{eqn:tpe}. Using tree-based kernel density
estimators~\citep{silverman:density:1986} to estimate $\ell$ and $b$, results
in the tree-structured Parzen estimator (TPE)~\citep{bergstra:tpe:2011}, and
reduces the computational cost to $\bigO(n)$.

\citet{tiao:bore:2021} provide a full discussion on the weaknesses of TPE.
Here, we highlight the most important. Density estimation in higher dimensions
is notoriously difficult~\citep{sugiyama:dre:2012}, and even the estimating the
kernel's bandwidth and choosing the correct kernel, is far from
trivial~\citep{terrell:vkde:1992}. We also note that even with correct density
estimation, the optimisation of \eqref{eqn:tpe} is often numerically
unstable~\citep{yamada:relDRE:2011}.

Instead of trying to first estimate the densities and then their ratio, one can
directly estimate their ratio by exploiting results in class-probability
estimation~\citep{cheng:dre:2004,bickel:dre:2007,sugiyama:dre:2012,
menon:linkingdre:2016}. It can be shown \citep{tiao:bore:2021} that the
$\gamma$-relative density-ratio of \eqref{eqn:bore:g} and \eqref{eqn:bore:l},
\begin{equation}
    \label{eqn:bore:rde}
    r_\gamma(\bx) =
        \frac{ \ell(\bx) }  { \gamma \ell(\bx) + (1 - \gamma) b(\bx) },
\end{equation}
is proportional to the probability of improvement \eqref{eqn:pi}:
\begin{equation}
    \PI(\bx, \Phi^{-1}(\gamma)) \propto r_\gamma(\bx).
\end{equation}
We note that the TPE acquisition function \eqref{eqn:tpe} is a special case 
of \eqref{eqn:bore:rde}, \ie $\TPE(\bx, \tau) \equiv r_0(\bx)$.

Next, we introduce a binary class label $z$ such that
\begin{equation}
    z = 
    \begin{cases} 
        \, 1 & \text{if} \,\, f < \tau \\
        \, 0 & \text{if} \,\, f \geq \tau.
    \end{cases}
\end{equation}
If the class-posterior probability of $f < \tau$ given an observation $\bx$ is
defined to be $\pi(\bx) = p( z = 1 \given \bx )$, and noting that
$\ell(\bx) = p( \bx \given z = 1 )$ and $b(\bx) = p( \bx \given z = 0 )$,
it can also be shown~\citep{tiao:bore:2021} that
\begin{equation}
    r_\gamma(\bx) \equiv \gamma^{-1} \pi(\bx).
\end{equation}
A probabilistic classifier, \eg a neural network, can be used to estimate
$\pi(\bx)$ by learning a function $\pi_{\bxi} : \mX \mapsto [0, 1]$
with parameters $\bxi$. Provided that the classifier is trained with a proper
scoring rule~\citep{gneiting:scoring:2007}, such as the binary cross entropy
(log loss), the relative density-ratio is approximated by
\begin{equation}
    \label{eqn:bore:approx}
    \pi_{\bxi}(\bx) \approx \gamma r_\gamma(\bx).
\end{equation}

This result links PI \eqref{eqn:pi} to density-ratio estimation and onto
class-probablity estimation, leading to the BO by density-ratio (BORE)
framework. It transforms the problem from training a GP and maximising an
acquisition function, into one of training a probabilistic classifier and
maximising $\pi_{\bxi}(\bx)$. To carry out BORE, a proportion $\gamma$ is first
chosen, and the best $\gamma$-th proportion of solutions are labelled as class
$1$, with the remaining labelled as class $0$. A classifier $\pi_{\bxi}$ is
then trained on the two sets of solutions, and
$\xnext = \argmax_\bx \pi_{\bxi}(\bx)$ is chosen as the next location to be
expensively evaluated.

The two main hyperparameters of BORE are the proportion $\gamma$ of solutions
to include in class $1$ and the choice of $\pi_{\bxi}$. Increasing $\gamma$
encourages exploration because it will result in a worse threshold $\tau$ from
which to (effectively) calculate PI with, \ie the likelihood of a solution
exceeding a worse threshold will be higher. In the original BORE
formulation~\citet{tiao:bore:2021}, $\gamma$ is fixed throughout optimisation.
Although it is fixed, exploitation increases as more solutions are evaluated
because the class threshold will be more biased towards the better solutions
collected during optimisation. The choice of $\pi_{\bxi}$ controls the types of
functions that can be modelled during optimisation. Multi-layer perceptrons
(MLPs) are a natural choice due to their function approximation
guarantees~\citep{hornik:mlp:1989}, their ability to scale to arbitrary
problem dimensions, and for being end-to-end differentiable. This latter point
means that quasi-newton based methods, such as
L-BFGS-B~\citep{byrd:bfgsb:1995}, can be used to optimise $\pi_{\bxi}(\bx)$.
Ensemble-based methods are an attractive alternative to MLPs because they also
scale well. They are not, however, end-to-end differentiable meaning that
non-gradient-based optimisation methods must be used such as, \eg
CMA-ES \citep{hansen:bipopcmaes:2009}. In this work we use MLPs and an
ensemble method, gradient-boosted trees (XGBoost), because they can be both
trained using a proper scoring rule, thereby ensuring a good approximation to
\eqref{eqn:bore:approx}.

\section{\alg: Multi-objective BORE}
\label{sec:mbore}
In this section, we introduce Multi-objective BO by Density-Ratio Estimation
(\alg). It extends BORE to the multi-objective setting by scalarising the
previously-evaluated solutions and separating them into two classes via
thresholding the scalarised values, thereby enabling a probabilistic classifier
to be trained. In doing so, to the best of our knowledge, we present the first
classification-based multi-objective method using scalarisation for BO. Before
discussing \alg further, we first give an overview of related work in the field
of classification-based multi-objective optimisation.

Classification-based multi-objective optimisation methods using surrogate
models often focus on predicting the quality of an evolved population of
solutions. Two main schemes are proposed, either predicting which Pareto shell
a candidate solution would belong to, or whether the current population (set of
previously-evaluated solutions) dominates it. Rank-based support vector
machines (SVM)~\citep{joachims:ranksvm:2005,li:ordreg:2007}, for example, have
been proposed~\citep{loshchilov:svmrank:2010,seah:ranking:2012} to predict the
shell a candidate solution belongs to. However, dominance prediction is a much
more common strategy. Early works include using a one-class SVM to models
whether solutions were dominated~\citep{loshchilov:dompred:2010}, and using a
naive Bayes classifier to predict whether one solution dominates
another~\citep{guo:preddom:2012}. More recently, the direction of research has
moved towards learning to predict whether candidate solutions are dominated by
an existing set of solutions, for example, by using k-nearest
neighbours~\citep{zhang:preselection:2018} or MLP
classifiers~\citep{pan:clfdom:2019,yuan:dompred:2021}.

The classification-based approach we present in this paper, \ie separating
solutions into two classes based on their scalarisation, is conceptually
similar to previous work on, \eg, classification based on dominance. However,
there are some important differences and benefits from using scalarisation.
Most importantly is that using scalarisation and the subsequent thresholding
into classes, allows for PI~\eqref{eqn:pi} to be calculated via the BORE
framework. This provides a much-needed theoretical motivation as to why a
classification-based approach is suitable for multi-objective optimisation. The
size of the proportion $\gamma$ gives control over how many solutions are
placed into each class. Contrastingly, if a set of solutions were completely
non-dominated with respect to one another, using a domination-based approach
would result in an empty class.

\begin{algorithm} [t!]
\caption{Multi-objective BO by Density-Ratio Estimation}
\label{alg:mbore}
\begin{algorithmic}[]
    \State \textbf{Inputs}: 
        Number of initial samples $S$, 
        total budget $T$,
        scalarising function $g(\cdot)$, 
        probabilistic classifier $\pi_{\bxi}(\cdot)$,
        proportion $\gamma$.
\end{algorithmic}

\begin{algorithmic}[1]
    \Statex \textbf{Steps:}
    \State $X \gets \LatinHypercubeSampling(\mX, S)$
        \label{alg:mbore:lhs}
        \Comment{\small{Generate initial samples}}

    \State $\bff_i \gets \bFF(\bx_i) \text{~for~} i \in \{1, \dots, S\}$
        \Comment{\small{Expensively evaluate samples}}

    \For{$t = S+1 \rightarrow T$}
        \State $g_i \gets g(\bff_i) \text{~for~} i \in \{1, \dots, t\}$
            \label{alg:mbore:scalarise}
            \Comment{\small{Scalarise objective values}}

        \State $\tau \gets \Phi^{-1}(\gamma)$
            \label{alg:mbore:tau}
            \Comment{\small{Calculate $\gamma$-th quantile of $\{ g_i \}_{i=1}^t$ }}

        \State $z_i \gets \mathbb{I}[g_i < \tau] \text{~for~} i \in \{1, \dots, t\}$
            \label{alg:mbore:label}
            \Comment{\small{Assign class labels}}

        \State $\bxi^{\star} \gets  \text{TrainClassifier}(\pi_{\bxi}, \{ (\bx_i, z_i) \}_{i=1}^t)$ 
            \label{alg:mbore:train}

        \State $\bx_{t+1} \gets \argmax_{\bx \in \mX} \pi_{\bxi^{\star}} (\bx)$ 
            \label{alg:mbore:xnext}
            \Comment{\small{Maximise class-posterior probability}}

        \State $\bff_{t+1} \gets \bFF(\bx_{t+1})$ 
            \label{alg:mbore:eval}
            \Comment{\small{Expensively evaluate new solution}}

    \EndFor
\end{algorithmic}
\end{algorithm}

Motivated by the success of BORE in the single-objective setting, we present
\alg. Algorithm~\ref{alg:mbore} outlines its general structure. It starts
(line~\ref{alg:mbore:lhs}), identically to BO, with a space-filling design such
as Latin hypercube sampling~\citep{mckay:lhs:1979}. These samples
$X = \{\bx_i\}_{i=1}^S$ are then expensively evaluated with the objective
functions, \ie $\bff_i = \bFF(\bx_i)$. At each subsequent iteration of the
algorithm, the objective values are scalarised (line~\ref{alg:mbore:scalarise})
and the $\gamma$-th quantile $\tau$ of the scalarised objective values is
calculated (line~\ref{alg:mbore:tau}). The solutions are then split into two
classes (line~\ref{alg:mbore:label}), with solutions that have a scalarisation
of less $\tau$ labelled as class $1$, and the remaining labelled as class $0$.
Next, a probabilistic classifier is trained (line~\ref{alg:mbore:train}), and
the location that maximises its prediction is chosen as the next location to
evaluate (lines~\ref{alg:mbore:xnext}~and~\ref{alg:mbore:eval}). This process
is then repeated until budget depletion.

Given the clear link to PI, and the strengths of MLPs and XGBoost in function
approximation, one might expect \alg to perform similarly to the traditional
mono-surrogate approach (BO) on low-dimensional problems (\eg $d \leq 10$) and
to improve upon BO on problems that have a larger dimensionality. In the
following section we investigate this by comparing BO to \alg on a wide variety
of low- and high-dimensional synthetic and real-world problems.

\section{Experimental Evaluation}
\label{sec:results}
The performance of \alg is investigated using two probabilistic classifiers,
XGBoost~\citep{chen:xgboost:2016} (XGB) and a multi-layer perception (MLP). We
compare it to the standard mono-surrogate BO
approach~\citep{knowles:parego:2006}, often referred to as
ParEGO~\citep{knowles:parego:2006}, using a GP~\eqref{eqn:gp:post} surrogate
model and the EI~\eqref{eqn:ei} acquisition function. The methods are compared
using two popular synthetic benchmarks,
DTLZ~\citep{deb:dtlz:2005} and WFG~\citep{huband:wfg:2006}, along with a set of
real-world problems~\citep{tanabe:rwproblems:2020}. Experiments are carried out
using the typical number of problem dimensions, \eg $d \leq 10$, with varying
numbers of objectives $M$. A high-dimensional, \eg $d \in \{20, 50, 100\}$,
version of the WFG benchmark is also evaluated to assess optimisation
performance in more challenging scenarios. Experiments are repeated for the
scalarisation methods discussed in
Section~\ref{sec:bo:mobo:sca}:
augmented Tchebycheff (AT)~\citep{knowles:parego:2006},
hypervolume improvement (HYPI)~\citep{rahat:infill:2017},
dominance ranking (DomRank)~\citep{rahat:infill:2017}, and our novel
scalariser, Pareto hypervolume contribution (PHC)~\eqref{eqn:sca:phc}.

The two versions of \alg (XGB and MLP) were created and trained using the same
configurations as BORE~\citep[][Appendix~J]{tiao:bore:2021}. A zero-mean GP
with an ARD \Matern~$5/2$ kernel was used for BO. At each iteration, before new
locations were selected, the hyperparameters of the GP were optimised by
maximising the log likelihood~\citep{rasmussen:gpml:2006} using 
L-BFGS-B~\citep{byrd:bfgsb:1995} with a multi-restart
strategy~\citep{wilson:maxacq:2018}, and choosing the best set of
hyperparameters from $10$ restarts for the model. The weight vectors for the
AT scalarisation were calculated via the Riesz' \emph{s-Energy}
method~\citep{blank:rieszs:2021}; see the supplement for more details.

Input variables were normalised to reside in $[0, 1]^d$ before they were used
to train both \alg and BO. Objective values similarly normalised on a
per-objective basis. The models were initially trained on $S = 2d$ observations
generated by maximin Latin hyper-cube sampling~\citep{mckay:lhs:1979} and then
optimisation was carried out for a further $300$ function evaluations. Each
optimisation run was repeated $21$ times from a different set of Latin
hypercube samples. Initial sets of locations were common across all methods to
enable statistical comparison. The next location to evaluate,
\ie $\xnext = \argmax_\bx \pi_{\bxi}(\bx)$, was selected for the MLP as
in~\citep{tiao:bore:2021}, and for XGB using bi-pop
CMA-ES~\citep{hansen:bipopcmaes:2009} with $10$ restarts. EI~\eqref{eqn:ei} in
BO was optimised with a multi-restart strategy of first sampling $B = 1024d$
randomly-chosen locations and then optimising the best $10$ of these with
L-BFGS-B. The budget for optimising $\pi_{\bxi}$ for both the MLP and XGB
classifiers was also set to $B$ to ensure fair comparison.
Like~\citet{tiao:bore:2021}, we fix $\gamma = 1/3$ for all experiments.

Performance is reported in terms of the
hypervolume~(HV)~\citep{zitzler:indicators:2004} of the estimated Pareto set;
IGD+~\citep{ishibuchi:igdplus:2015} is also reported in the supplement. The BO
optimisation pipeline was constructed using
GPyTorch~\citep{gardner:gpytorch:2018} and
BoTorch~\citep{balandat:botorch:2020}. \alg uses
pagmo~\citep{biscani:pagmo:2020} for carrying out fast
dominated sorting and hypervolume calculation, as well as 
pymoo~\citep{blank:pymoo:2020} for the IGD+ calculation. Code for all methods,
as well as the initial starting solutions, reference vectors and the estimated
Pareto sets for HV and IGD+ is available
online\footnote{\url{http://www.github.com/georgedeath/mbore}}.

\subsection{Synthetic Benchmarks}
\label{sec:results:synth}
\begin{table}[t]
\renewcommand{\arraystretch}{0.9}
\begin{tabular}[t]{l c l}
\toprule
    Name & IDs & Problem configurations ($d$, $M$) \\
\midrule
    DTLZ & 1--7 & (2,2), (5,2), (5,3), (5,5), (10,2), (10,3), (10,5), (10,10) \\
    WFG  & 1--9 & (6,2), (6,3), (8,2), (8,3), (10,2), (10,3), (10,5) \\
\bottomrule
\end{tabular}
\caption{Benchmarks, problem IDs and configurations.}
\label{tbl:synth_benchmarks}
\end{table}
The popular DTLZ~\citep{deb:dtlz:2005} and WFG~\citep{huband:wfg:2006}
benchmark problem suites were selected to compare \alg and BO. The benchmarks
were chosen because both the problem dimensionality $d$ and the number of
objectives $M$ are configurable, allowing for a large range of problems to be
generated. Table~\ref{tbl:synth_benchmarks} outlines the benchmark problems
used, including the specific problem numbers (IDs) and combinations of
($d$, $M$) used from each suite, resulting in $56$ and $63$ distinct test
problems from DTLZ and WFG respectively. Following standard
practice~\citep{blank:pymoo:2020,chugh:scalarisers:2020}, the WFG position
and scale parameters $(k, l)$ were set to $(4, d - 4)$ for $M = 2$, and
$(2M - 1, d - (2M - 1))$ for $M > 2$.

\begin{table}[t]
\renewcommand{\arraystretch}{0.9}
\begin{tabular}[t]{l c x{1cm} x{1cm} x{1cm}}
\addlinespace[-\aboverulesep]\cmidrule[\heavyrulewidth]{3-5}
    &      &           XGB &            MLP &       GP  \\
\midrule
    \multirow{2}{*}{\rotatebox[origin=c]{90}{HYPI}}
    & DTLZ &     \best{36} &   \phantom{0}4 &       29  \\
    & WFG  &     \best{37} &   \phantom{0}0 &       36  \\
\midrule
    \multirow{2}{*}{\rotatebox[origin=c]{90}{DR}}
    & DTLZ &           21  &             11 & \best{37} \\
    & WFG  &           36  &   \phantom{0}0 & \best{45} \\
\midrule
    \multirow{2}{*}{\rotatebox[origin=c]{90}{PHC}} 
    & DTLZ &     \best{36} &             12 &       22  \\
    & WFG  &     \best{39} &   \phantom{0}0 &       36  \\
\midrule
    \multirow{2}{*}{\rotatebox[origin=c]{90}{AT}}
    & DTLZ &           13  &   \phantom{0}8 & \best{43} \\
    & WFG  & \phantom{0}1  &   \phantom{0}0 & \best{63} \\
\bottomrule
\end{tabular}
\caption{%
    Performance summary of \alg (MLP and XGB) and BO (GP) for a given
    scalariser on the benchmarks. Table values correspond to the number of
    times each model was the best or statistically equivalent to the best
    model.
}
\label{tbl:summary}
\end{table}
We start by comparing the performance of \alg using the XGB and MLP 
classification models to mono-surrogate BO with a GP surrogate model.
Table~\ref{tbl:summary} shows the number of times a model (XGB, MLP, and GP) is
best on each benchmark for a given scalarisation method. Specifically, a model
is counted as being the best if it has the largest median hypervolume over the
$21$ optimisation runs, or it is statistically indistinguishable from the best
method, according to a one-sided, paired Wilcoxon signed-rank
test~\citep{knowles:testing:2006} with Holm-Bonerroni
correction~\citep{holm:test:1979}. Models that are the best the highest number
of times on each benchmark for a given scalariser (table rows) are highlighted
in grey. Interestingly, either XGB or GP is the best for each scalarisation
method, with results relatively close for HYPI, DomRank (DR) and PHC. This
highlights the efficacy of \alg for general-purpose use in multi-objective BO,
and suggests that it should be preferred over BO when using the HYPI and PHC
scalarisation methods. Note that \alg is outperforming BO even though it is
approximating PI, an acquisition strategy that is generally regarded as being
inferior to EI~\citep{death:egreedy:2021,garnett:bobook:2022}. We suspect that
this is because the increased modelling capacity of XGB is outweighing the
marginally worse performance of PI.

\begin{figure}[t]
\includegraphics[width=\linewidth, clip, trim={0 13 0 0}]{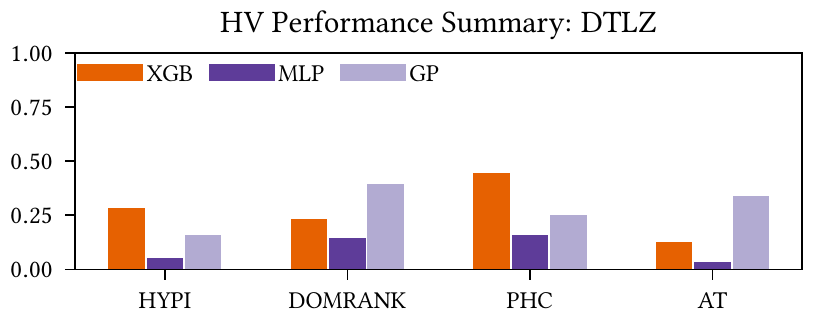}\\
\includegraphics[width=\linewidth]{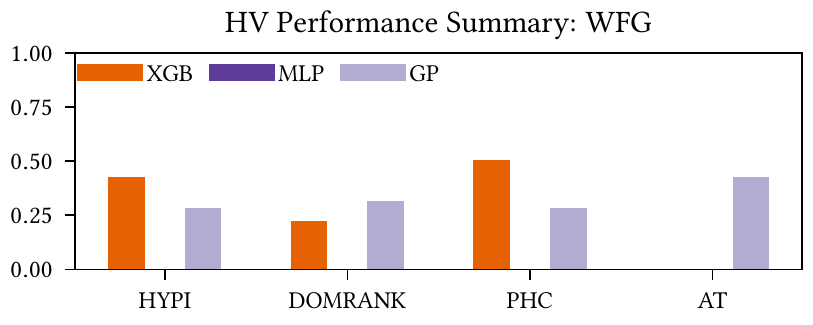}%
\caption{%
    Hypervolume (HV) performance summary for the DTLZ and WFG benchmarks. Bar
    heights correspond to the proportion of times that a model and scalariser
    combination is best or statistically equivalent to the best method.
}
\label{fig:synth_summary_barchart}
\end{figure}
Given that Table~\ref{tbl:summary} only compares performance between models for
a given scalarisation, we now evaluate which combination of scalarisation and
model performs the best on both the DTLZ and WFG benchmarks.
Figure~\ref{fig:synth_summary_barchart} summarises the performance for all
combinations of model and scalariser. Bar heights correspond to the proportion
of times each model and scalarisation combination was the best on each test
problem. As can be seen from the figure, \alg with XGB and our novel PHC
scalariser~\eqref{eqn:sca:phc} has the best overall performance for both
benchmarks. Surprisingly, \alg with the MLP classification method performs
worse than the best performing method on all the $63$ WFG test
problems. However, this reflects the results presented
in~\citep{tiao:bore:2021} that also show that XGB tends outperform MLP. We note
that relative rankings of methods remain the same when using IGD+; see the
supplement for details.

\begin{figure}[t] 
\centering
\includegraphics[width=\linewidth]{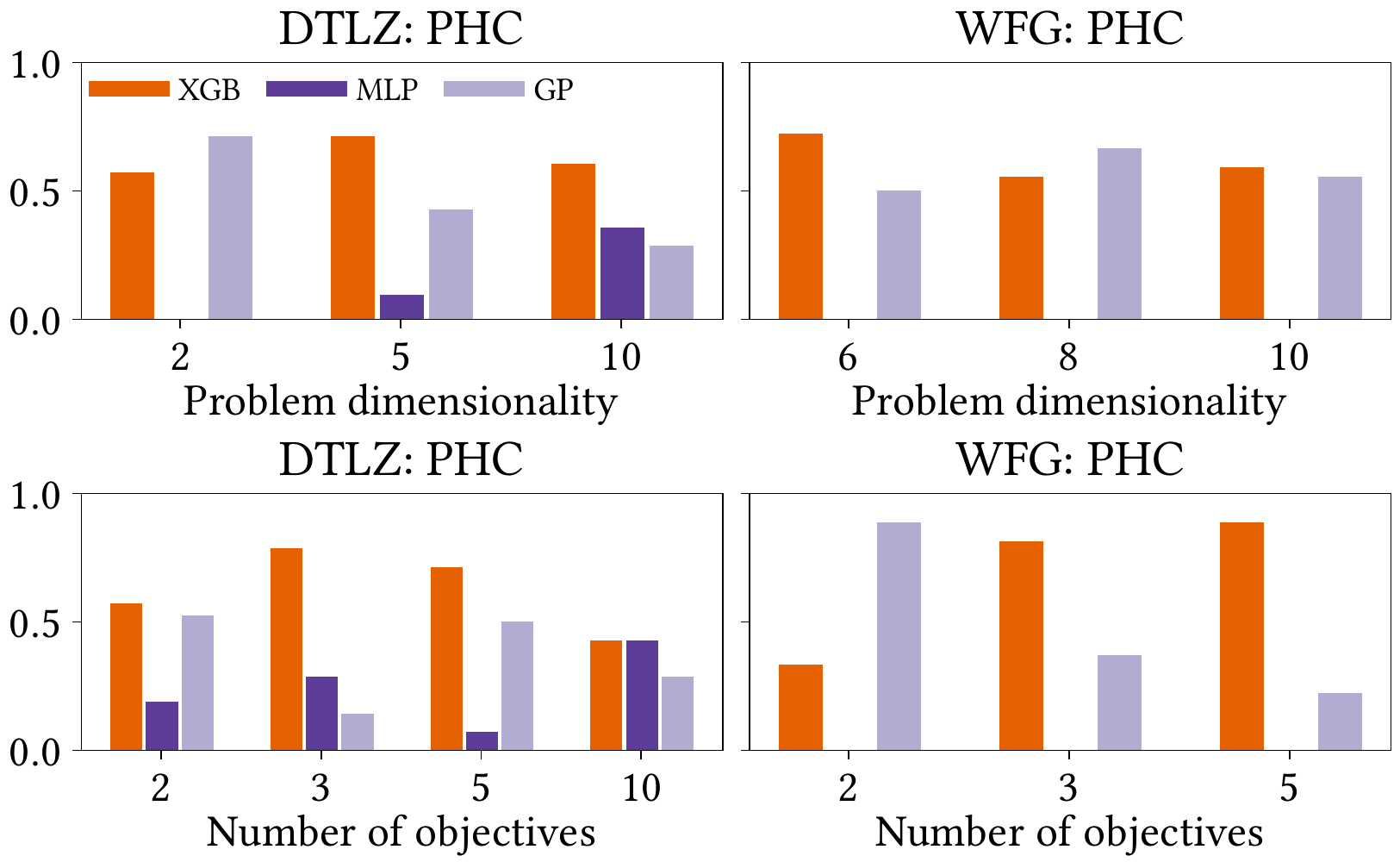}%
\caption{%
    Hypervolume performance summary for the PHC scalariser on the
    DTLZ (\emph{left}) and WFG (\emph{right}) benchmarks given the problem's
    dimensionality (\emph{upper}) and number of objectives (\emph{lower}).
}
\label{fig:synth_phc}
\end{figure}
Next, we investigate the performance of the models with respect to a problem's
dimensionality~$d$ and number of objectives~$M$. Since PHC was the
best-performing method, we limit this discussion to its results. Results for
all scalarisers are available in the supplement. Figure~\ref{fig:synth_phc}
summarises the performance of each combination of model and scalariser for the
two benchmarks in terms of $d$~and~$M$. The DTLZ results show that BO's
performance deteriorates as~$d$ increases, a trend mirrored for the other
scalarisations. Conversely, there is no clear pattern regarding $M$ for DLTZ.
In the WFG benchmark the reverse is true, changing $d$ does not result in a
consistent trend, but the performance of BO decreases and XGB increases as $M$
increases across the majority of scalarisation methods. We conjecture that this
is related to the increased complexity of the WFG
problems~\citep{huband:wfg:2006}, compared to DTLZ. Therefore, increasing~$M$
leads to an increase in the scalarisation landscape's complexity, resulting in
features that GPs model poorly, such as discontinuities.

\subsection{Real-world Benchmark}
\label{sec:results:rw}
Hand-crafted benchmarks, such as DTLZ and WFG, are known to have properties
that are unlikely to appear in real-world
applications~\citep{ishibuchi:perf:2017,zapotecasmartinez:perf:2019}.
Therefore, to investigate the performance of \alg in more realistic settings,
we turn to the real-world benchmark of \citet{tanabe:rwproblems:2020}. It has
$11$ continuous-valued test problems taken from real-world problems, such as
car side impact design~\citep{jain:carsideimpact:2014} and water resource
planning~\citep{ray:waterplanning:2001}; see~\citep{tanabe:rwproblems:2020} for
detailed descriptions. The problem \emph{RE3-4-3} was not included due to
numerical instabilities; we focus on the remaining~$10$ problems.

\begin{figure}[t]
\includegraphics[width=\linewidth]{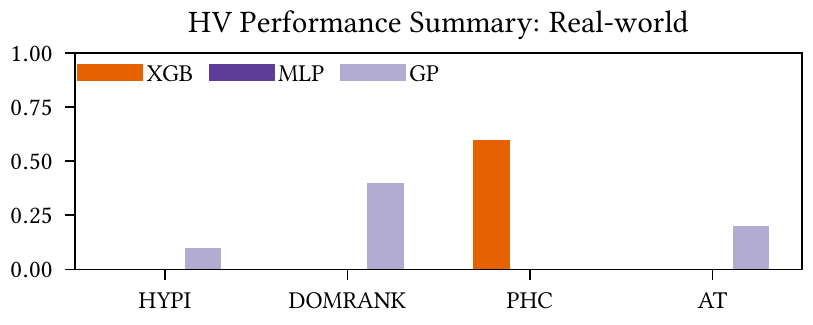}\\
\caption{%
    Hypervolume (HV) performance summary for the real-world benchmark problems.
    Bar heights correspond to the proportion of times that a model and
    scalariser combination is best or statistically equivalent to the best
    method.
}
\label{fig:rw_summary_barchart}
\end{figure}
Figure~\ref{fig:rw_summary_barchart} summarises the performance of each
combination of model and scalariser on the real-world problem benchmark. As can
be seen from the figure, \alg with XGB and using the PHC scalarisation is by
far the best method for optimising the real-world problems based on
hypervolume. Interestingly, for DomRank and AT, using BO (GP) was more
effective than \alg, with HYPI and PHC roughly equal between the two methods;
see the supplement for full details. These results again show that using
\alg is suitable for multi-objective optimisation and outperforms BO when using
more effective scalarisation methods.

\subsection{High-dimensional Benchmark}
\label{sec:results:hd}
\begin{figure}[t]
\includegraphics[width=\linewidth]{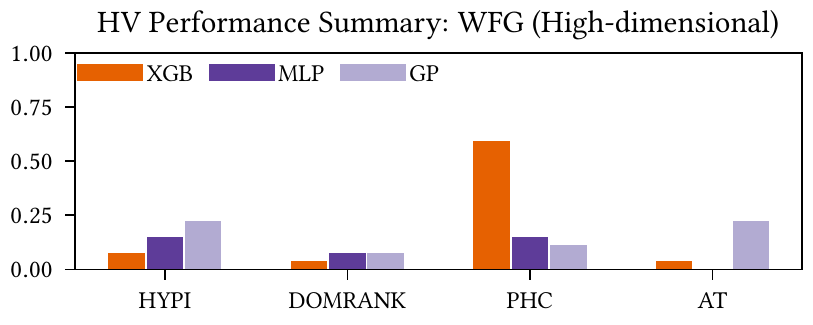}\\
\caption{%
    Hypervolume (HV) performance summary for the high-dimensional WFG 
    benchmark. Bar heights correspond to the proportion of times that a model
    and scalariser combination is best or statistically equivalent to the best
    method.
}
\label{fig:hd_summary_barchart}
\end{figure}
BO often struggles on high-dimensional problems due to the surrogate models of
choice, GPs, decreasing in modelling accuracy as the problem dimensionality
increases~\citep{gyorfi:lowerbounds:2002}. Consequentially, comparisons of
multi-objective methods often focus on lower numbers of both objectives $M$
and $d$. In contrast to this, and in order to evaluate \alg in more realistic
settings where both $d$ and $M$ are comparatively large, we investigate
optimisation performance on the WFG benchmark~\citep{huband:wfg:2006} with a
large dimensionality~$d \in \{20, 50, 100\}$. Due to the computational costs of
training GPs in high dimensions, we limit the number of objectives to one
configuration ($M=10$); see the following section for a comparison of the
computational costs.

\looseness=-1 
Figure~\ref{fig:hd_summary_barchart} summarises the performance on these
problems. The efficacy of \alg is repeated for the high-dimesional versions of
the WFG problems, with the combination of \alg, XGB and the PHC scalariser
being the best on roughly two thirds of the problems. This is somewhat expected
because, in the high-dimensional ($d \geq 20$) setting due to the
aforementioned difficulties GPs can have with larger problem dimensions.
Intriguingly, as shown in the supplement, the comparative performance for
MLP-based \alg increases with $d$ for all four scalarisations.

\subsection{Computational Timing}
\label{sec:results:additional}
\begin{figure}[t] 
\includegraphics[width=\linewidth]{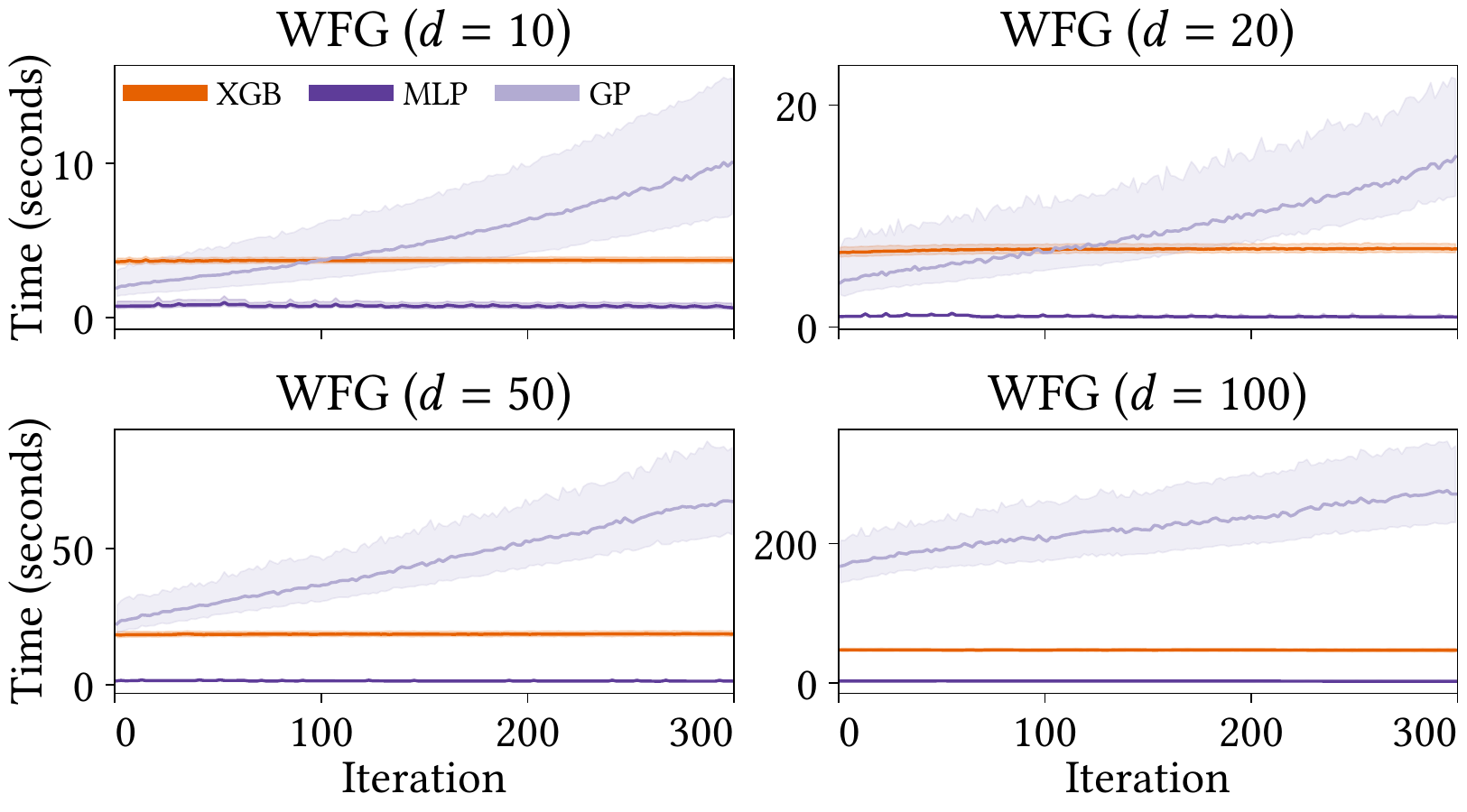}\\
\caption{%
    Computational time taken per iteration on the WFG benchmark 
    for~$d \in \{10, 20, 50 ,100\}$. The median computation time over all
    scalarisations is shown as the solid lines, with the corresponding
    interquartile ranges shaded.
}
\label{fig:timing}
\end{figure}
We also investigate the computational performance of the three methods, \alg
with XGB and MLP, and BO with a GP surrogate model. The cost of performing one
iteration of each method was recorded for all scalarisers on the WFG benchmark
with $d \in \{10, 20, 50, 100\}$. To ensure a fair comparison, each optimisation
run was carried out on one core of an Intel Xeon E5-2640 v4 CPU.
Figure~\ref{fig:timing} shows the timing results. The median computation time
across all scalarisations for each iteration is shown, with shaded regions
corresponding to the interquartile range. As theory necessarily
dictates~\citep{rasmussen:gpml:2006}, the GP's computation time increases with
the number of solutions. Conversely, the computation time of XGB and MLP are
roughly constant for all iterations for a given dimensionality.

\section{Conclusions}
\label{sec:conc}
In this work, we presented \alg: a novel multi-objective algorithm for
expensive optimisation problems by scalarisation. It replaces the traditional
mono-surrogate BO pipeline by training a probabilistic classifier on
previously-evaluated solutions that were thresholded into two classes based on
their objective value's scalarised form. The classifier's predictions can then
be shown to approximate the probability of improving over the given threshold.
In addition to \alg, we also introduce PHC, a dominance-preserving
scalarisation method that uses a modified form of each solution's hypervolume
contribution to its Pareto shell.

As demonstrated throughout, \alg provides a strong alternative to
mono-surrogate BO using GPs. This is particularly true for more difficult
problems, such as the WFG and real-world benchmarks, as well as for problems
with higher dimensionalities. Additionally, the computational costs of \alg
remain approximately constant as the number of solutions included in the model
increases. We note that, we are not able to recommend \alg over BO (or vice
versa) for an arbitrary scalarisation, as shown by BO being consistently better
with the AT scalarisation and \alg with PHC. However, PHC consistently
outperformed all other scalarisation methods. Therefore, we recommend the use
of \alg with XGB using the PHC scalarisation method as the new \emph{de facto}
choice for mono-surrogate-based multi-objective optimisation.

Future work includes extending \alg to non-continuous spaces by using random
forests. These are able to naturally adapt to discrete and categorical data
without the need for, \eg one-hot encoding. Additionally, we also seek to
extend \alg to the multi-surrogate setting by replacing its multiple GP models
with one or more probabilistic classifiers, thereby substantially reducing
computational complexity.

\begin{acks}
The authors would like to thank Roman Garnett for valuable discussions
regarding the original BORE derivation. The authors would also like to
acknowledge the use of the University of Exeter High-Performance Computing
(HPC) facility in carrying out this work. Dr. Rahat was supported by the
\grantsponsor{}{Engineering and Physical Research Council}{}
[grant number \grantnum{}{EP/W01226X/1}].
\end{acks}

\bibliographystyle{ACM-Reference-Format}
\bibliography{ref}

\end{document}


\title[%
    \alg: Multi-objective BO by Density-Ratio Estimation%
]{%
    \alg: Multi-objective Bayesian Optimisation by Density-Ratio Estimation
          (Supplementary Material)
}

\author{George {De Ath}}
\email{g.de.ath@exeter.ac.uk}
\orcid{0000-0003-4909-0257}
\affiliation{%
  \department{Department of Computer Science}
  \institution{University of Exeter}
  \city{Exeter}
  \country{United Kingdom}
}

\author{Tinkle Chugh}
\email{t.chugh@exeter.ac.uk}
\orcid{0000-0001-5123-8148}
\affiliation{%
  \department{Department of Computer Science}
  \institution{University of Exeter}
  \city{Exeter}
  \country{United Kingdom}
}

\author{Alma A. M. Rahat}
\email{a.a.m.rahat@swansea.ac.uk}
\orcid{0000-0002-5023-1371}
\affiliation{%
  \department{Department of Computer Science}
  \institution{Swansea University}
  \city{Swansea}
  \country{United Kingdom}
}

%

\ccsdesc[500]{Computing methodologies~Modeling methodologies}
\ccsdesc[500]{Theory of computation~Gaussian processes}
\ccsdesc[500]{Applied computing~Multi-criterion optimization and decision-making}

\keywords{%
    Bayesian optimisation,
    Surrogate modelling,
    Scalarisation methods,
    Efficient multi-objective optimisation,
    Acquisition function
}


\maketitle
\appendix

In this supplementary materials document we provide the results for all the
experiments carried out in this work. In Section~\ref{sec:exp_setup} we provide
additional details on the benchmarks and methods used. 
Section~\ref{sec:summary} summarises the performance of each method based on
model (XGB, MLP or GP) with respect to a given benchmark and scalariser, as
well as for problem dimensionality and number of objectives.
Section~\ref{sec:comp} shows the computational timing per problem
dimensionality for the WFG and DTLZ benchmarks.
Finally, in Section~\ref{sec:conv} we provide convergence plots in terms of
hypervolume and IGD+ for all test problems.

\section{Experimental set up}
\label{sec:exp_setup}

In the following sections we give the reference and ideal points for all three
benchmarks used, as well as the number of weight vectors used for the augmented
Tchebycheff scalarisation method.

\subsection{Benchmark details}
Reference points and ideal points were used normalise objective values before
the hypervolume and IGD+ indicators were calculated. This ensures that
objectives are weighted equally in both measures. The $i$-th element of a given
vector of solutions $\bff = \{f^1, \dots, f^M\}$ is normalised by calculating
$\hat{f}^i = (f^i - q^i) / (r^i - q^i)$, where $\bq = \{q^1, \dots, q^M \}$ and
$\brr = \{r^1, \dots, r^M \}$ are the ideal and reference points respectively
for a given problem. The $i$-th element of $\bq$ and $\brr$ was found by
optimising the $i$-th objective of the corresponding test problem with
CMA-ES~\citep{hansen:bipopcmaes:2009}, either minimising it for $\bq$ or
maximising for $\brr$. For completeness, we include the ideal and reference
points used for all test problems evaluated in this work. Note that,
for conciseness, in the following tables we use a python-like vector-building
notation to denote vectors, \eg $[10]*M$ is a vector of length $M$ containing
with each element containing the value $10$, and $[10] * (M-1) + [20]$ is
another $M$-length vector with its last value being $20$ and its $M-1$
other values being $10$.

\begin{table}[H]
\begin{tabular}[h]{l r r r}
    \toprule
    Problem ID(s) & $d$ & Reference Point & Ideal Point\\
    \midrule
    1 &  2 & $[120]  * M$ & $[0] * M$\\
      &  5 & $[450]  * M$ & $[0] * M$\\
      & 10 & $[1000] * M$ & $[0] * M$\\
    \midrule
    2, 4, 5 &  2 & $[2]  * M$ & $[0] * M$\\
            &  5 & $[2]  * M$ & $[0] * M$\\
            & 10 & $[4] * M$ & $[0] * M$\\
    \midrule
    3   &  2 & $[250]  * M$ & $[0] * M$\\
        &  5 & $[1000] * M$ & $[0] * M$\\
        & 10 & $[2000] * M$ & $[0] * M$\\
    \midrule
    6   &  2 & $[2.5] * M$ & $[0] * M$\\
        &  5 & $[5]   * M$ & $[0] * M$\\
        & 10 & $[10]  * M$ & $[0] * M$\\
    \midrule
    7   &  2 & $[1.5] * (M-1) + [23]$ & See below \\
        &  5 & $[1.5] * (M-1) + [60]$ & See below \\
        & 10 & $[1.5] * (M-1) + [110]$ & See below \\
    \bottomrule
\end{tabular}
\caption{%
    DTLZ~\citep{deb:dtlz:2005} reference and ideal points for a given
    dimensionality $d$ and number of objectives $M$.
}
\end{table}

\begin{table}[H]
\begin{tabular}[h]{l r}
    \toprule
    $M$ & Ideal point \\
    \midrule
    2  & $[0, 2.307]$ \\
    3  & $[0, 0, 2.614]$ \\
    5  & $[0, 0, 0, 0, 3.228]$ \\
    10 & $[0, 0, 0, 0, 0, 0, 0, 0, 0, 4.763]$ \\
    \bottomrule
\end{tabular}
\caption{%
    Ideal points for the DTLZ~\citep{deb:dtlz:2005} test problem number $7$ for
    a given number of objectives and $d \leq 10$.
}
\end{table}

\begin{table}[H]
\begin{tabular}[h]{l r r}
    \toprule
    $M$ & Reference Point & Ideal Point \\
    \midrule
    $2$ & $[3, 5]$ & $[0, 0]$ \\
    $3$ & $[3, 5, 7]$ & $[0, 0, 0]$ \\
    $5$ & $[3, 5, 7, 9, 11]$ & $[0, 0, 0, 0, 0]$ \\
    $10$ & $[3, 5, 7, 9, 11, 13, 15, 17, 19, 21]$ & 
           $[0, 0, 0, 0, 0, 0, 0, 0, 0, 0]$ \\
    \bottomrule
\end{tabular}
\caption{%
    WFG~\citep{huband:wfg:2006} reference and ideal points for a given number
    of objectives $M$ and $d \in \{6, 8, 10, 20,50, 100\}$.
}
\end{table}

\begin{table}[H]
\begin{adjustbox}{width=1\textwidth}
\begin{tabular}[h]{l r r}
    \toprule
    Problem $(d, M)$ & Reference Point & Ideal Point \\
    \midrule
    RE2-4-1 ($4, 2$) & $[2995, 0.051]$ 
                         & $[1237, 0.002]$ \\
    RE2-2-4 ($2, 2$) & $[6005, 45]$ & $[60.5, 0]$ \\
    RE3-3-1 ($3, 3$) & $[817, 8250000, 19360000]$ 
                         & $[0, 0.3, 0]$ \\
    RE3-4-2 ($4, 3$) & $[334, 17600, 425100000]$ 
                         & $[0.01, 0.0004, 0]$ \\
    RE3-5-4 ($5, 3$) & $[1705, 11.8, 0.27]$ 
                         & $[-0.73, 1.13, 0]$ \\
    RE3-4-7 ($4, 3$) & $[1.01, 1.25, 1.1]$ 
                         & $[0, 0, -0.44]$ \\
    RE4-7-1 ($7, 4$) & $[43, 4.5, 13.1, 14.2]$ 
                         & $[15.5, 3.5, 10.6, 0]$ \\
    RE4-6-2 ($6, 4$) & $[0, 20100, 31100, 15.4]$ 
                         & $[-2757, 3962, 1947, 0]$ \\
    RE6-3-1 ($3, 6$) & $[83100, 1351, 2854000, 16028000, 358000, 99800]$ 
                         & $[63840, 30, 285346, 183749, 7.2, 0]$\\
    RE9-7-1 ($7, 9$) & $[42.7, 1.6, 350, 1.1, 1.57, 1.75, 1.25, 1.3, 1.06]$
                         & $[15.5, 0, 0, 0.09, 0.36, 0.52, 0.73, 0.61, 0.66]$\\
    \bottomrule
\end{tabular}
\end{adjustbox}
\caption{%
    Real-world benchmark~\citep{tanabe:rwproblems:2020} reference and ideal
    points.
}
\end{table}

\subsection{Augmented Tchebycheff Scalarisation}
We used $\rho = 0.05$ in the scalariser's calculations. The weight vectors for
the scalarisation were calculated via the Riesz' s-Energy
method~\citep{blank:rieszs:2021}. The following table details the number of
weight vectors used for a given number of objectives~$M$.
\begin{table}[H]
\begin{tabular}[h]{l r r r r r r r r r r}
    \toprule
    Number of objectives     &   2 &   3 &   4 &   5 &   6 &   7 &   8 &  9 &  10 \\
    \midrule
    Number of weight vectors & 100 & 105 & 120 & 126 & 132 & 112 & 156 & 90 & 275 \\
    \bottomrule
\end{tabular}
\caption{%
    Number of weight vectors used for a given number of objectives.
}
\end{table}

\subsection{Gaussian Process Priors}
\label{sec:gppriors}
A zero-mean GP was used for BO, alongside an ARD \Matern kernel with
$\nu = 5/2$:
\begin{align}
    \label{eq:gp:matern52}
        \kappa_{Matern}(\bx, \xnext \given \btheta) =
        & \gpoutputscale^2
        \frac{2^{1-\nu}}{\Gamma(\nu)}
        \left(
          \sqrt{2\nu }r
        \right)^\nu
        K_{\nu}\left(
          \sqrt{2\nu } r
        \right),
\end{align}
where $K_{\nu}$ is a modified Bessel function and 
$r^2 = \sum_{i=1}^d (x_i - x_i')^2 / \omega_i^2$ is the squared distance
between $\bx$ and $\bx'$ scaled by the length-scales $\omega_i$. We place
uniform priors on the kernel's hyperparameters
$\btheta = \{\omega_1, \dots, \omega_d, \sigma_o \}$. Given that the
locations with which the GPs are trained on are rescaled to the unit hypercube
and that its target function values are standardised, we define them to be
$\omega \sim \Uniform(10^{-4}, \sqrt{d})$ and
$\sigma_o \sim \Uniform(10^{-4}, 10)$ respectively. This allows for
interactions to occur on a sensible length-scale, \ie the largest distance in
$[0, 1]^d$ is $\sqrt{d}$, and a plausible output-scale, \ie not too large given
the standardised target function.

\section{Overall summary of performance for all problem sets}
\label{sec:summary}
\begin{table}[H]
\begin{tabular}[h]{l c | c c c | c c c }
\toprule
    &            & \multicolumn{3}{c|}{Hypervolume} & \multicolumn{3}{c}{IGD+} \\
\midrule
    & Benchmark & XGB & MLP & GP                  & XGB & MLP & GP \\
\midrule
    \multirow{4}{*}{\rotatebox[origin=c]{90}{HYPI}}
    & DTLZ     & \best{36} &  4 &       29  & \best{40} & 1 &       28  \\
    & WFG      & \best{37} &  0 &       36  & \best{36} & 1 &       34  \\
    & RW       &        4  &  0 &  \best{6} &        5  & 0 &  \best{6} \\
    & WFG (HD) &       10  &  7 & \best{18} &       10  & 9 & \best{18} \\
\midrule
    \multirow{4}{*}{\rotatebox[origin=c]{90}{DomRank}}
    & DTLZ       & 21 & 11 & \best{37}   &       30  &        5  & \best{36} \\
    & WFG        & 36 &  0 & \best{45}   & \best{40} &        1  &       35  \\
    & RW         &  1 &  1 &  \best{8}   &        2  &        1  &  \best{8} \\
    & WFG (HD)   &  5 & 15 & \best{17}   &        7  & \best{16} &       10  \\
\midrule
    \multirow{4}{*}{\rotatebox[origin=c]{90}{PHC}} 
    & DTLZ     & \best{36} & 12 & 22     & \best{41} &  8 & 22 \\
    & WFG      & \best{39} &  0 & 36     & \best{38} &  0 & 35 \\
    & RW       &  \best{7} &  0 &  5     &  \best{9} &  0 &  3 \\
    & WFG (HD) & \best{18} &  6 &  7     & \best{17} &  5 &  6 \\
\midrule
    \multirow{4}{*}{\rotatebox[origin=c]{90}{AT}}
    & DTLZ     & 13 &  8 & \best{43}     & 27 &  2 & \best{42} \\
    & WFG      &  1 &  0 & \best{63}     &  4 &  2 & \best{62} \\
    & RW       &  2 &  0 &  \best{8}     &  2 &  0 &  \best{8} \\
    & WFG (HD) &  0 &  4 & \best{23}     &  0 &  8 & \best{23} \\
\bottomrule
\end{tabular}
\caption{%
    Performance summary of \alg (MLP and XGB) and BO (GP) for a given
    scalarisation method (HYPI, DomRank, PHC and AT), on each benchmark.
    Table values correspond to the number of times each model was the best or
    statistically equivalent to the best method for a given benchmark and
    scalarisation method. Note how the mono-surrogate-based BO (GP) performs
    the best across all problems for AT scalariser, whereas \alg with XGB does
    for PHC.
}
\label{tbl:summary_table}
\end{table}

\begin{figure}[H] 
\includegraphics[width=0.5\linewidth]{figs/hv_DTLZ_summary_barchart}%
\includegraphics[width=0.5\linewidth]{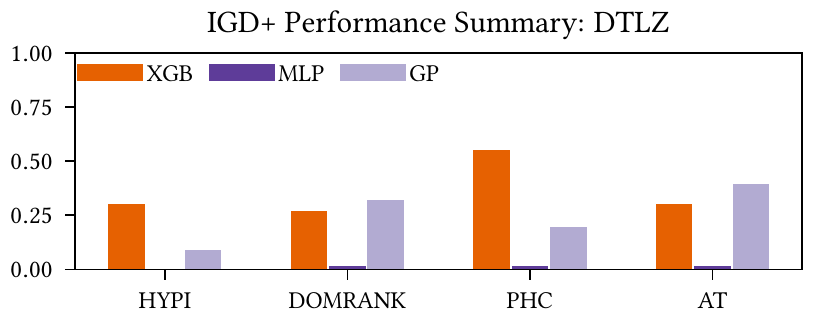}\\
\includegraphics[width=0.5\linewidth]{figs/hv_WFG_summary_barchart}%
\includegraphics[width=0.5\linewidth]{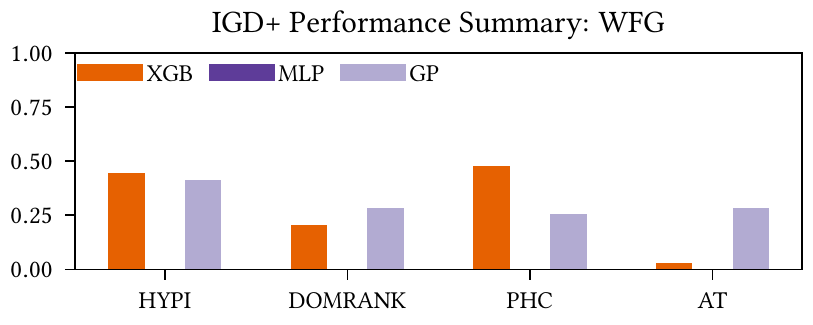}\\
\includegraphics[width=0.5\linewidth]{figs/hv_RW_summary_barchart}%
\includegraphics[width=0.5\linewidth]{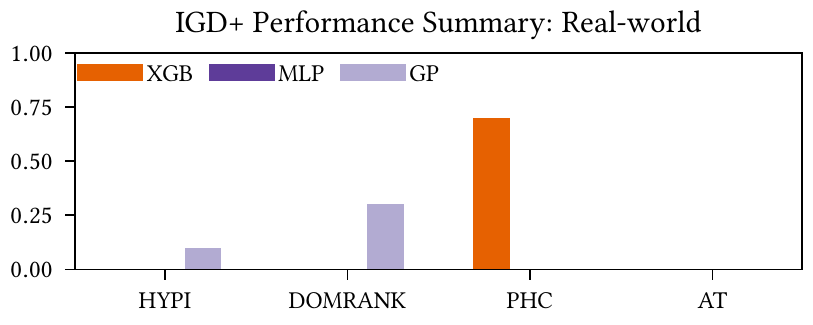}\\
\includegraphics[width=0.5\linewidth]{figs/hv_WFG_HD_summary_barchart}%
\includegraphics[width=0.5\linewidth]{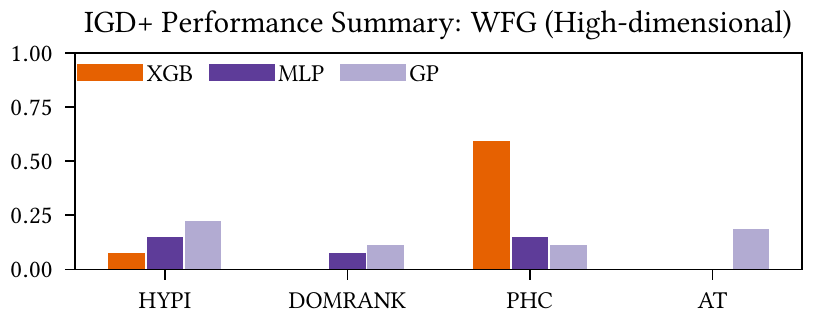}\\
\caption{%
    Performance summary for the sets of benchmark problems evaluated:
    (\emph{left column}) hypervolume (\emph{right column}) IGD+.
    Bar heights correspond to the  proportion of times that a method is best or
    statistically equivalent to the best method the benchmark's problems.
}
\label{fig:dtlz_summary}
\end{figure}

\subsection{Performance Summary based on dimensionality/objectives: DTLZ}
\begin{figure}[H] 
\includegraphics[width=0.5\linewidth]{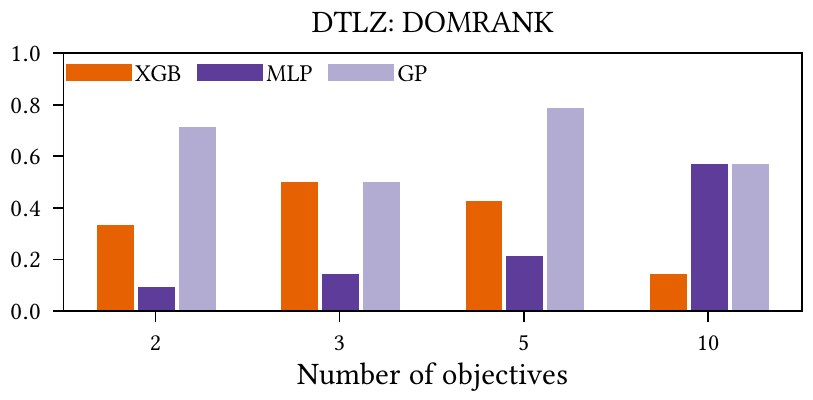}%
\includegraphics[width=0.5\linewidth]{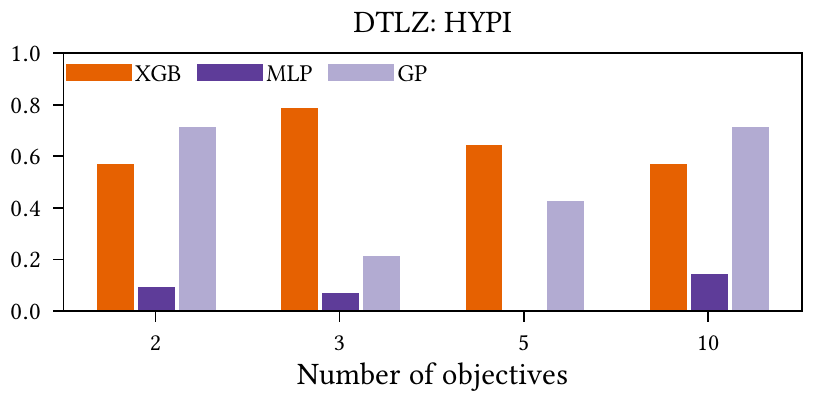}\\
\includegraphics[width=0.5\linewidth]{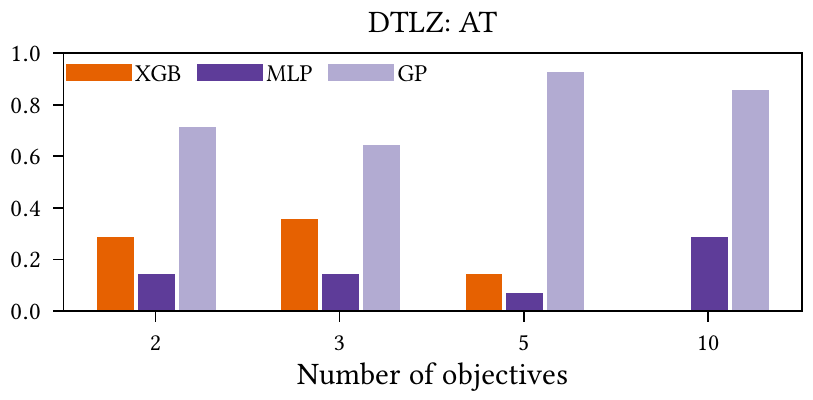}%
\includegraphics[width=0.5\linewidth]{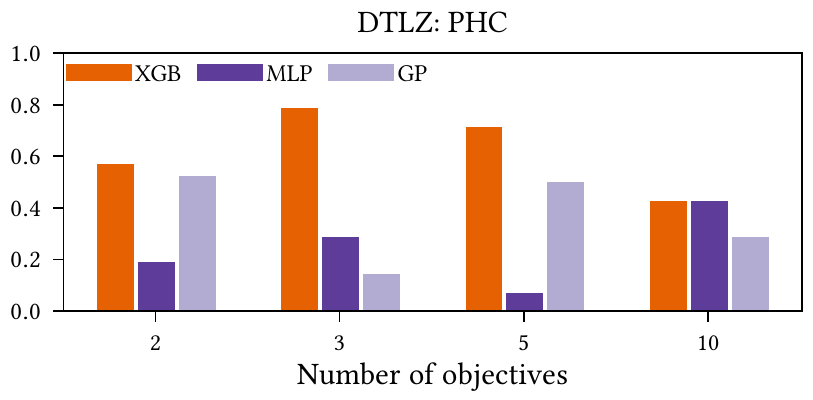}\\
\rule{\linewidth}{0.4pt}
\includegraphics[width=0.5\linewidth]{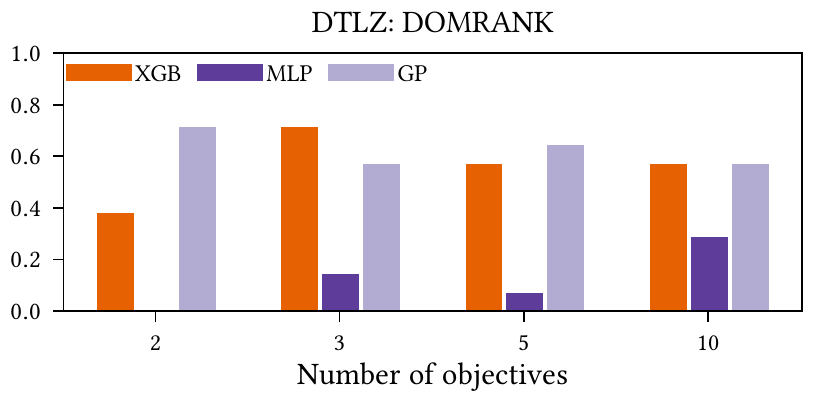}%
\includegraphics[width=0.5\linewidth]{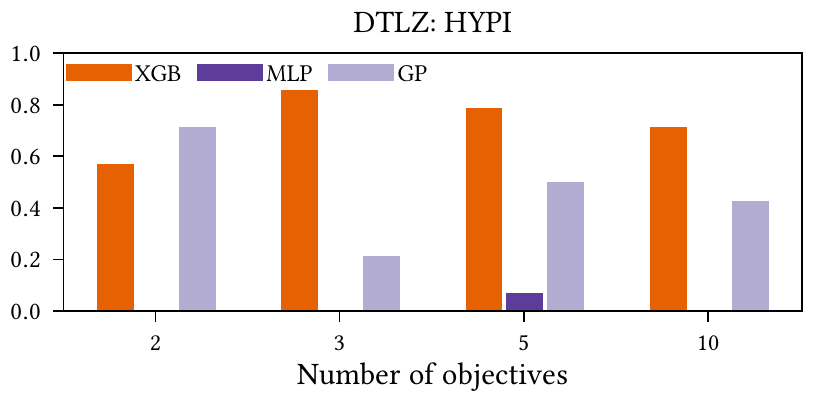}\\
\includegraphics[width=0.5\linewidth]{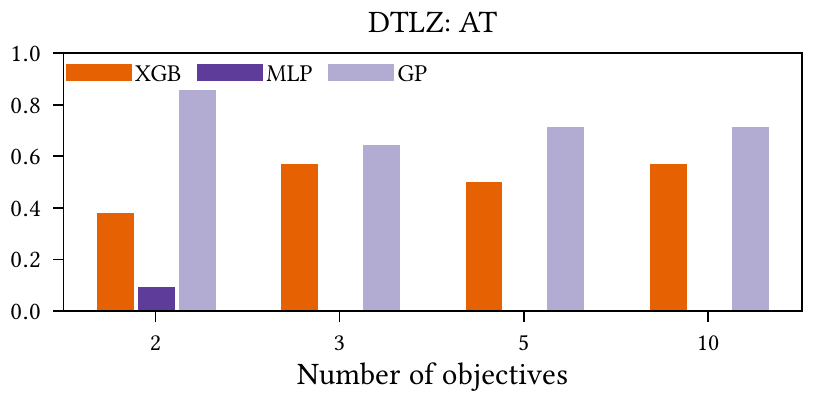}%
\includegraphics[width=0.5\linewidth]{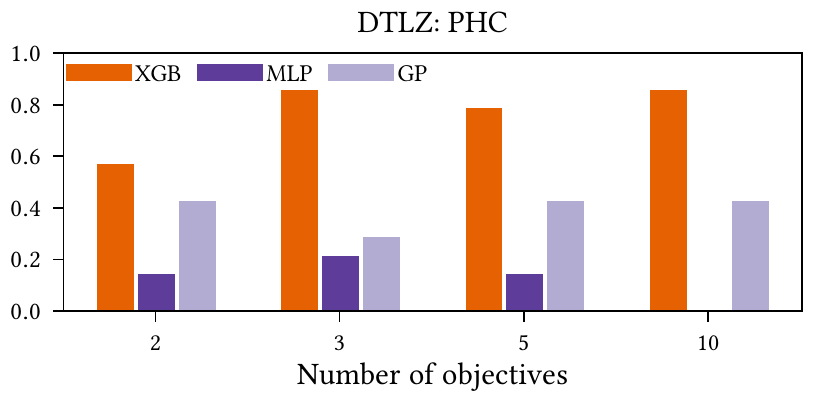}\\
\caption{%
    Performance summary for the DTLZ test problems based on the number of
    objectives a problem has: (\emph{upper}) hypervolume (\emph{lower}) IGD+.
    Bar heights correspond to the  proportion of times that a method is best or
    statistically equivalent to the best method the benchmark's problems.
}
\label{fig:dtlz_obj}
\end{figure}

\begin{figure}[H] 
\includegraphics[width=0.5\linewidth]{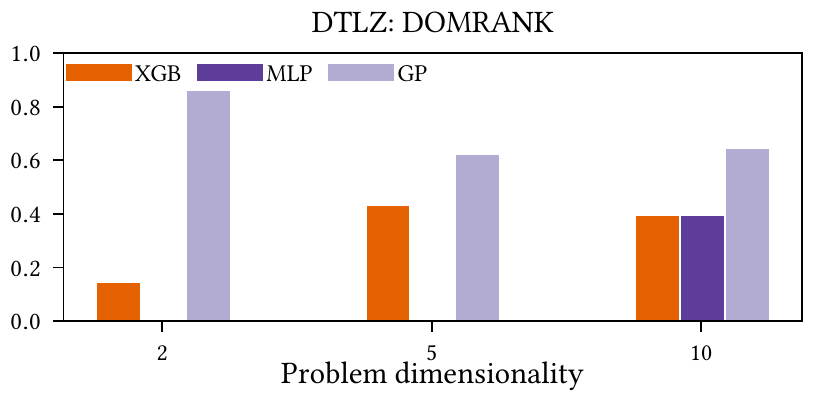}%
\includegraphics[width=0.5\linewidth]{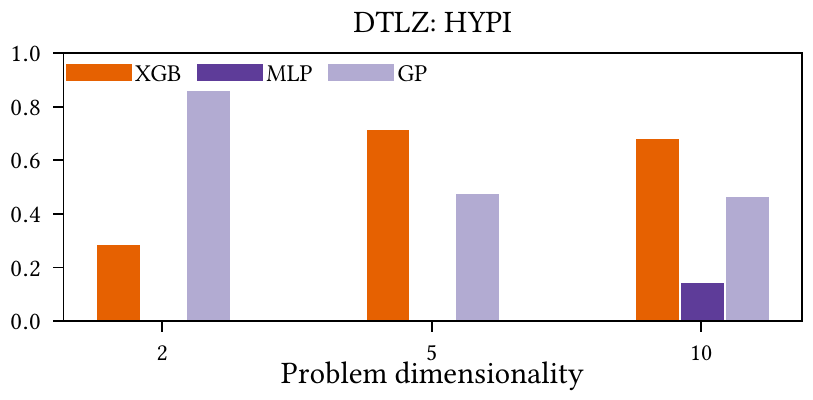}\\
\includegraphics[width=0.5\linewidth]{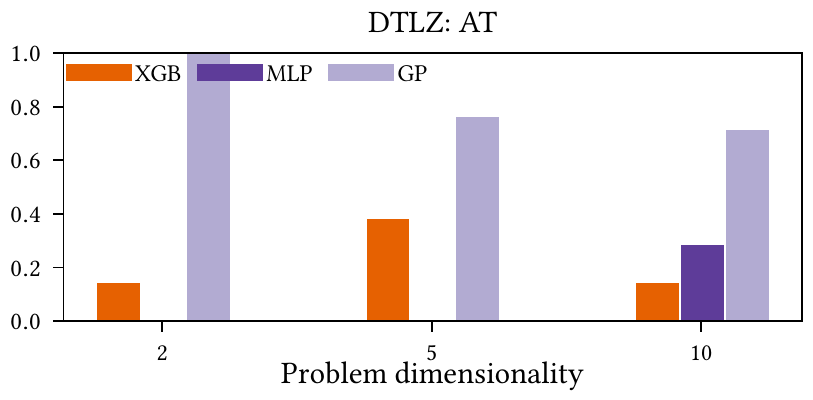}%
\includegraphics[width=0.5\linewidth]{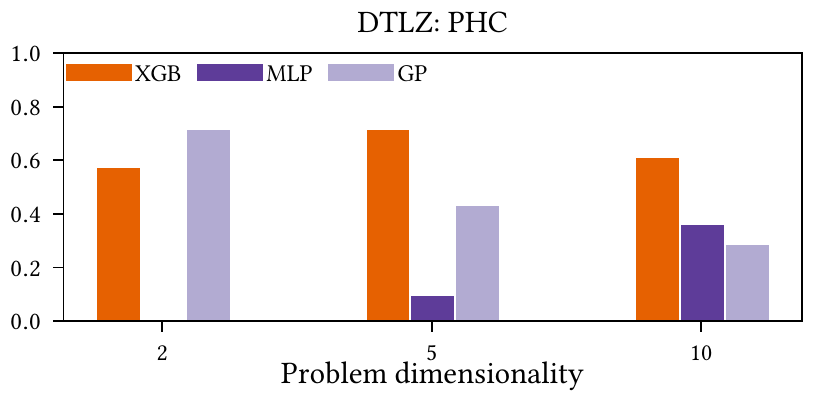}\\
\rule{\linewidth}{0.4pt}
\includegraphics[width=0.5\linewidth]{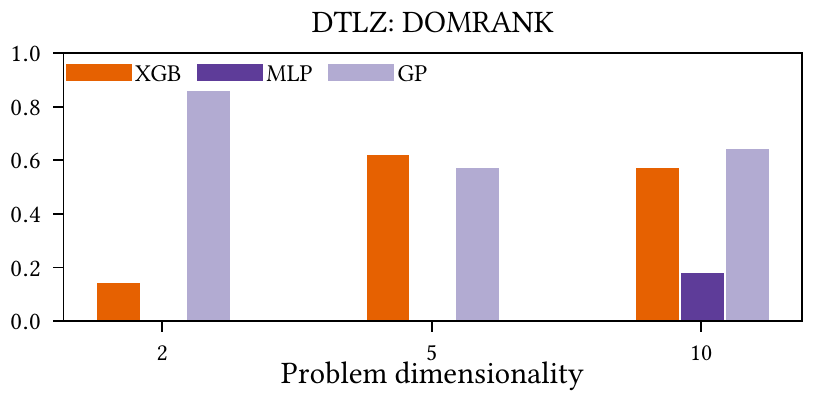}%
\includegraphics[width=0.5\linewidth]{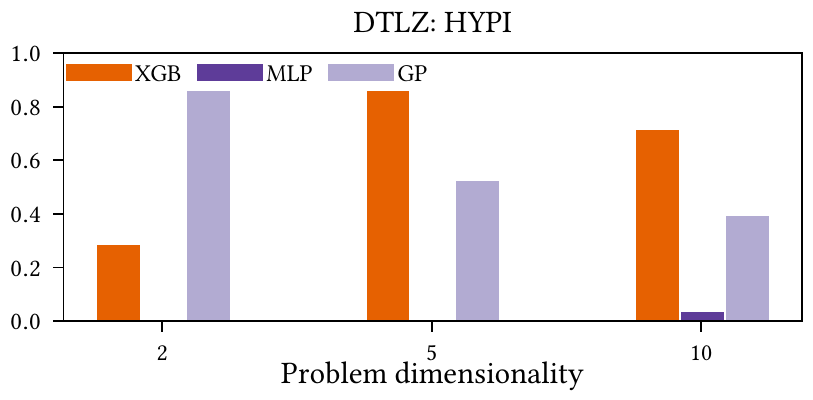}\\
\includegraphics[width=0.5\linewidth]{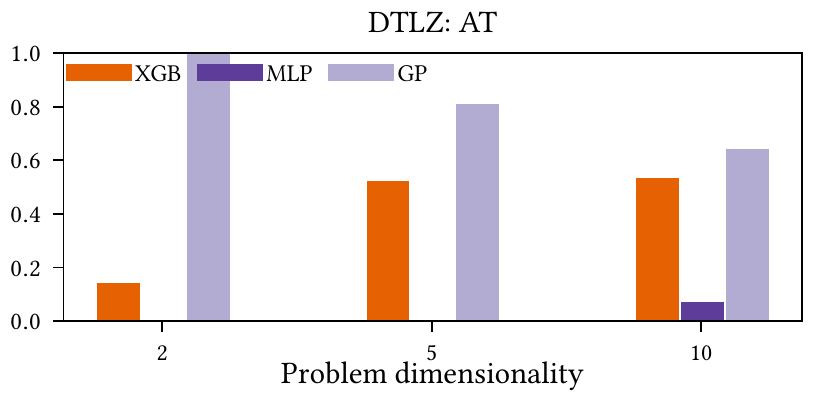}%
\includegraphics[width=0.5\linewidth]{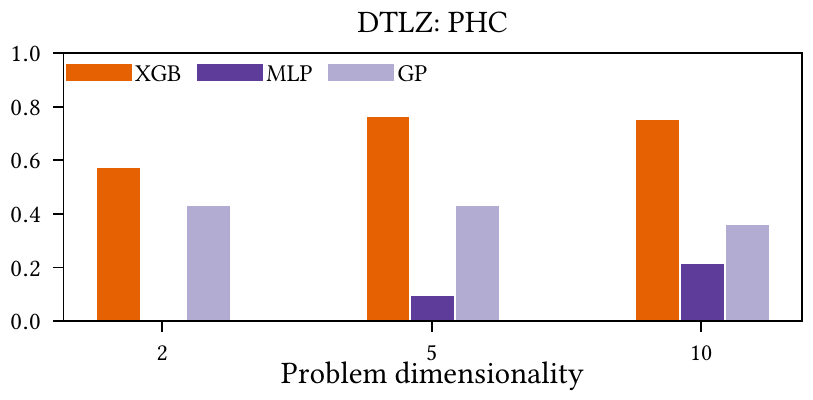}\\
\caption{%
    Performance summary for the DTLZ test problems based on the problem's
    dimensionality: (\emph{upper}) hypervolume (\emph{lower}) IGD+.
    Bar heights correspond to the  proportion of times that a method is best or
    statistically equivalent to the best method the benchmark's problems.
}
\label{fig:dtlz_dim}
\end{figure}

\subsection{Performance Summary based on dimensionality/objectives: WFG}
\begin{figure}[H] 
\includegraphics[width=0.5\linewidth]{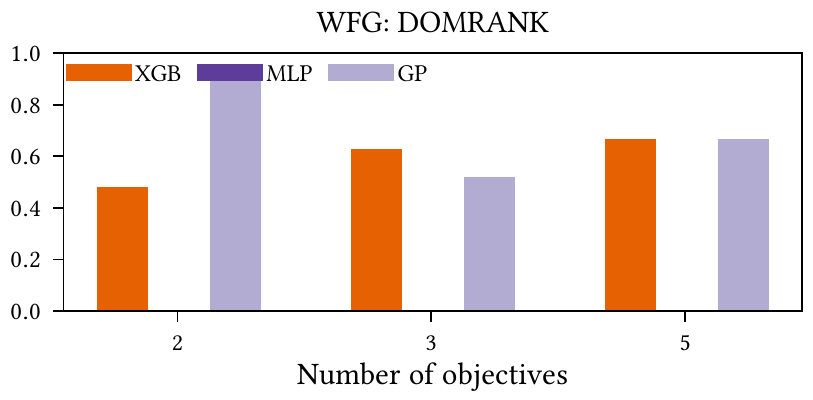}%
\includegraphics[width=0.5\linewidth]{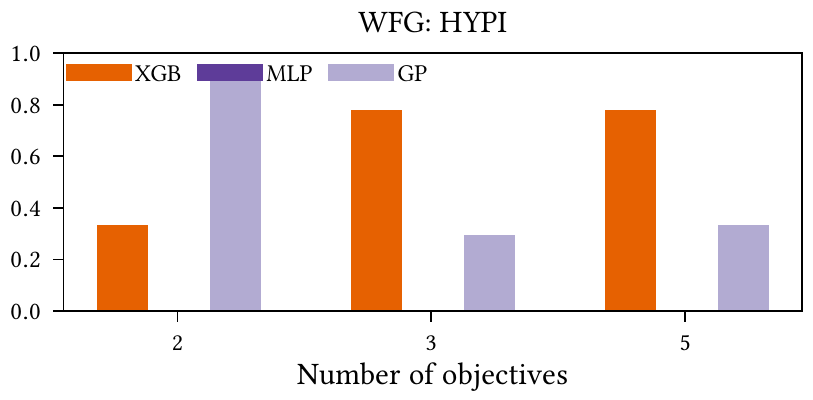}\\
\includegraphics[width=0.5\linewidth]{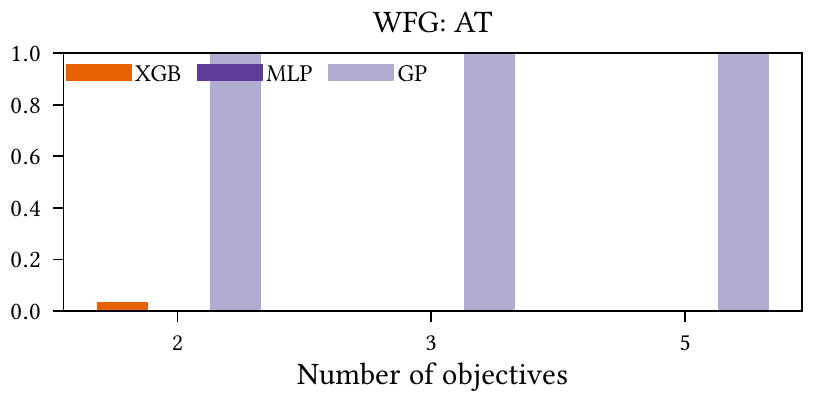}%
\includegraphics[width=0.5\linewidth]{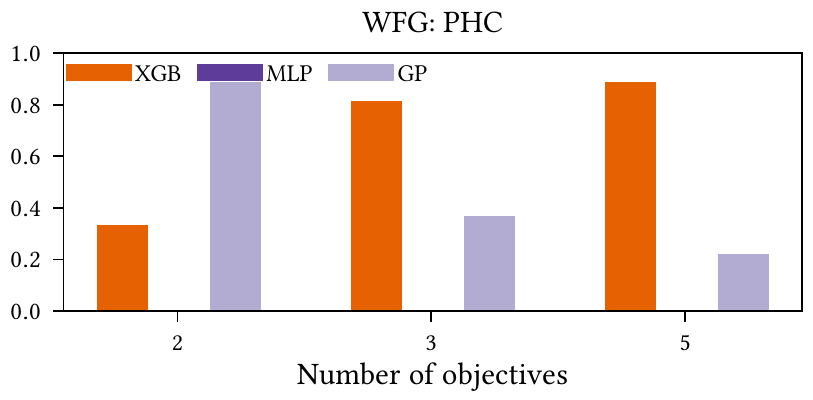}\\
\rule{\linewidth}{0.4pt}
\includegraphics[width=0.5\linewidth]{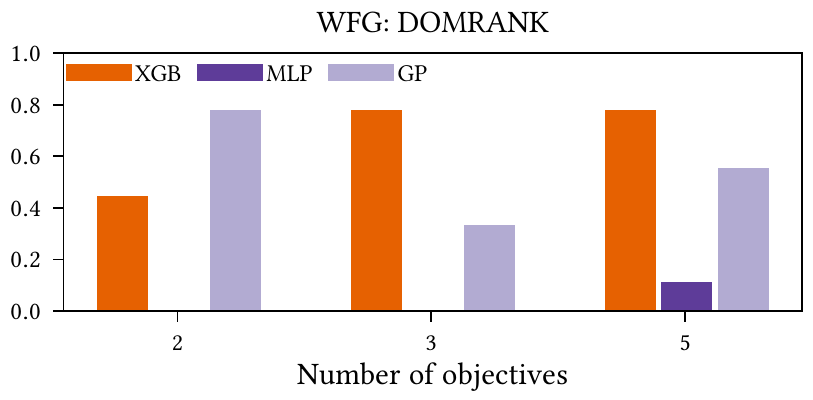}%
\includegraphics[width=0.5\linewidth]{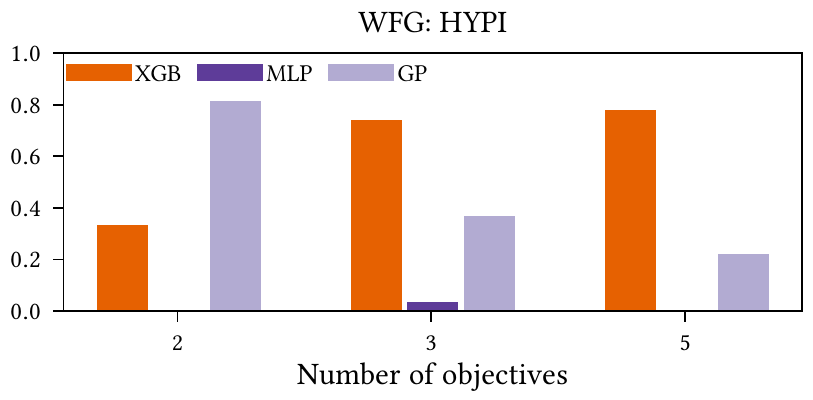}\\
\includegraphics[width=0.5\linewidth]{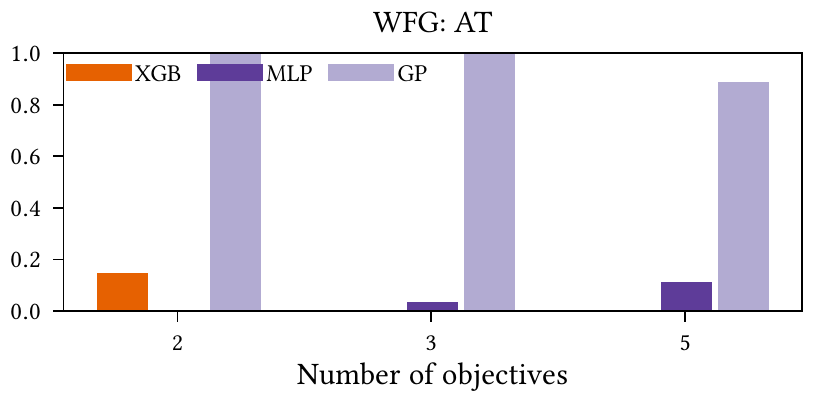}%
\includegraphics[width=0.5\linewidth]{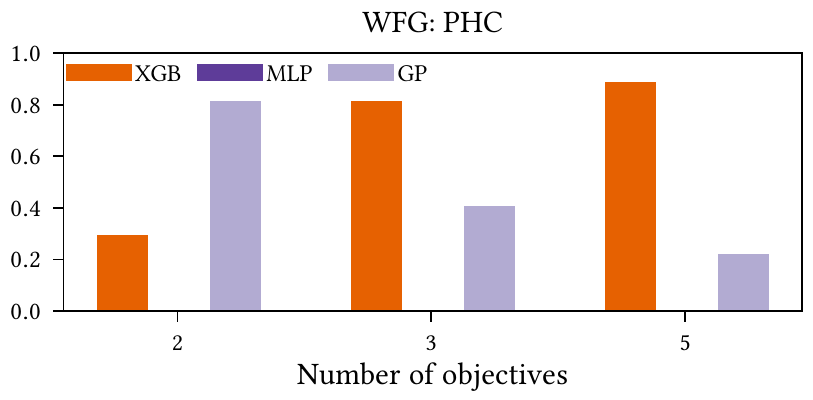}\\
\caption{%
    Performance summary for the WFG test problems based on the number of
    objectives a problem has: (\emph{upper}) hypervolume (\emph{lower}) IGD+.
    Bar heights correspond to the  proportion of times that a method is best or
    statistically equivalent to the best method the benchmark's problems.
}
\label{fig:wfg_obj}
\end{figure}

\begin{figure}[H] 
\includegraphics[width=0.5\linewidth]{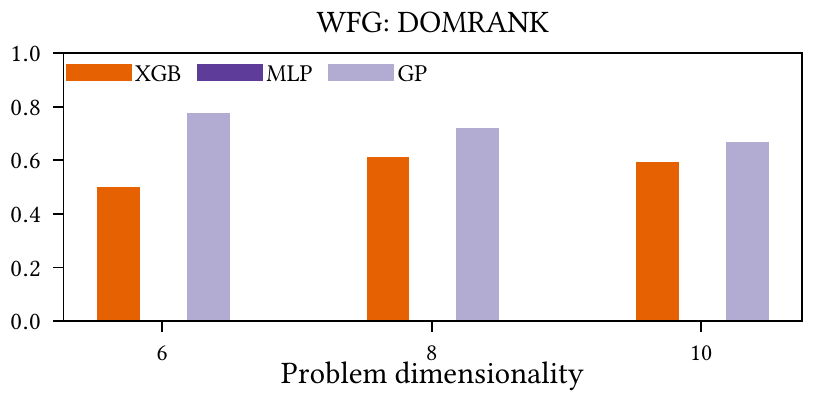}%
\includegraphics[width=0.5\linewidth]{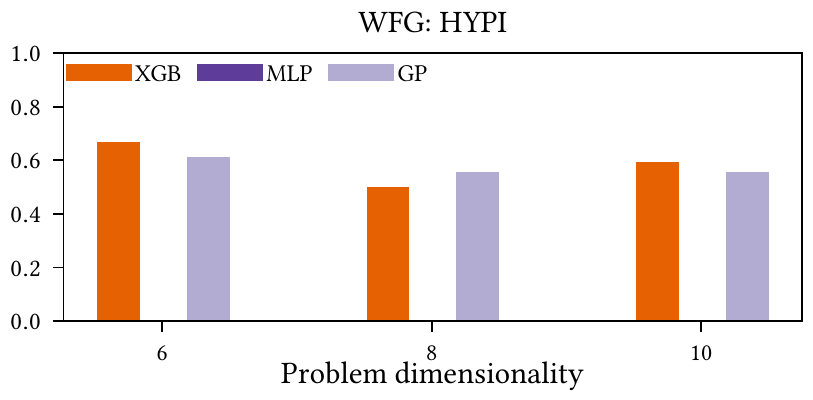}\\
\includegraphics[width=0.5\linewidth]{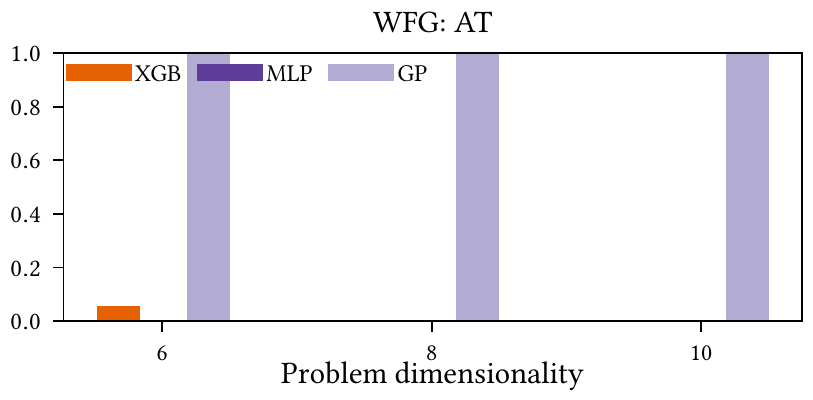}%
\includegraphics[width=0.5\linewidth]{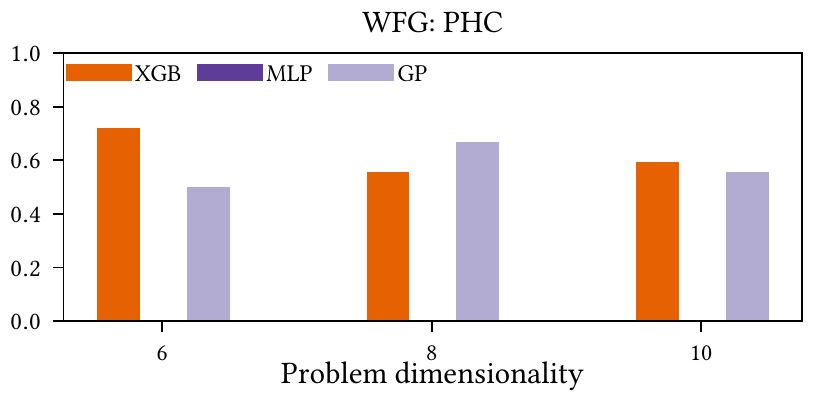}\\
\rule{\linewidth}{0.4pt}
\includegraphics[width=0.5\linewidth]{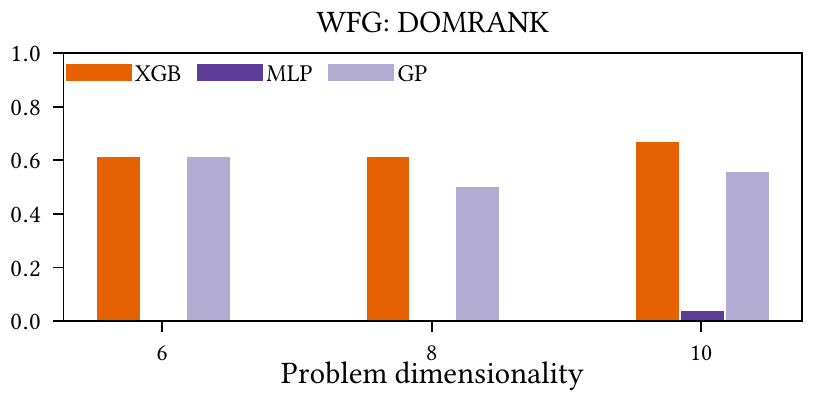}%
\includegraphics[width=0.5\linewidth]{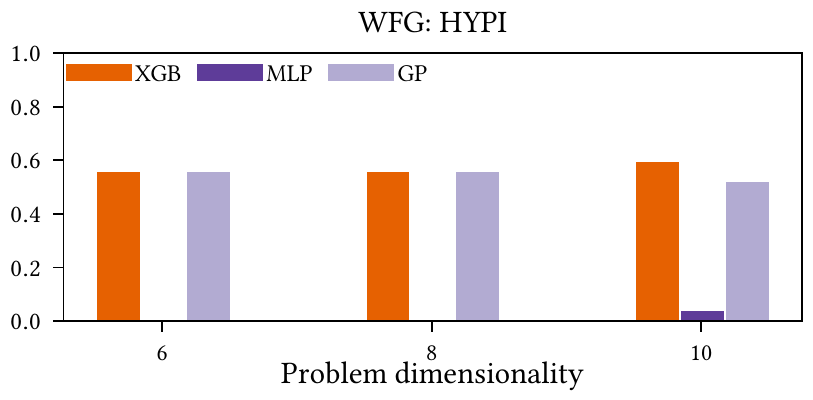}\\
\includegraphics[width=0.5\linewidth]{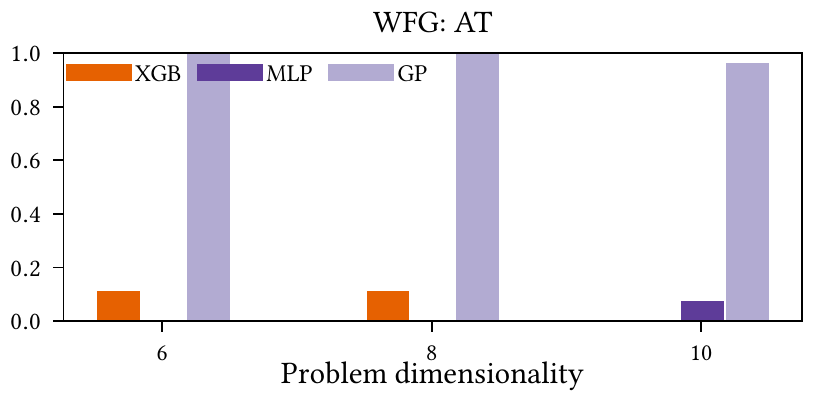}%
\includegraphics[width=0.5\linewidth]{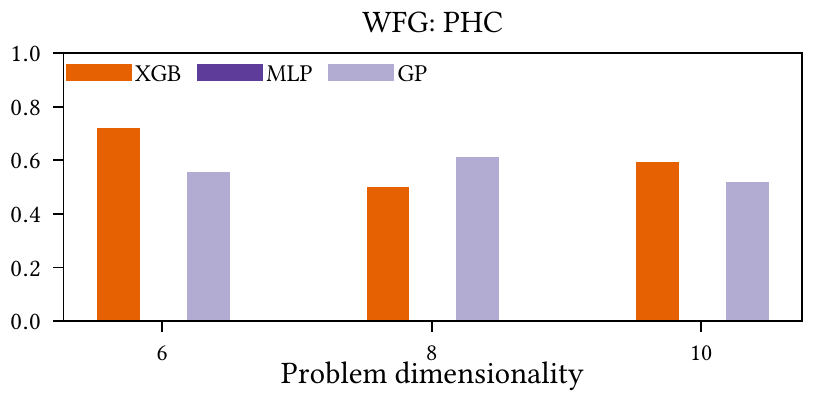}\\
\caption{%
    Performance summary for the WFG test problems based on the problem's
    dimensionality: (\emph{upper}) hypervolume (\emph{lower}) IGD+.
    Bar heights correspond to the  proportion of times that a method is best or
    statistically equivalent to the best method the benchmark's problems.
}
\label{fig:WFG_dim}
\end{figure}

\subsection{Performance Summary based on dimensionality/objectives: WFG (high-dimensional)}
\begin{figure}[H] 
\includegraphics[width=0.5\linewidth]{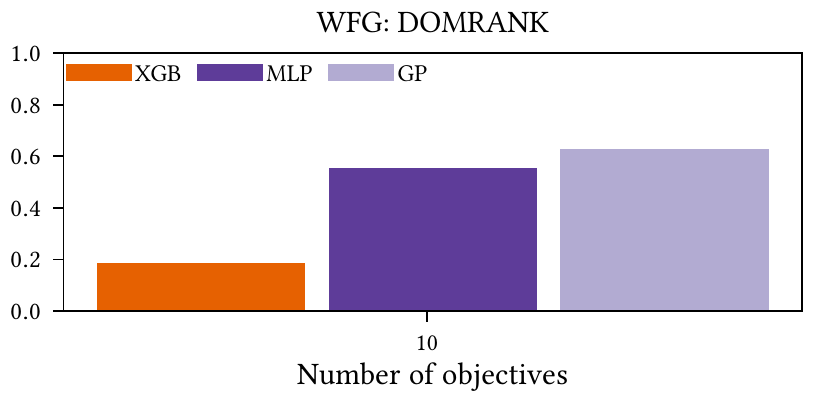}%
\includegraphics[width=0.5\linewidth]{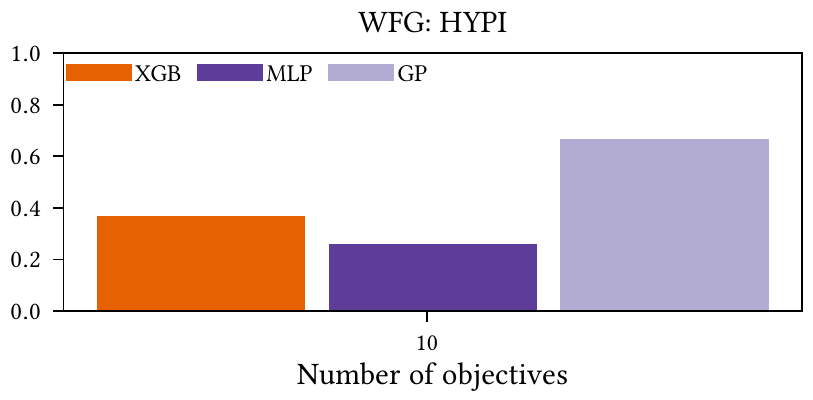}\\
\includegraphics[width=0.5\linewidth]{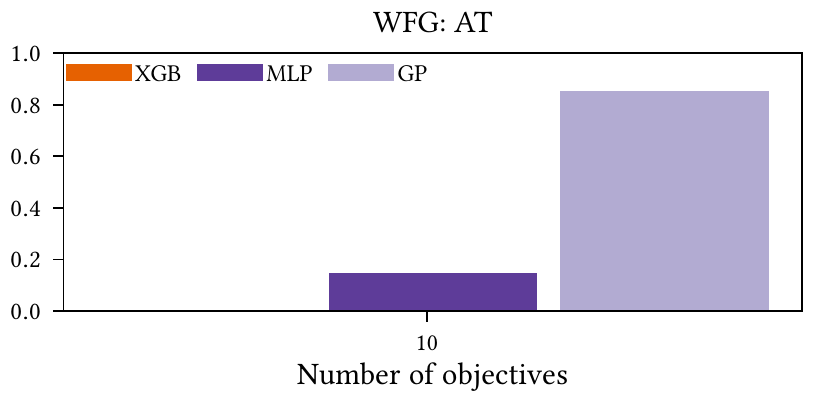}%
\includegraphics[width=0.5\linewidth]{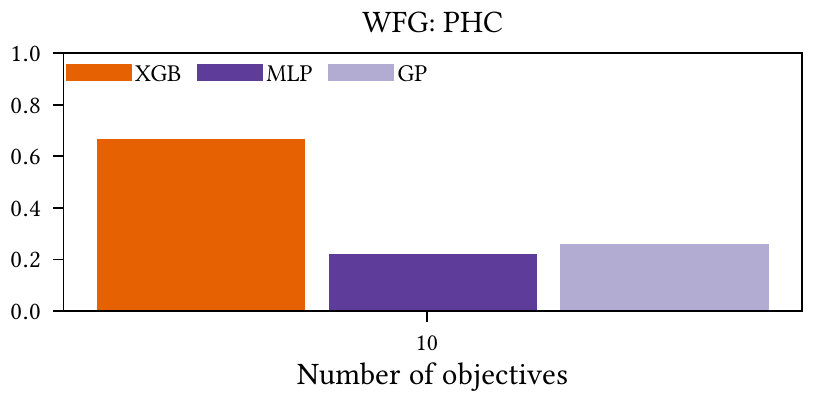}\\
\rule{\linewidth}{0.4pt}
\includegraphics[width=0.5\linewidth]{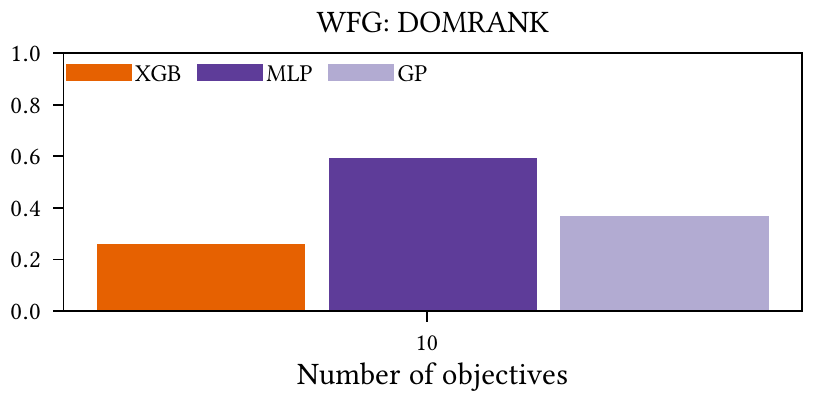}%
\includegraphics[width=0.5\linewidth]{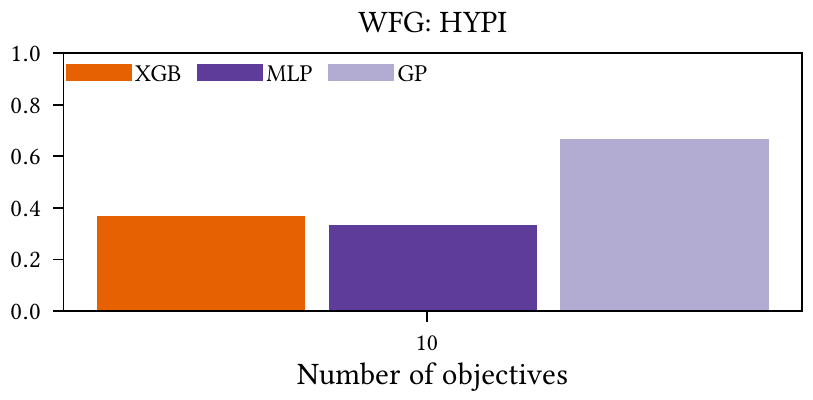}\\
\includegraphics[width=0.5\linewidth]{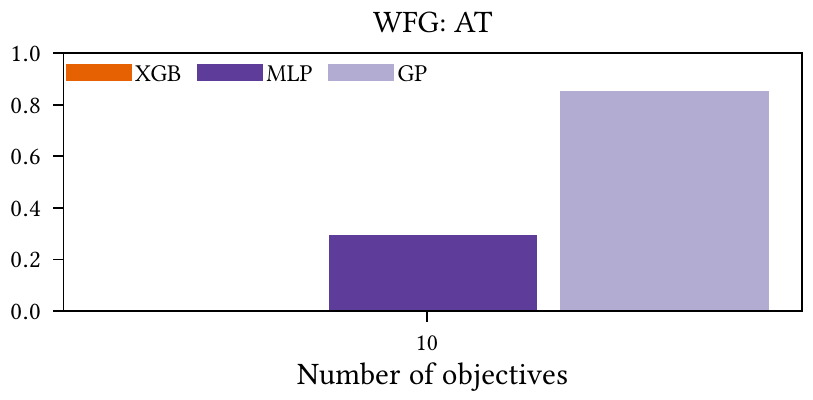}%
\includegraphics[width=0.5\linewidth]{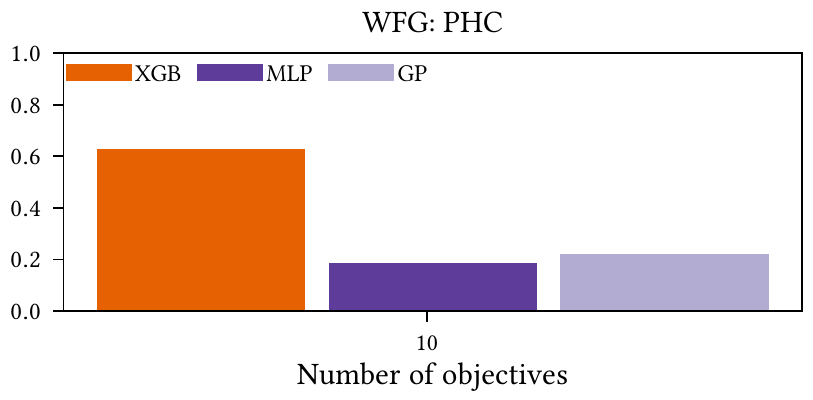}\\
\caption{%
    Performance summary for the high-dimensional WFG test problems based on the
    number of objectives a problem has: (\emph{upper}) hypervolume
    (\emph{lower}) IGD+. Bar heights correspond to the  proportion of times
    that a method is best or statistically equivalent to the best method the
    benchmark's problems.
}
\label{fig:WFG_HD_obj}
\end{figure}

\begin{figure}[H] 
\includegraphics[width=0.5\linewidth]{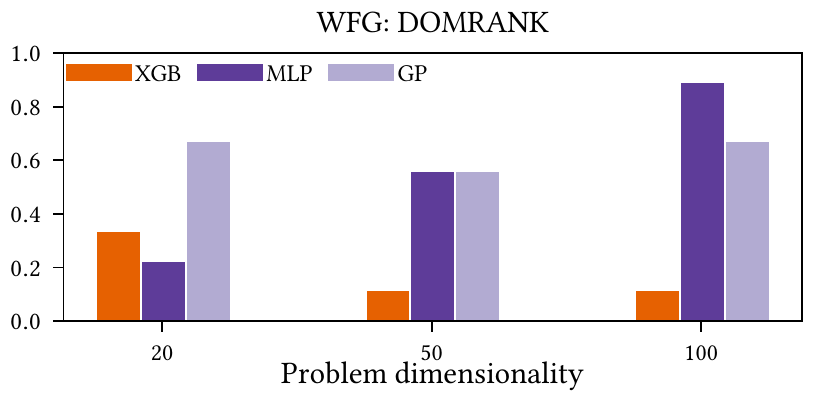}%
\includegraphics[width=0.5\linewidth]{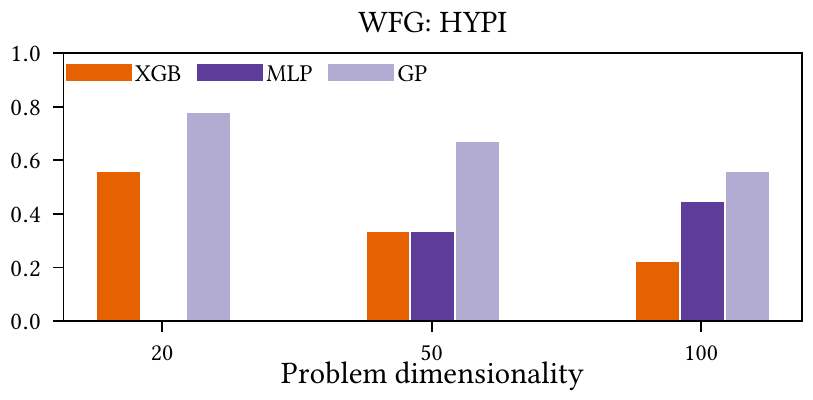}\\
\includegraphics[width=0.5\linewidth]{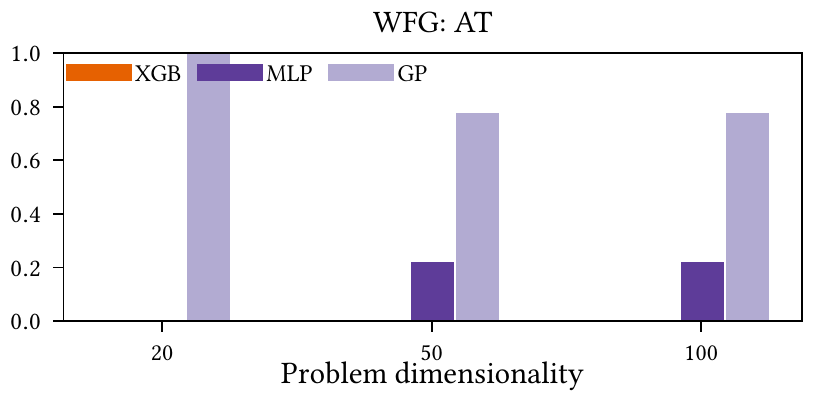}%
\includegraphics[width=0.5\linewidth]{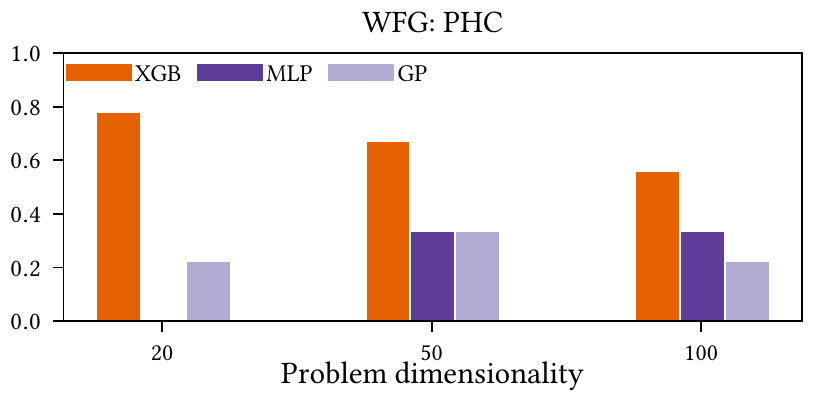}\\
\rule{\linewidth}{0.4pt}
\includegraphics[width=0.5\linewidth]{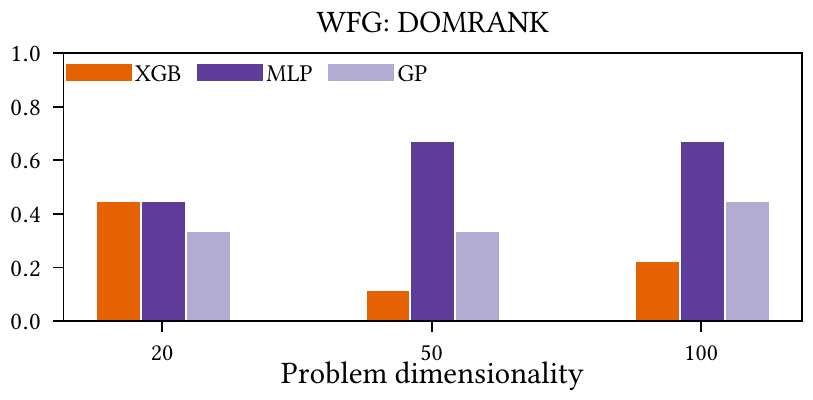}%
\includegraphics[width=0.5\linewidth]{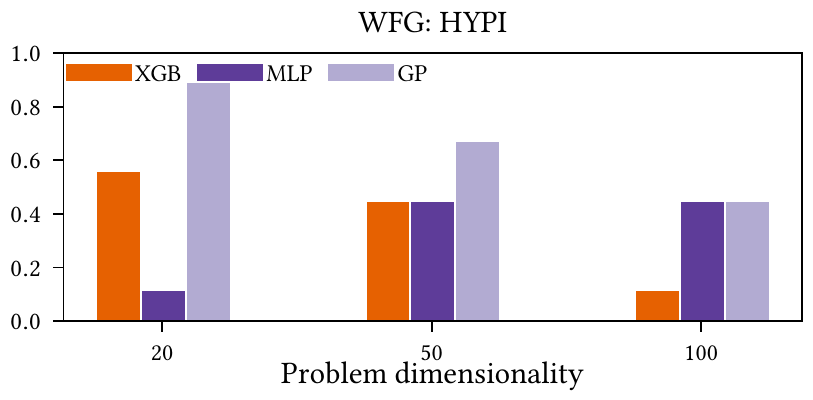}\\
\includegraphics[width=0.5\linewidth]{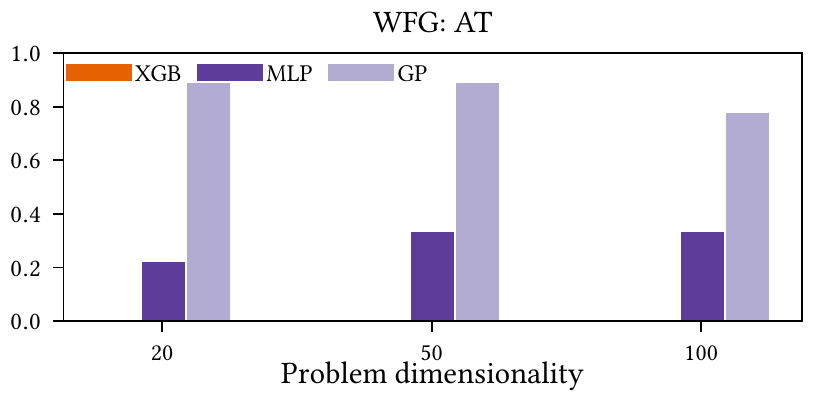}%
\includegraphics[width=0.5\linewidth]{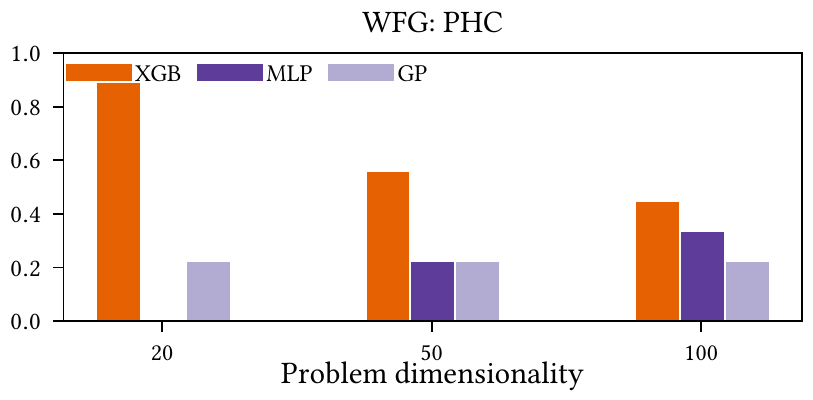}\\
\caption{%
    Performance summary for the high-dimensional WFG test problems based on the
    problem's dimensionality: (\emph{upper}) hypervolume (\emph{lower}) IGD+.
    Bar heights correspond to the  proportion of times that a method is best or
    statistically equivalent to the best method the benchmark's problems.
}
\label{fig:WFG_HD_dim}
\end{figure}

\newpage
\section{Computational timing for DTLZ and WFG}
\label{sec:comp}
The following plots show the computation time for each problem dimensionality
across the DTLZ ($d \in \{2, 5, 10\}$) and WFG
($d \in \{6, 8, 10, 20, 50, 100\}$) benchmarks. Note that the timings of the
real-world problems are not included because they are not configurable, \ie
only one dimensionality can be used per problem.
\begin{figure}[H] 
\includegraphics[width=0.33\linewidth]{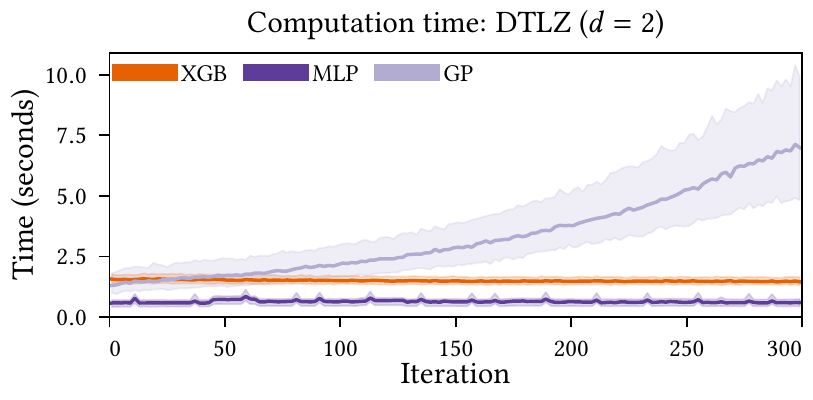}%
\includegraphics[width=0.33\linewidth]{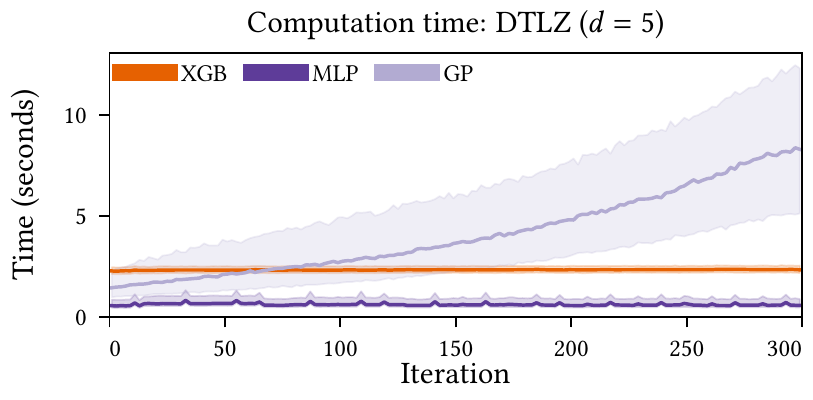}%
\includegraphics[width=0.33\linewidth]{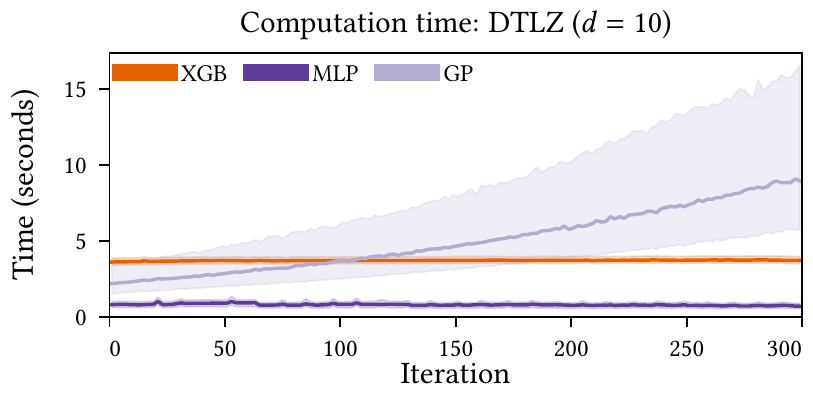}\\
\includegraphics[width=0.33\linewidth]{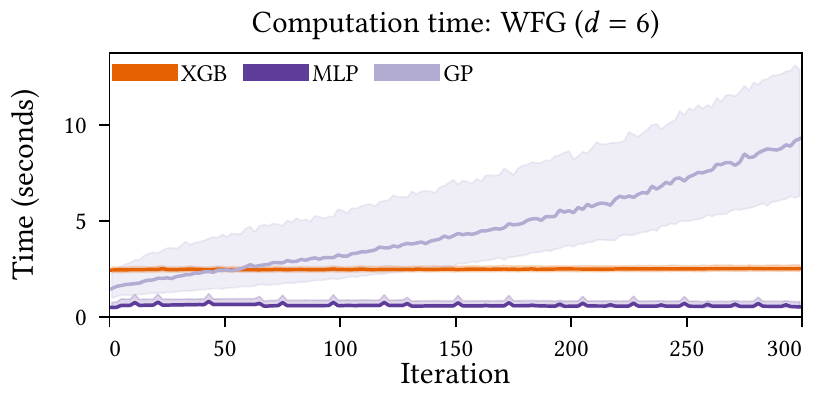}%
\includegraphics[width=0.33\linewidth]{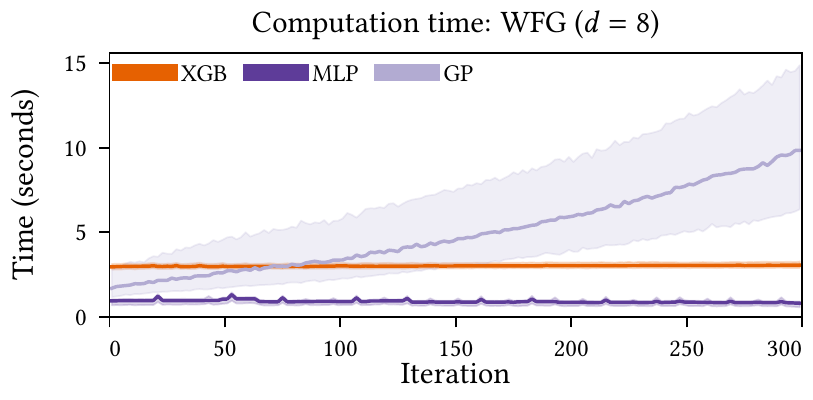}%
\includegraphics[width=0.33\linewidth]{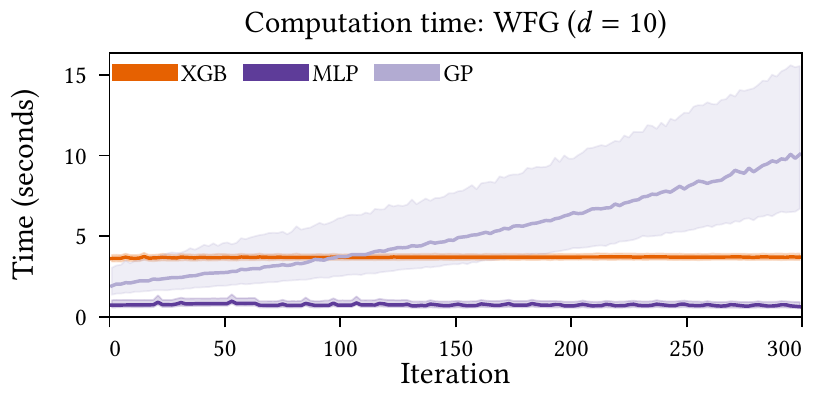}\\
\includegraphics[width=0.33\linewidth]{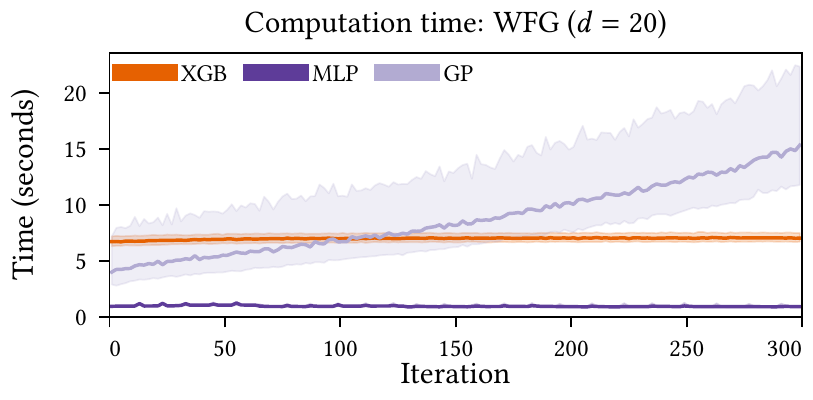}%
\includegraphics[width=0.33\linewidth]{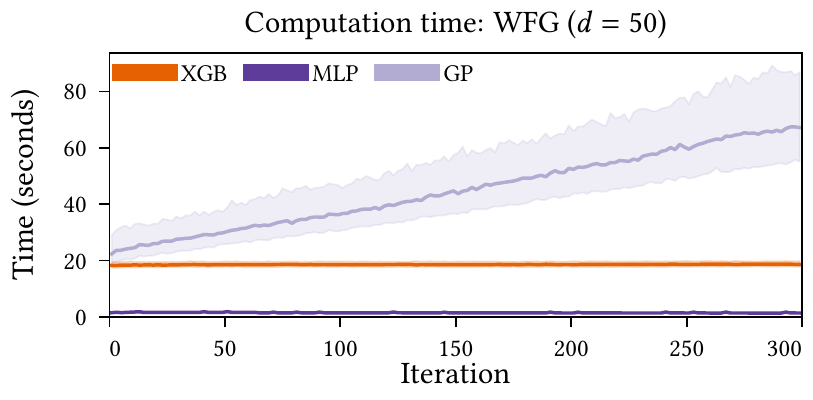}%
\includegraphics[width=0.33\linewidth]{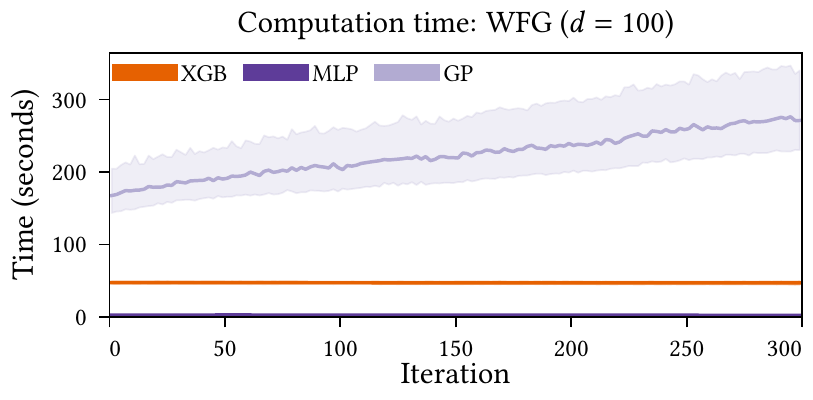}\\
\caption{%
    Computation time taken per iteration on the DTLZ and WFG benchmark
    problems. The median computation time over all scalarisers and problems
    is shown as the solid lines, with their corresponding interquartile ranges
    shaded.
}
\label{fig:timing}
\end{figure}

\newpage
\section{Convergence plots}
\label{sec:conv}

\subsection{DTLZ Convergence Plots}
\label{sec:conv:dtlz}
\begin{figure}[H]
\includegraphics[width=0.25\linewidth]{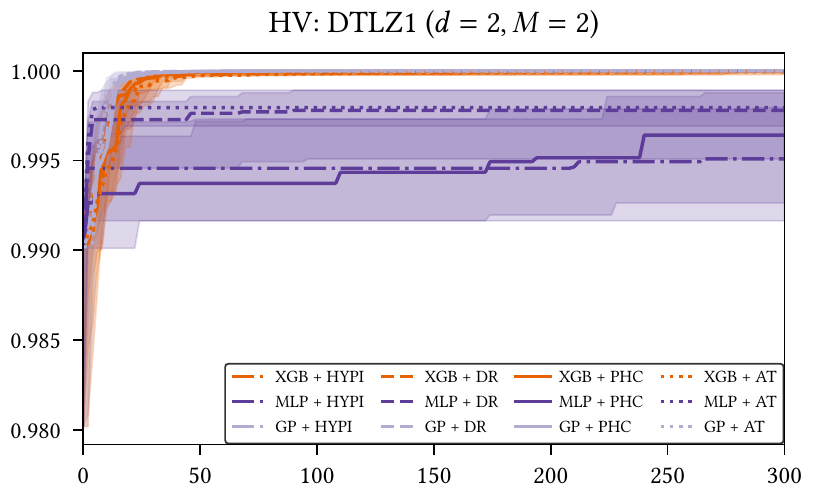}%
\includegraphics[width=0.25\linewidth]{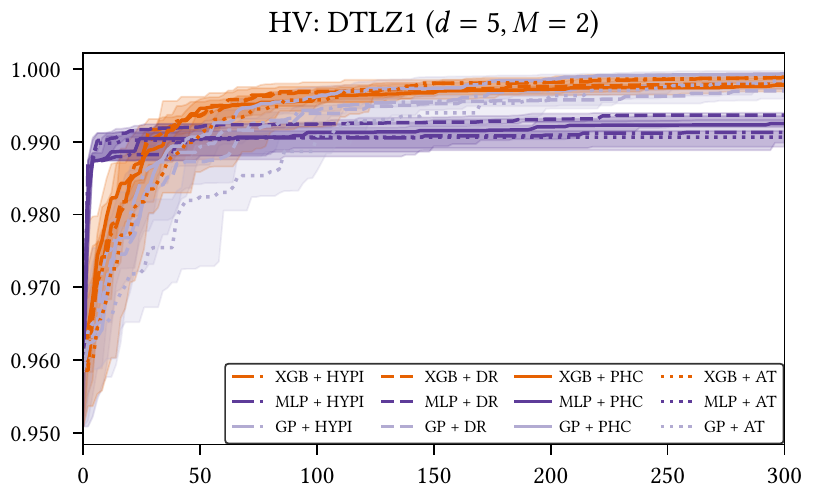}%
\includegraphics[width=0.25\linewidth]{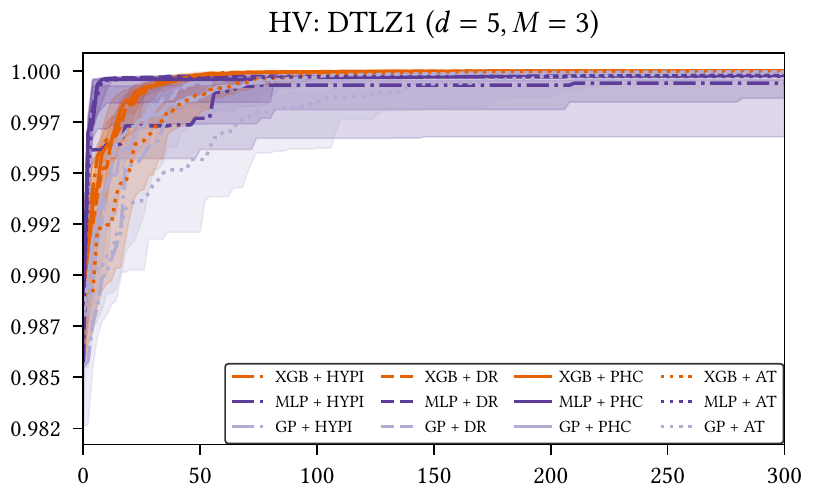}%
\includegraphics[width=0.25\linewidth]{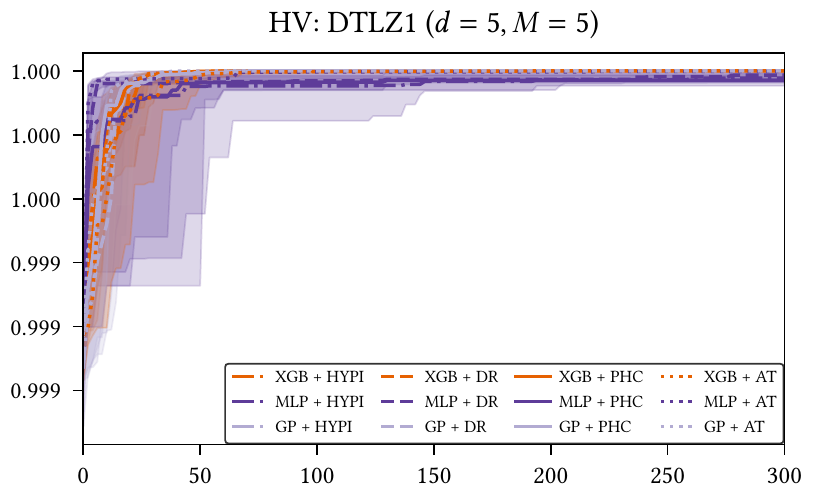}\\
\includegraphics[width=0.25\linewidth]{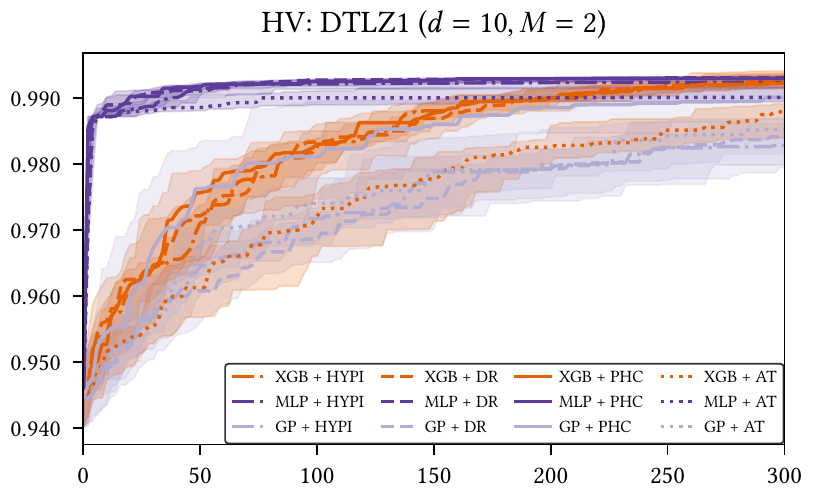}%
\includegraphics[width=0.25\linewidth]{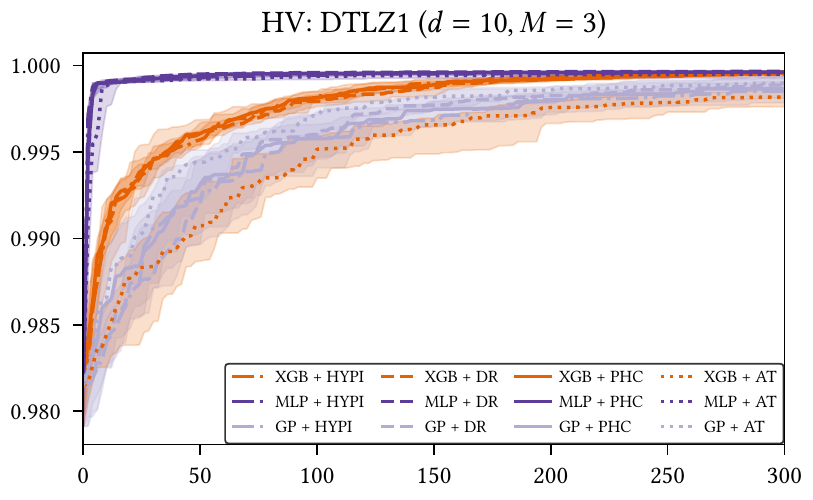}%
\includegraphics[width=0.25\linewidth]{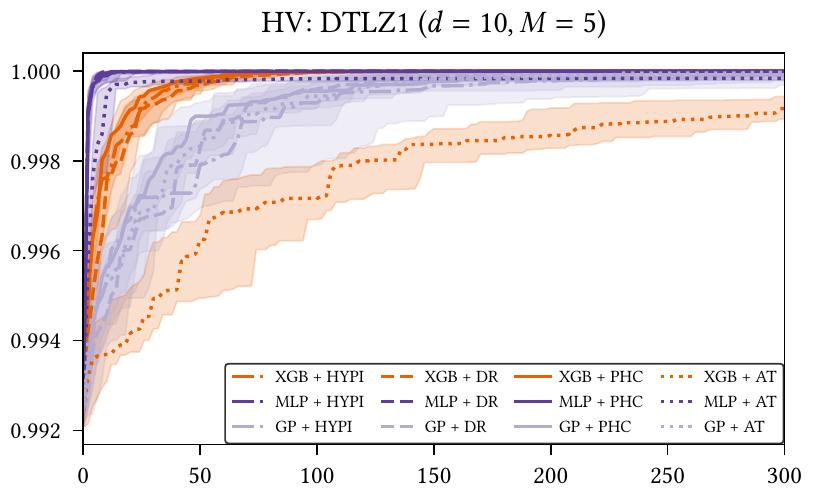}%
\includegraphics[width=0.25\linewidth]{figs/conv_hv_DTLZ1_10_3}\\
\includegraphics[width=0.25\linewidth]{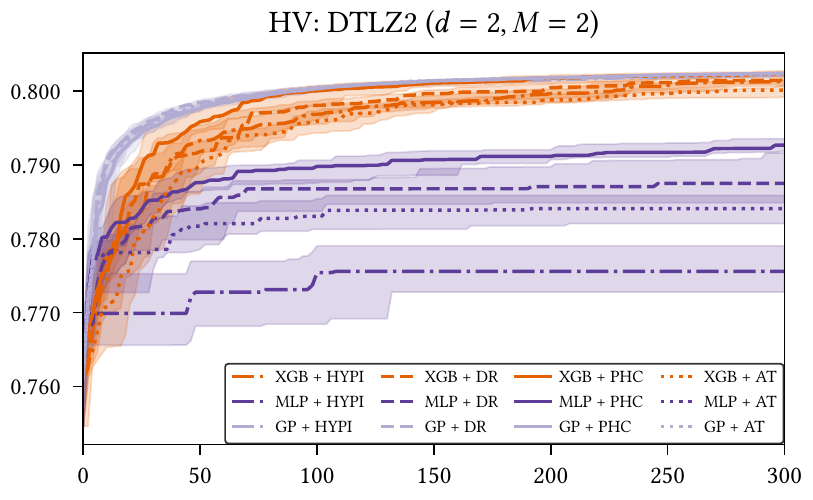}%
\includegraphics[width=0.25\linewidth]{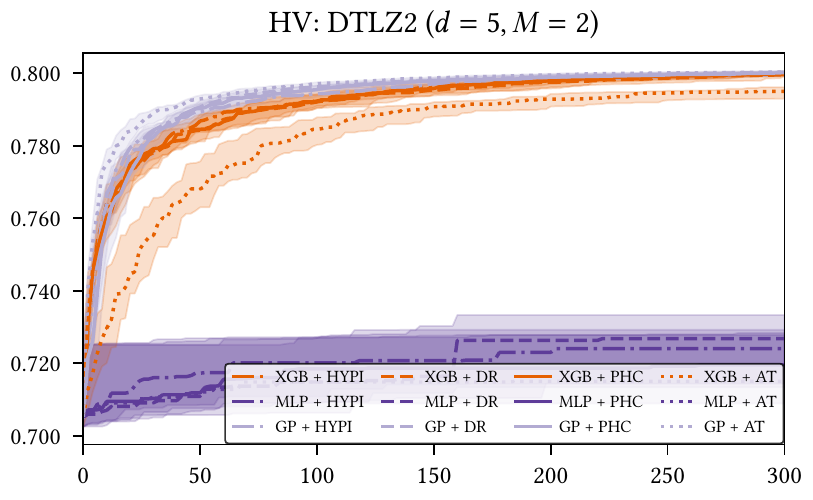}%
\includegraphics[width=0.25\linewidth]{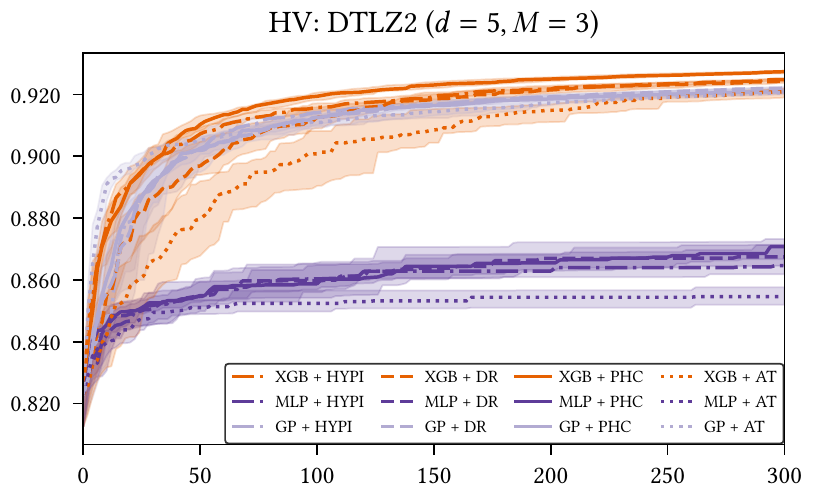}%
\includegraphics[width=0.25\linewidth]{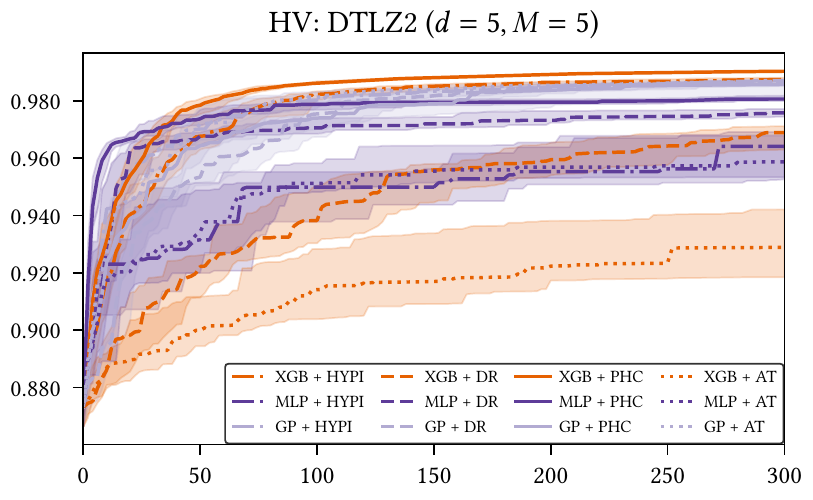}\\
\includegraphics[width=0.25\linewidth]{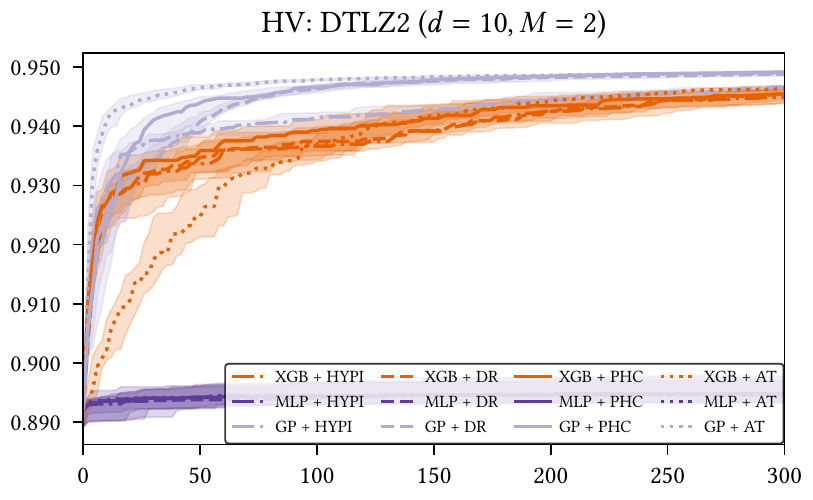}%
\includegraphics[width=0.25\linewidth]{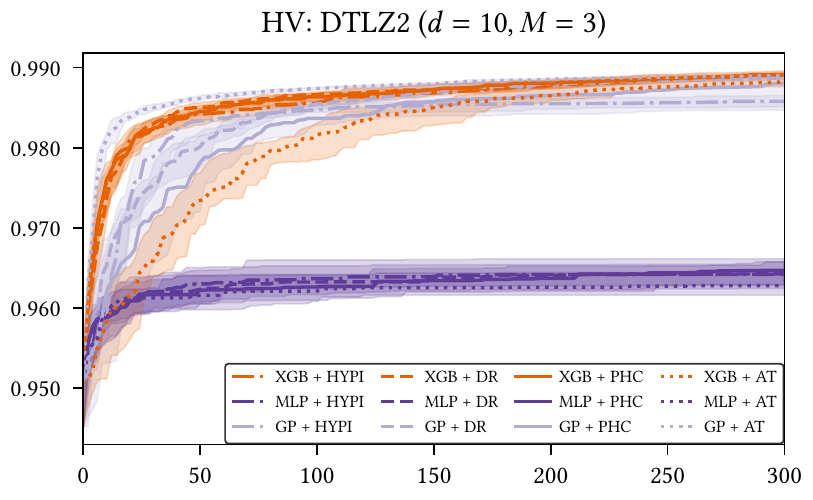}%
\includegraphics[width=0.25\linewidth]{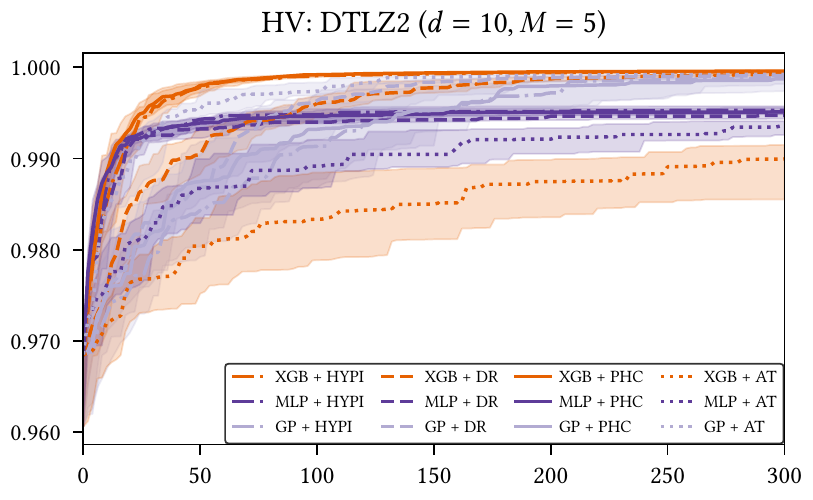}%
\includegraphics[width=0.25\linewidth]{figs/conv_hv_DTLZ2_10_3}\\
\rule{\linewidth}{0.4pt}
\includegraphics[width=0.25\linewidth]{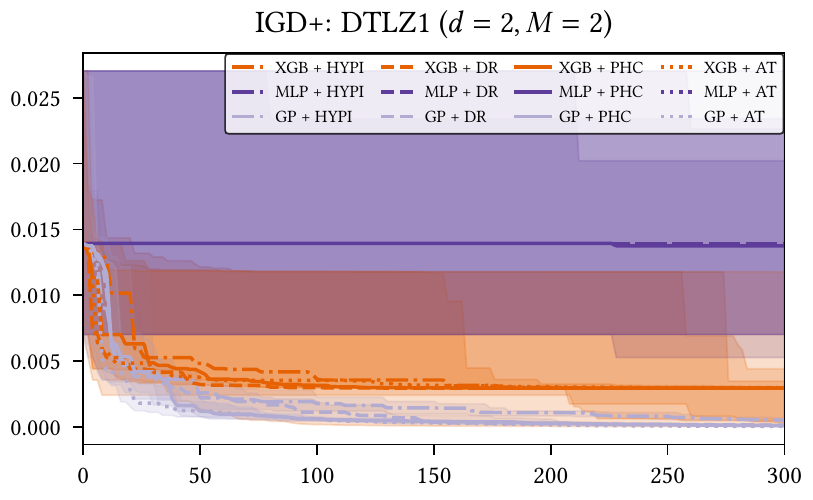}%
\includegraphics[width=0.25\linewidth]{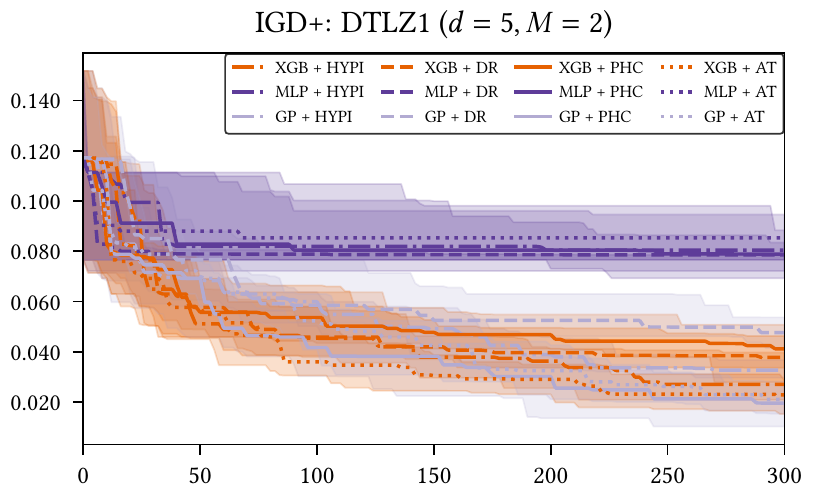}%
\includegraphics[width=0.25\linewidth]{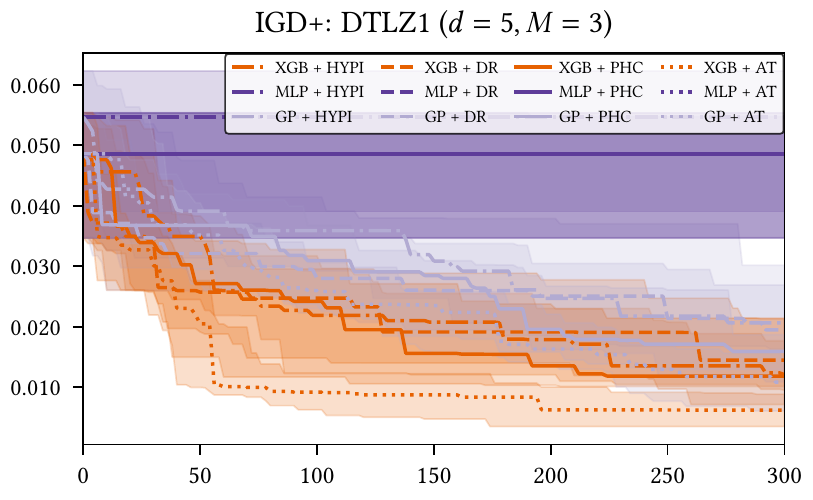}%
\includegraphics[width=0.25\linewidth]{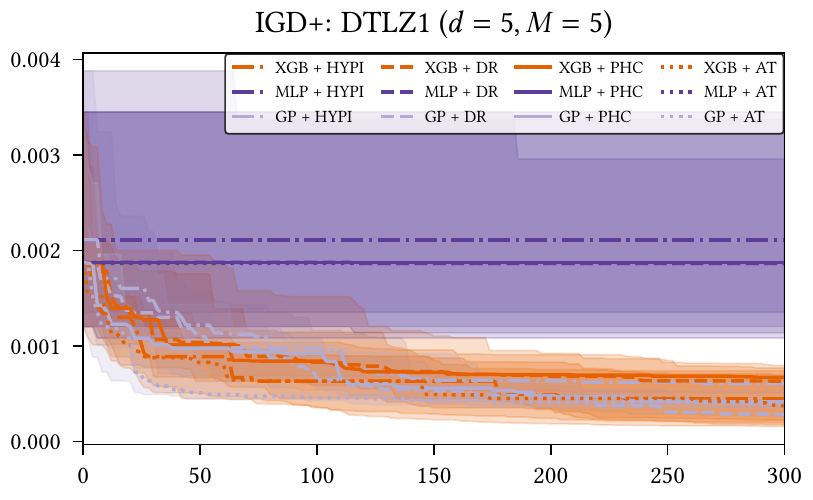}\\
\includegraphics[width=0.25\linewidth]{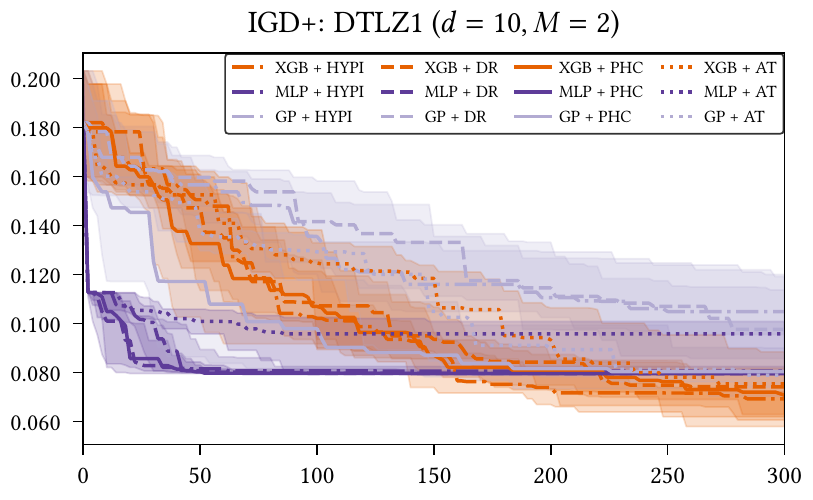}%
\includegraphics[width=0.25\linewidth]{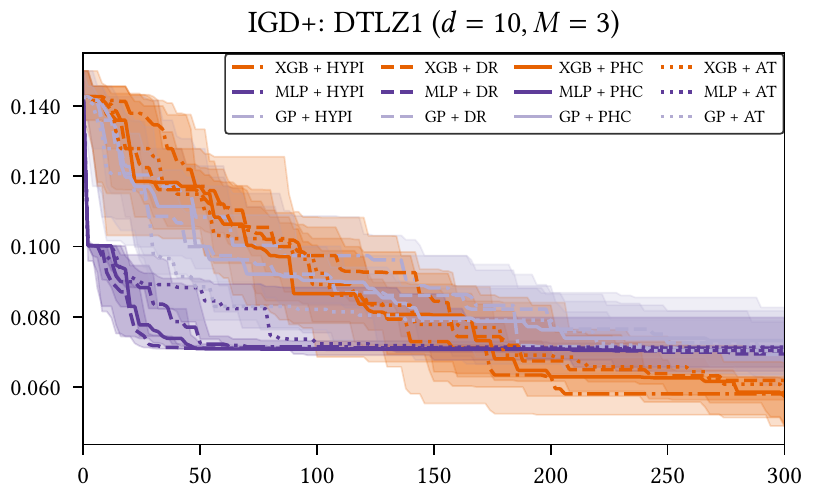}%
\includegraphics[width=0.25\linewidth]{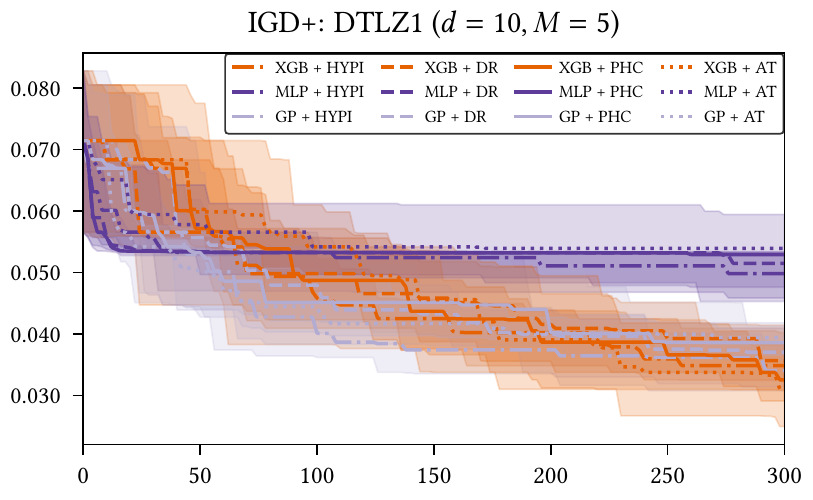}%
\includegraphics[width=0.25\linewidth]{figs/conv_igd+_DTLZ1_10_3}\\
\includegraphics[width=0.25\linewidth]{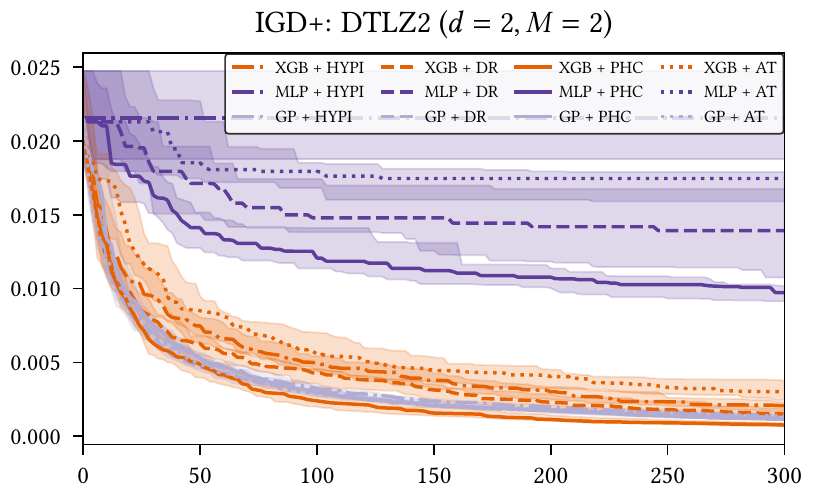}%
\includegraphics[width=0.25\linewidth]{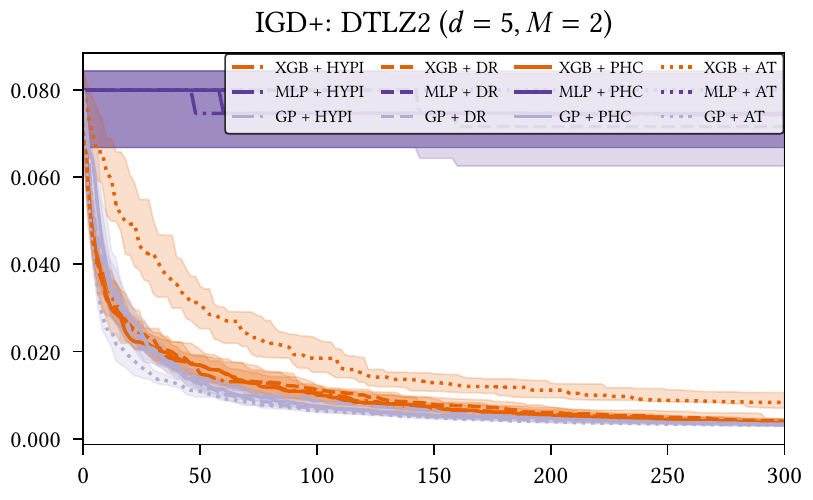}%
\includegraphics[width=0.25\linewidth]{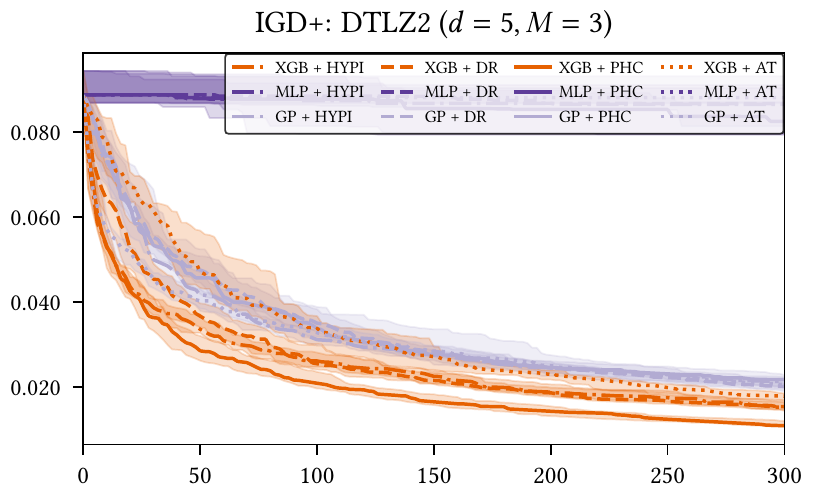}%
\includegraphics[width=0.25\linewidth]{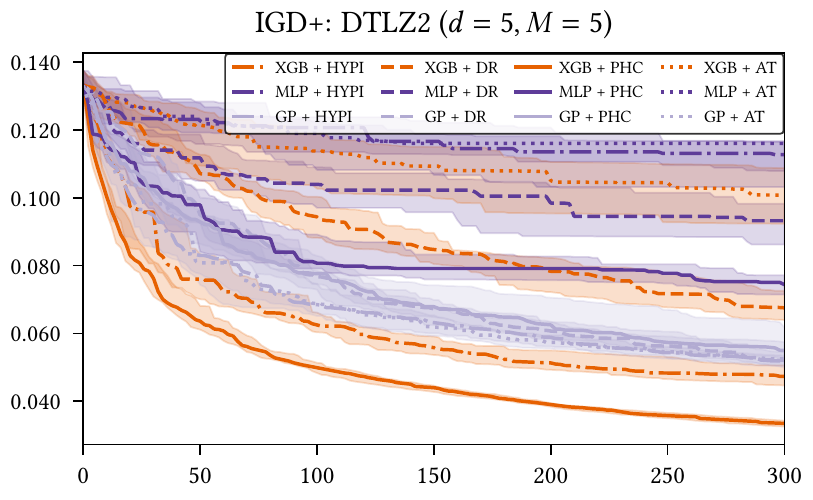}\\
\includegraphics[width=0.25\linewidth]{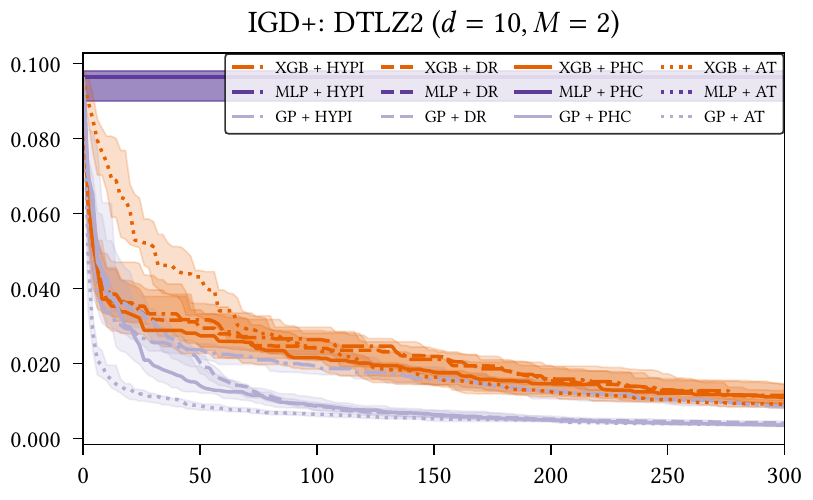}%
\includegraphics[width=0.25\linewidth]{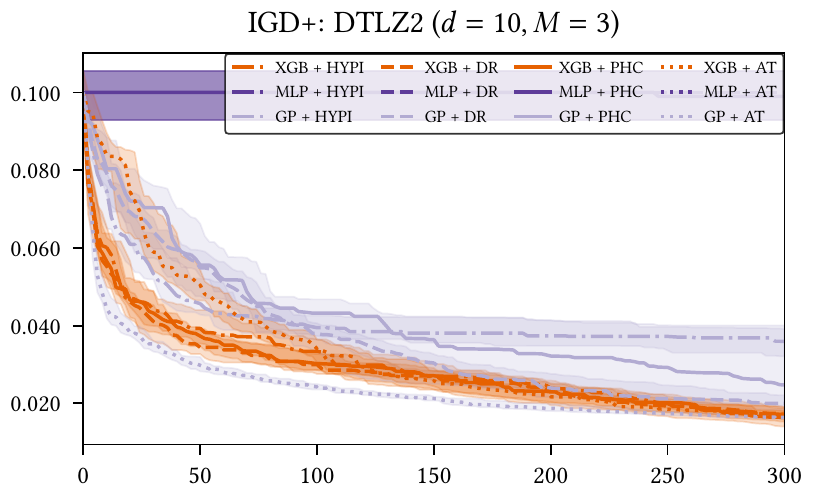}%
\includegraphics[width=0.25\linewidth]{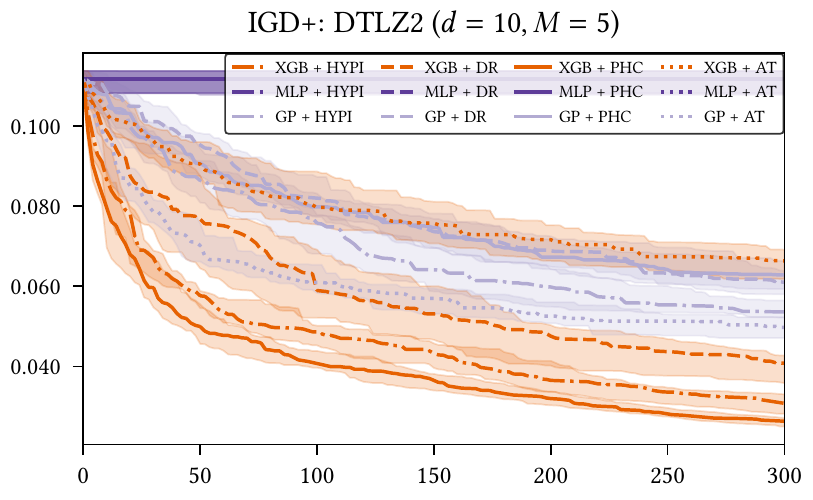}%
\includegraphics[width=0.25\linewidth]{figs/conv_igd+_DTLZ2_10_3}\\
\caption{%
    Hypervolume (\emph{upper}) and IGD+ (\emph{lower})
    convergence plots for DTLZ1 and DTLZ2.}
\label{fig:conv_DTLZ12}
\end{figure}

\begin{figure}[H]
\includegraphics[width=0.25\linewidth]{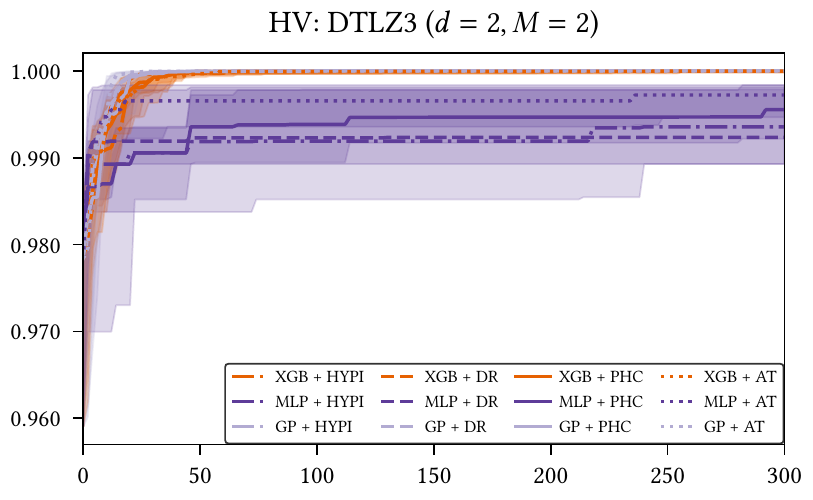}%
\includegraphics[width=0.25\linewidth]{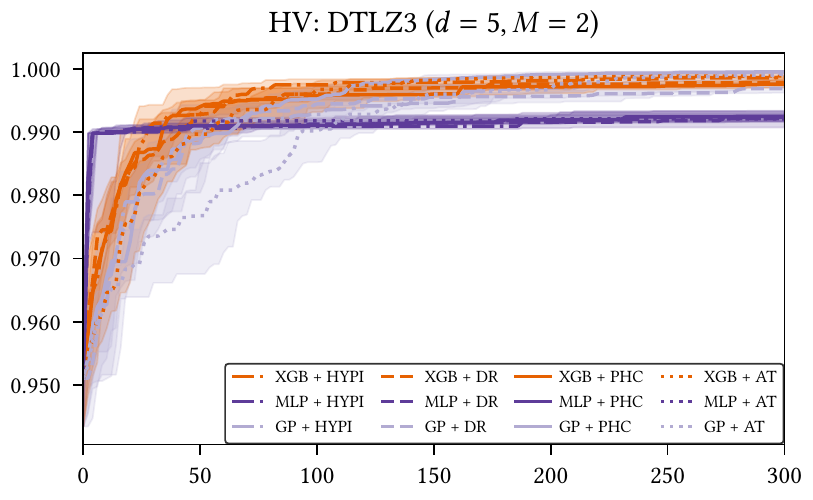}%
\includegraphics[width=0.25\linewidth]{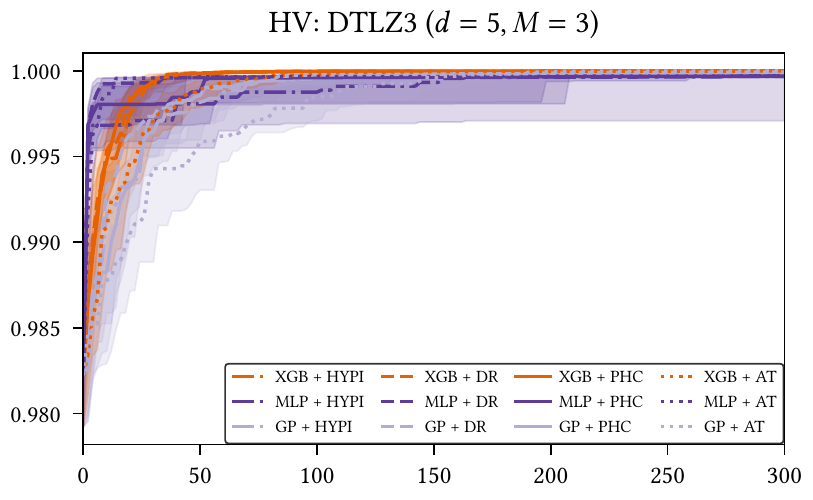}%
\includegraphics[width=0.25\linewidth]{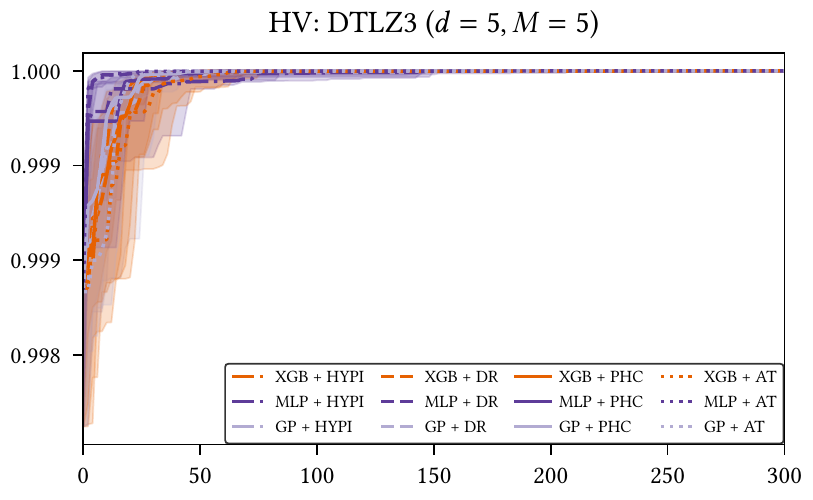}\\
\includegraphics[width=0.25\linewidth]{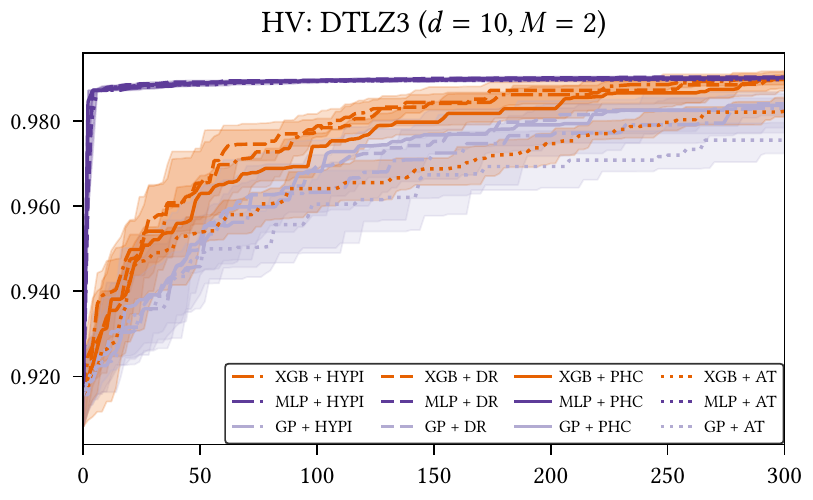}%
\includegraphics[width=0.25\linewidth]{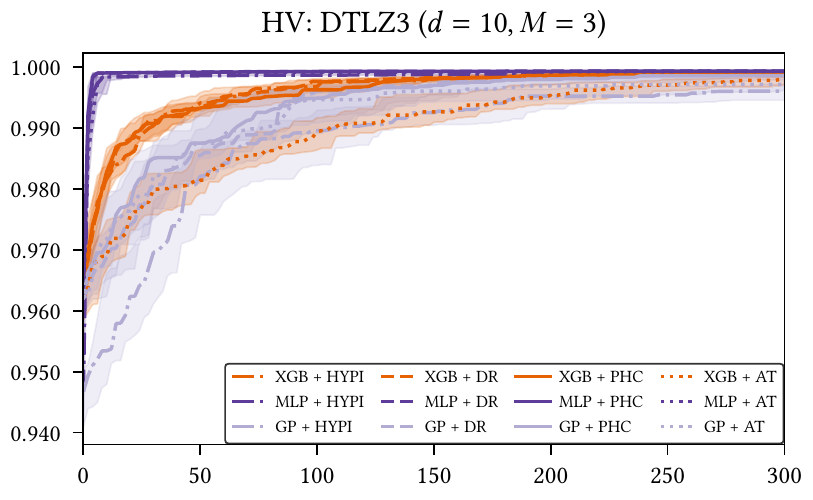}%
\includegraphics[width=0.25\linewidth]{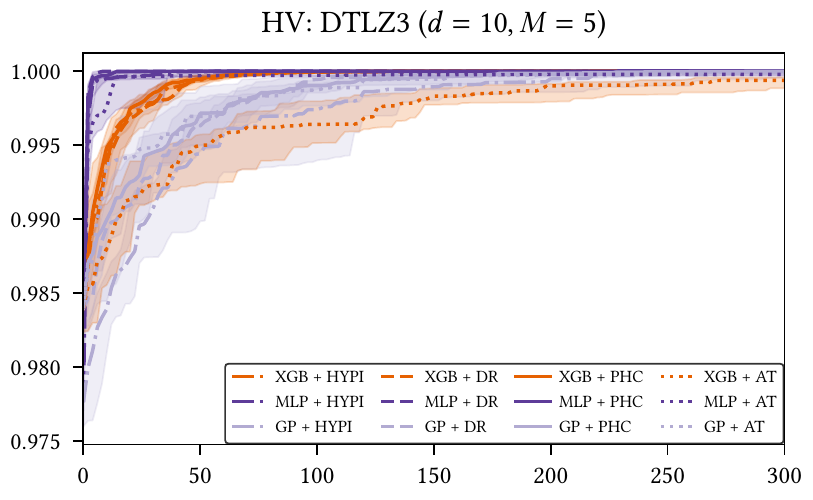}%
\includegraphics[width=0.25\linewidth]{figs/conv_hv_DTLZ3_10_3}\\
\includegraphics[width=0.25\linewidth]{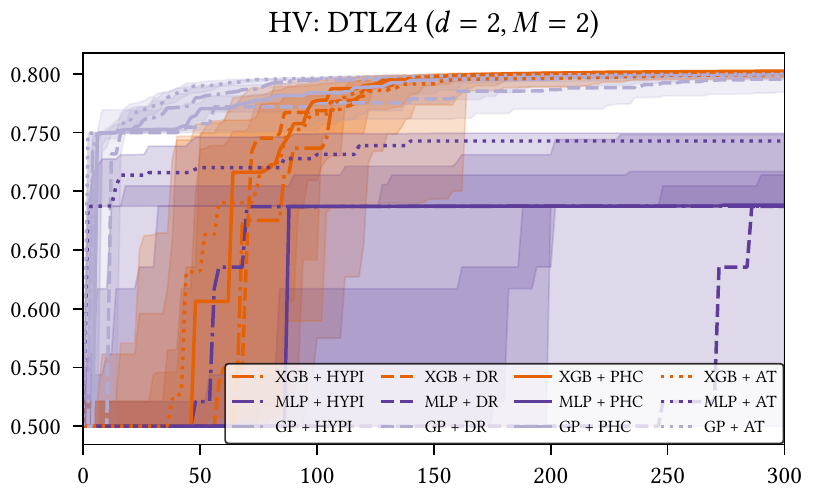}%
\includegraphics[width=0.25\linewidth]{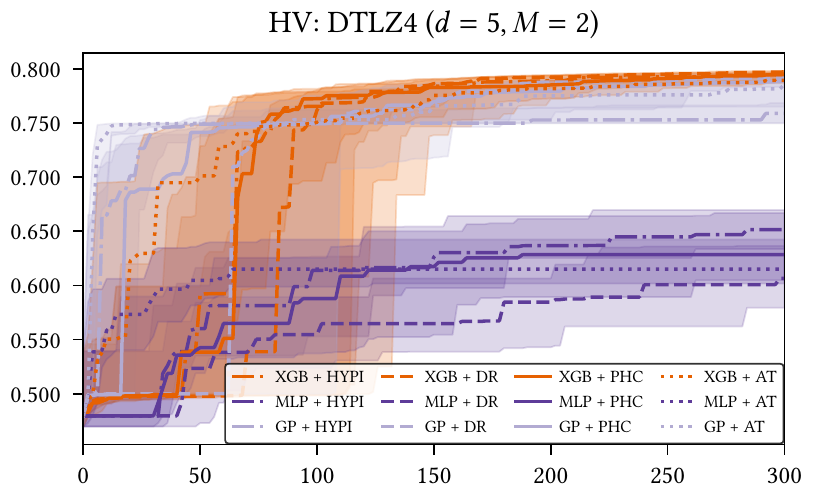}%
\includegraphics[width=0.25\linewidth]{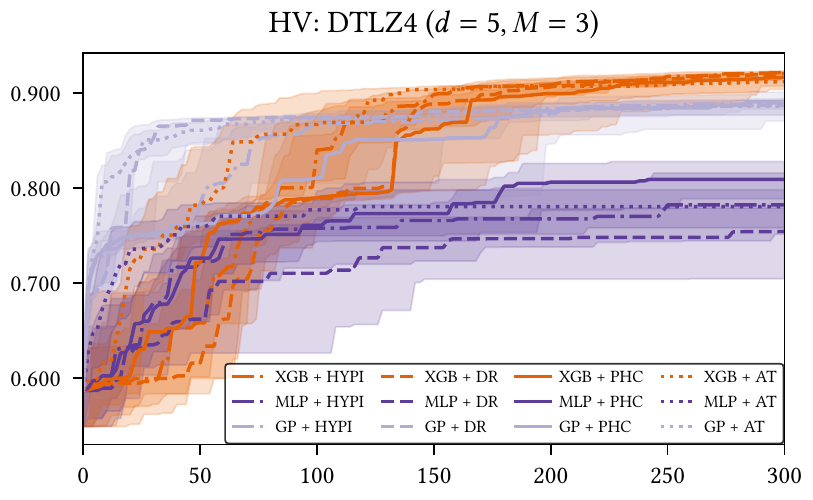}%
\includegraphics[width=0.25\linewidth]{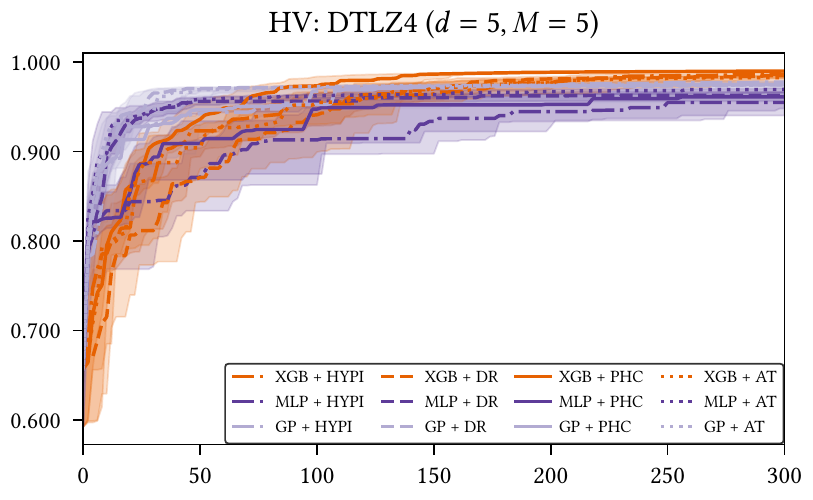}\\
\includegraphics[width=0.25\linewidth]{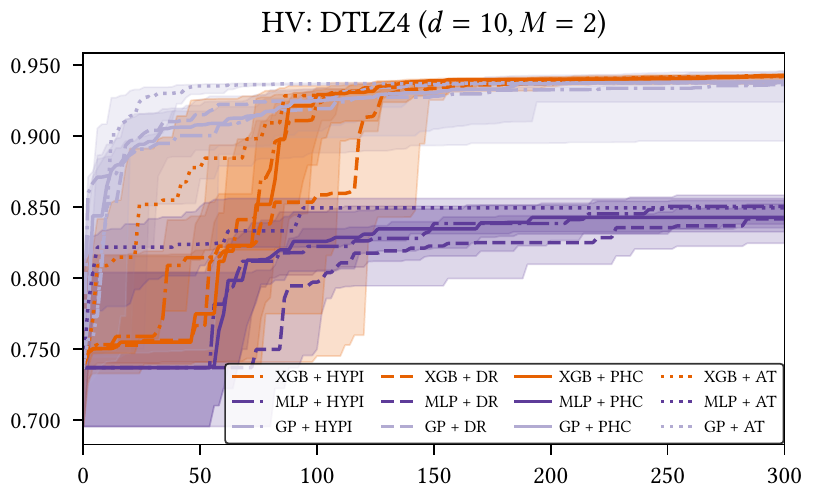}%
\includegraphics[width=0.25\linewidth]{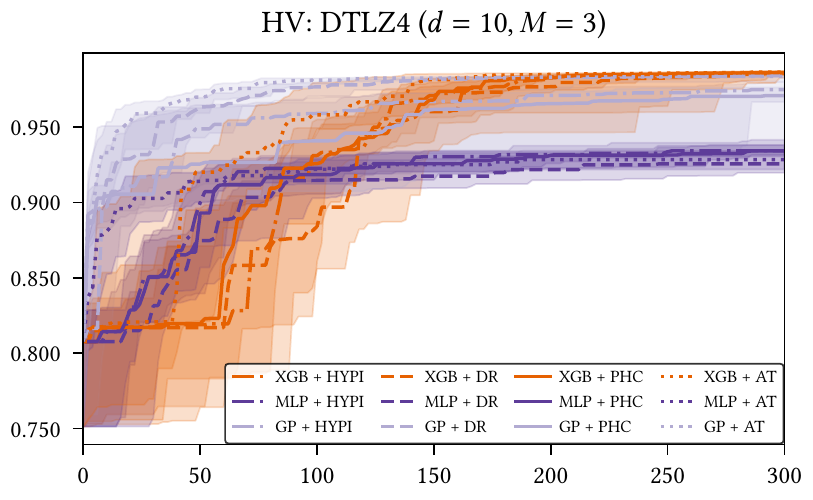}%
\includegraphics[width=0.25\linewidth]{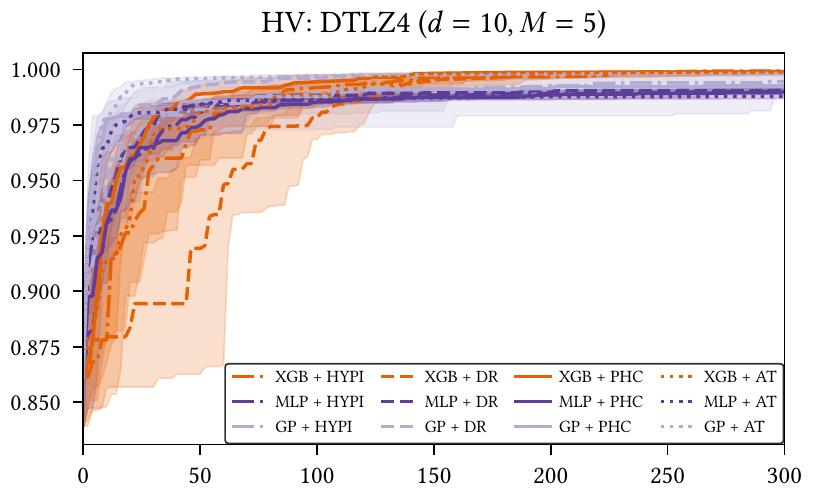}%
\includegraphics[width=0.25\linewidth]{figs/conv_hv_DTLZ4_10_3}\\
\rule{\linewidth}{0.4pt}
\includegraphics[width=0.25\linewidth]{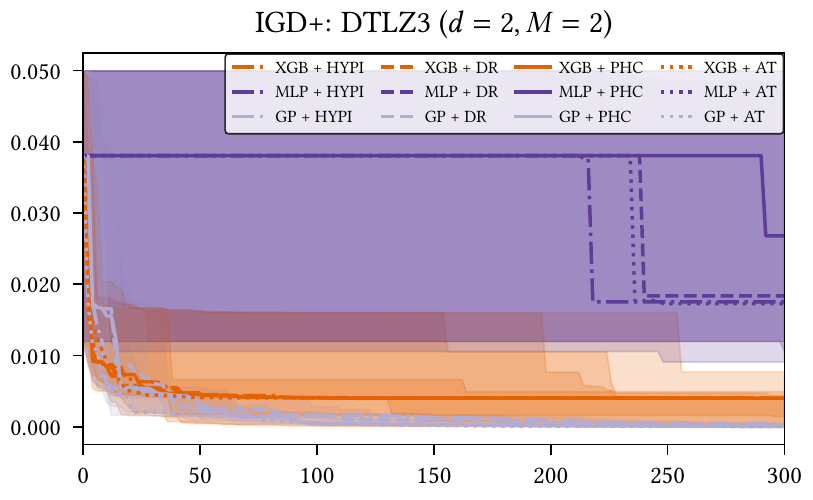}%
\includegraphics[width=0.25\linewidth]{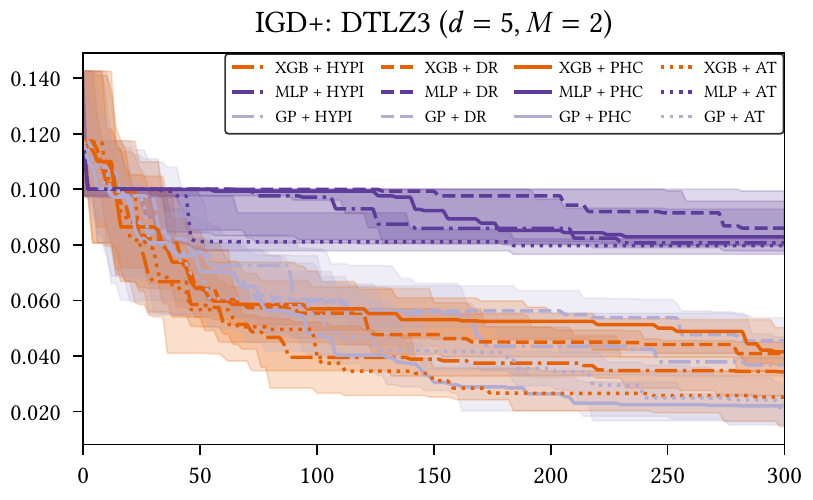}%
\includegraphics[width=0.25\linewidth]{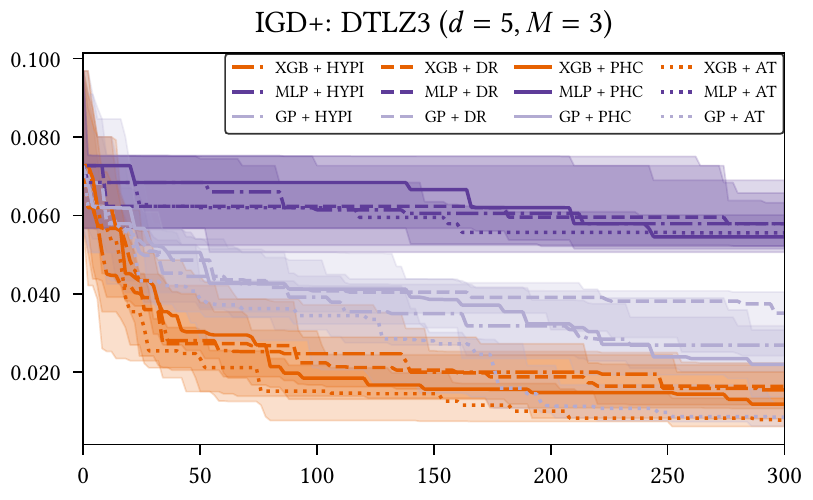}%
\includegraphics[width=0.25\linewidth]{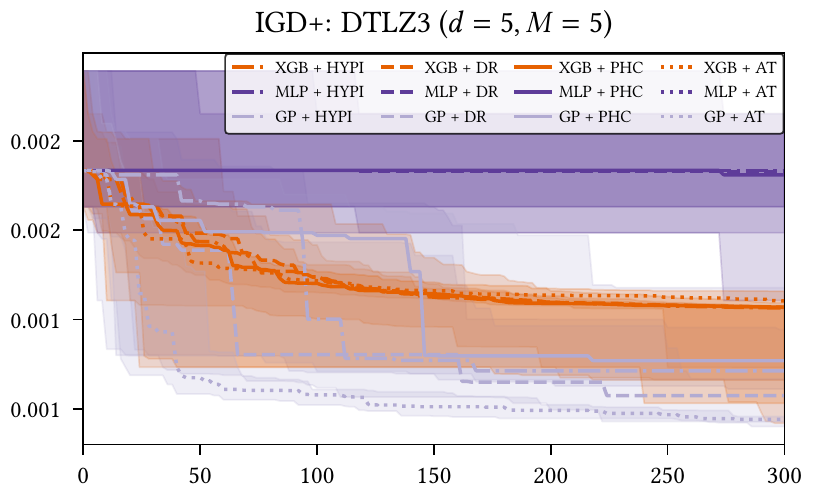}\\
\includegraphics[width=0.25\linewidth]{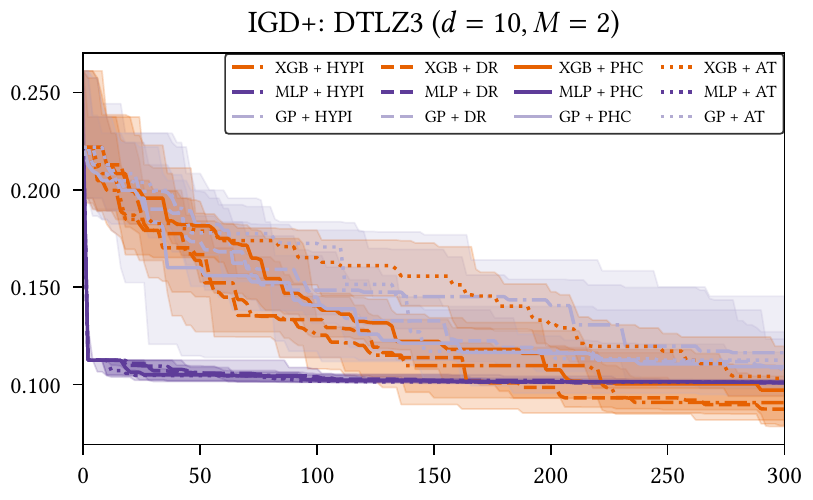}%
\includegraphics[width=0.25\linewidth]{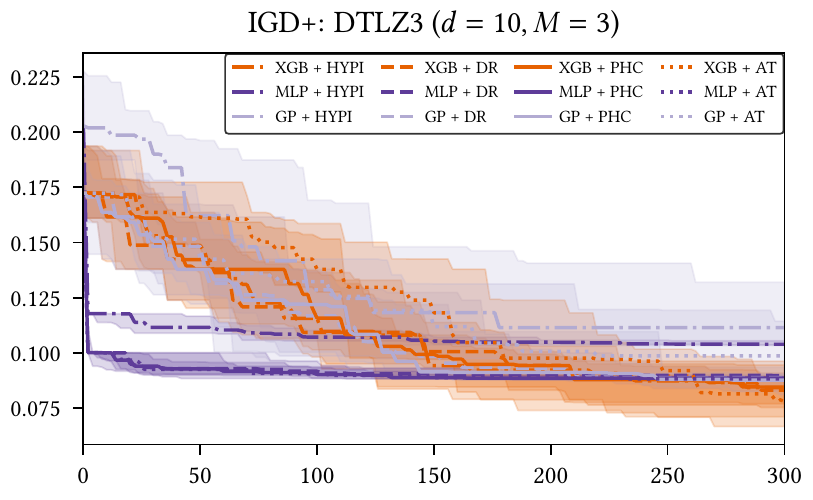}%
\includegraphics[width=0.25\linewidth]{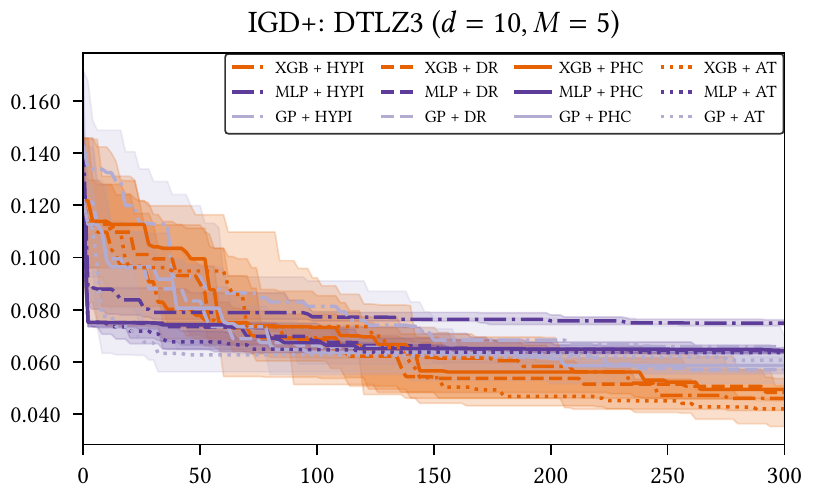}%
\includegraphics[width=0.25\linewidth]{figs/conv_igd+_DTLZ3_10_3}\\
\includegraphics[width=0.25\linewidth]{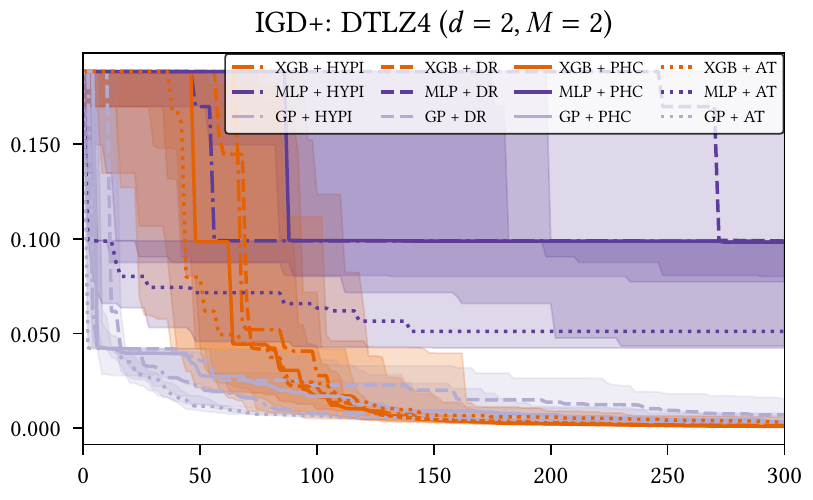}%
\includegraphics[width=0.25\linewidth]{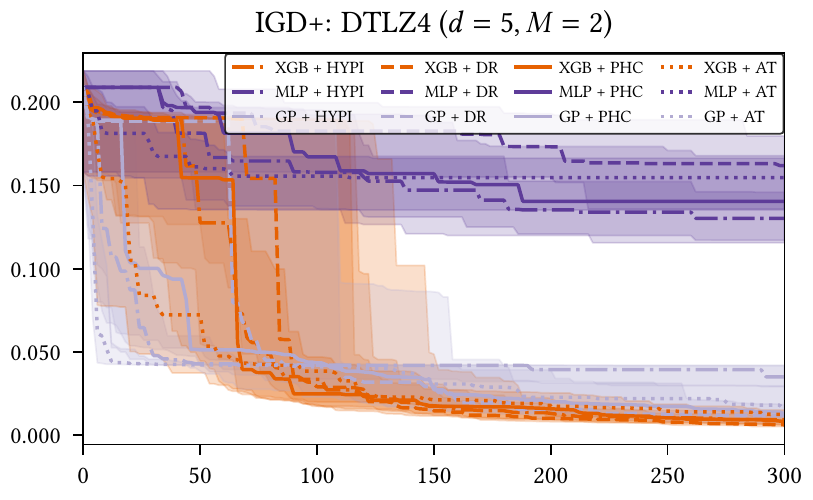}%
\includegraphics[width=0.25\linewidth]{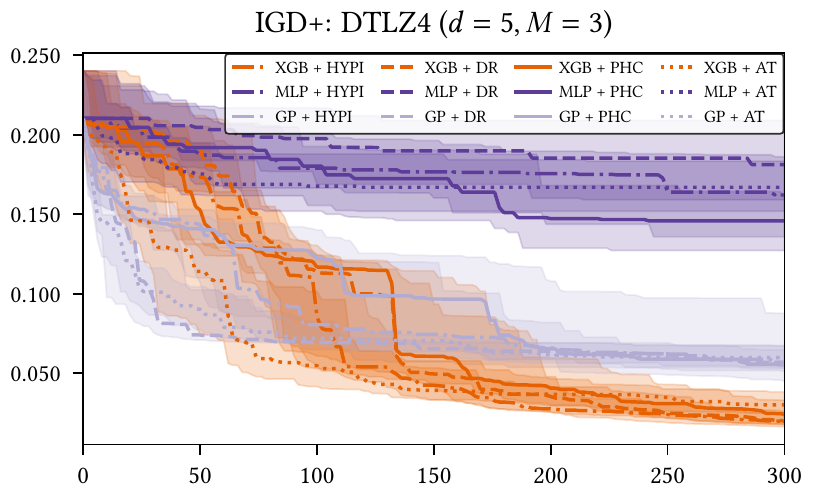}%
\includegraphics[width=0.25\linewidth]{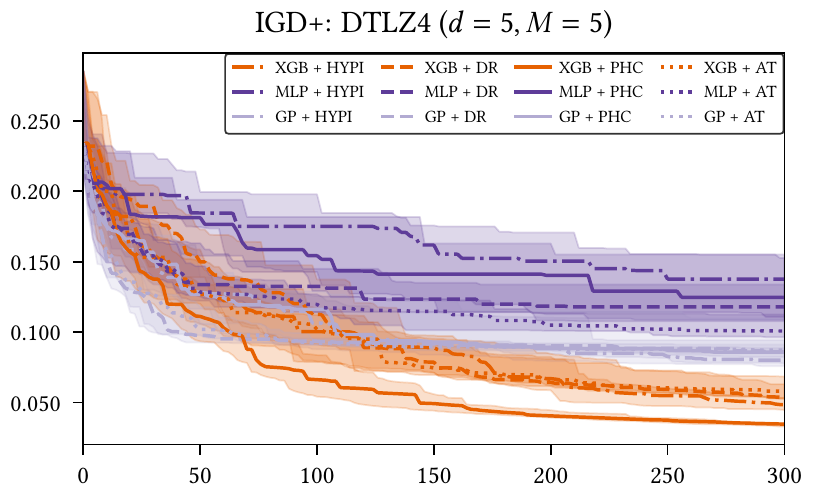}\\
\includegraphics[width=0.25\linewidth]{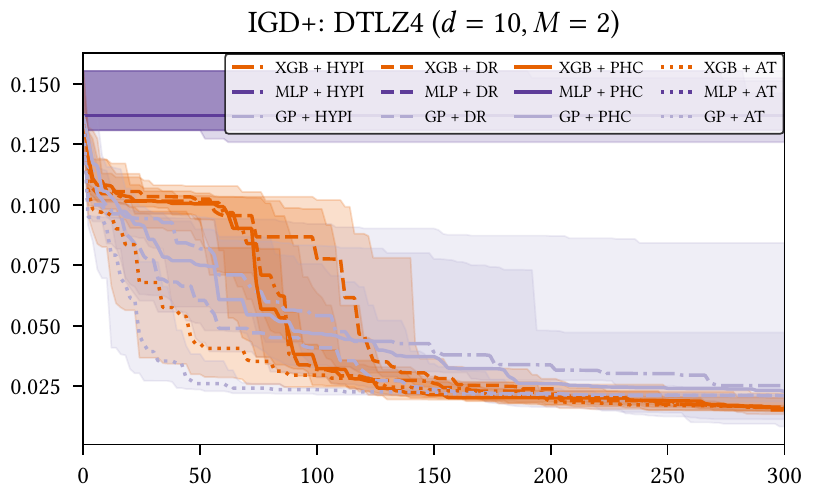}%
\includegraphics[width=0.25\linewidth]{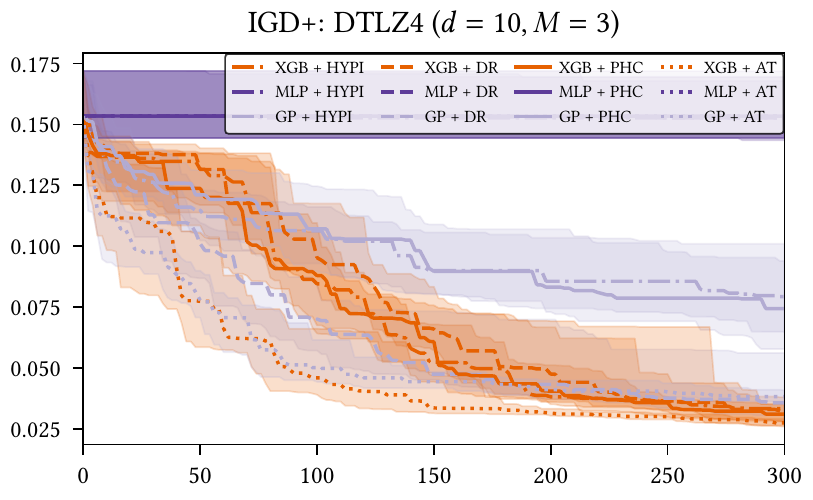}%
\includegraphics[width=0.25\linewidth]{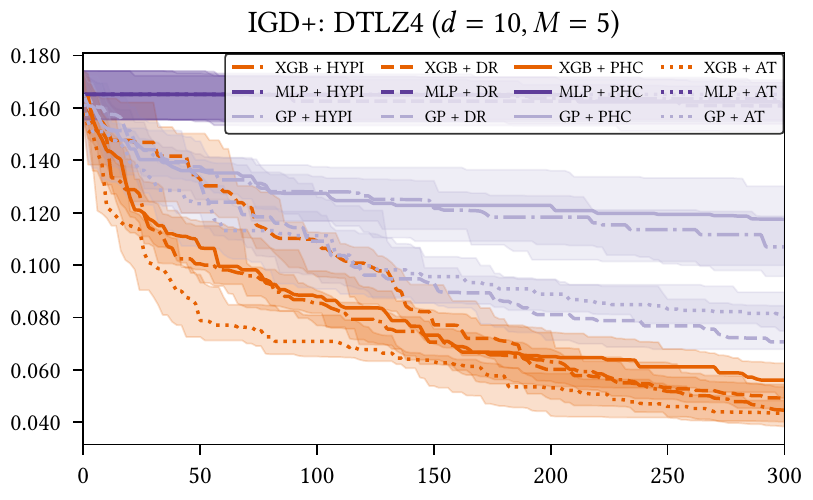}%
\includegraphics[width=0.25\linewidth]{figs/conv_igd+_DTLZ4_10_3}\\
\caption{%
    Hypervolume (\emph{upper}) and IGD+ (\emph{lower})
    convergence plots for DTLZ3 and DTLZ4.}
\label{fig:conv_DTLZ34}
\end{figure}

\begin{figure}[H]
\includegraphics[width=0.25\linewidth]{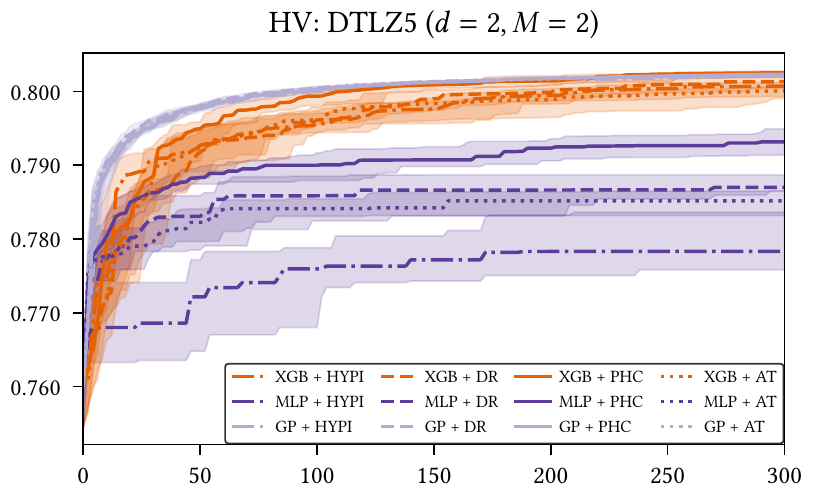}%
\includegraphics[width=0.25\linewidth]{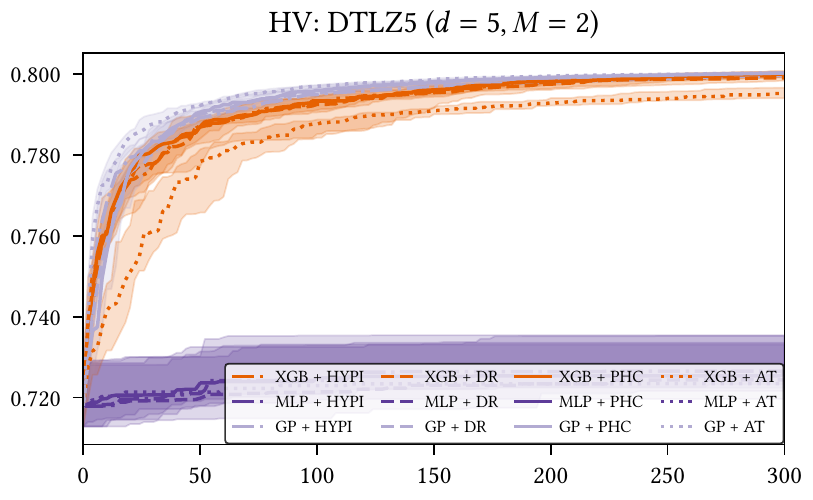}%
\includegraphics[width=0.25\linewidth]{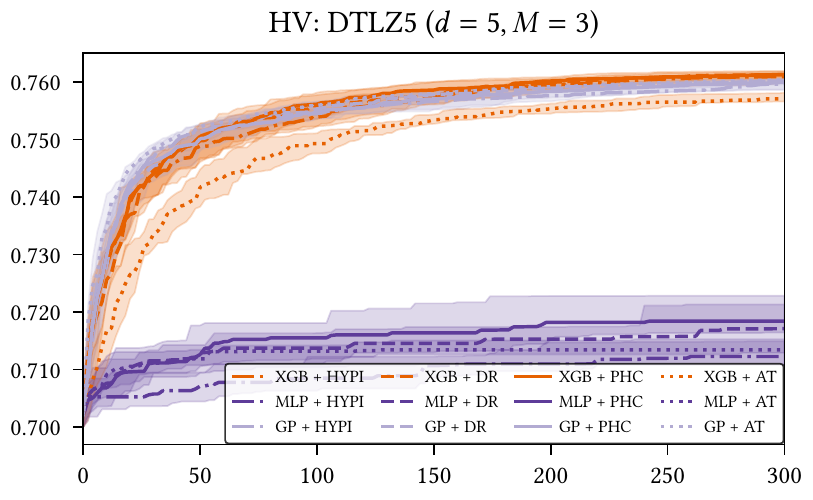}%
\includegraphics[width=0.25\linewidth]{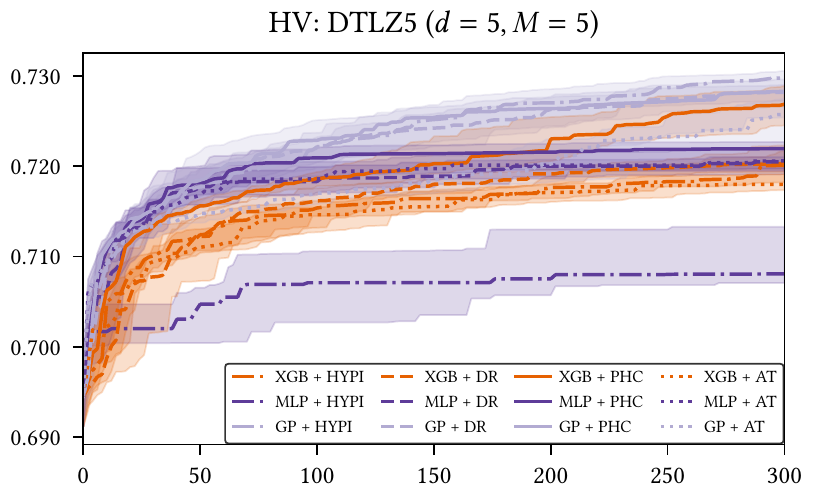}\\
\includegraphics[width=0.25\linewidth]{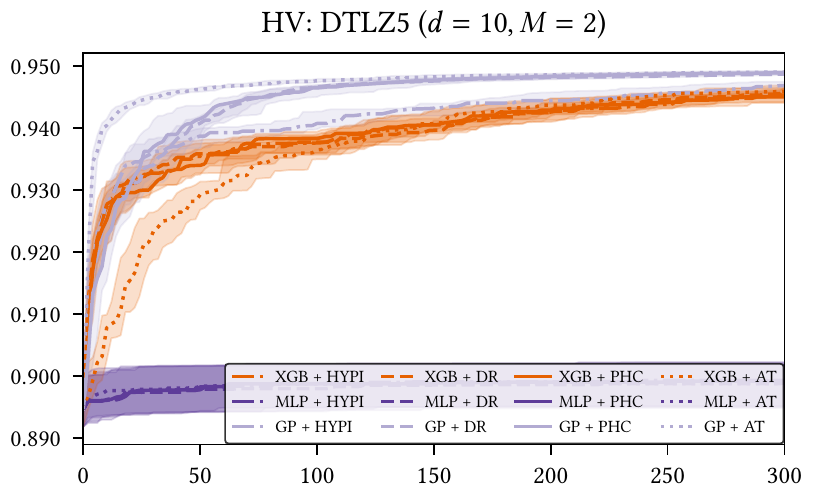}%
\includegraphics[width=0.25\linewidth]{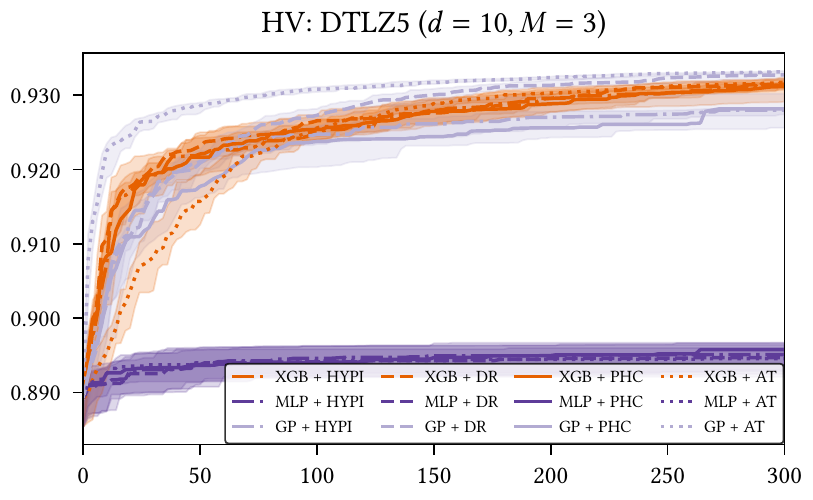}%
\includegraphics[width=0.25\linewidth]{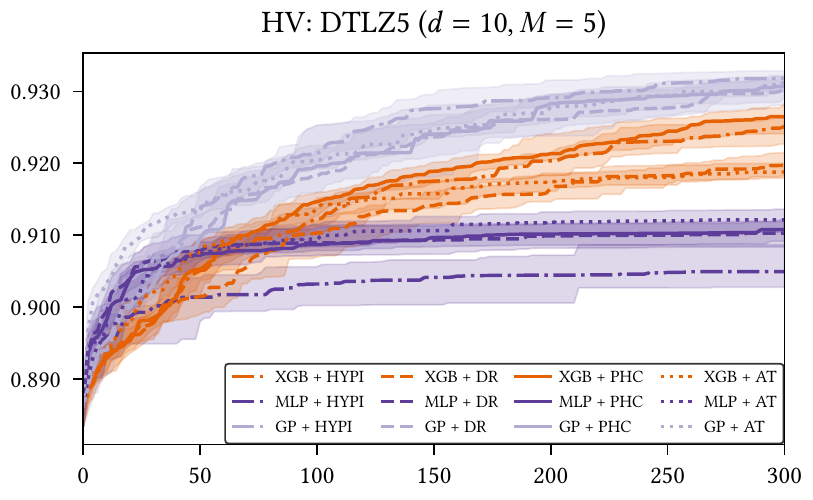}%
\includegraphics[width=0.25\linewidth]{figs/conv_hv_DTLZ5_10_3}\\
\includegraphics[width=0.25\linewidth]{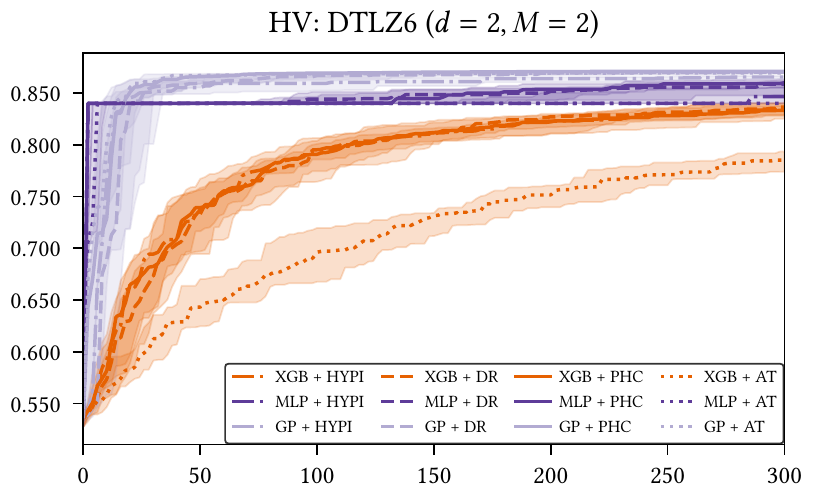}%
\includegraphics[width=0.25\linewidth]{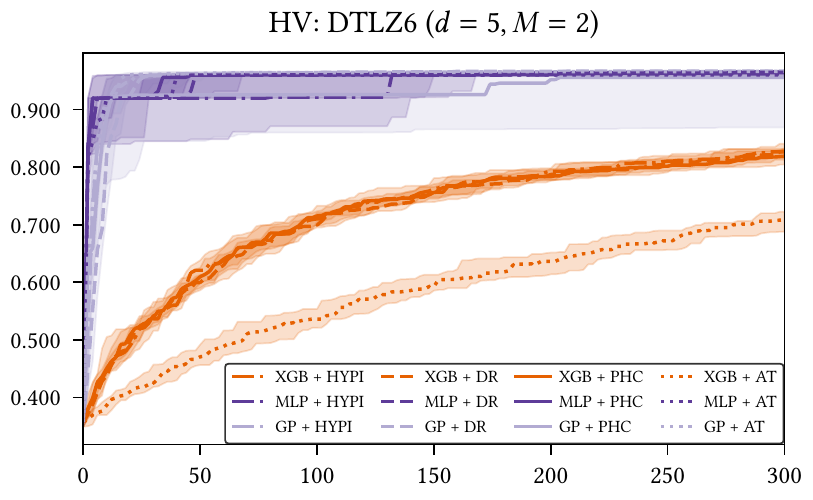}%
\includegraphics[width=0.25\linewidth]{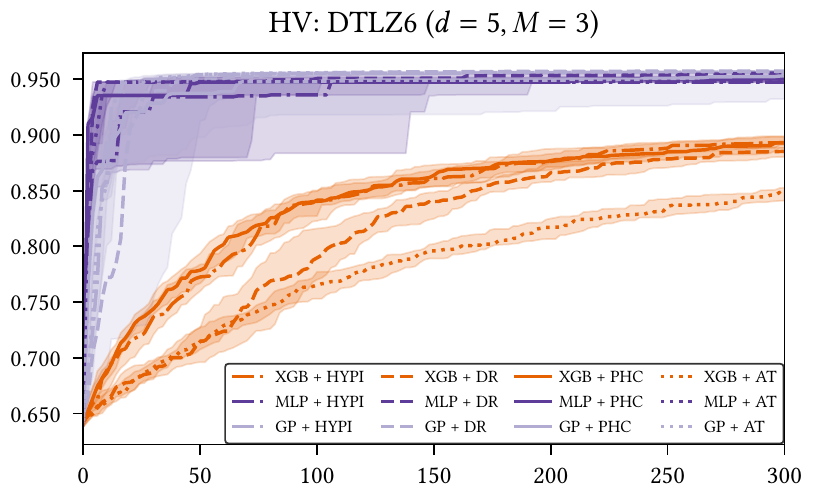}%
\includegraphics[width=0.25\linewidth]{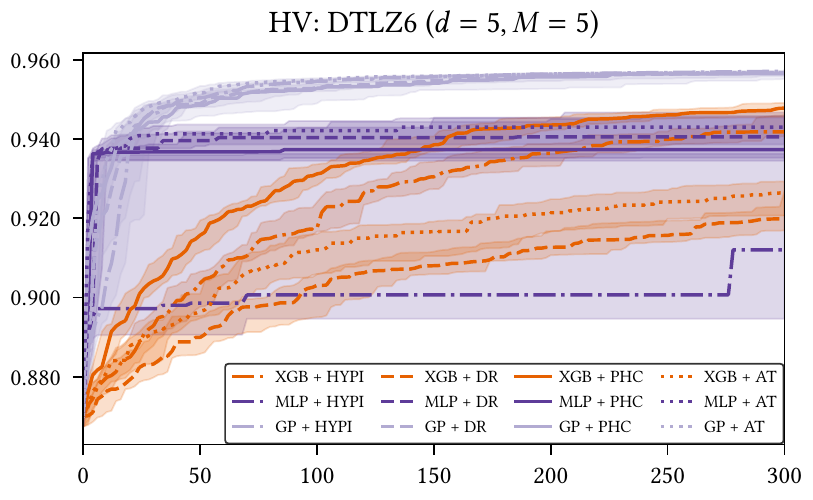}\\
\includegraphics[width=0.25\linewidth]{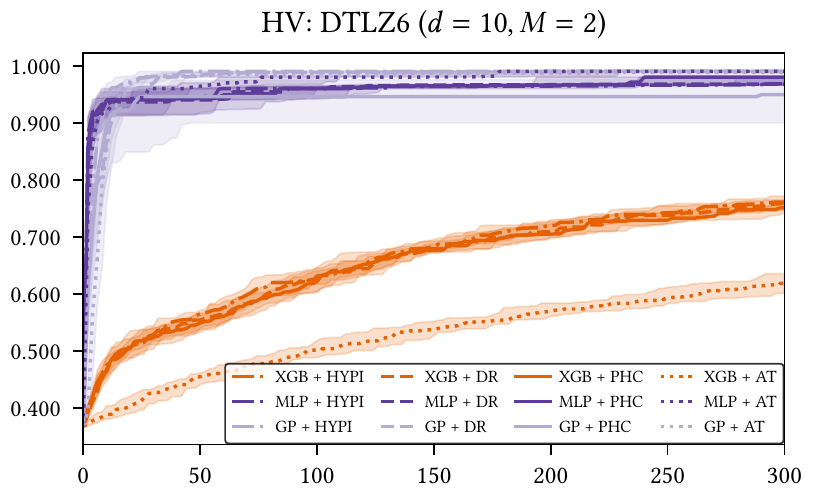}%
\includegraphics[width=0.25\linewidth]{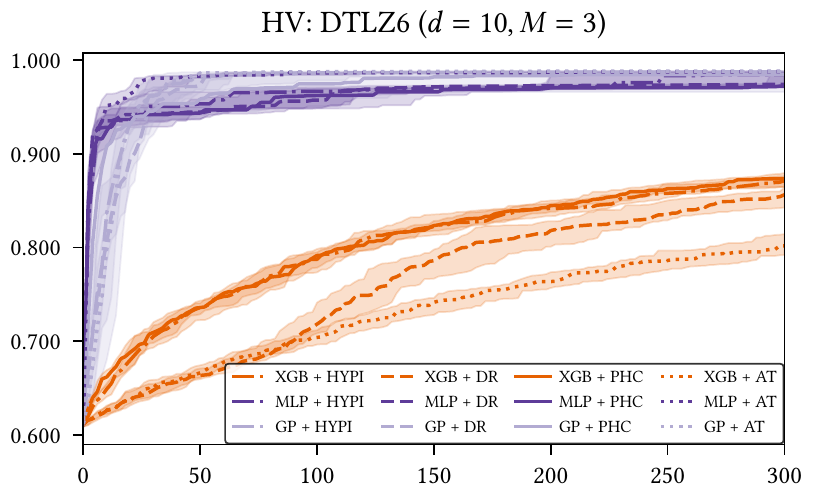}%
\includegraphics[width=0.25\linewidth]{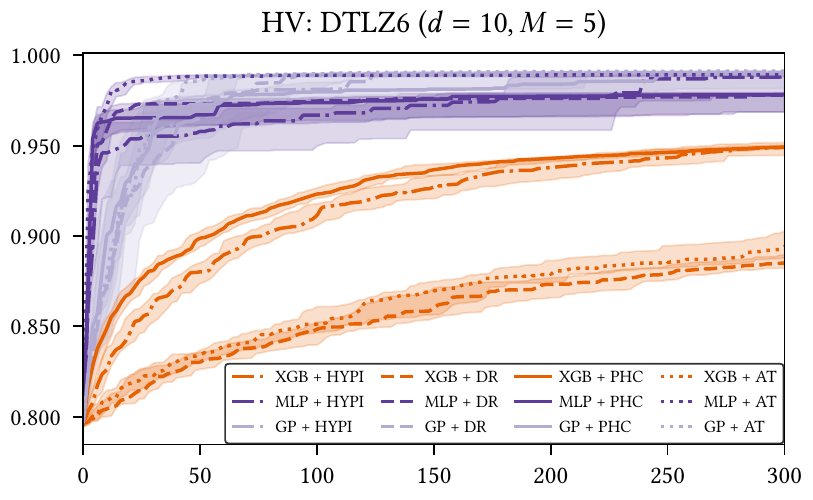}%
\includegraphics[width=0.25\linewidth]{figs/conv_hv_DTLZ6_10_3}\\
\rule{\linewidth}{0.4pt}
\includegraphics[width=0.25\linewidth]{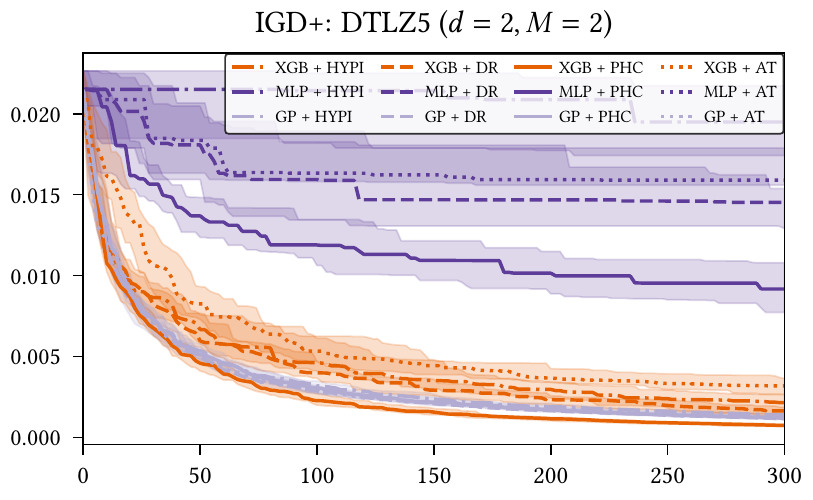}%
\includegraphics[width=0.25\linewidth]{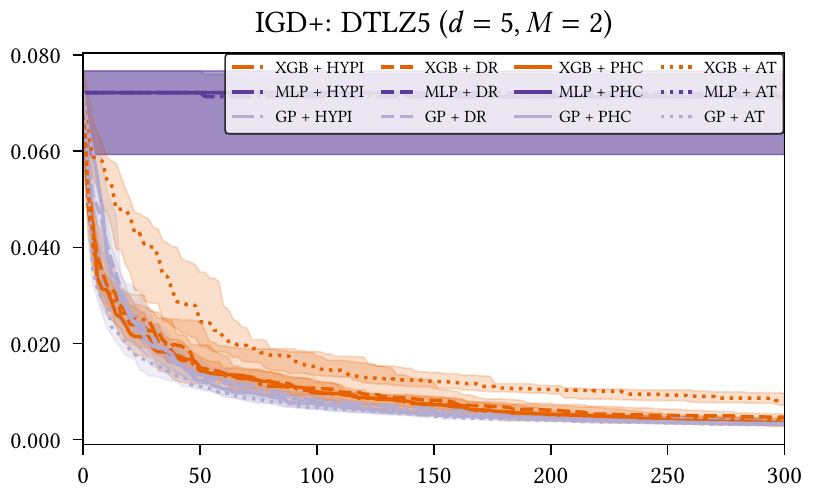}%
\includegraphics[width=0.25\linewidth]{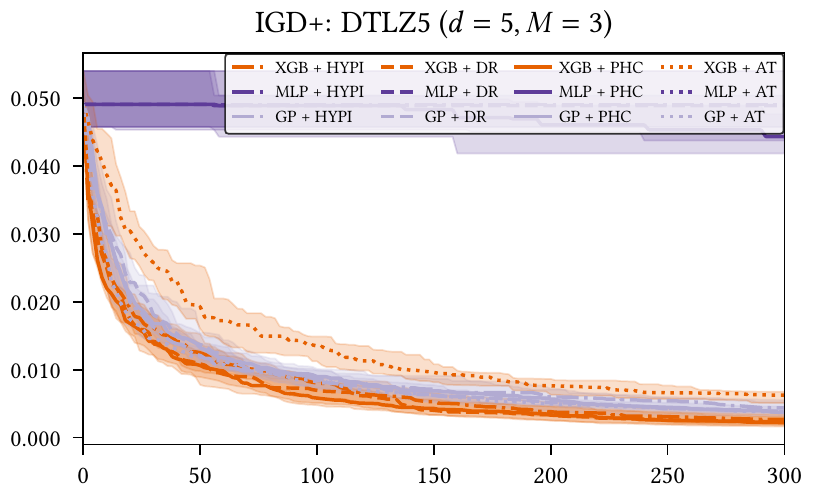}%
\includegraphics[width=0.25\linewidth]{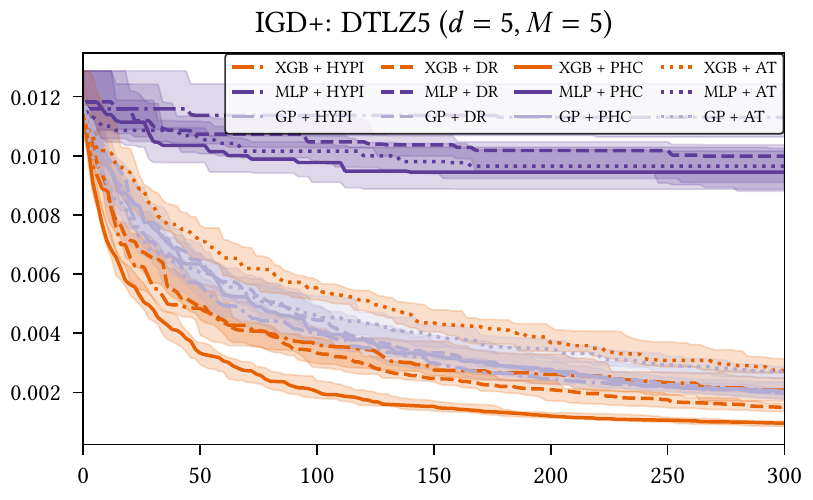}\\
\includegraphics[width=0.25\linewidth]{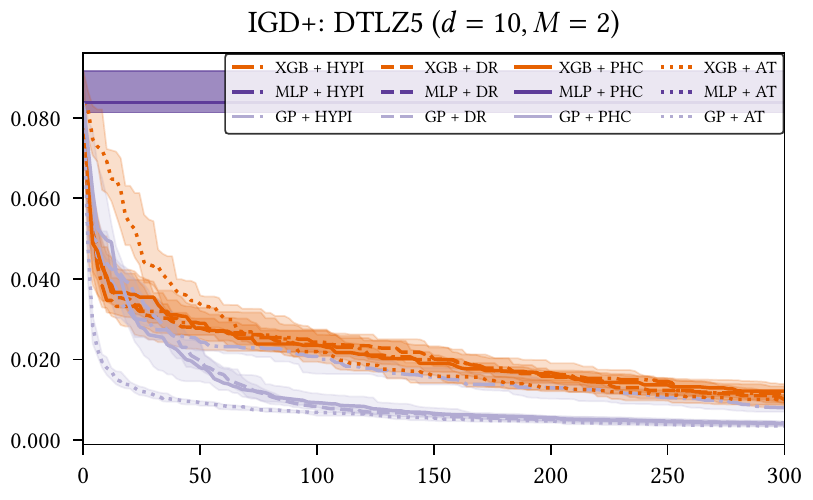}%
\includegraphics[width=0.25\linewidth]{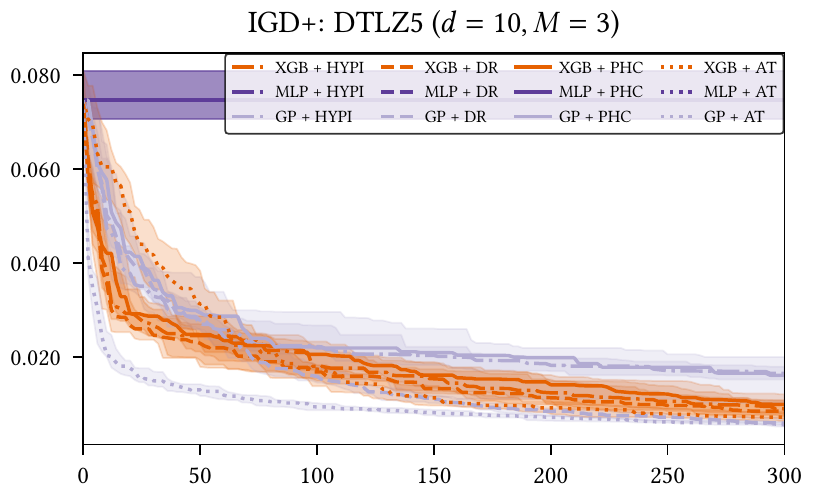}%
\includegraphics[width=0.25\linewidth]{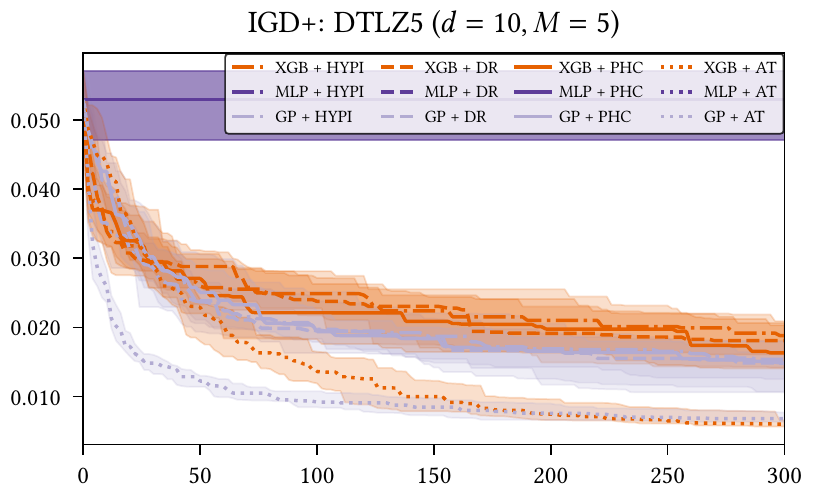}%
\includegraphics[width=0.25\linewidth]{figs/conv_igd+_DTLZ5_10_3}\\
\includegraphics[width=0.25\linewidth]{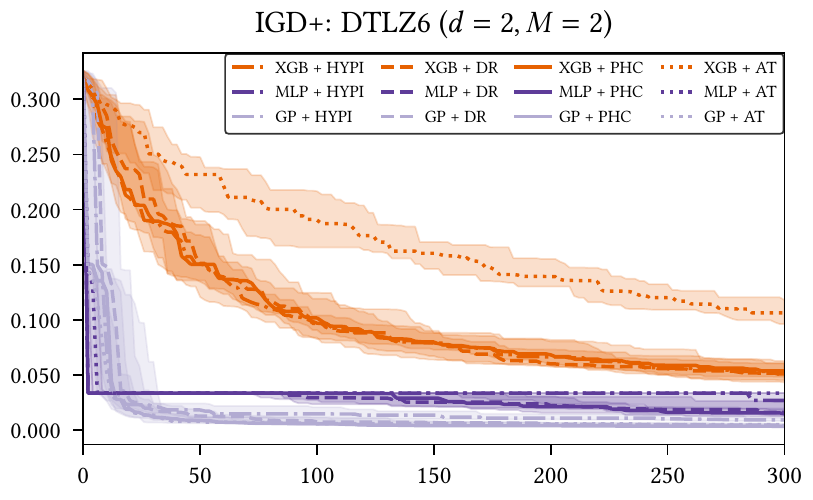}%
\includegraphics[width=0.25\linewidth]{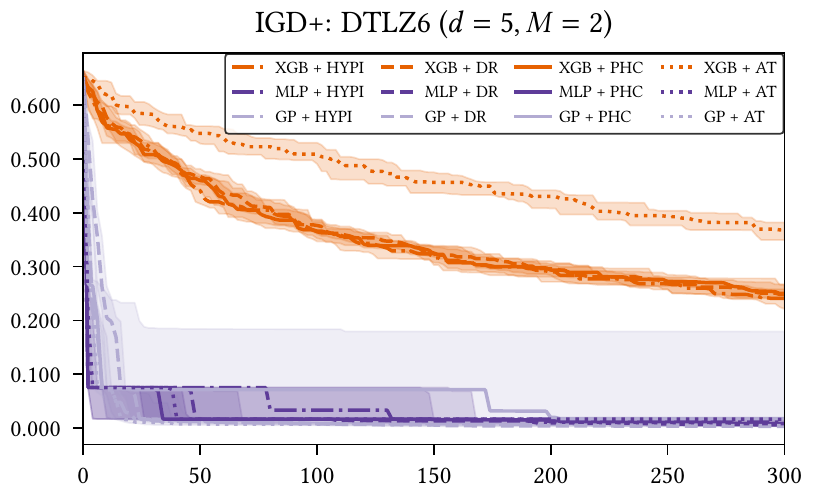}%
\includegraphics[width=0.25\linewidth]{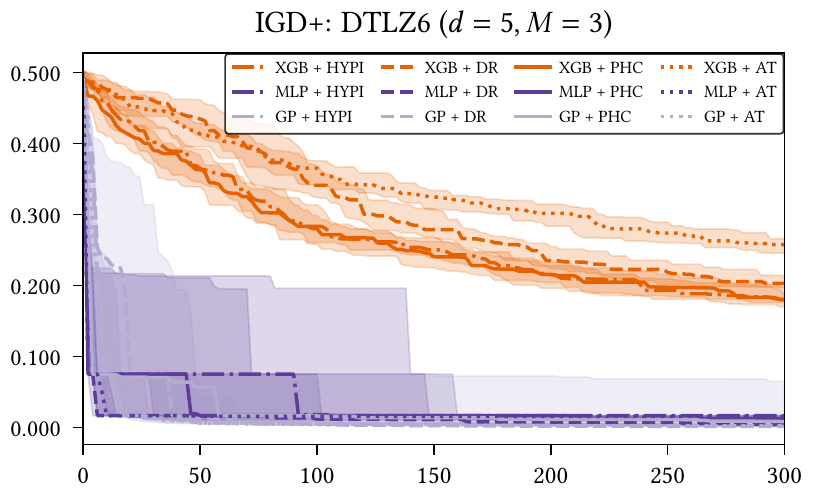}%
\includegraphics[width=0.25\linewidth]{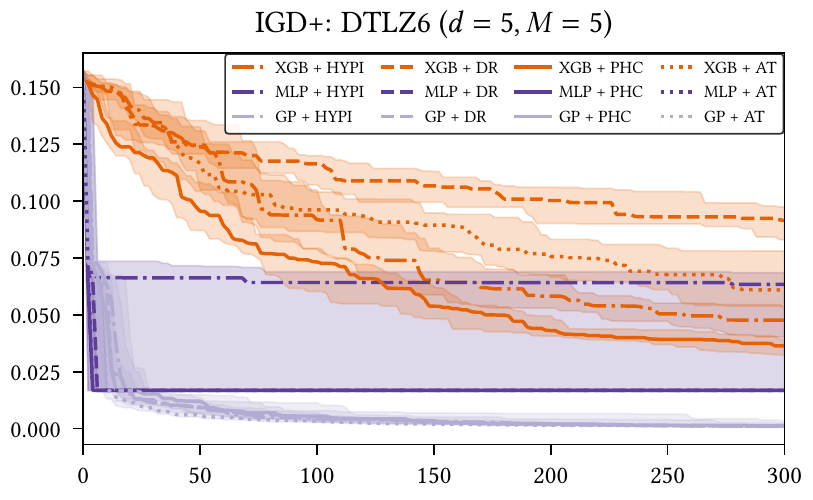}\\
\includegraphics[width=0.25\linewidth]{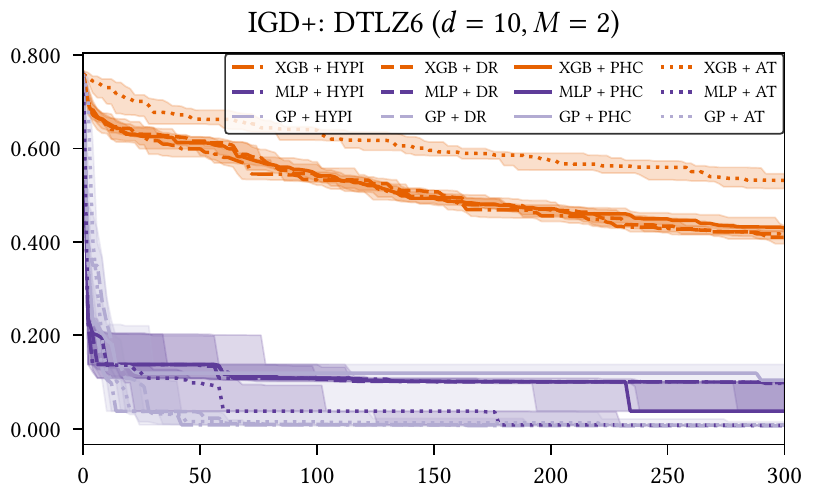}%
\includegraphics[width=0.25\linewidth]{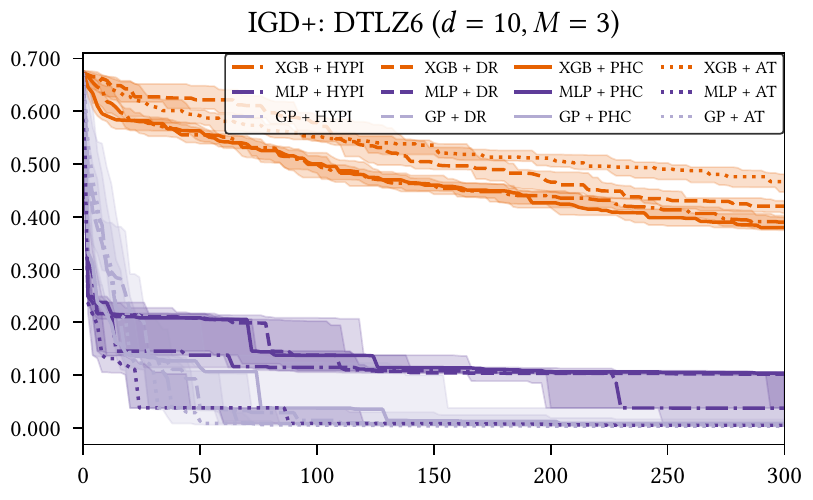}%
\includegraphics[width=0.25\linewidth]{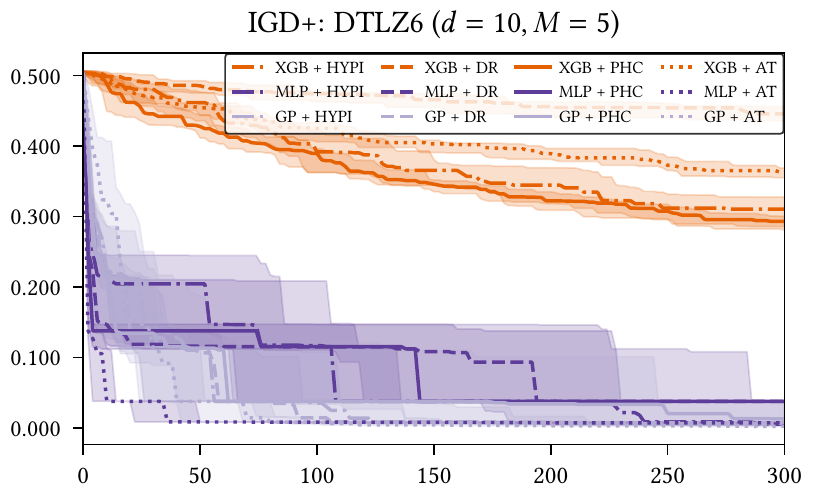}%
\includegraphics[width=0.25\linewidth]{figs/conv_igd+_DTLZ6_10_3}\\
\caption{%
    Hypervolume (\emph{upper}) and IGD+ (\emph{lower})
    convergence plots for DTLZ5 and DTLZ6.}
\label{fig:conv_DTLZ56}
\end{figure}

\begin{figure}[H]
\includegraphics[width=0.25\linewidth]{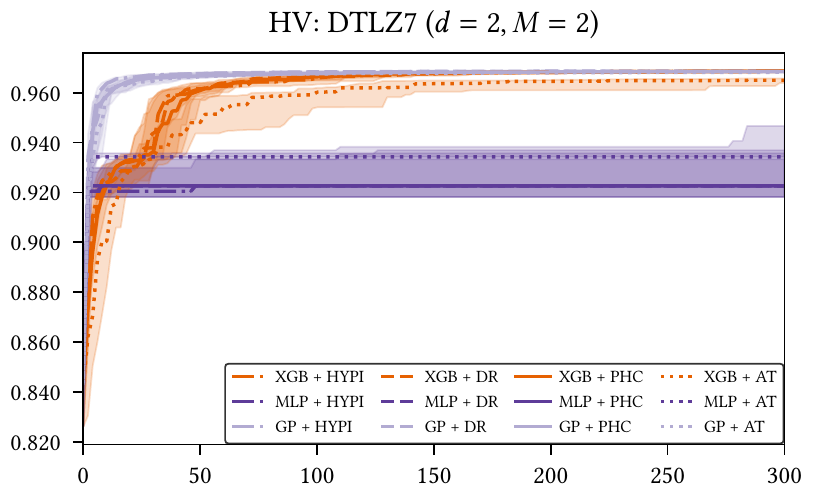}%
\includegraphics[width=0.25\linewidth]{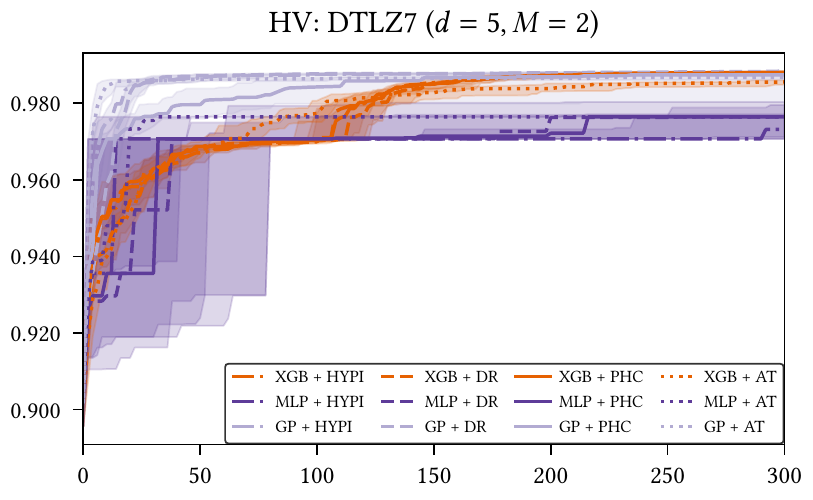}%
\includegraphics[width=0.25\linewidth]{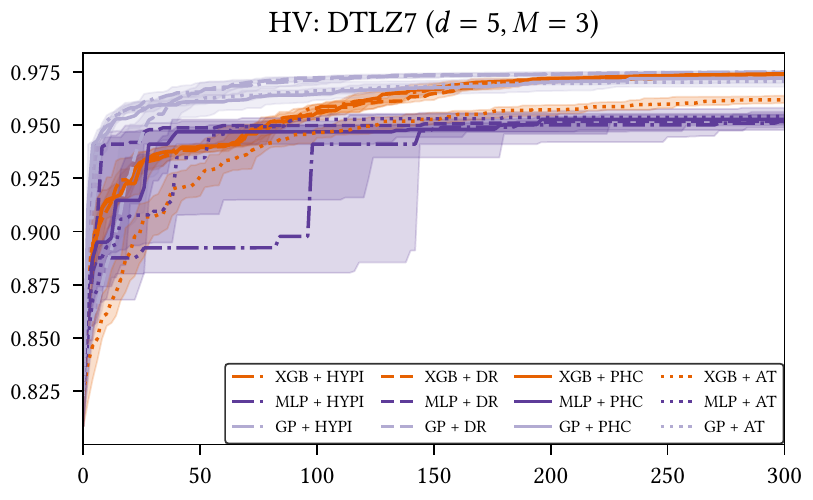}%
\includegraphics[width=0.25\linewidth]{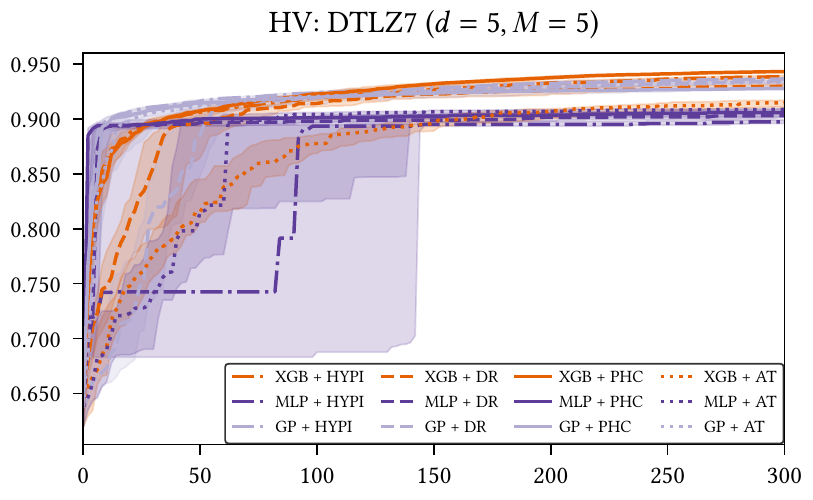}\\
\includegraphics[width=0.25\linewidth]{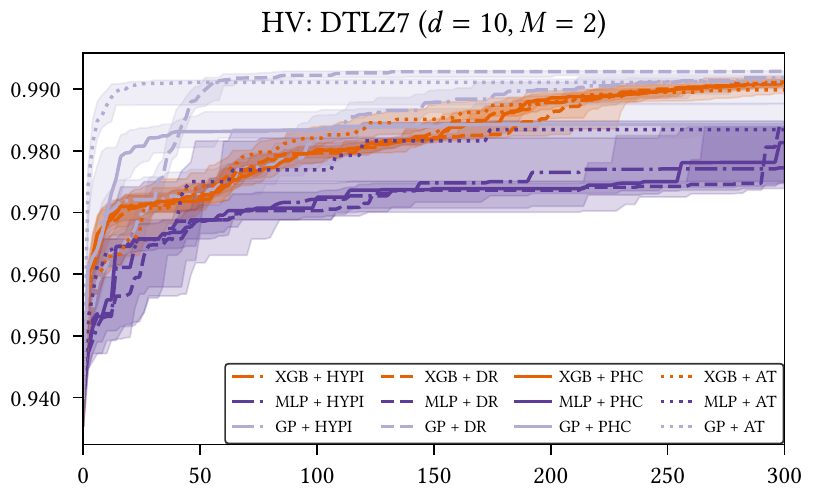}%
\includegraphics[width=0.25\linewidth]{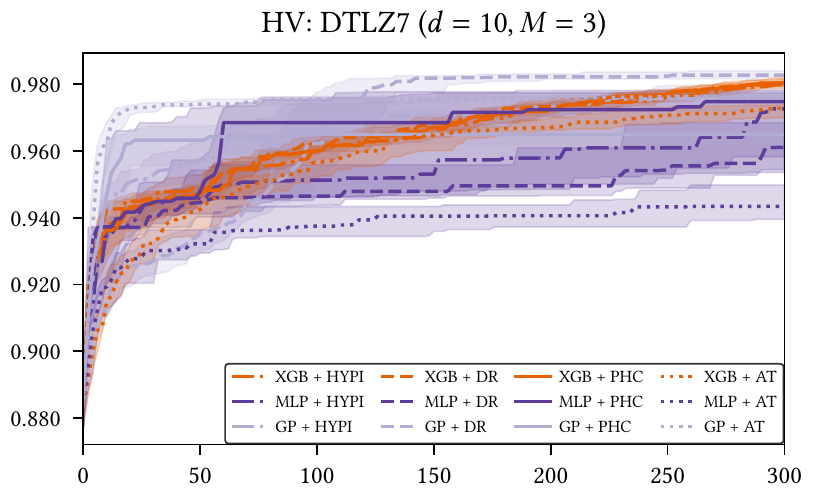}%
\includegraphics[width=0.25\linewidth]{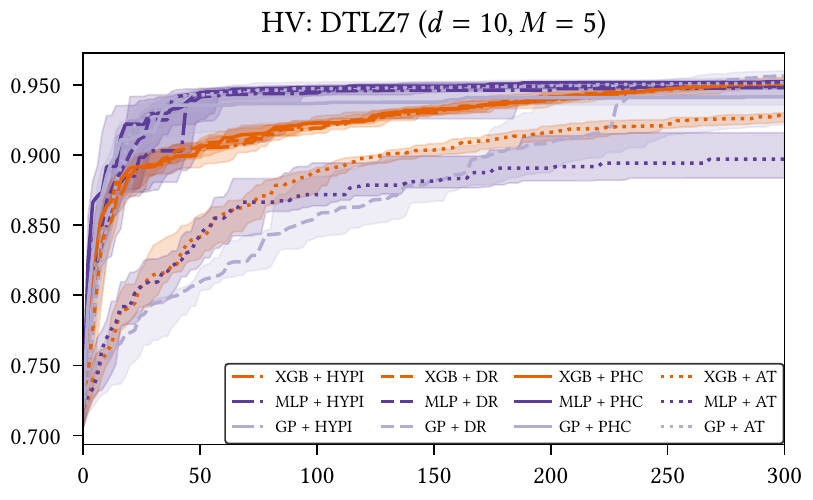}%
\includegraphics[width=0.25\linewidth]{figs/conv_hv_DTLZ7_10_3}\\
\includegraphics[width=0.25\linewidth]{figs/conv_hv_DTLZ6_2_2}%
\includegraphics[width=0.25\linewidth]{figs/conv_hv_DTLZ6_5_2}%
\includegraphics[width=0.25\linewidth]{figs/conv_hv_DTLZ6_5_3}%
\includegraphics[width=0.25\linewidth]{figs/conv_hv_DTLZ6_5_5}\\
\includegraphics[width=0.25\linewidth]{figs/conv_hv_DTLZ6_10_2}%
\includegraphics[width=0.25\linewidth]{figs/conv_hv_DTLZ6_10_3}%
\includegraphics[width=0.25\linewidth]{figs/conv_hv_DTLZ6_10_5}%
\includegraphics[width=0.25\linewidth]{figs/conv_hv_DTLZ6_10_3}\\
\caption{%
    Hypervolume (\emph{upper}) and IGD+ (\emph{lower})
    convergence plots for DTLZ7.}
\label{fig:conv_DTLZ7}
\end{figure}

\newpage
\subsection{WFG Convergence Plots}
\label{sec:conv:WFG}

\begin{figure}[H]
\includegraphics[width=0.25\linewidth]{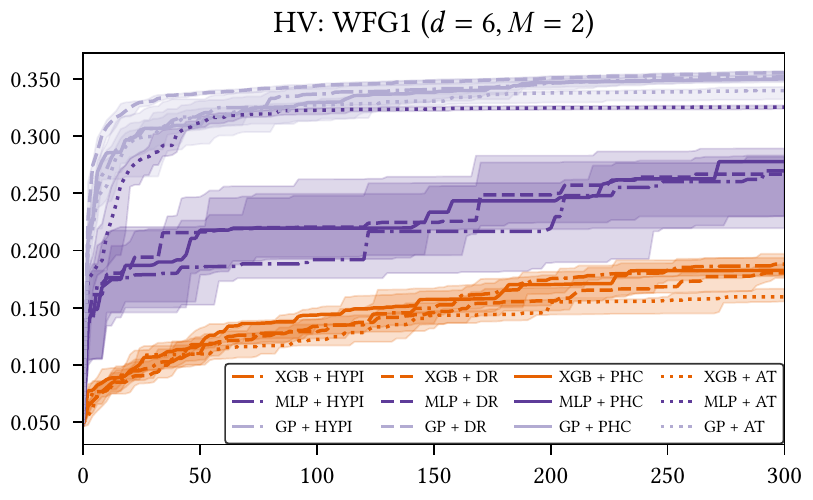}%
\includegraphics[width=0.25\linewidth]{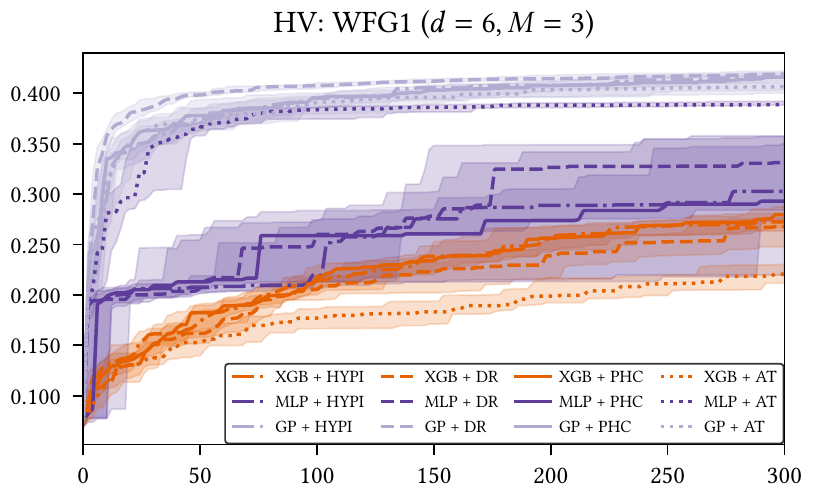}%
\includegraphics[width=0.25\linewidth]{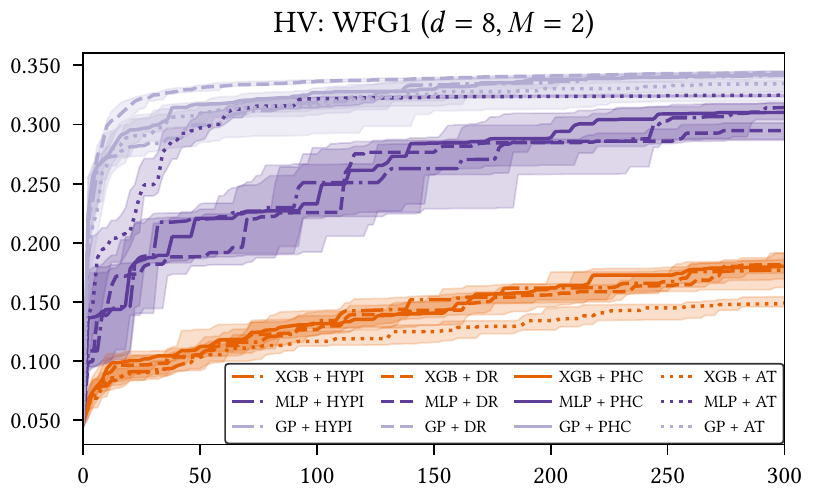}%
\includegraphics[width=0.25\linewidth]{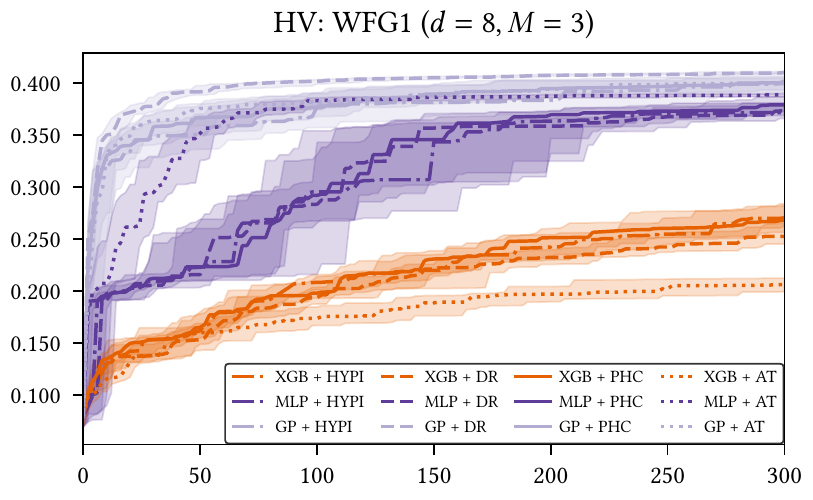}\\
\includegraphics[width=0.25\linewidth]{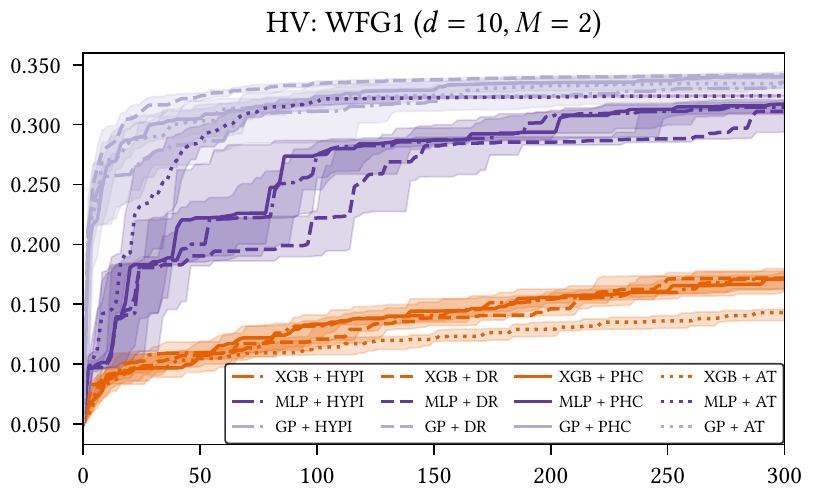}%
\includegraphics[width=0.25\linewidth]{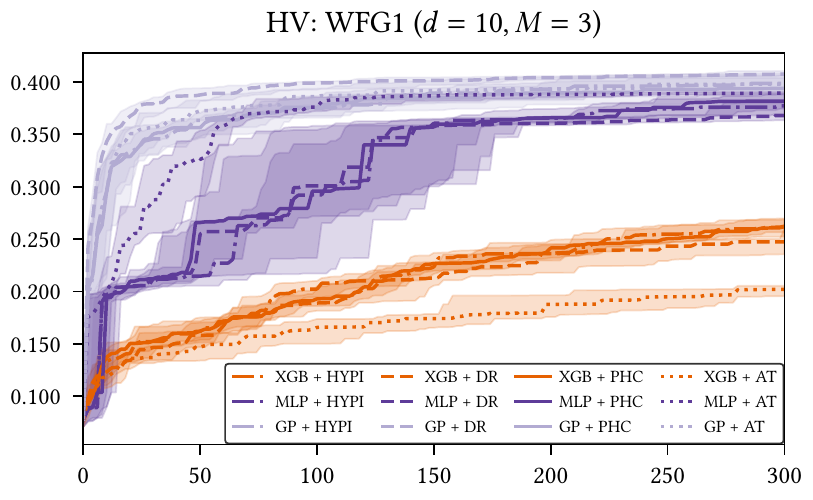}%
\includegraphics[width=0.25\linewidth]{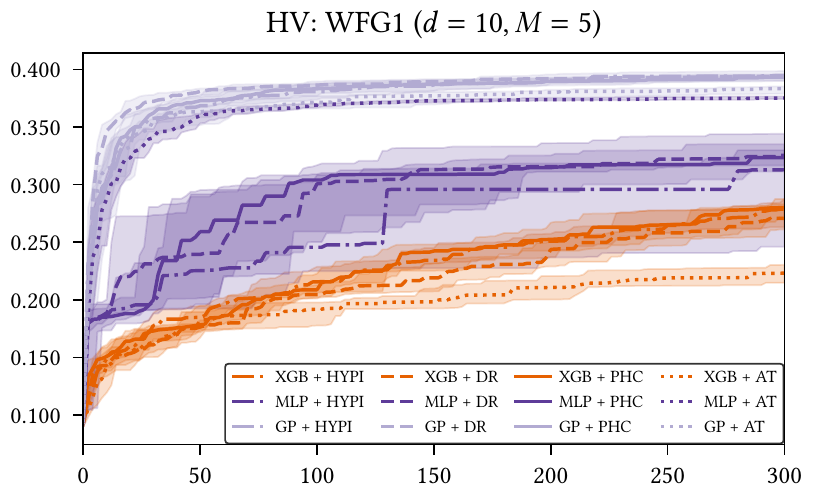}\\
\includegraphics[width=0.25\linewidth]{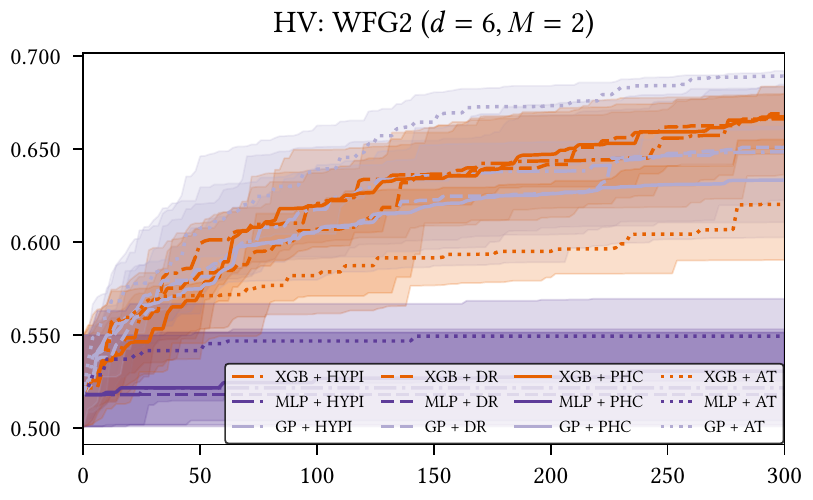}%
\includegraphics[width=0.25\linewidth]{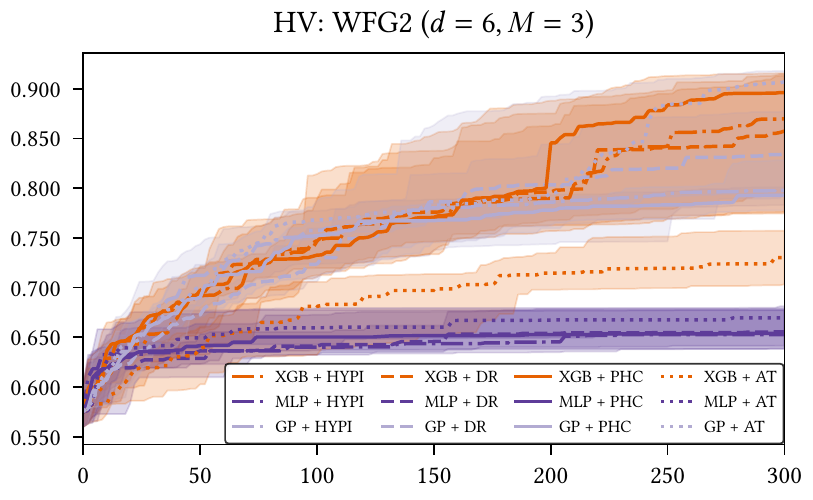}%
\includegraphics[width=0.25\linewidth]{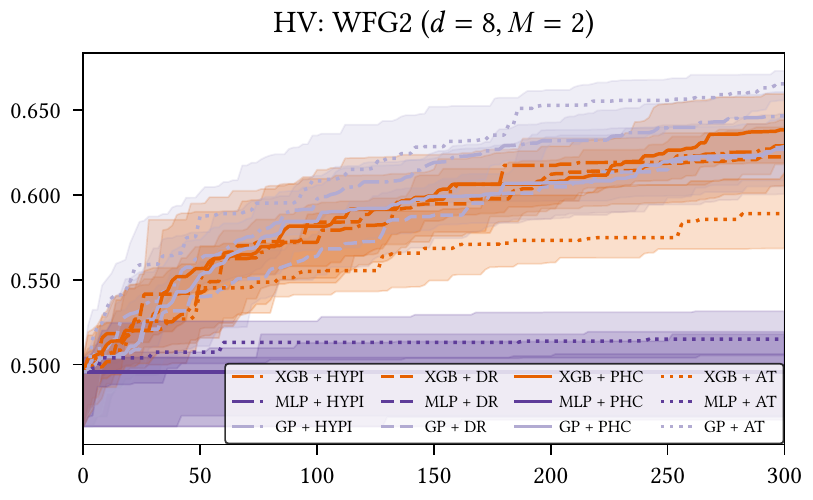}%
\includegraphics[width=0.25\linewidth]{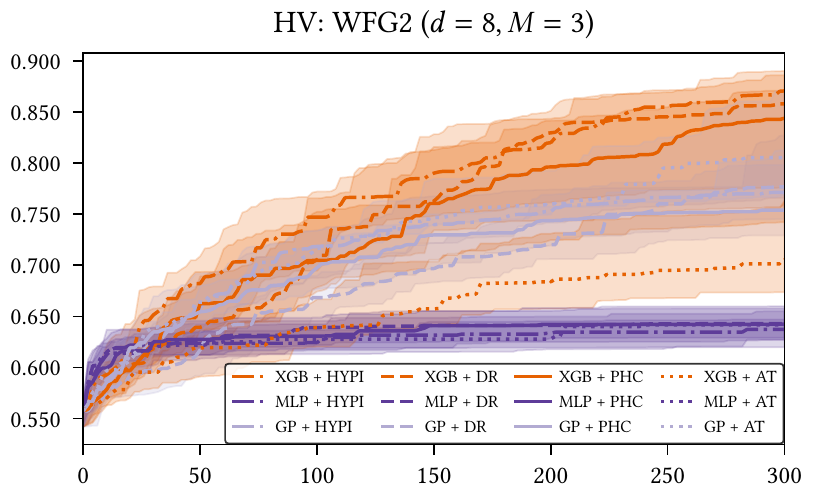}\\
\includegraphics[width=0.25\linewidth]{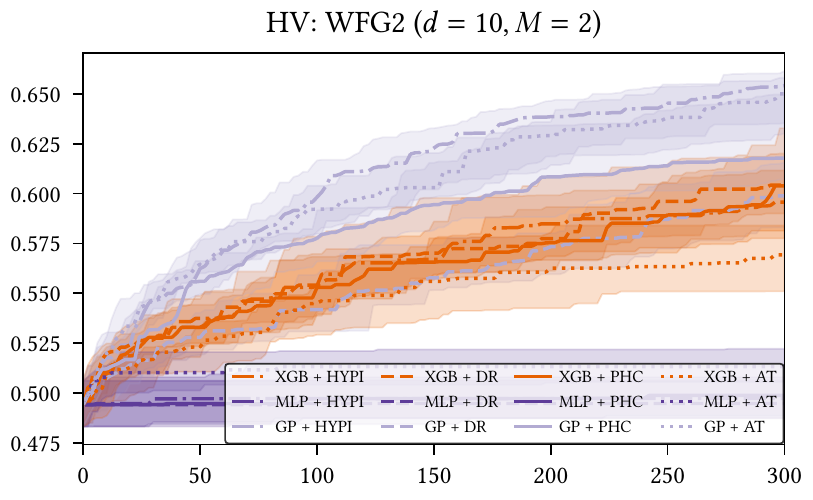}%
\includegraphics[width=0.25\linewidth]{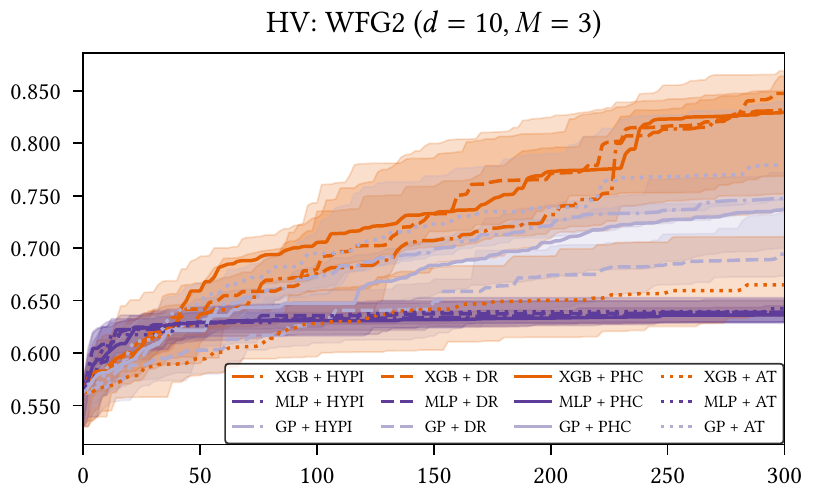}%
\includegraphics[width=0.25\linewidth]{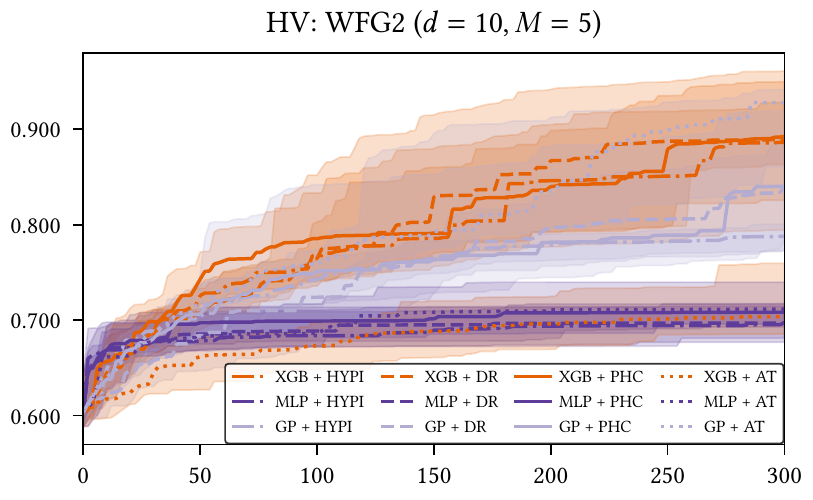}\\
\rule{\linewidth}{0.4pt}
\includegraphics[width=0.25\linewidth]{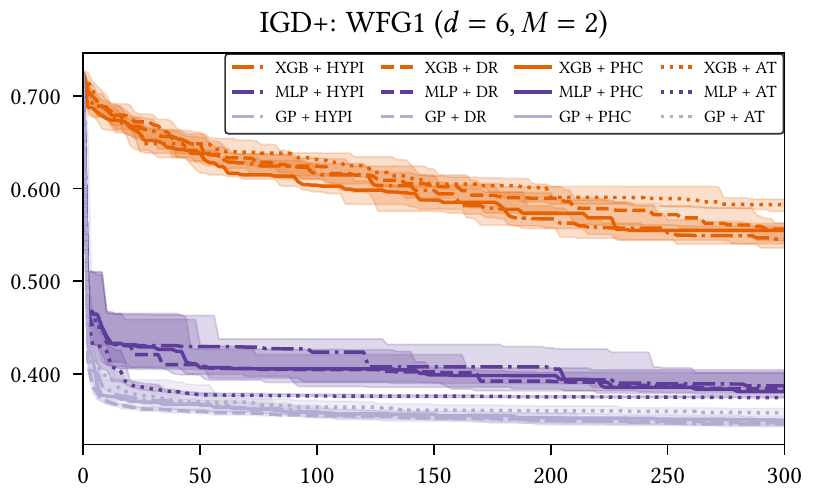}%
\includegraphics[width=0.25\linewidth]{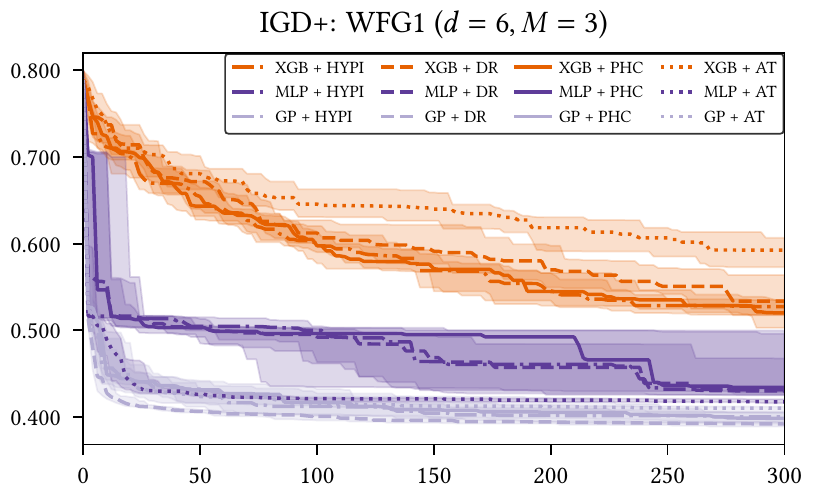}%
\includegraphics[width=0.25\linewidth]{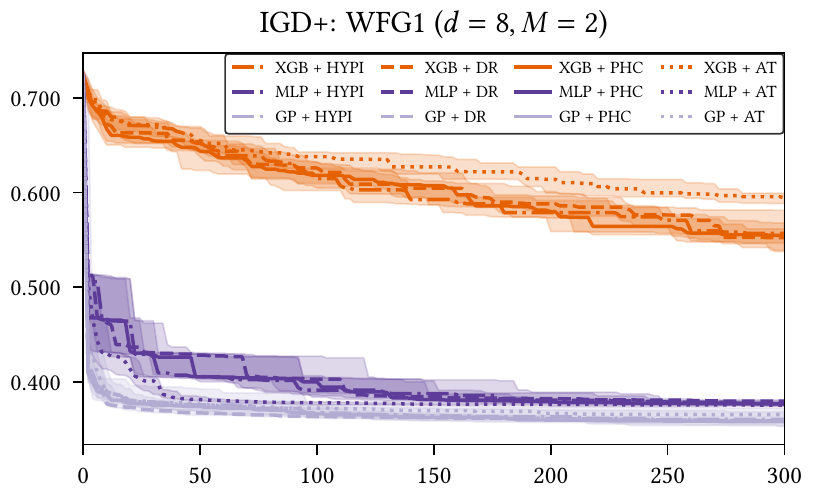}%
\includegraphics[width=0.25\linewidth]{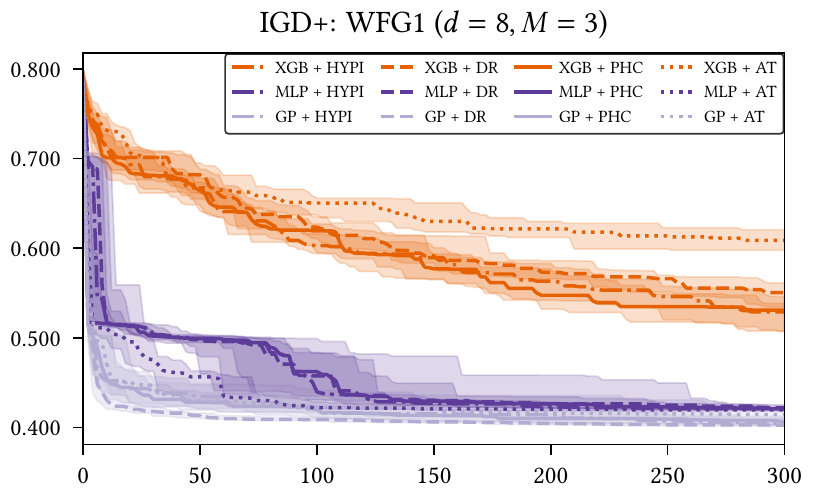}\\
\includegraphics[width=0.25\linewidth]{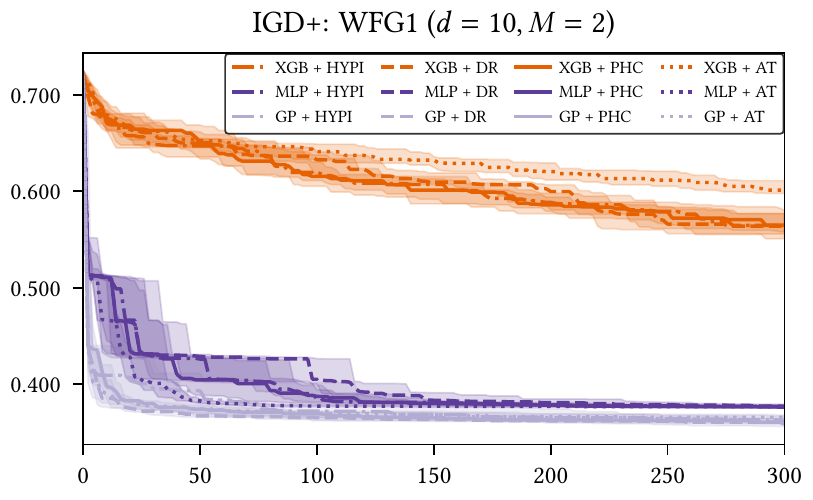}%
\includegraphics[width=0.25\linewidth]{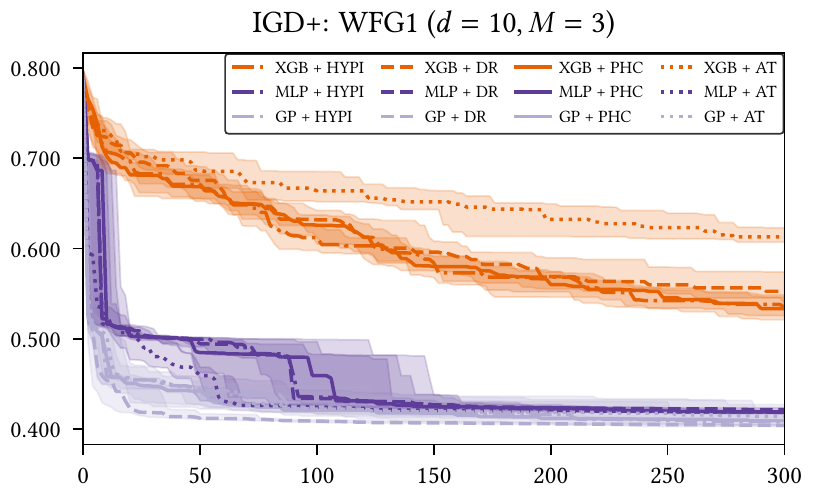}%
\includegraphics[width=0.25\linewidth]{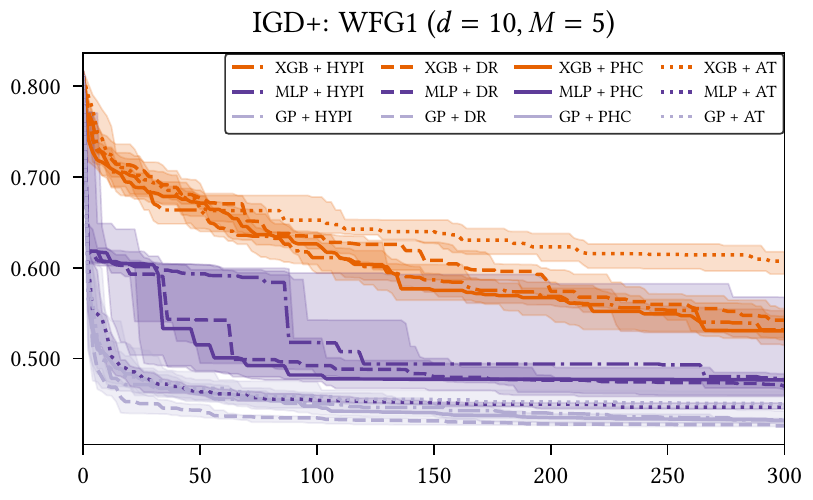}\\
\includegraphics[width=0.25\linewidth]{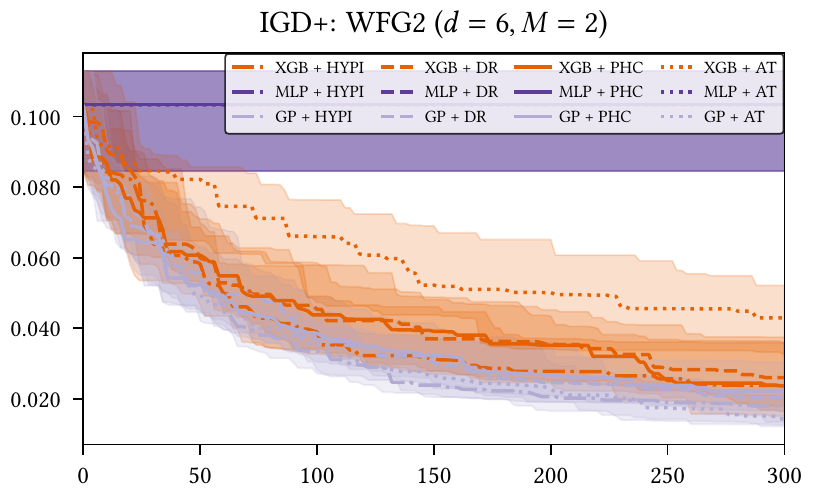}%
\includegraphics[width=0.25\linewidth]{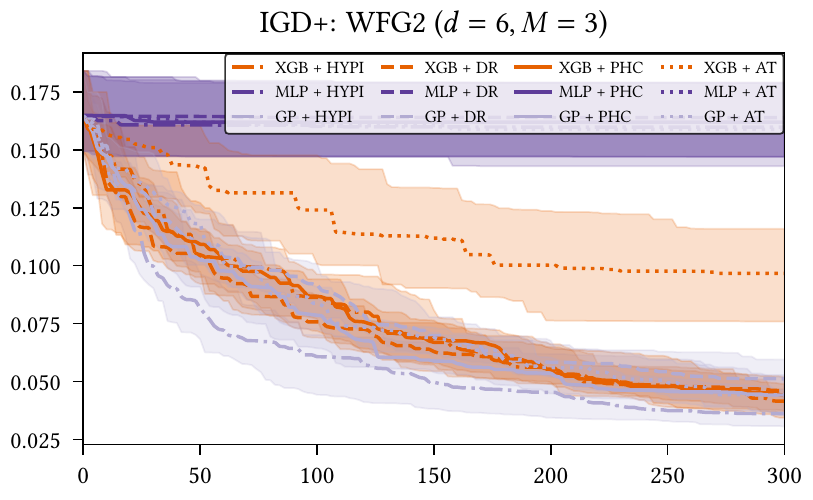}%
\includegraphics[width=0.25\linewidth]{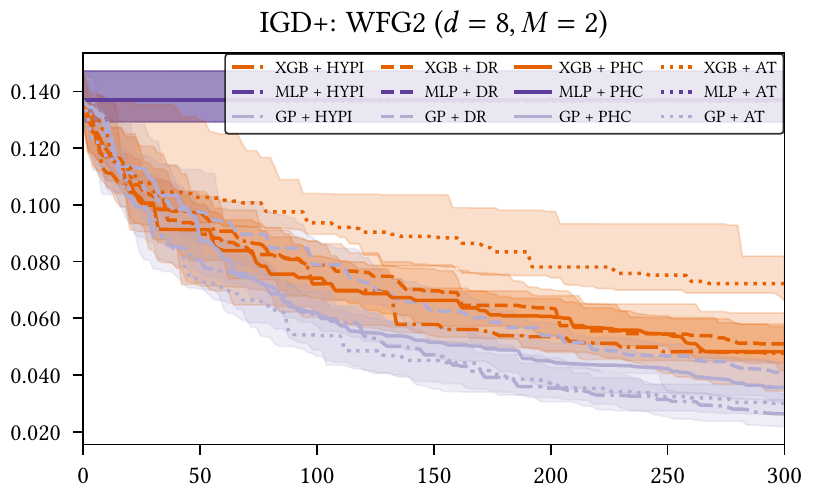}%
\includegraphics[width=0.25\linewidth]{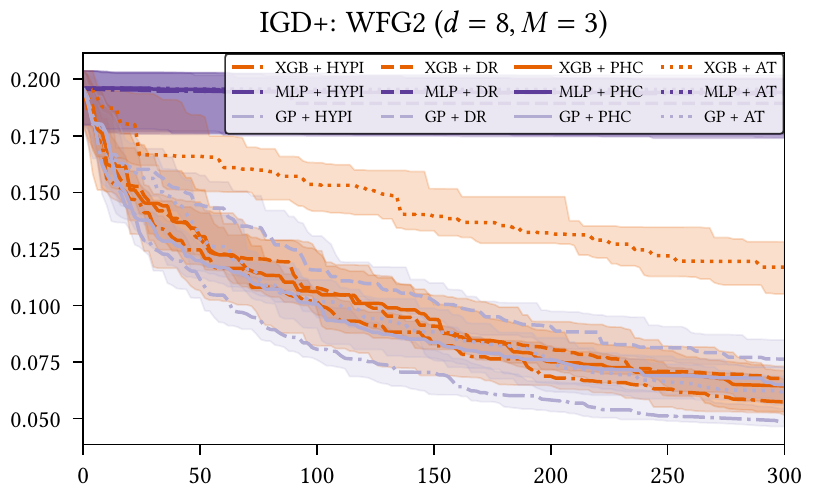}\\
\includegraphics[width=0.25\linewidth]{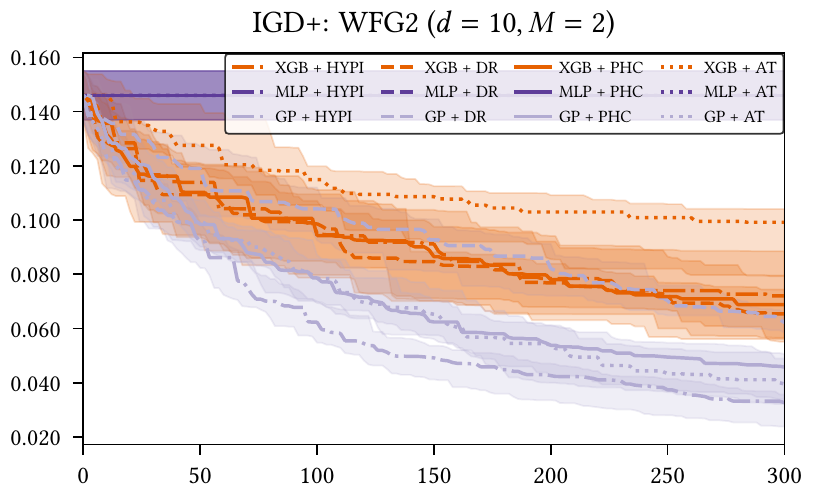}%
\includegraphics[width=0.25\linewidth]{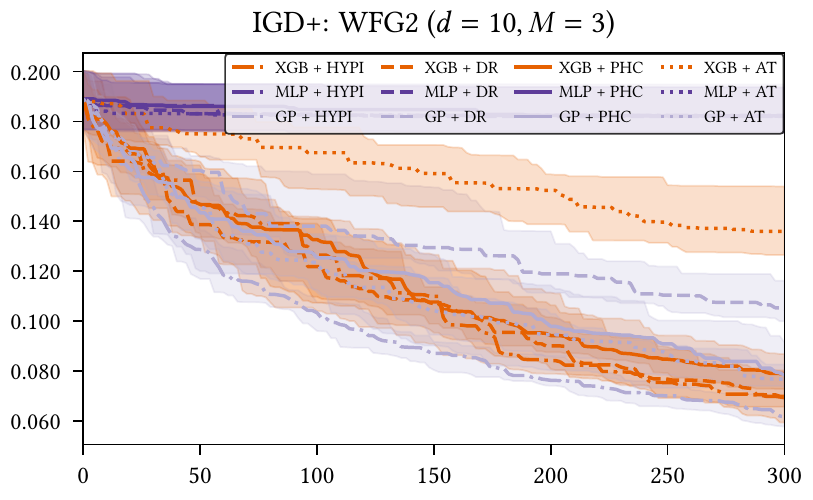}%
\includegraphics[width=0.25\linewidth]{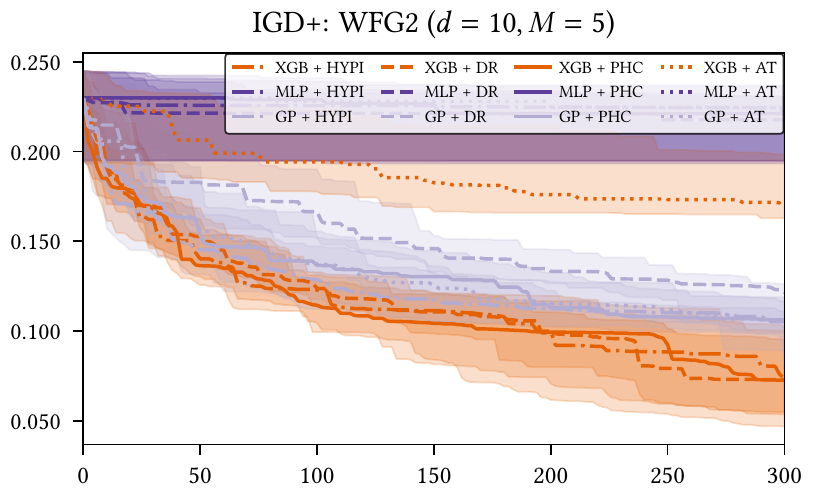}\\
\caption{%
    Hypervolume (\emph{upper}) and IGD+ (\emph{lower})
    convergence plots for WFG1 and WFG2.}
\label{fig:conv_WFG12}
\end{figure}

\begin{figure}[H]
\includegraphics[width=0.25\linewidth]{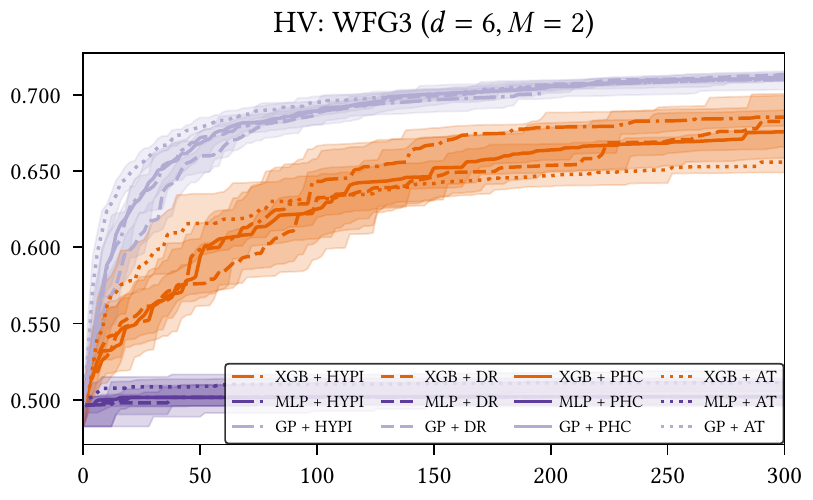}%
\includegraphics[width=0.25\linewidth]{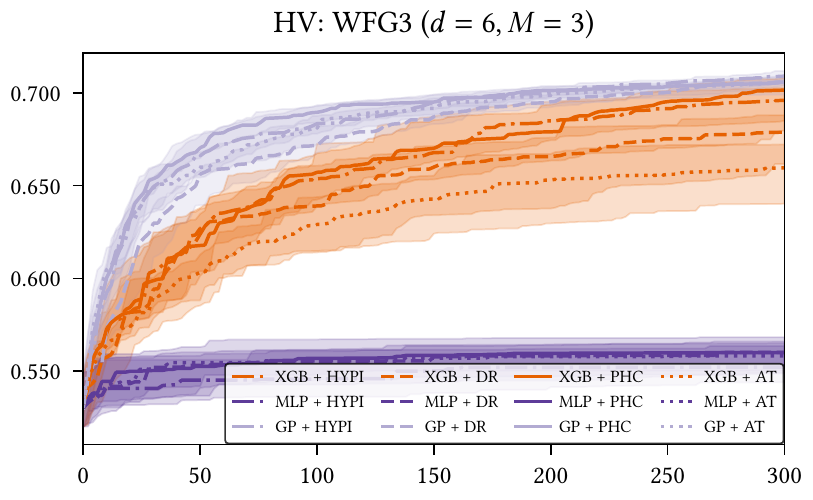}%
\includegraphics[width=0.25\linewidth]{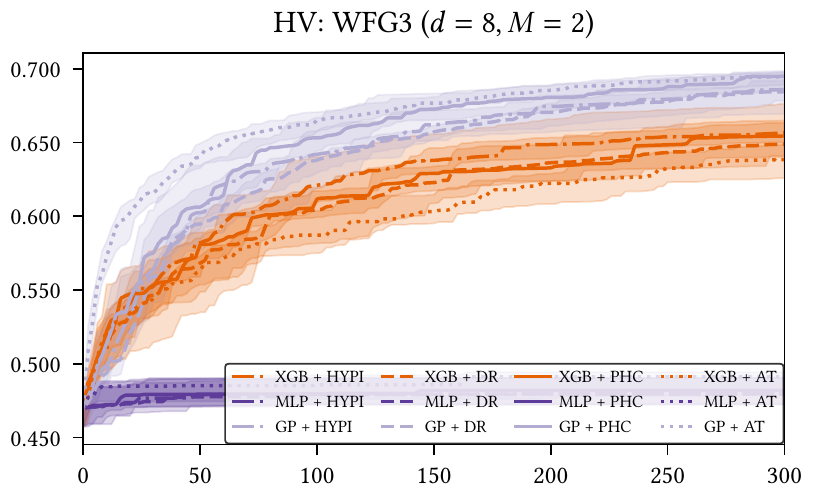}%
\includegraphics[width=0.25\linewidth]{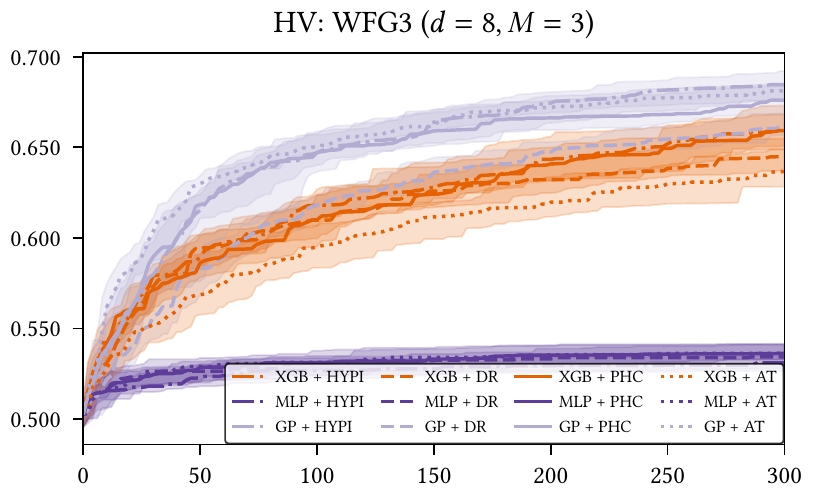}\\
\includegraphics[width=0.25\linewidth]{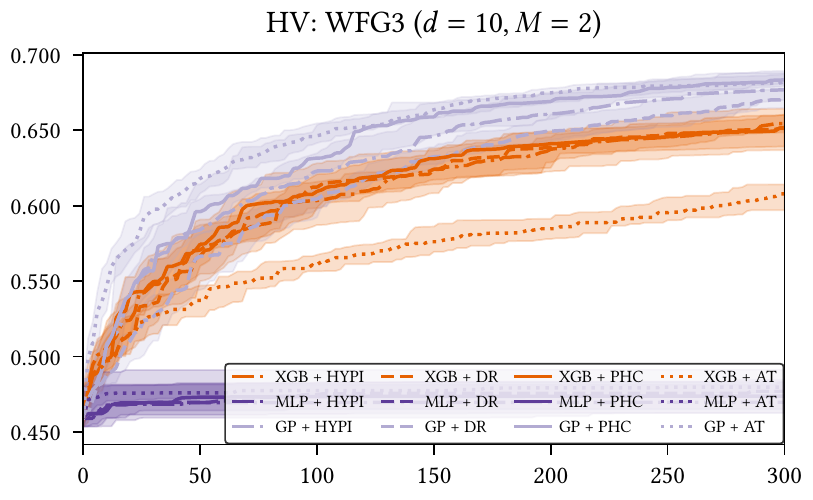}%
\includegraphics[width=0.25\linewidth]{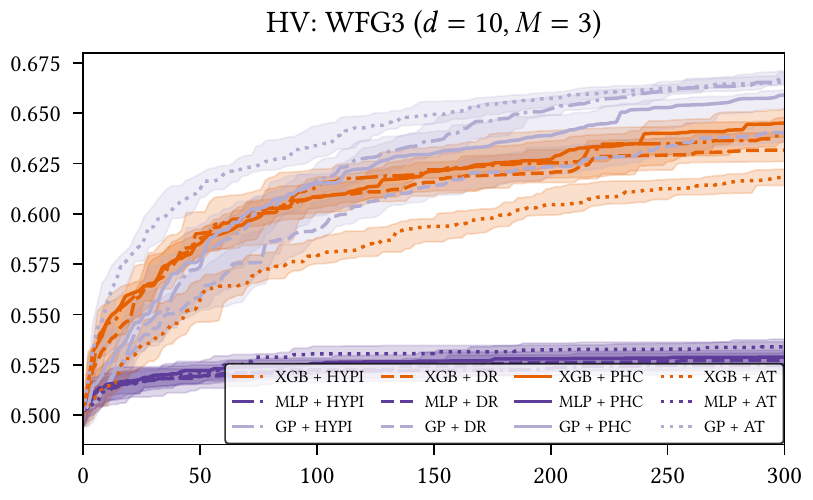}%
\includegraphics[width=0.25\linewidth]{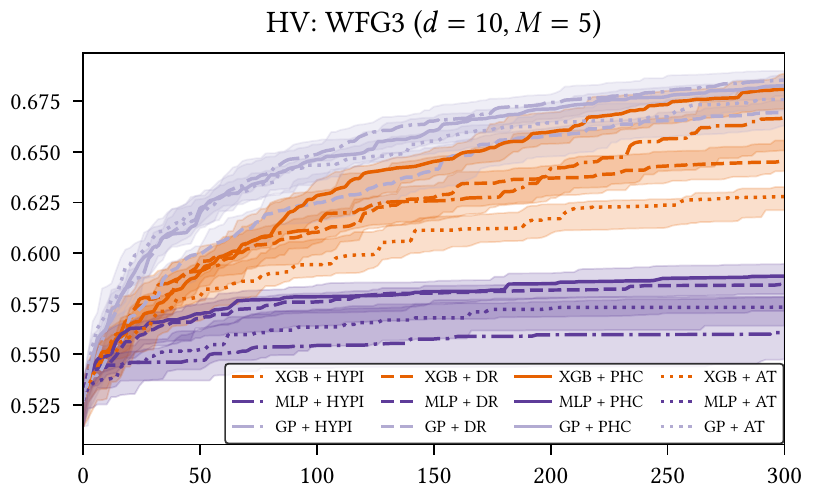}\\
\includegraphics[width=0.25\linewidth]{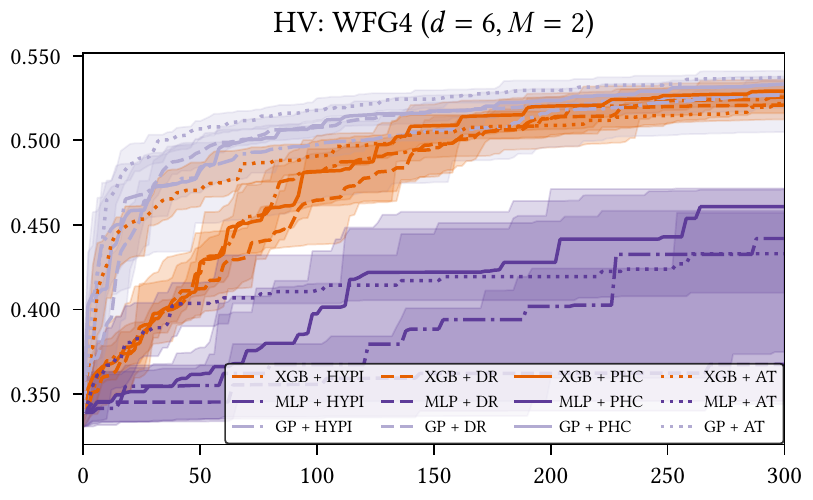}%
\includegraphics[width=0.25\linewidth]{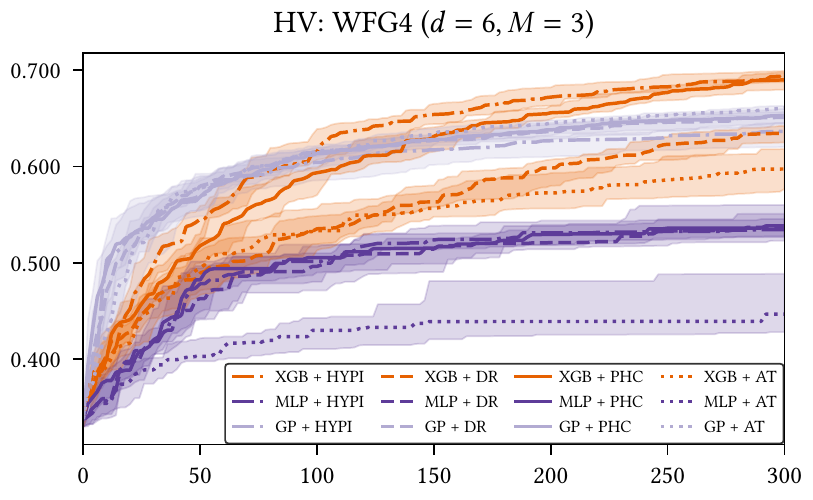}%
\includegraphics[width=0.25\linewidth]{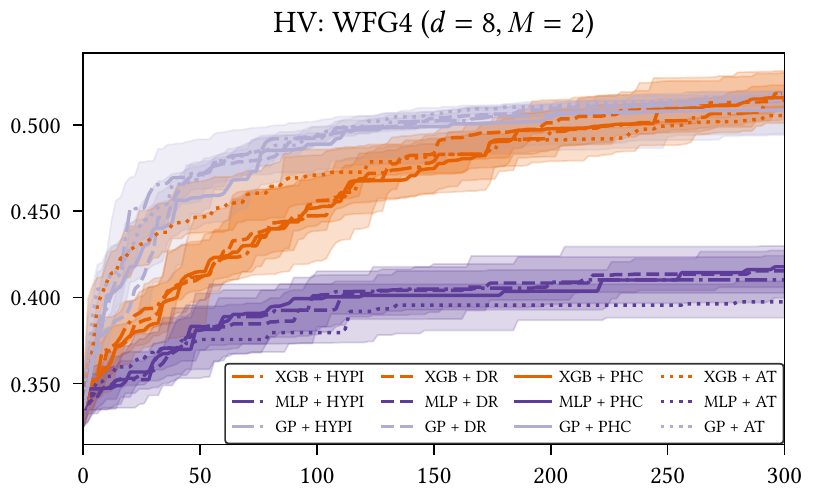}%
\includegraphics[width=0.25\linewidth]{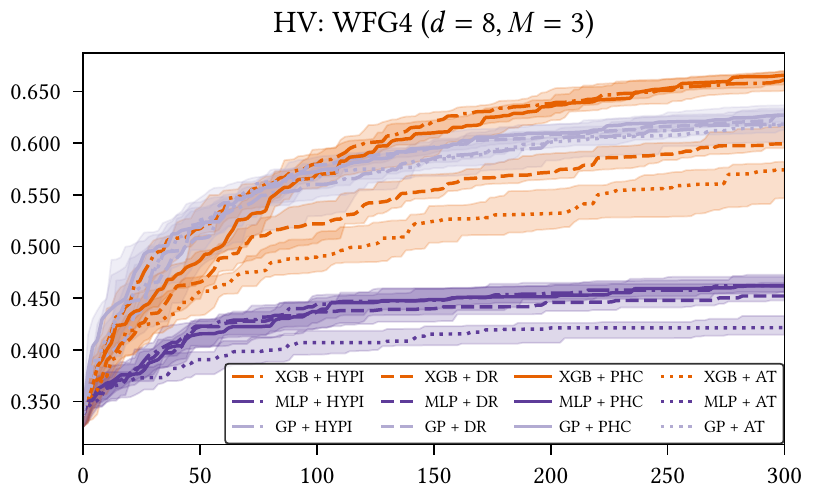}\\
\includegraphics[width=0.25\linewidth]{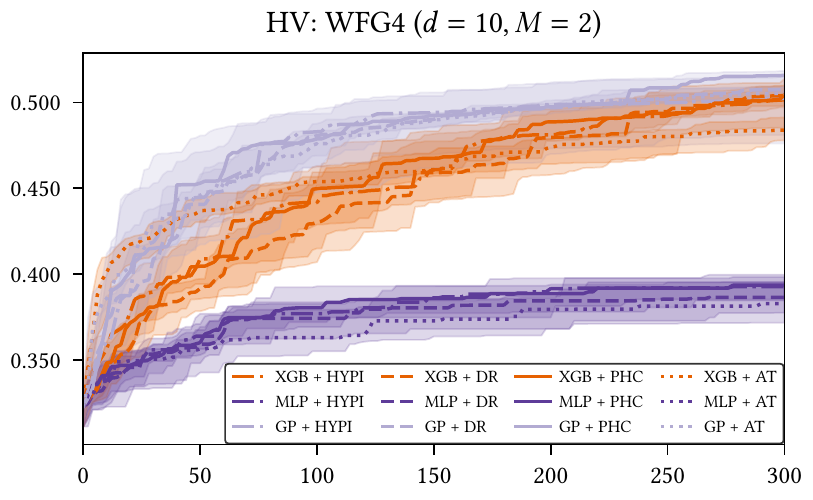}%
\includegraphics[width=0.25\linewidth]{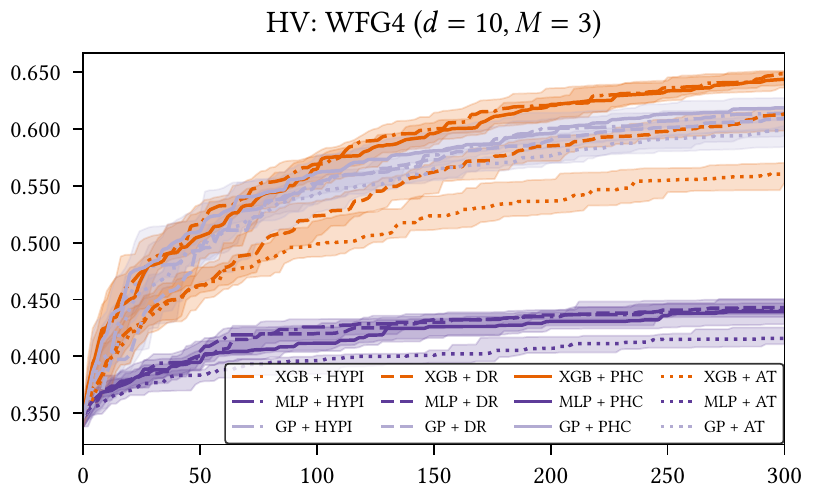}%
\includegraphics[width=0.25\linewidth]{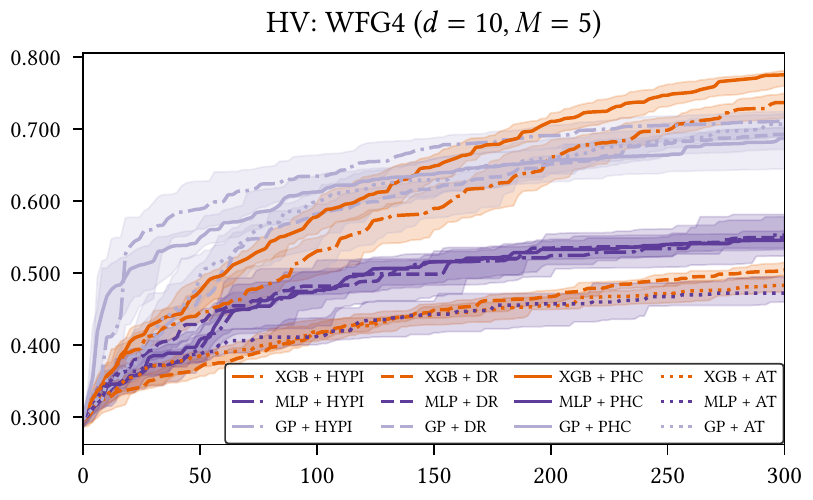}\\
\rule{\linewidth}{0.4pt}
\includegraphics[width=0.25\linewidth]{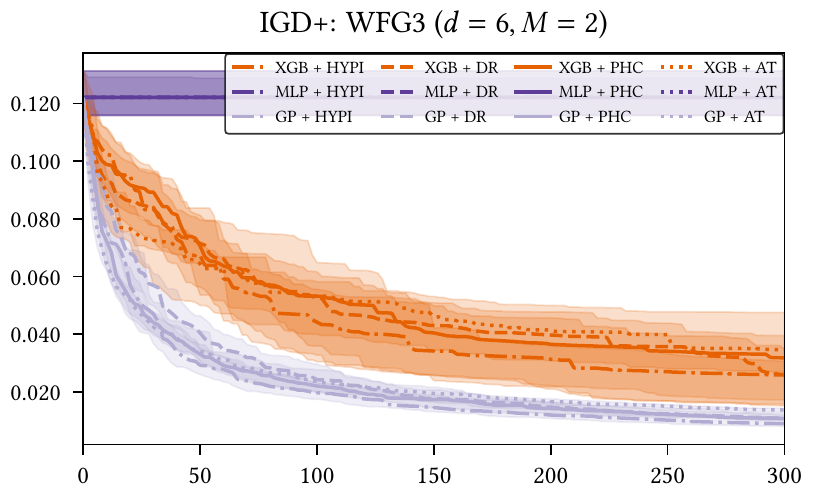}%
\includegraphics[width=0.25\linewidth]{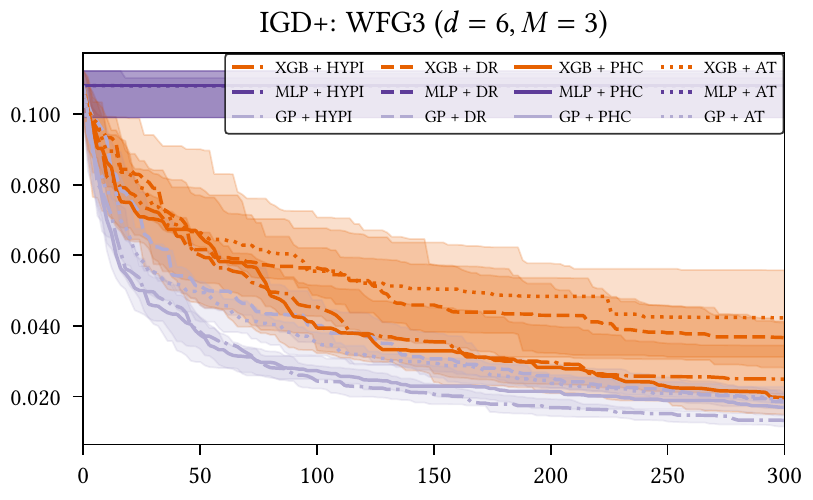}%
\includegraphics[width=0.25\linewidth]{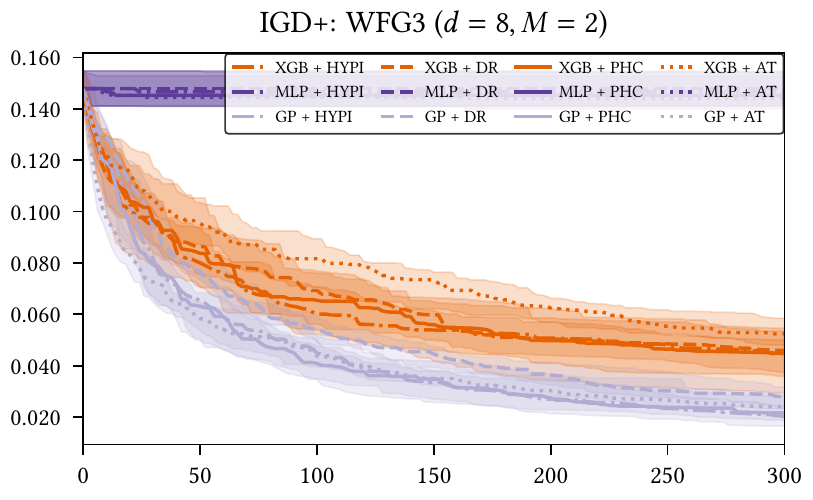}%
\includegraphics[width=0.25\linewidth]{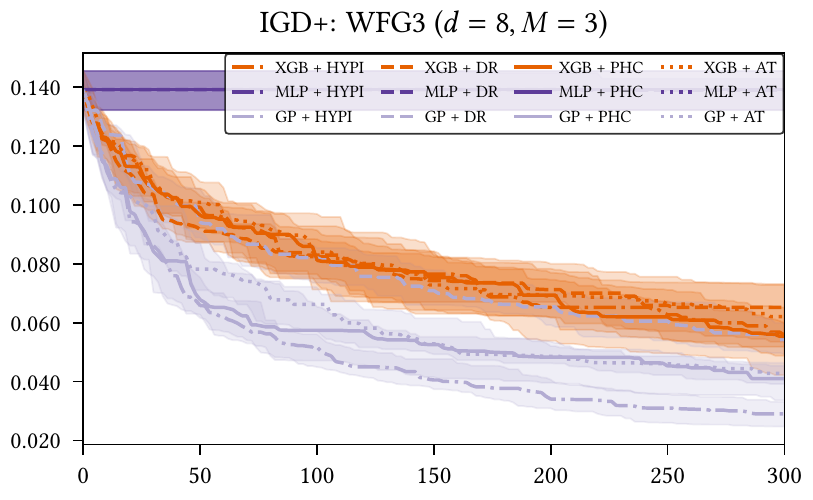}\\
\includegraphics[width=0.25\linewidth]{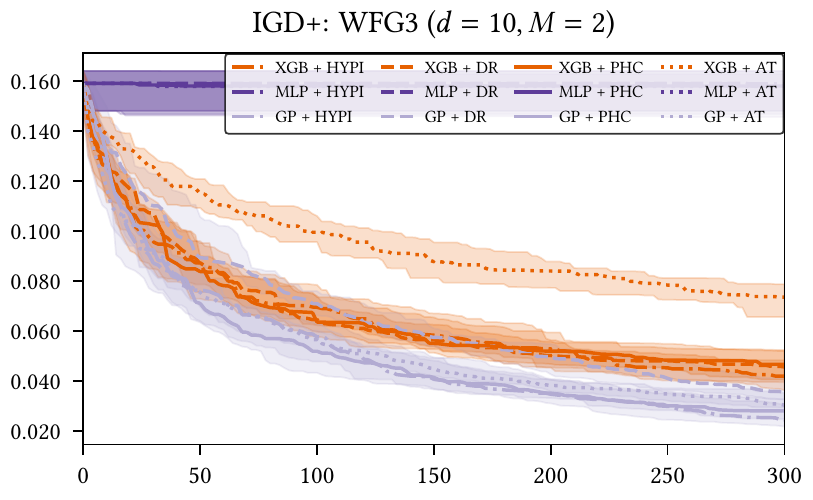}%
\includegraphics[width=0.25\linewidth]{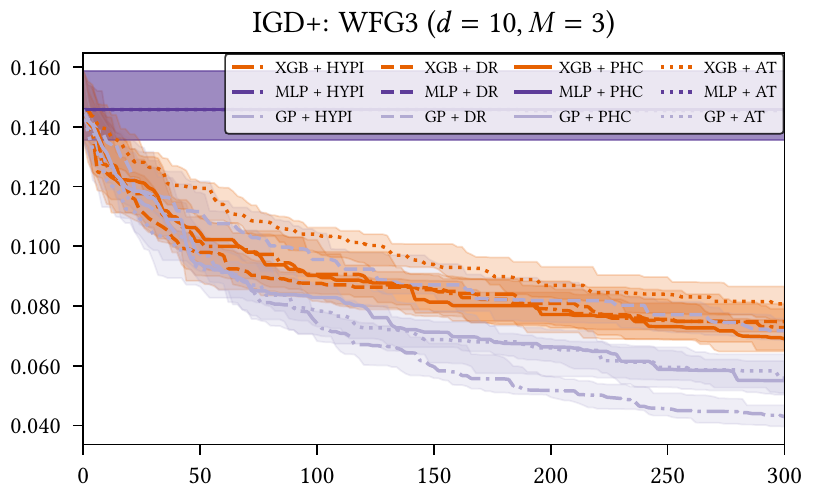}%
\includegraphics[width=0.25\linewidth]{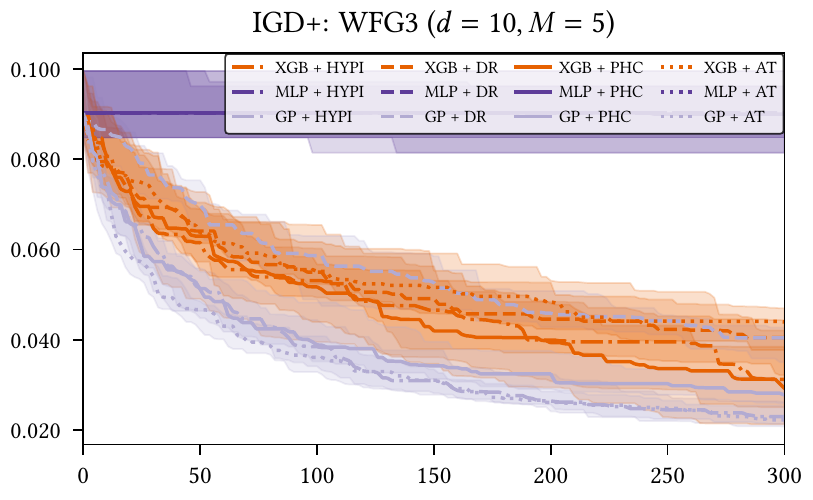}\\
\includegraphics[width=0.25\linewidth]{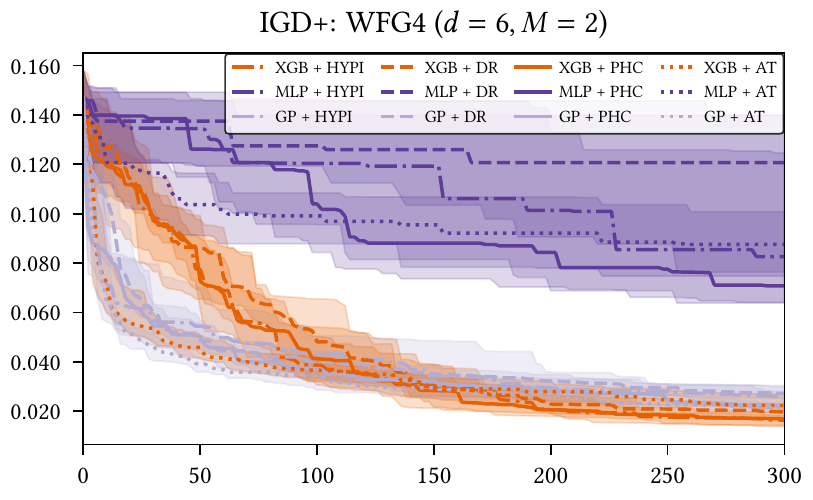}%
\includegraphics[width=0.25\linewidth]{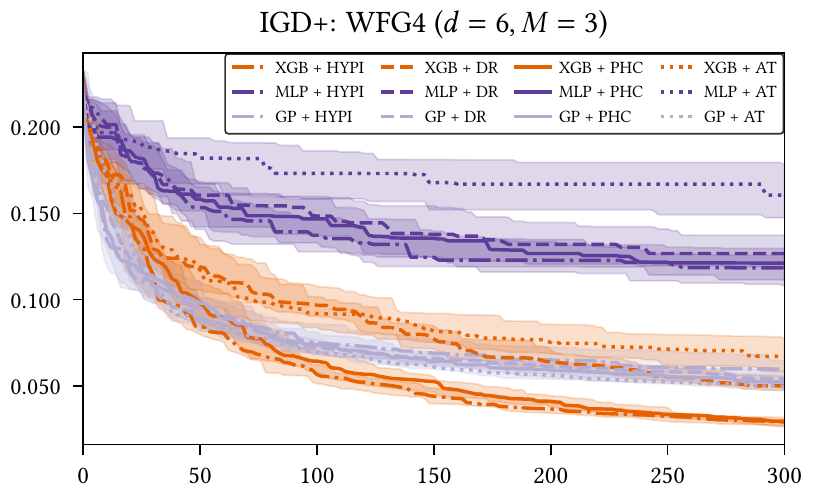}%
\includegraphics[width=0.25\linewidth]{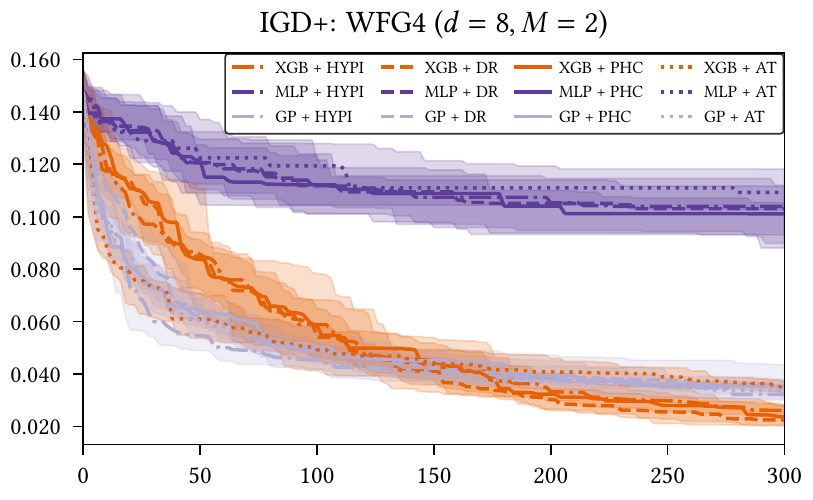}%
\includegraphics[width=0.25\linewidth]{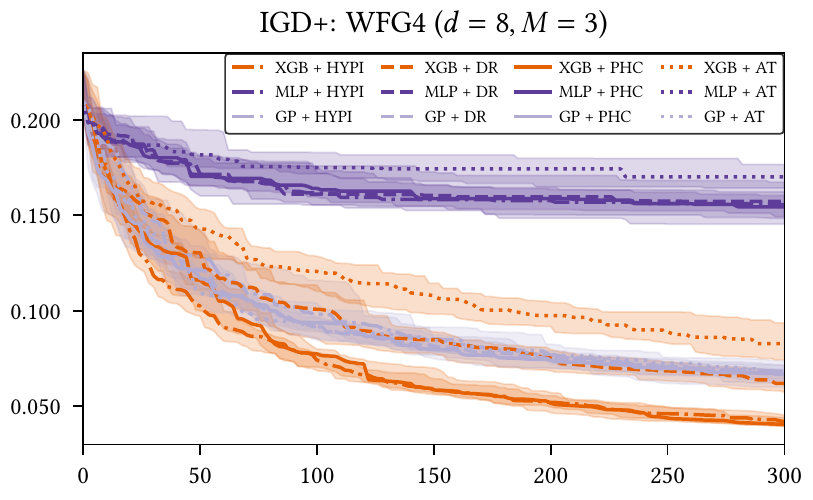}\\
\includegraphics[width=0.25\linewidth]{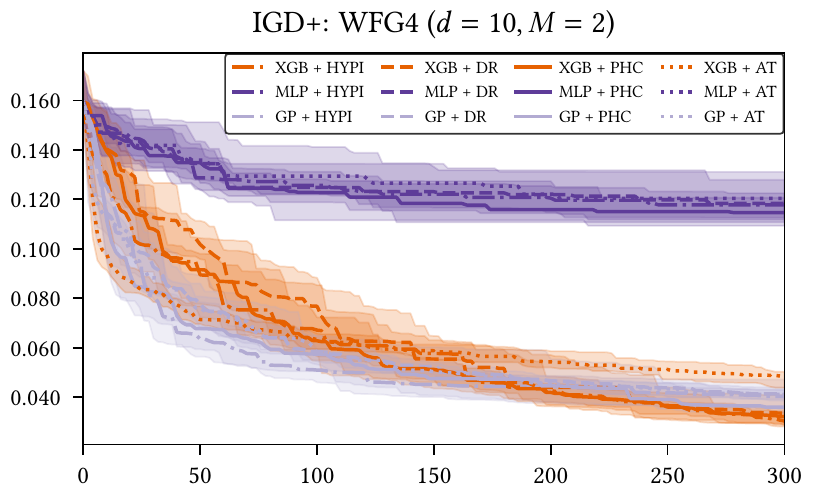}%
\includegraphics[width=0.25\linewidth]{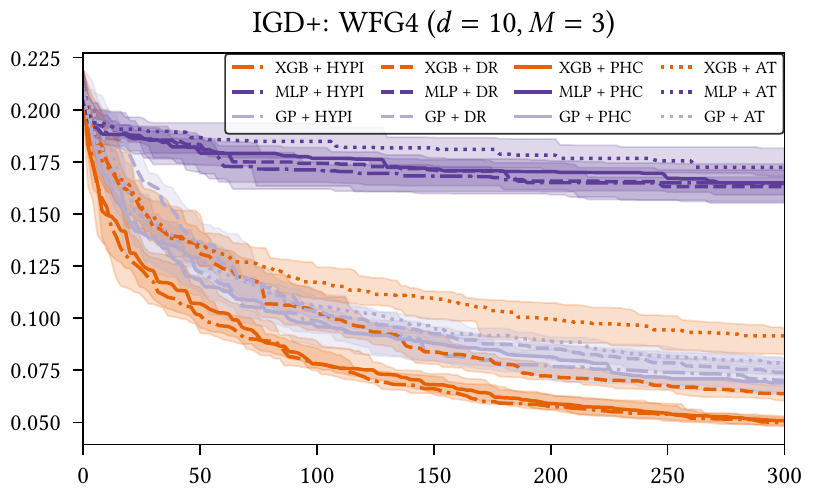}%
\includegraphics[width=0.25\linewidth]{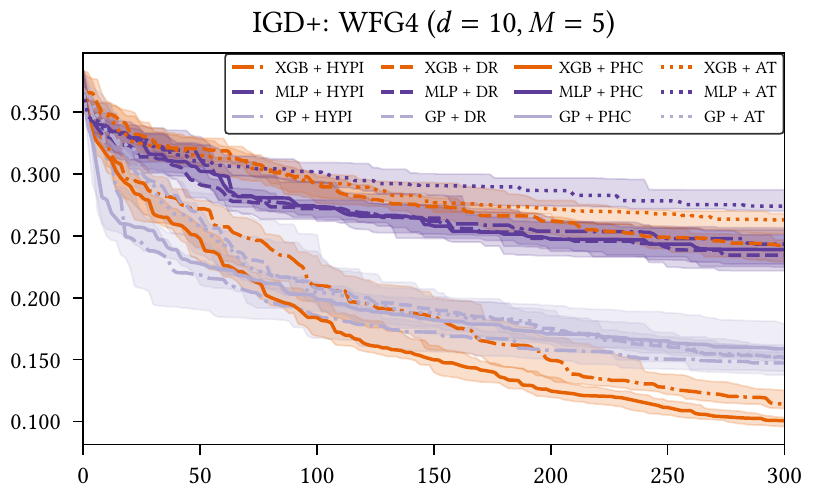}\\
\caption{%
    Hypervolume (\emph{upper}) and IGD+ (\emph{lower})
    convergence plots for WFG3 and WFG4.}
\label{fig:conv_WFG34}
\end{figure}

\begin{figure}[H]
\includegraphics[width=0.25\linewidth]{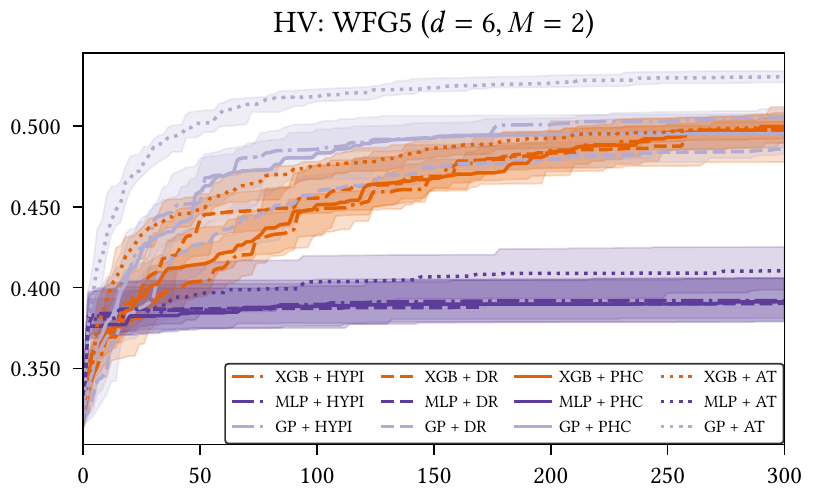}%
\includegraphics[width=0.25\linewidth]{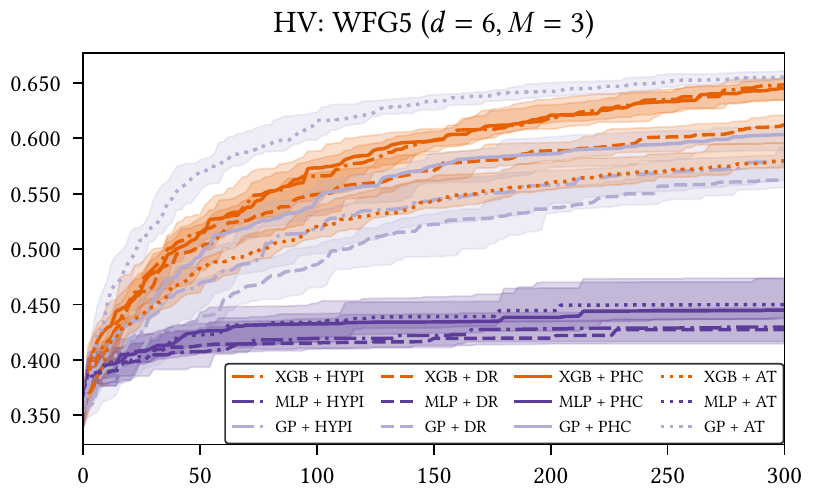}%
\includegraphics[width=0.25\linewidth]{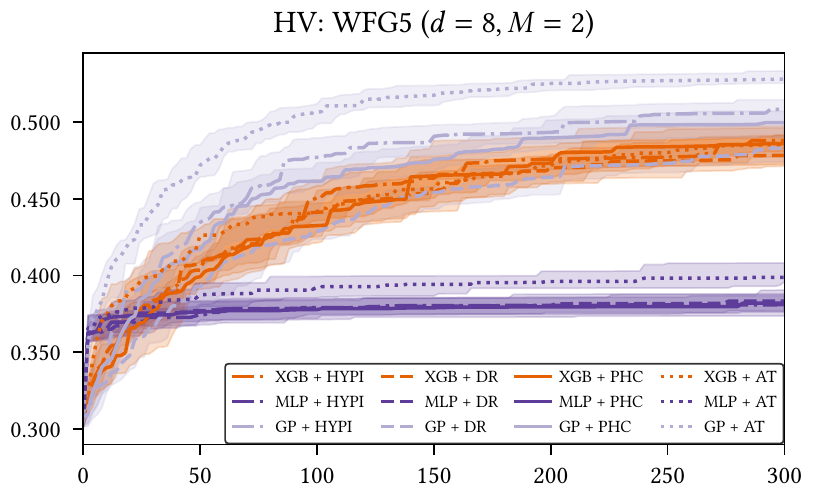}%
\includegraphics[width=0.25\linewidth]{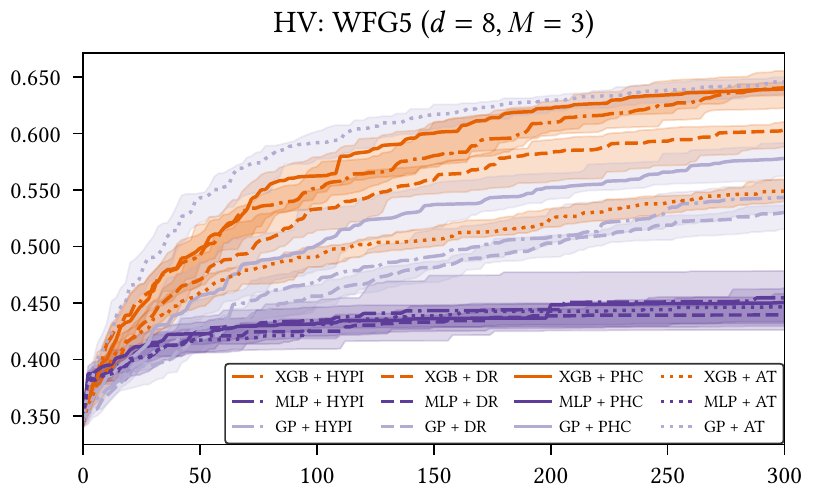}\\
\includegraphics[width=0.25\linewidth]{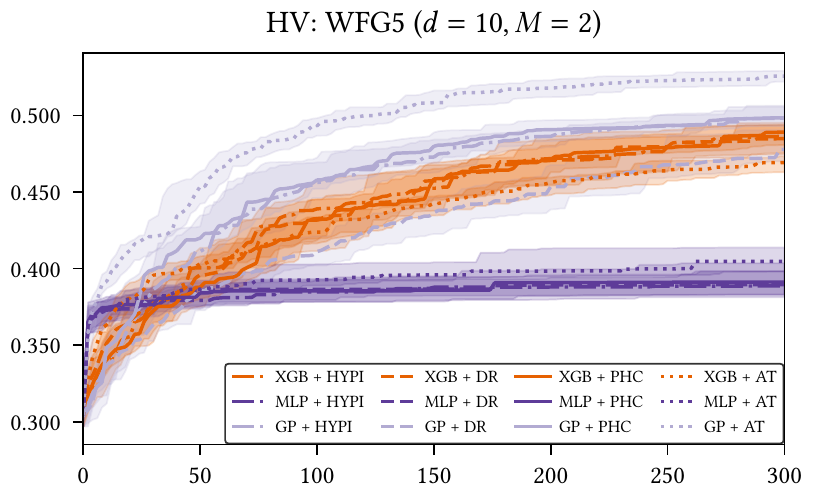}%
\includegraphics[width=0.25\linewidth]{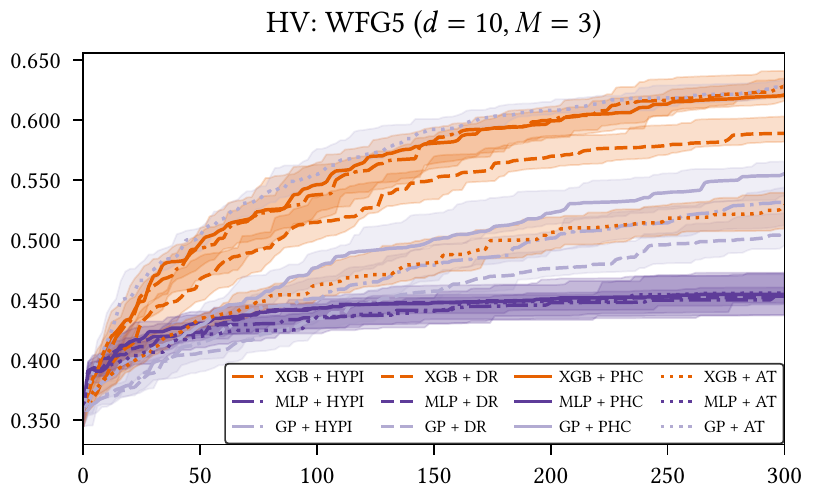}%
\includegraphics[width=0.25\linewidth]{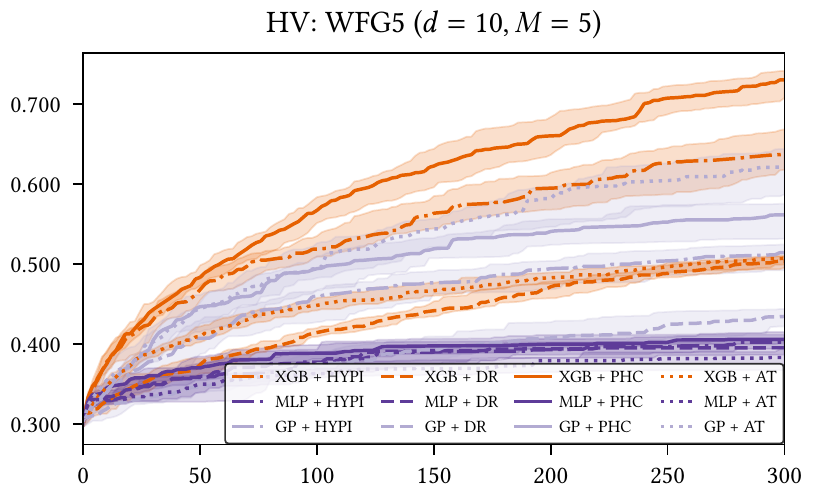}\\
\includegraphics[width=0.25\linewidth]{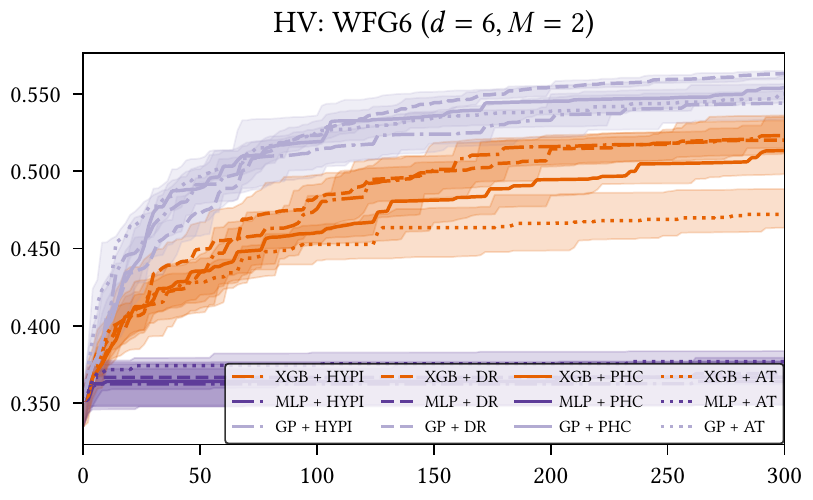}%
\includegraphics[width=0.25\linewidth]{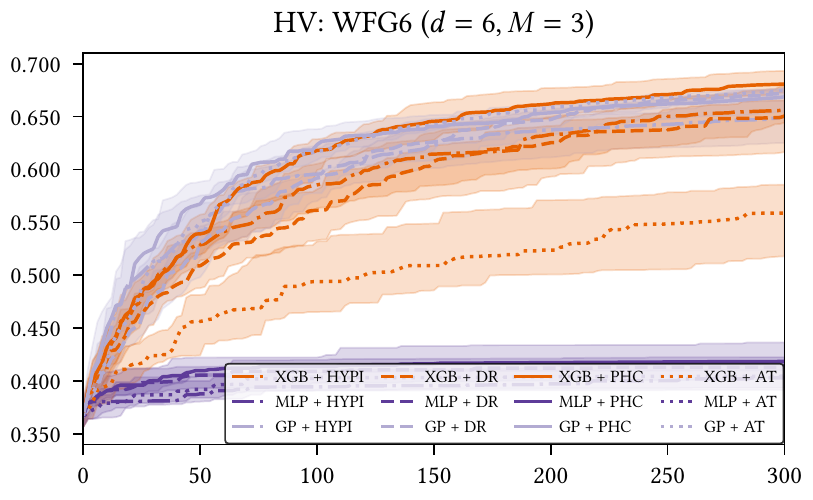}%
\includegraphics[width=0.25\linewidth]{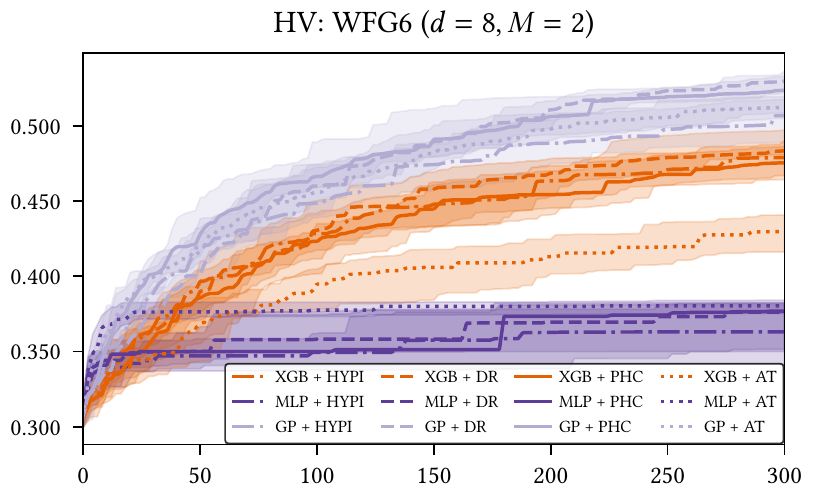}%
\includegraphics[width=0.25\linewidth]{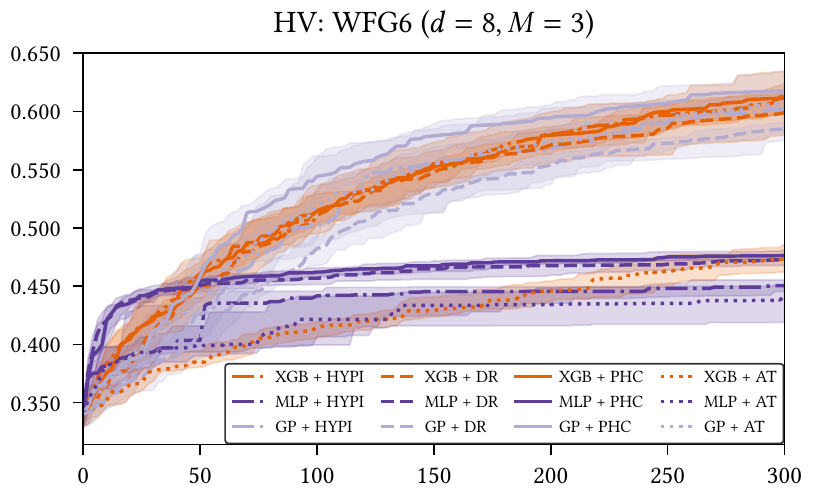}\\
\includegraphics[width=0.25\linewidth]{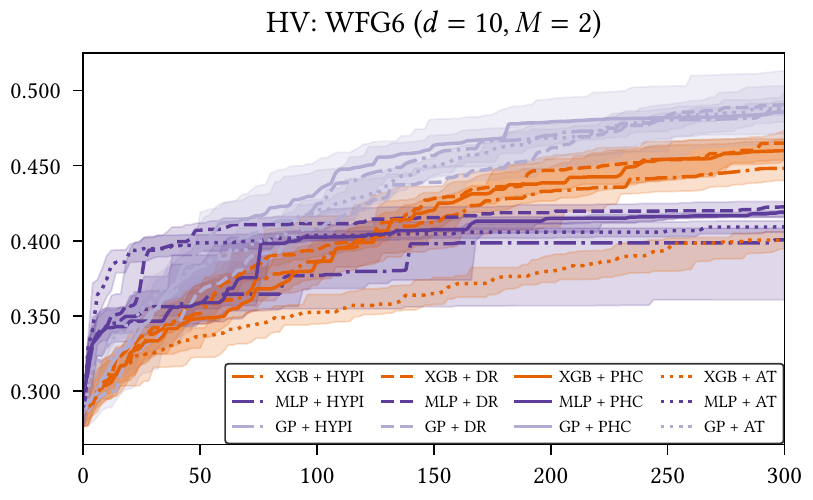}%
\includegraphics[width=0.25\linewidth]{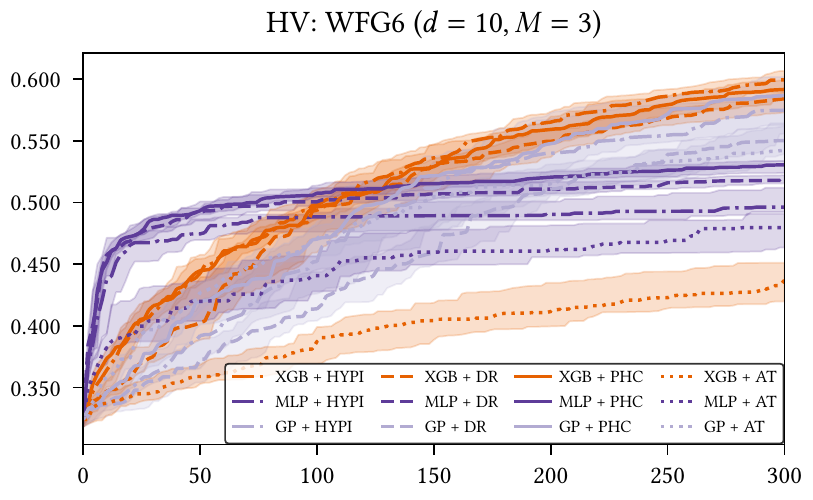}%
\includegraphics[width=0.25\linewidth]{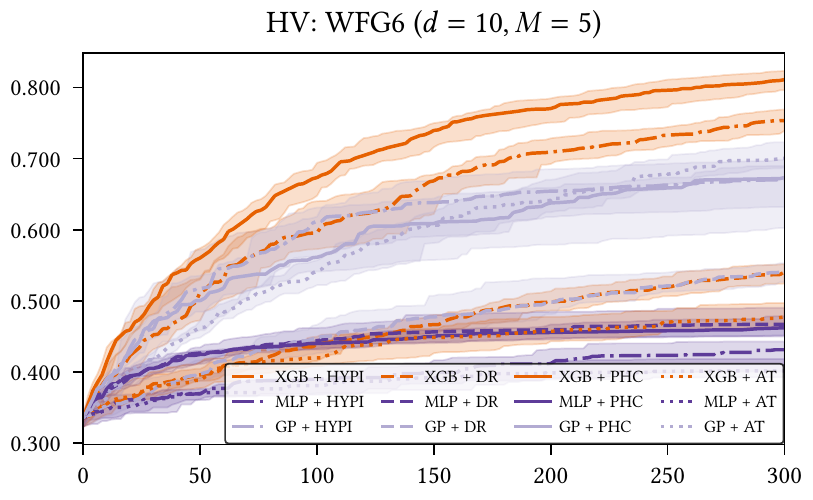}\\
\rule{\linewidth}{0.4pt}
\includegraphics[width=0.25\linewidth]{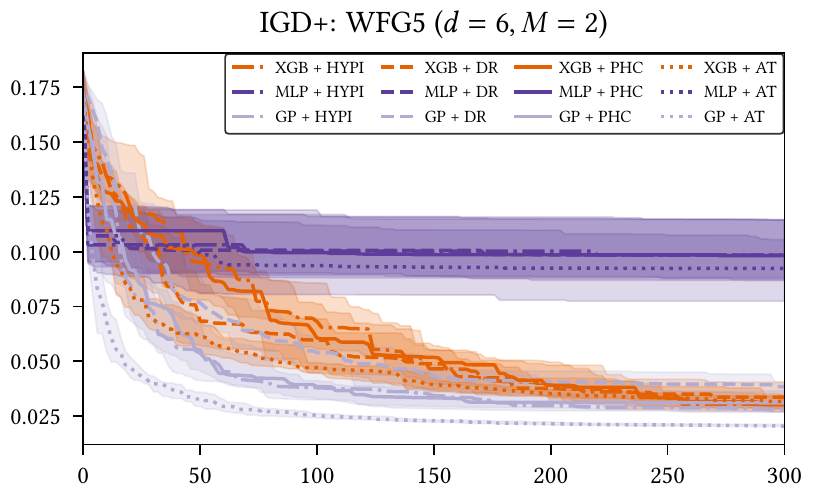}%
\includegraphics[width=0.25\linewidth]{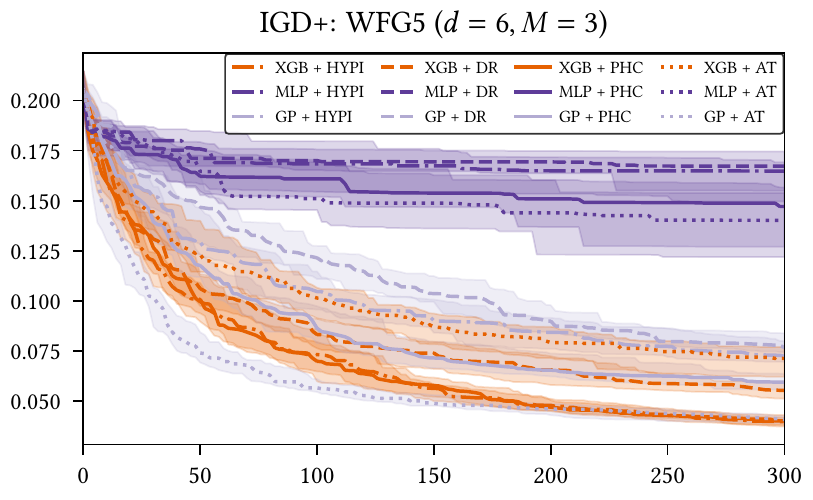}%
\includegraphics[width=0.25\linewidth]{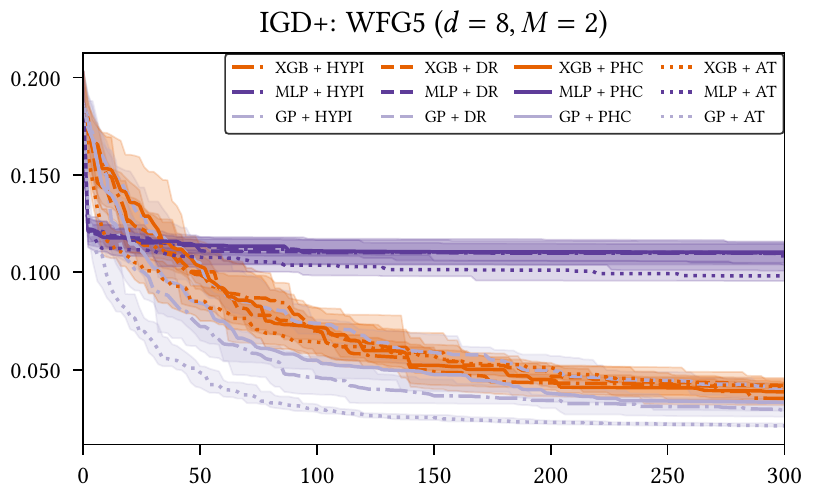}%
\includegraphics[width=0.25\linewidth]{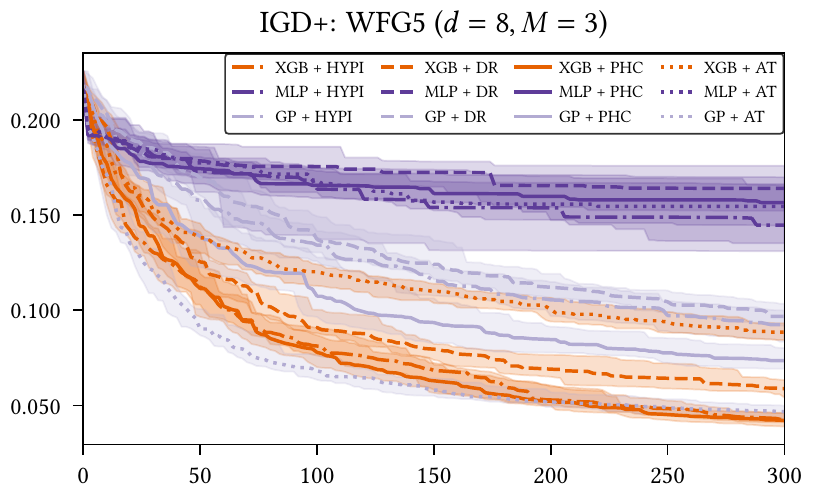}\\
\includegraphics[width=0.25\linewidth]{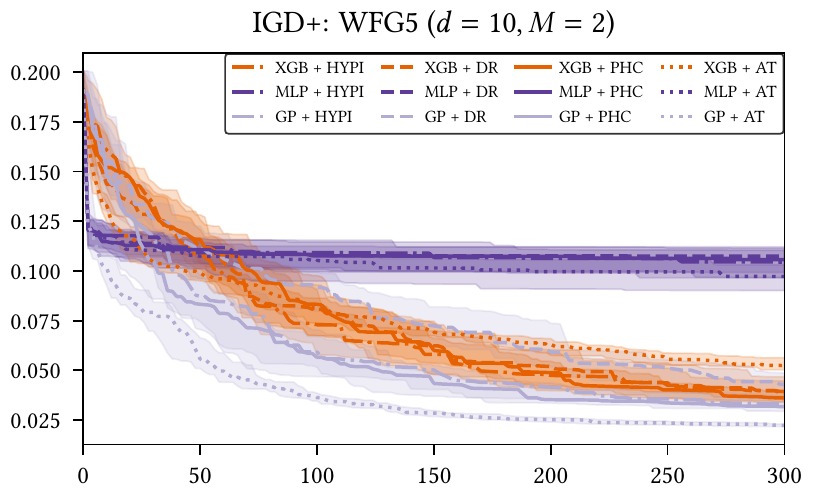}%
\includegraphics[width=0.25\linewidth]{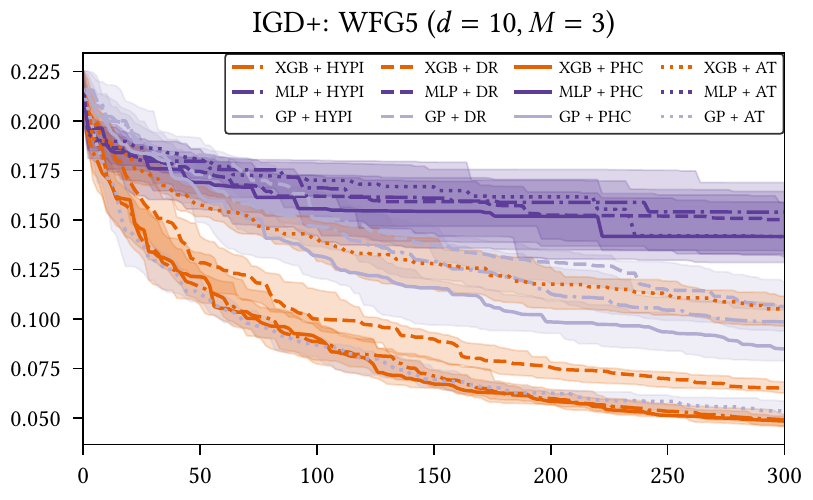}%
\includegraphics[width=0.25\linewidth]{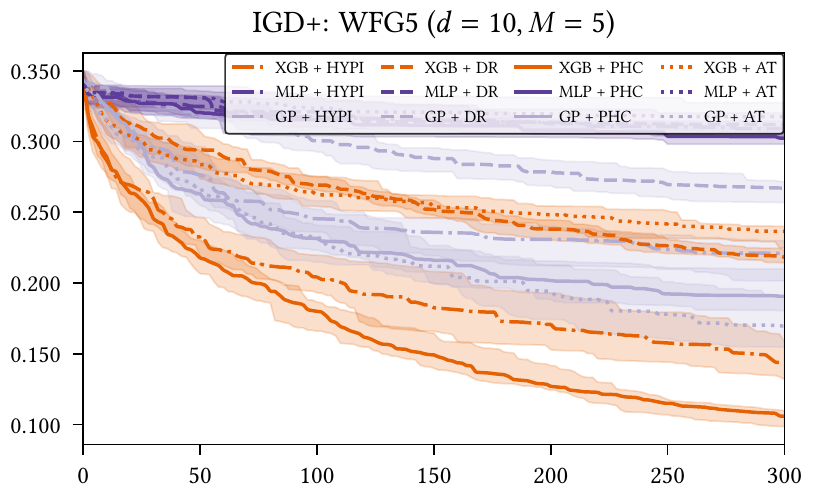}\\
\includegraphics[width=0.25\linewidth]{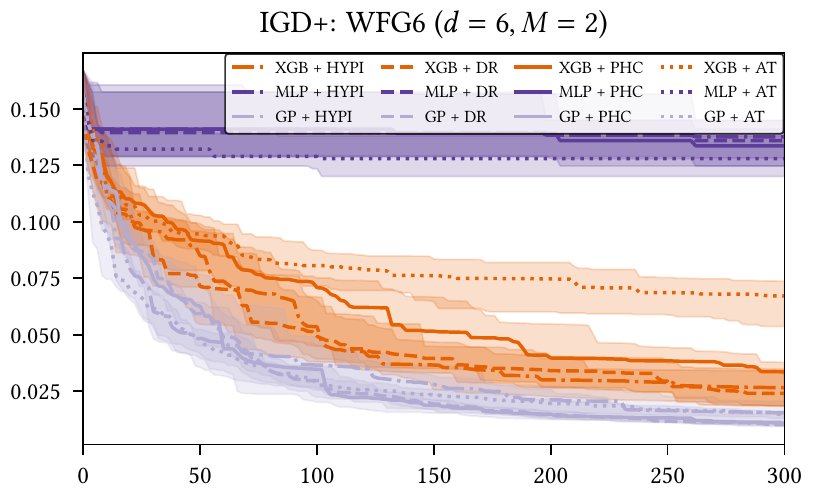}%
\includegraphics[width=0.25\linewidth]{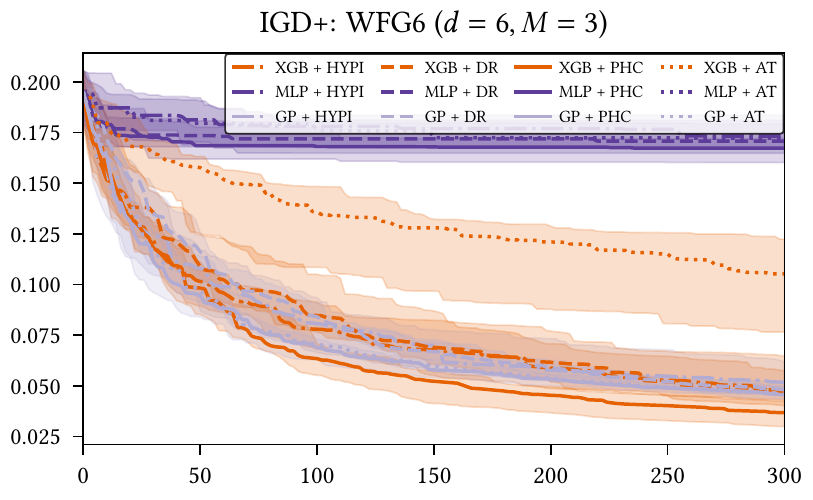}%
\includegraphics[width=0.25\linewidth]{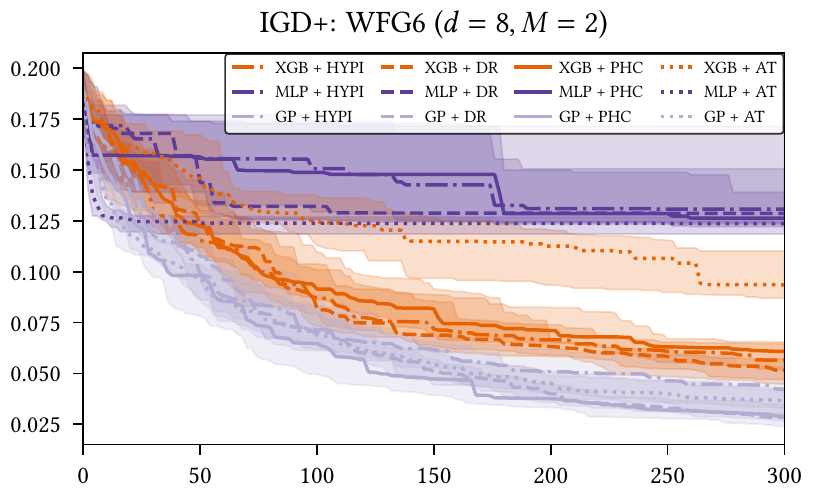}%
\includegraphics[width=0.25\linewidth]{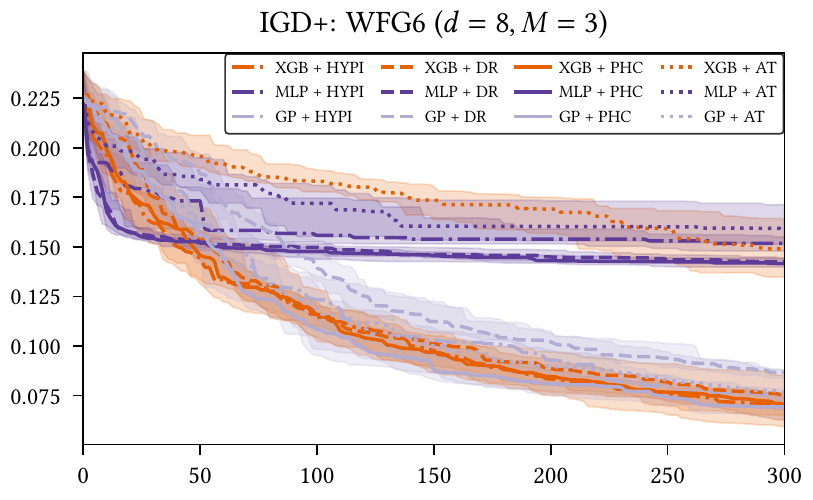}\\
\includegraphics[width=0.25\linewidth]{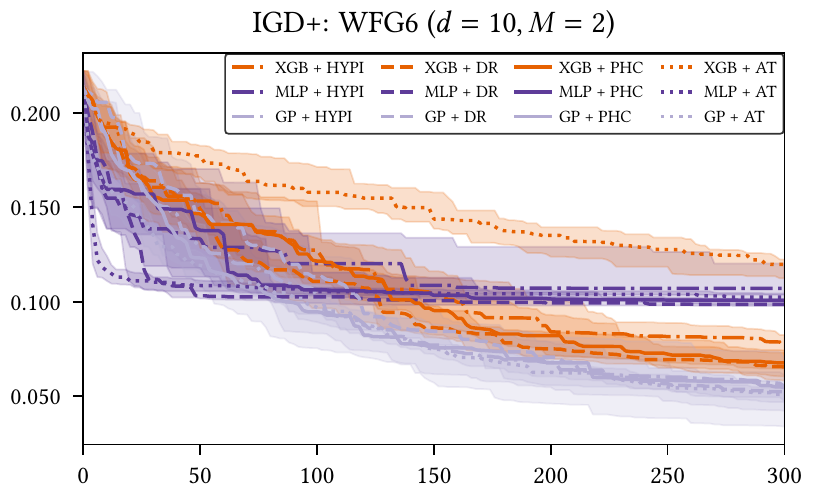}%
\includegraphics[width=0.25\linewidth]{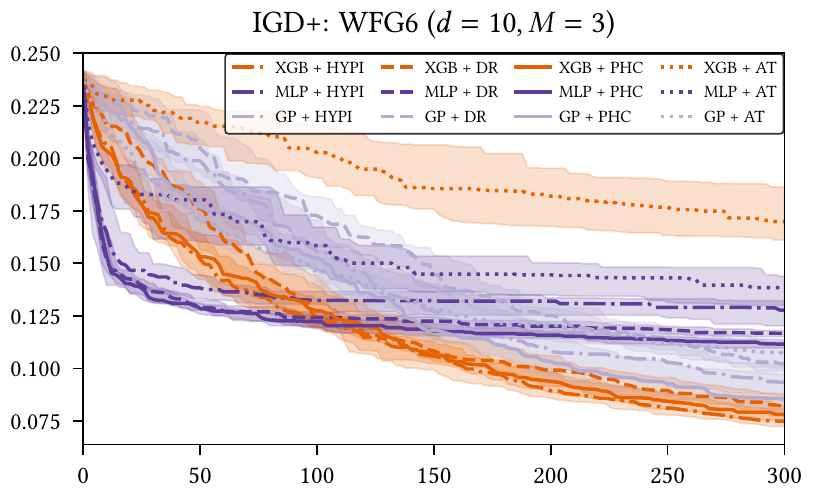}%
\includegraphics[width=0.25\linewidth]{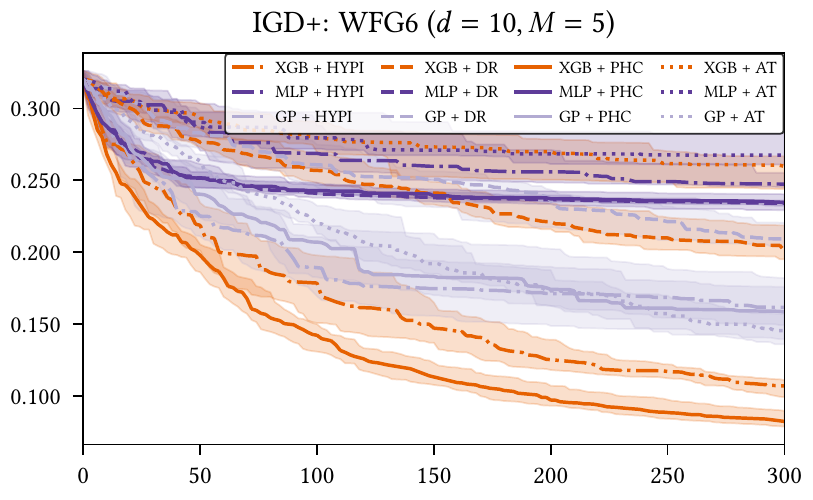}\\
\caption{%
    Hypervolume (\emph{upper}) and IGD+ (\emph{lower})
    convergence plots for WFG5 and WFG6.}
\label{fig:conv_WFG56}
\end{figure}

\begin{figure}[H]
\includegraphics[width=0.25\linewidth]{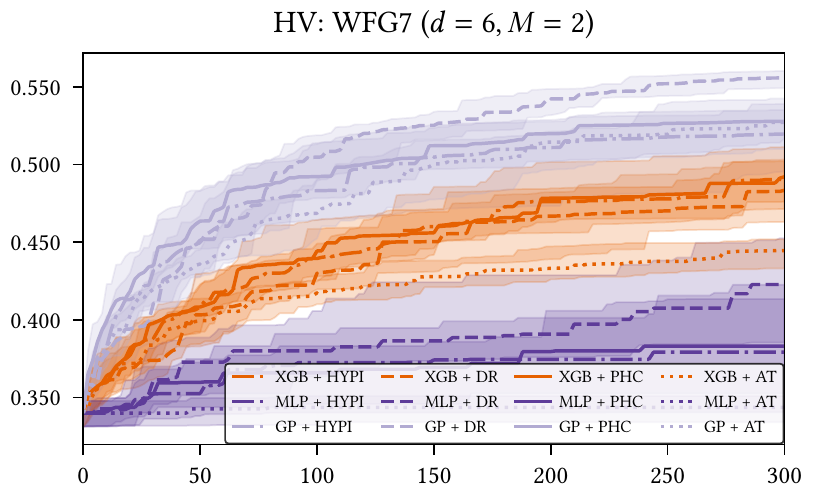}%
\includegraphics[width=0.25\linewidth]{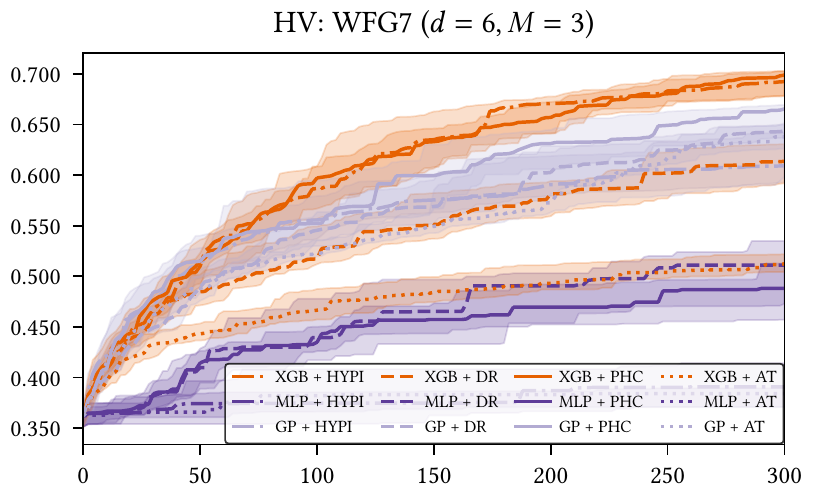}%
\includegraphics[width=0.25\linewidth]{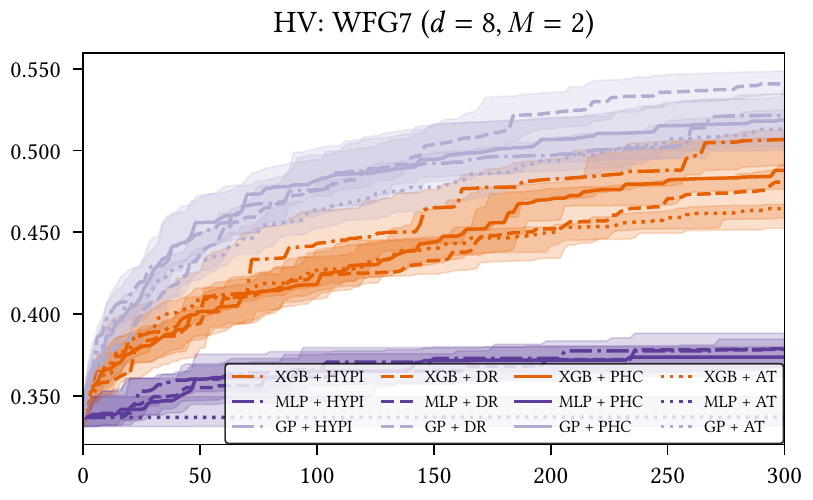}%
\includegraphics[width=0.25\linewidth]{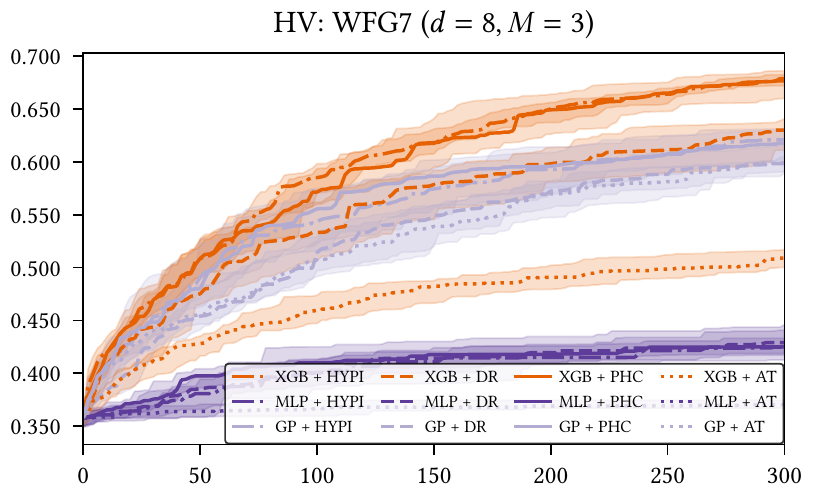}\\
\includegraphics[width=0.25\linewidth]{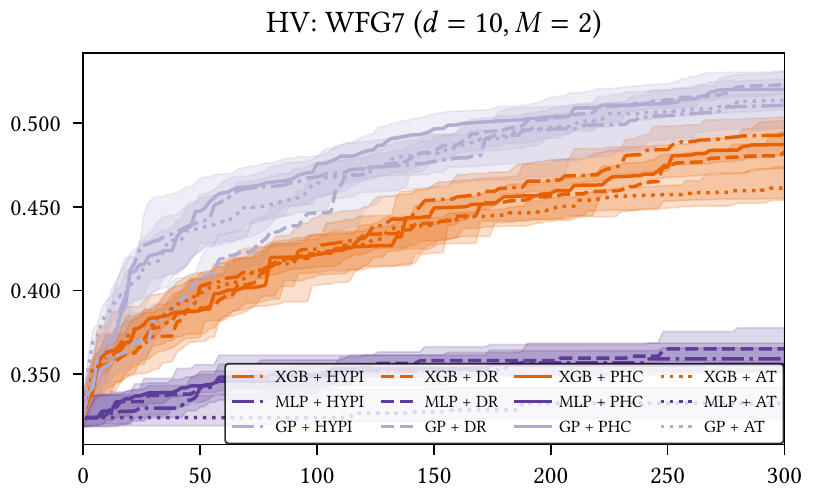}%
\includegraphics[width=0.25\linewidth]{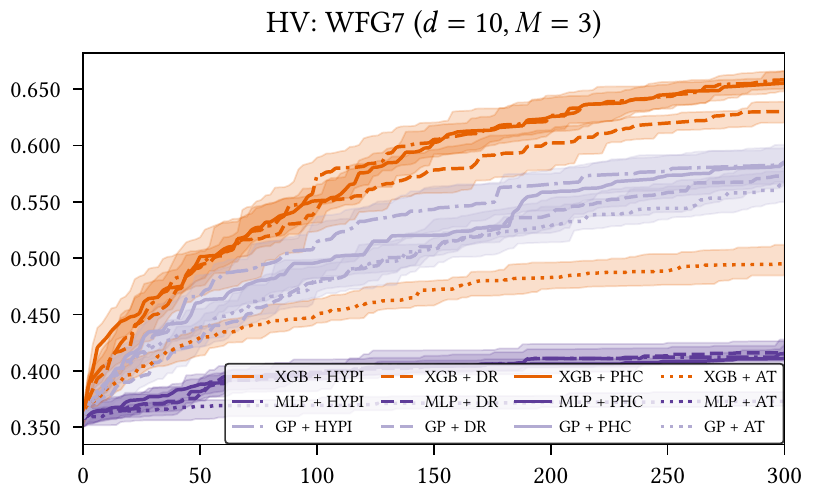}%
\includegraphics[width=0.25\linewidth]{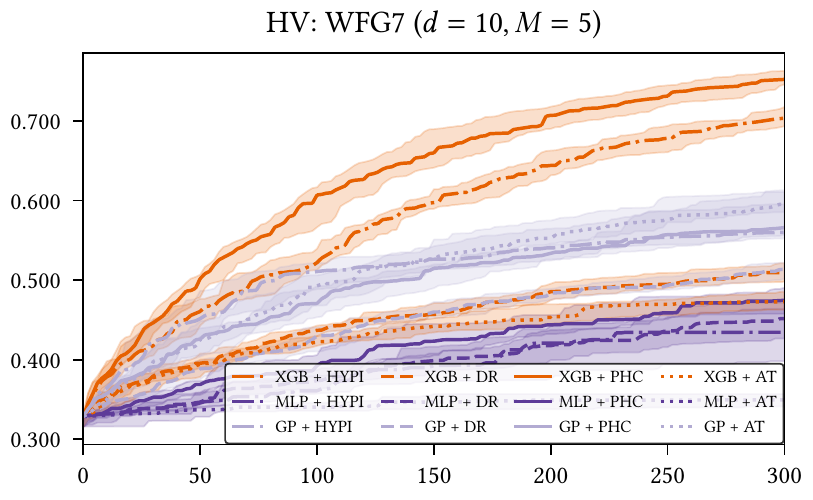}\\
\includegraphics[width=0.25\linewidth]{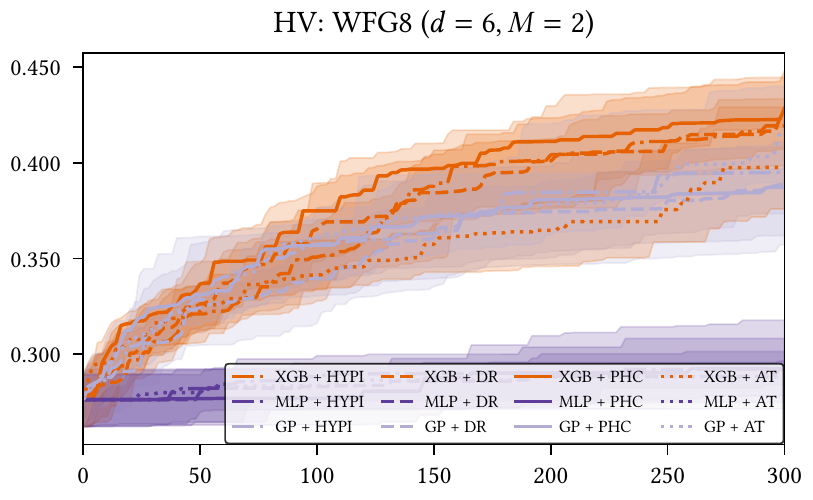}%
\includegraphics[width=0.25\linewidth]{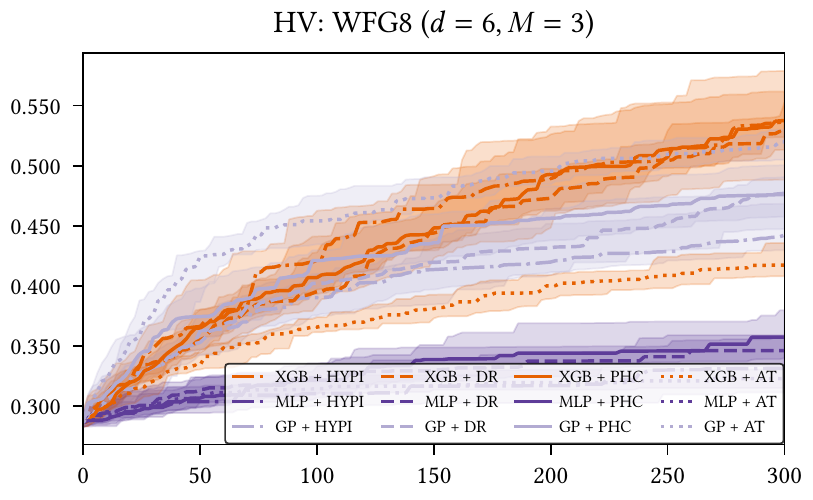}%
\includegraphics[width=0.25\linewidth]{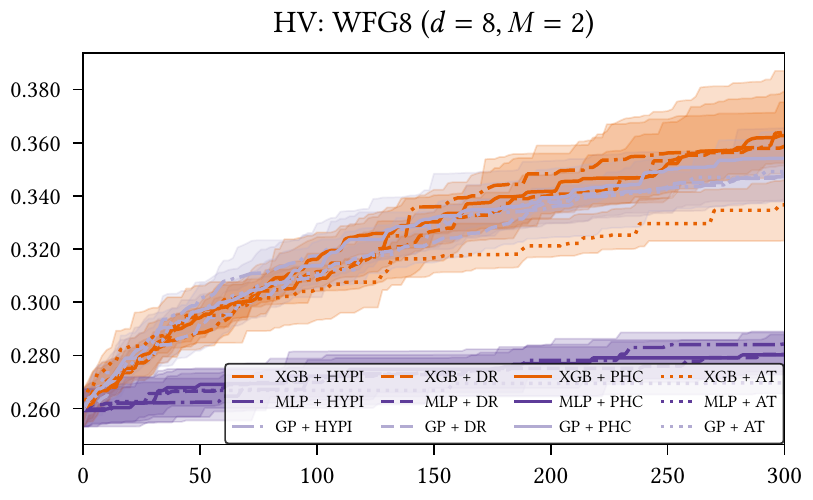}%
\includegraphics[width=0.25\linewidth]{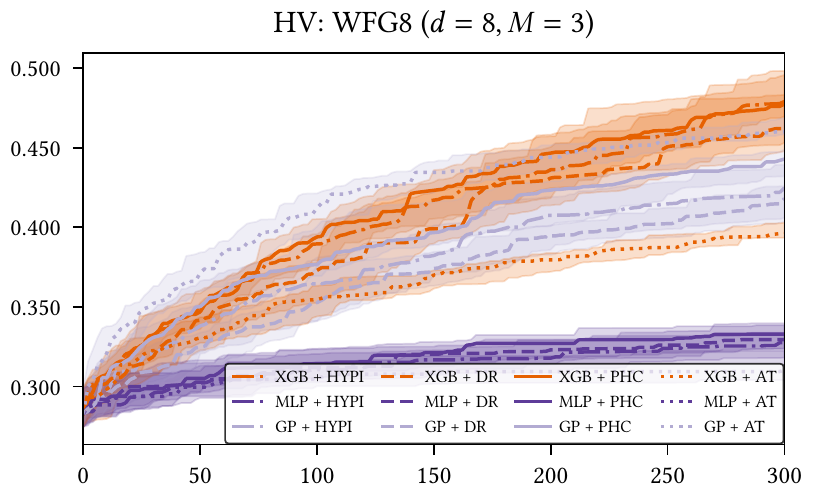}\\
\includegraphics[width=0.25\linewidth]{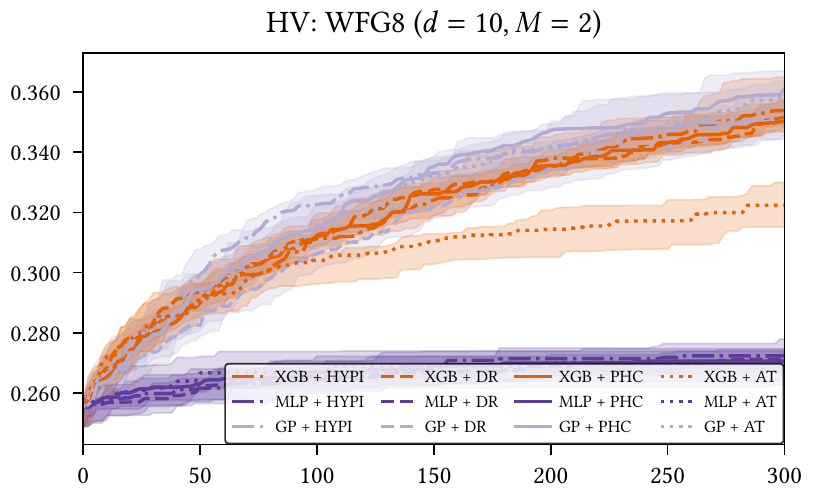}%
\includegraphics[width=0.25\linewidth]{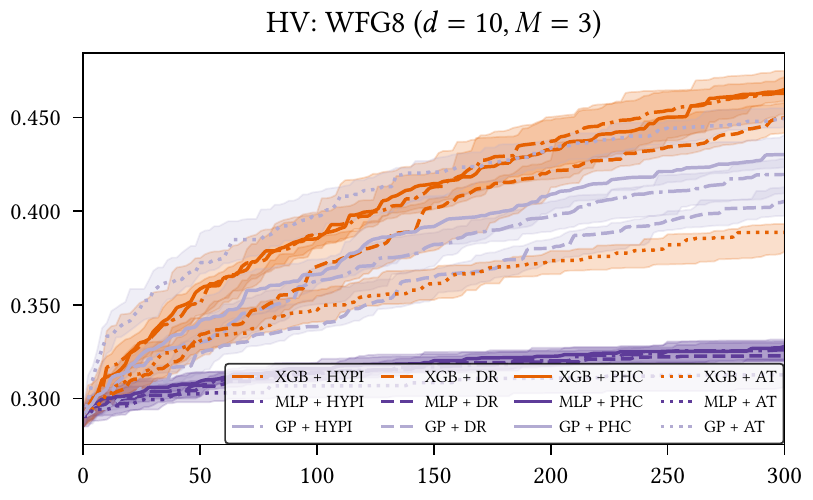}%
\includegraphics[width=0.25\linewidth]{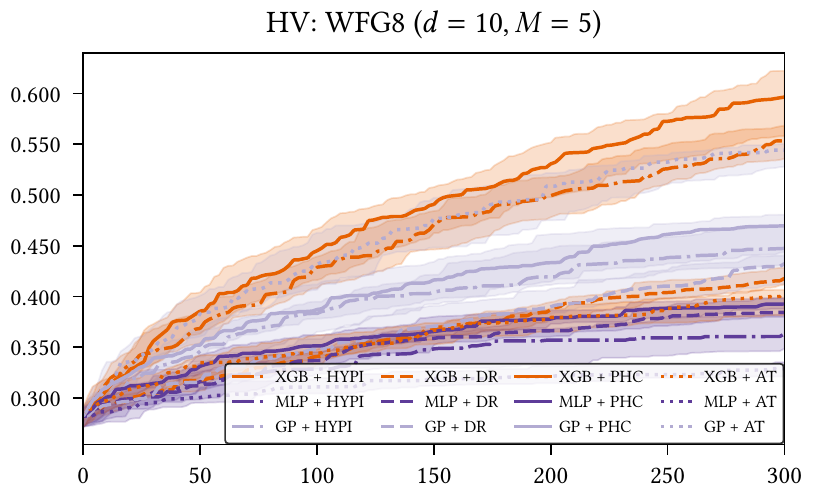}\\
\rule{\linewidth}{0.4pt}
\includegraphics[width=0.25\linewidth]{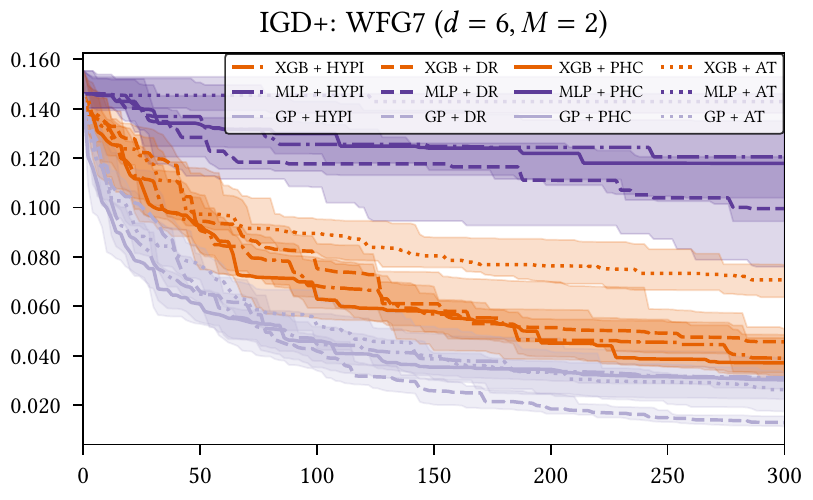}%
\includegraphics[width=0.25\linewidth]{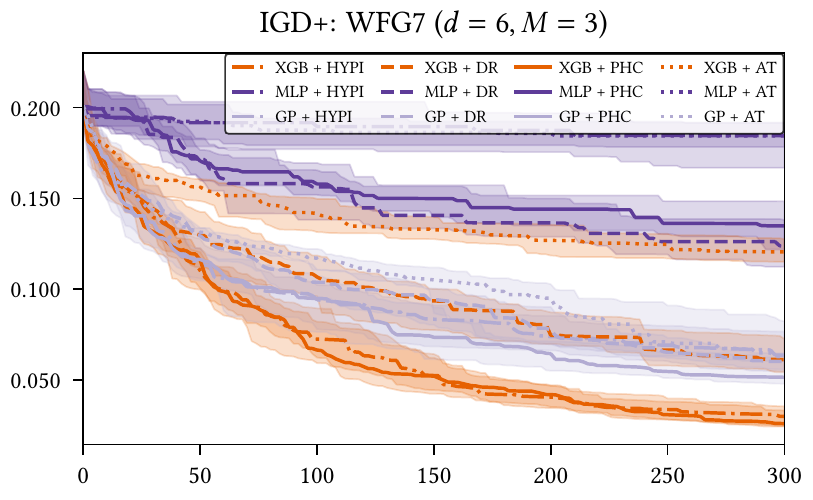}%
\includegraphics[width=0.25\linewidth]{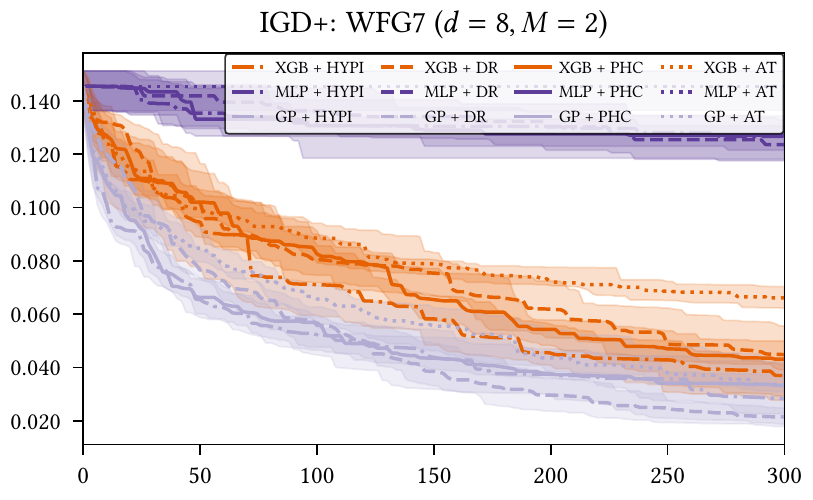}%
\includegraphics[width=0.25\linewidth]{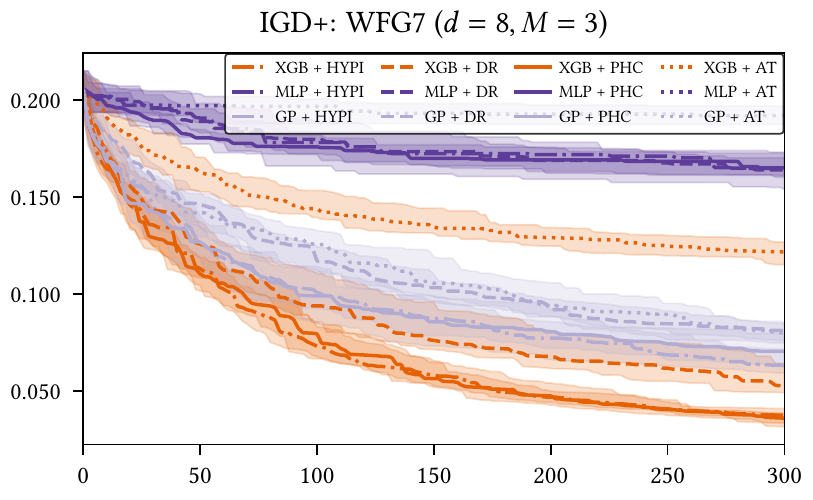}\\
\includegraphics[width=0.25\linewidth]{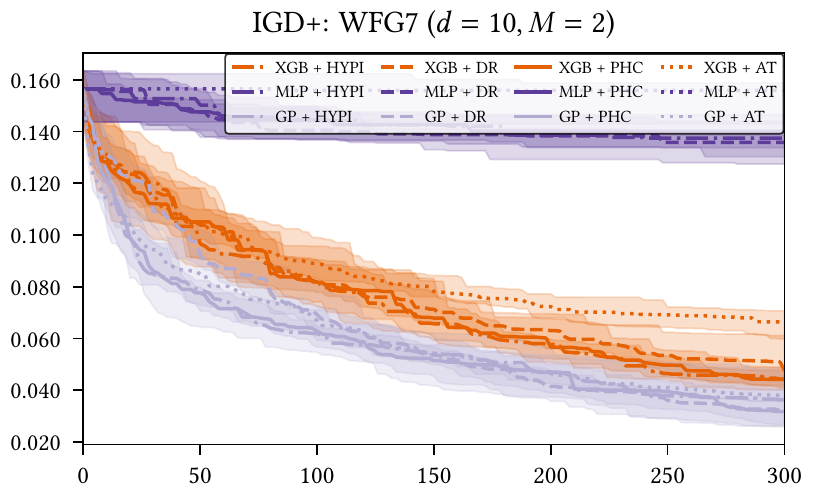}%
\includegraphics[width=0.25\linewidth]{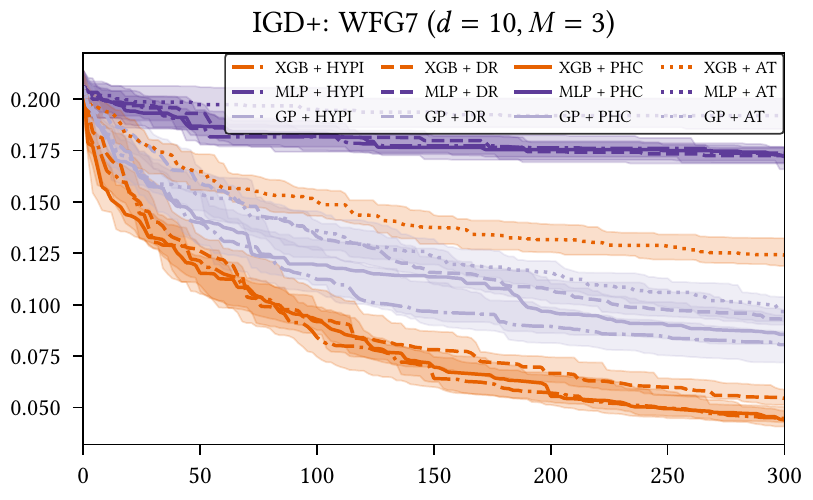}%
\includegraphics[width=0.25\linewidth]{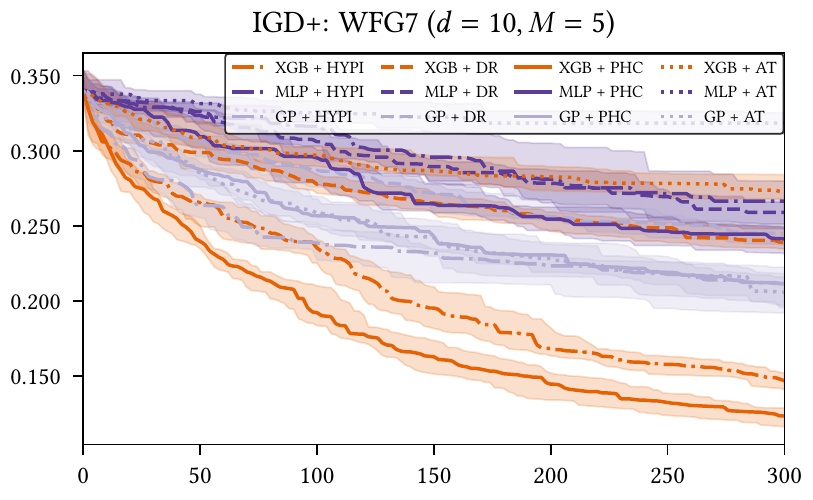}\\
\includegraphics[width=0.25\linewidth]{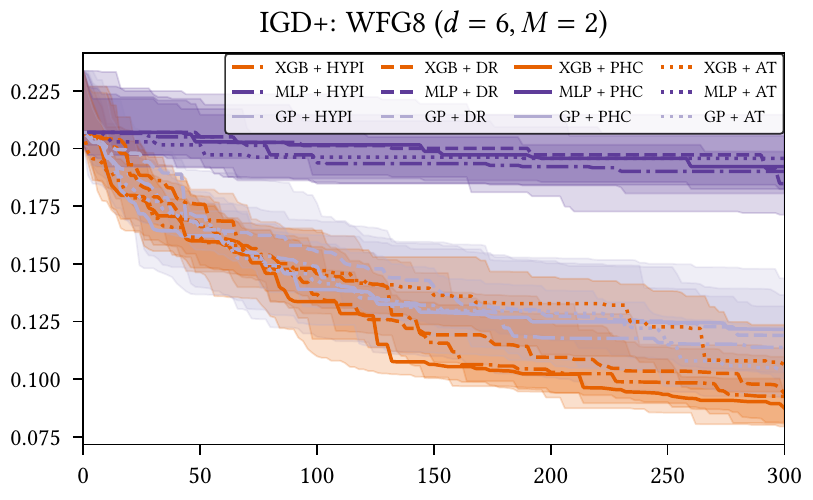}%
\includegraphics[width=0.25\linewidth]{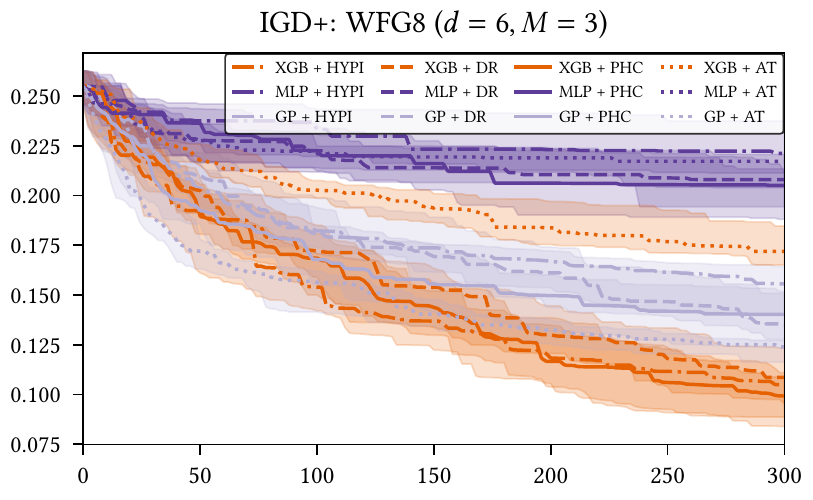}%
\includegraphics[width=0.25\linewidth]{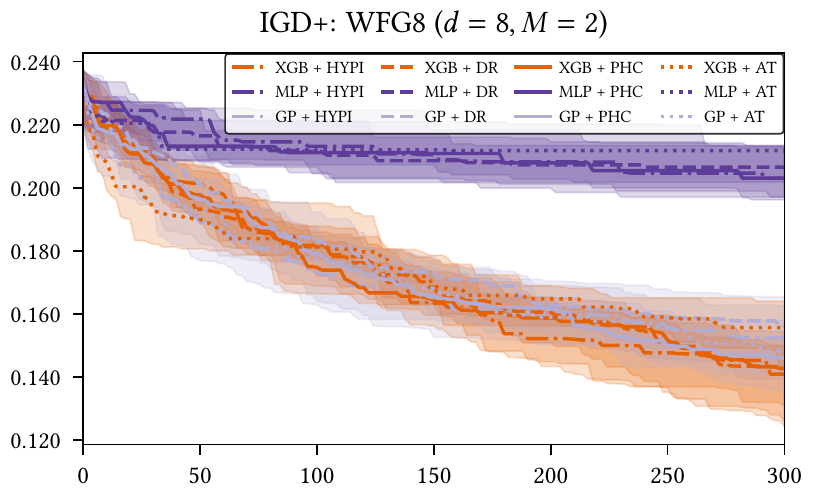}%
\includegraphics[width=0.25\linewidth]{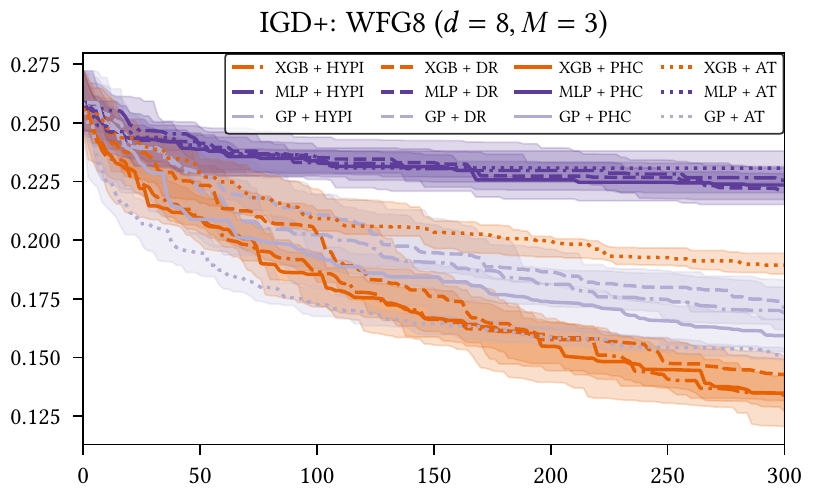}\\
\includegraphics[width=0.25\linewidth]{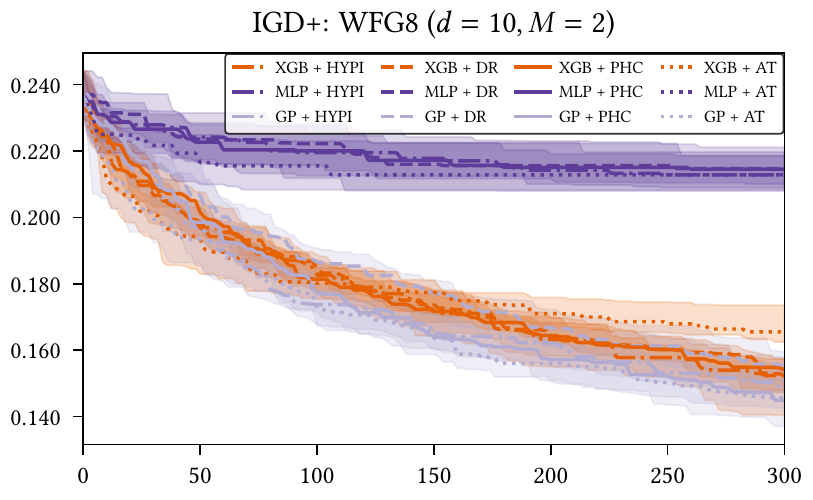}%
\includegraphics[width=0.25\linewidth]{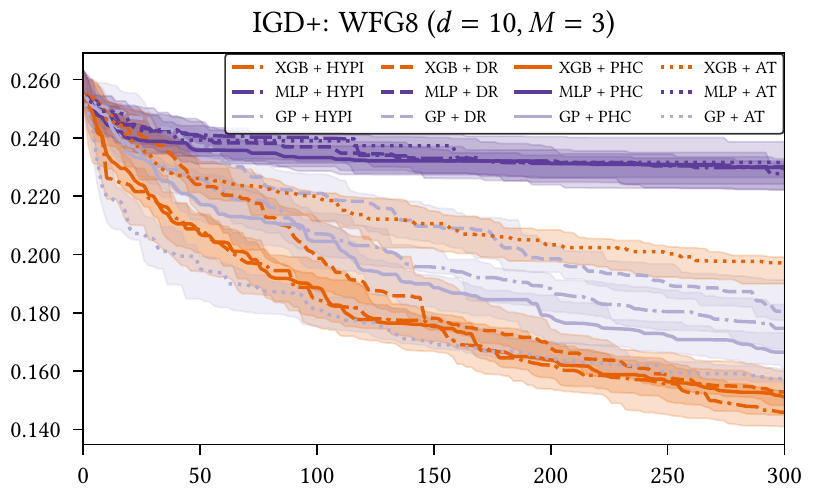}%
\includegraphics[width=0.25\linewidth]{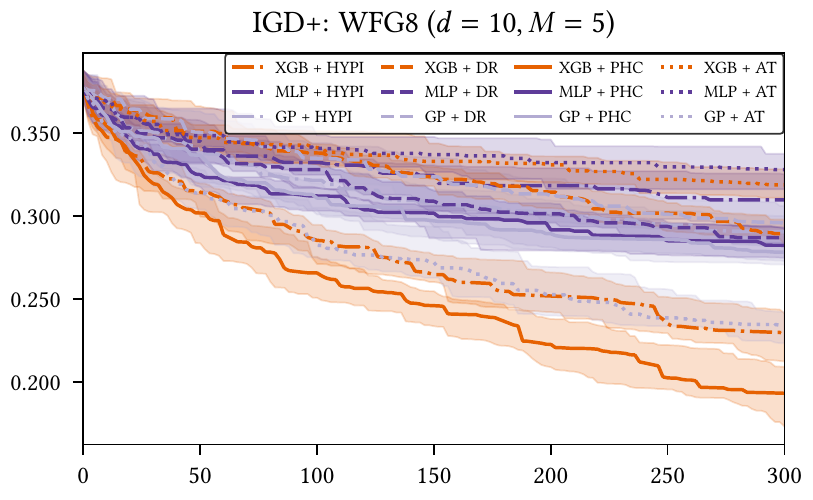}\\
\caption{%
    Hypervolume (\emph{upper}) and IGD+ (\emph{lower})
    convergence plots for WFG7 and WFG8.}
\label{fig:conv_WFG78}
\end{figure}

\begin{figure}[H]
\includegraphics[width=0.25\linewidth]{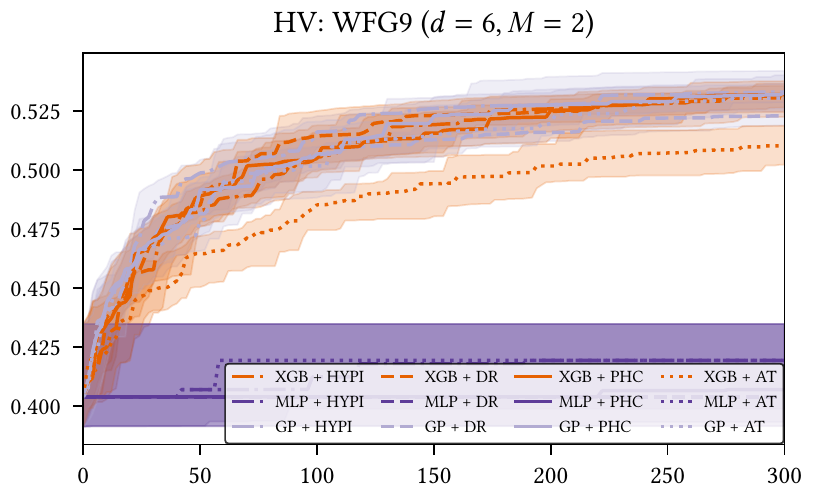}%
\includegraphics[width=0.25\linewidth]{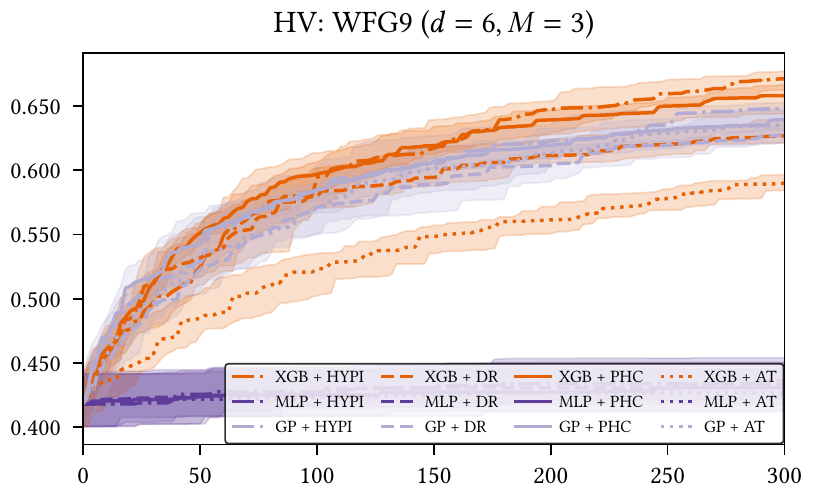}%
\includegraphics[width=0.25\linewidth]{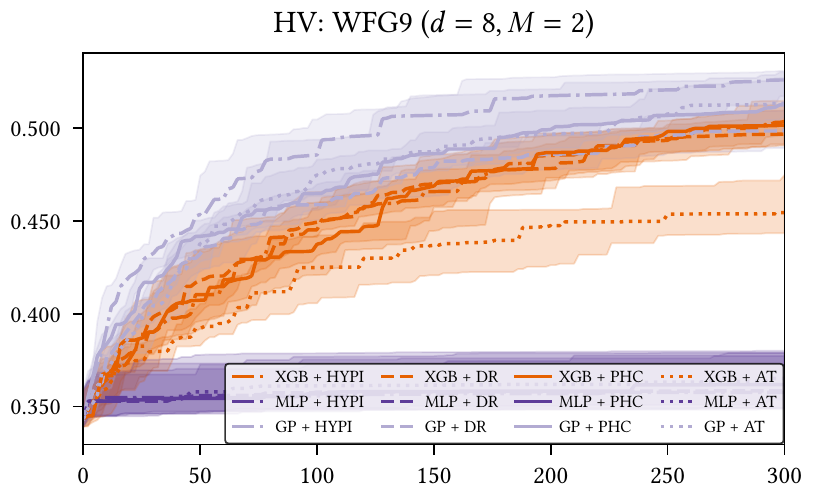}%
\includegraphics[width=0.25\linewidth]{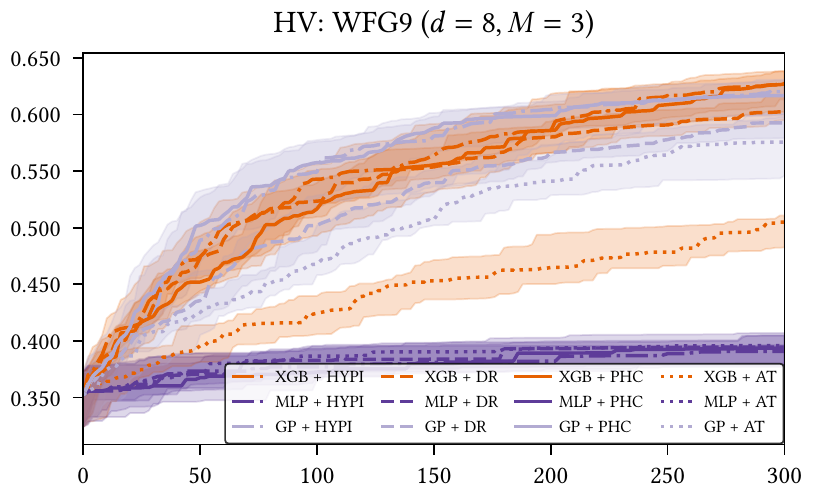}\\
\includegraphics[width=0.25\linewidth]{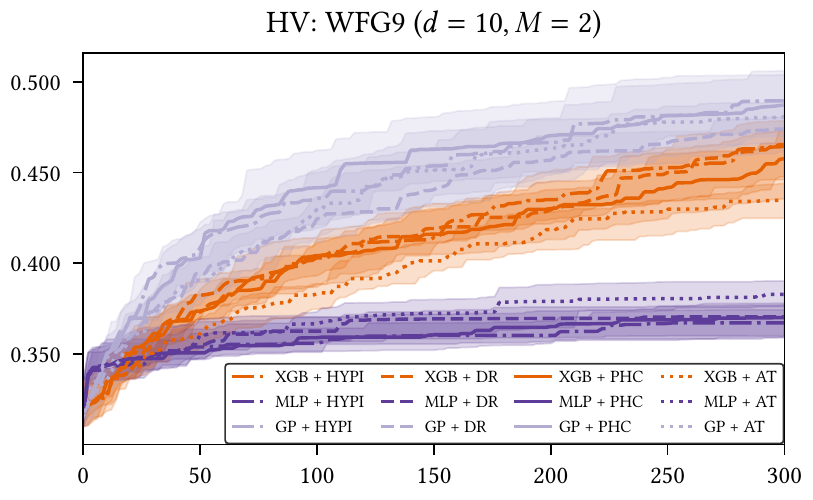}%
\includegraphics[width=0.25\linewidth]{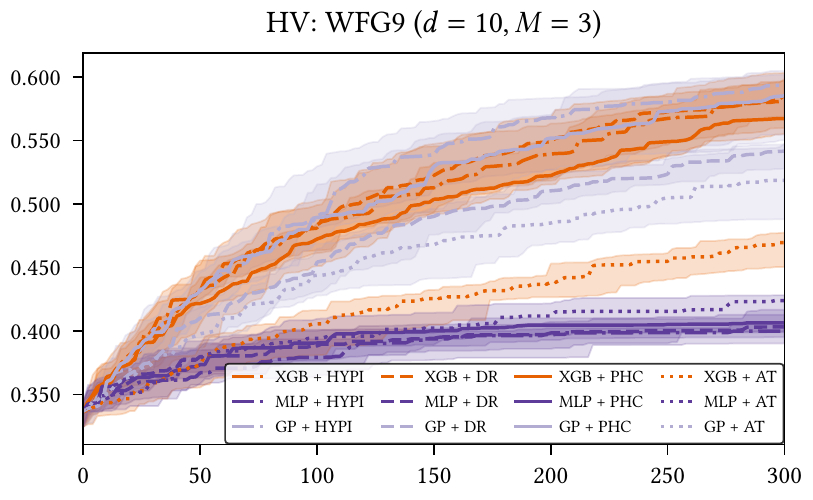}%
\includegraphics[width=0.25\linewidth]{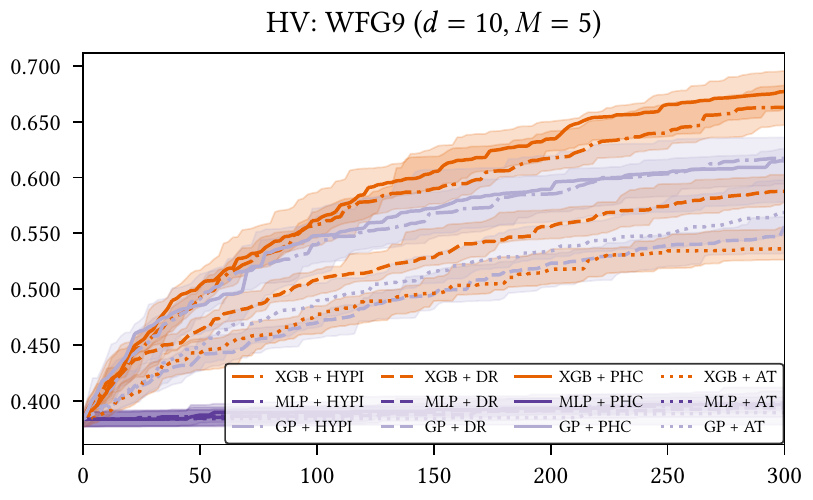}\\
\rule{\linewidth}{0.4pt}
\includegraphics[width=0.25\linewidth]{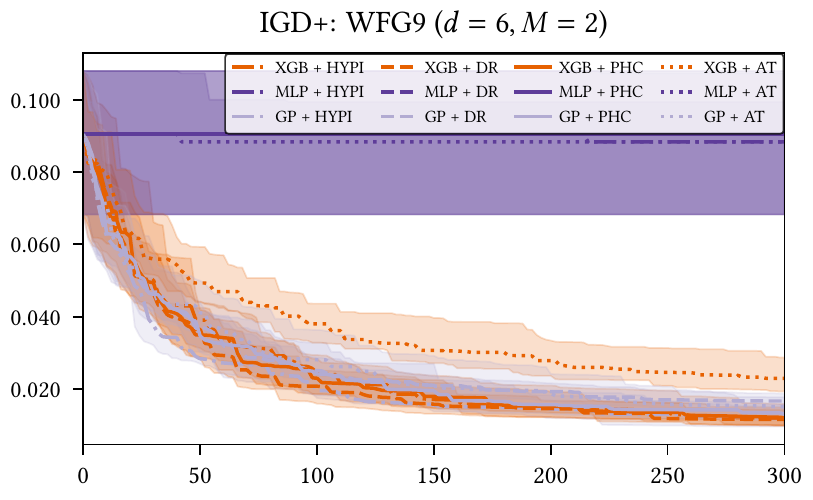}%
\includegraphics[width=0.25\linewidth]{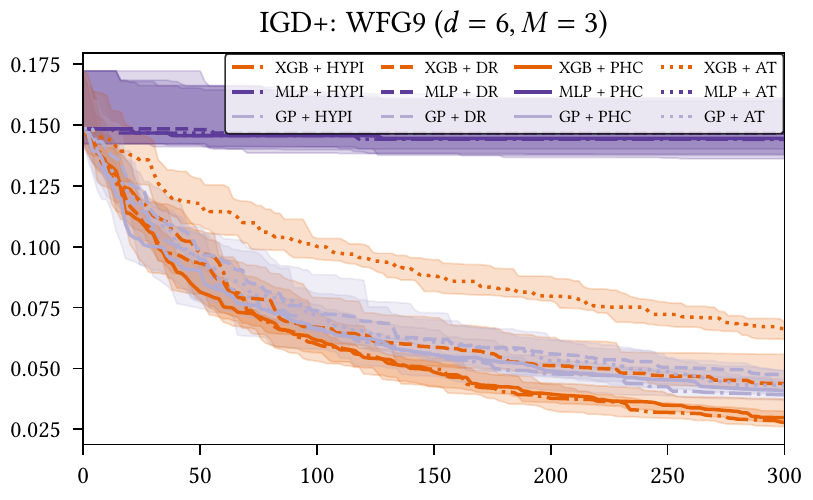}%
\includegraphics[width=0.25\linewidth]{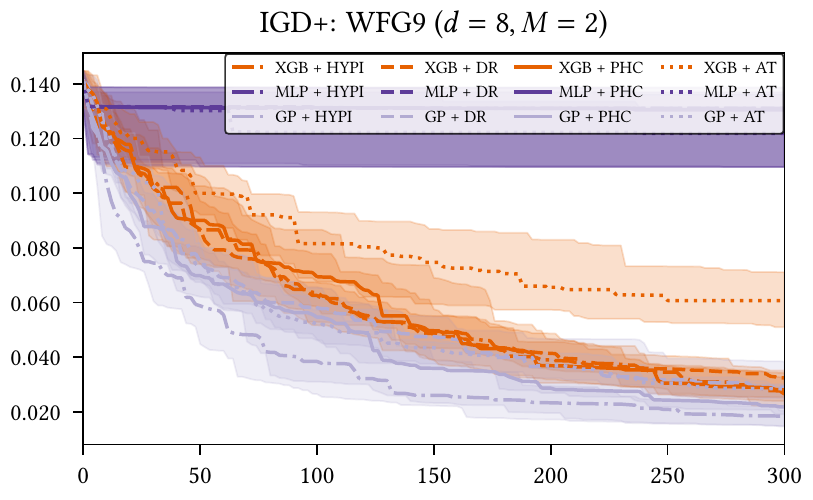}%
\includegraphics[width=0.25\linewidth]{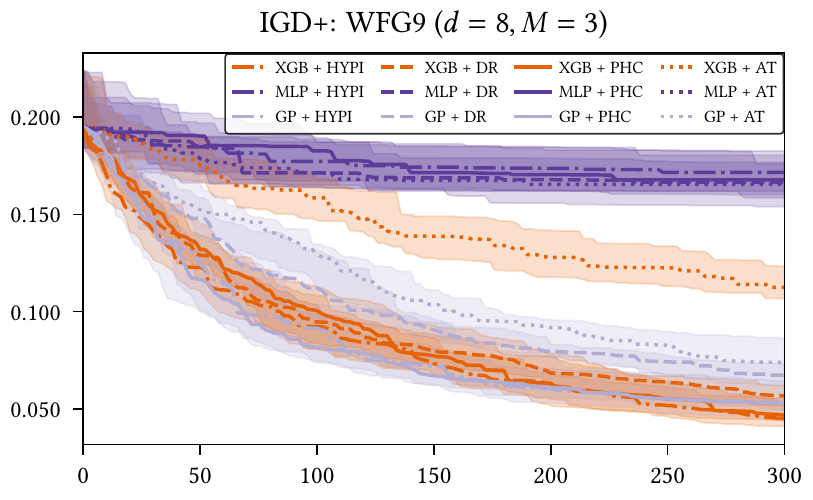}\\
\includegraphics[width=0.25\linewidth]{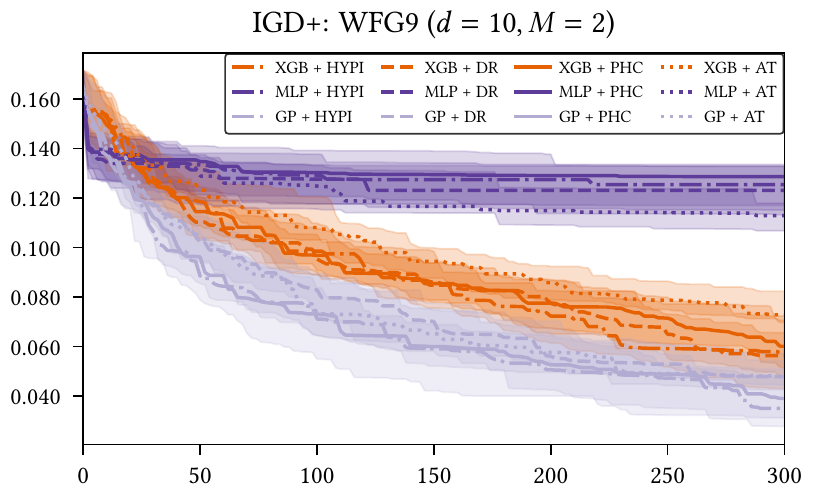}%
\includegraphics[width=0.25\linewidth]{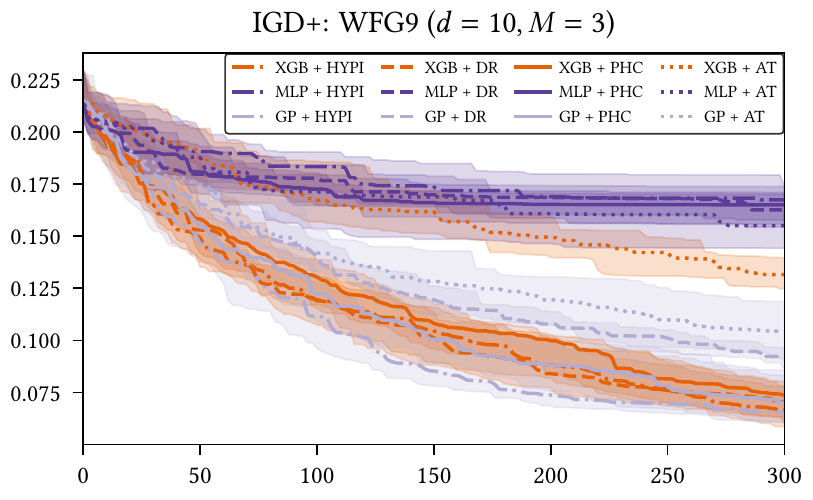}%
\includegraphics[width=0.25\linewidth]{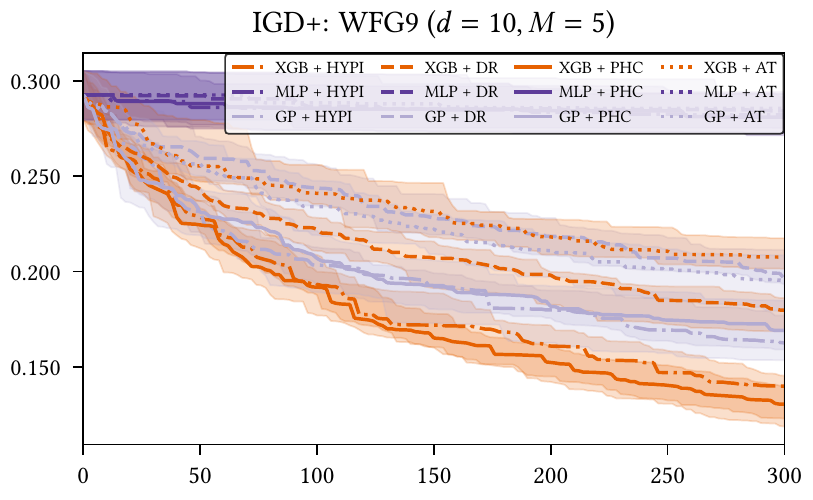}\\
\caption{%
    Hypervolume (\emph{upper}) and IGD+ (\emph{lower})
    convergence plots for WFG9.}
\label{fig:conv_WFG9}
\end{figure}

\newpage
\subsection{Real-world Convergence Plots}
\label{sec:conv:RW}
\begin{figure}[H]
\includegraphics[width=0.25\linewidth]{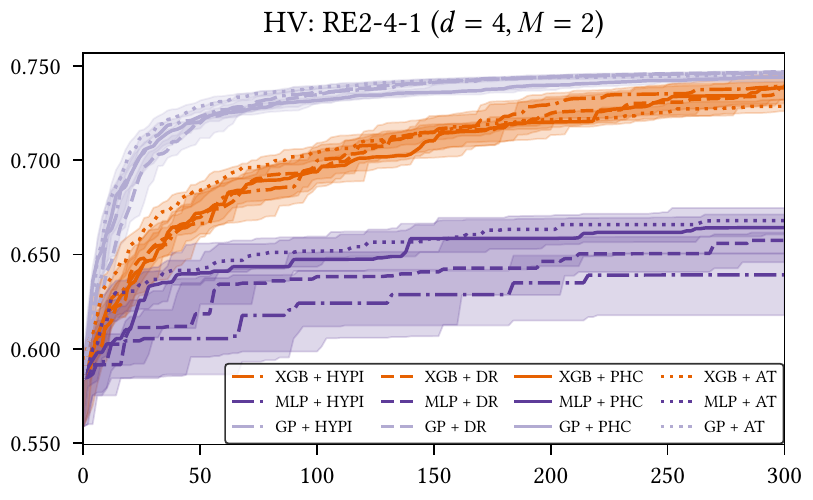}%
\includegraphics[width=0.25\linewidth]{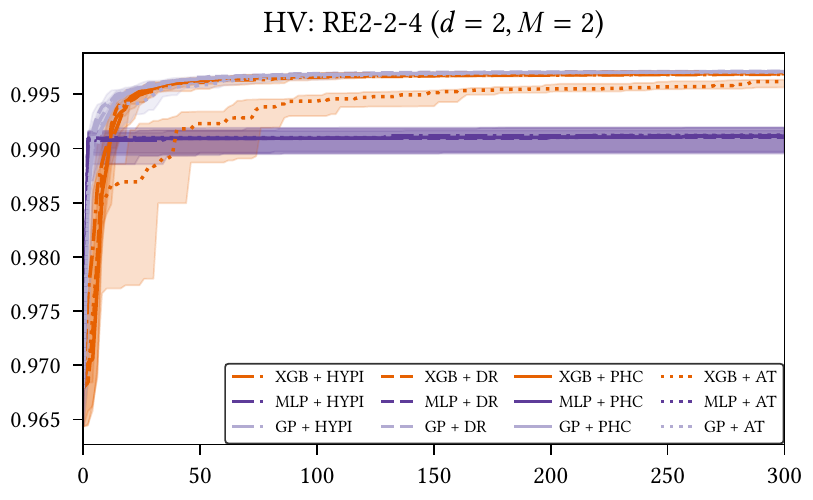}%
\includegraphics[width=0.25\linewidth]{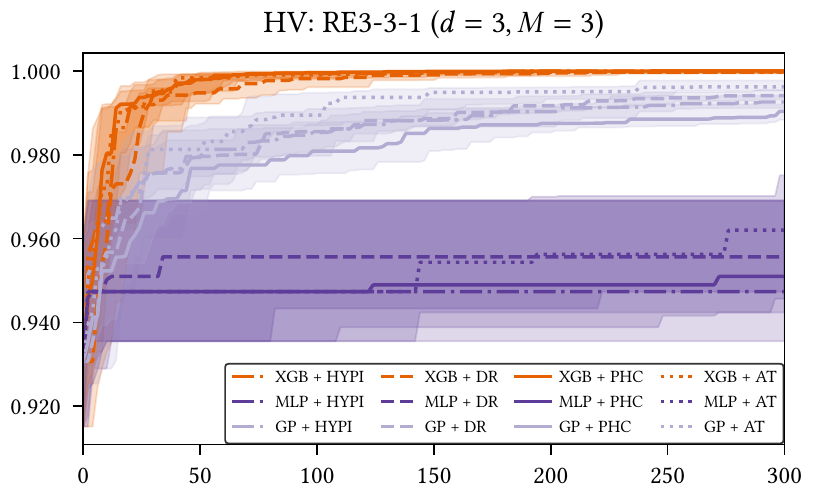}%
\includegraphics[width=0.25\linewidth]{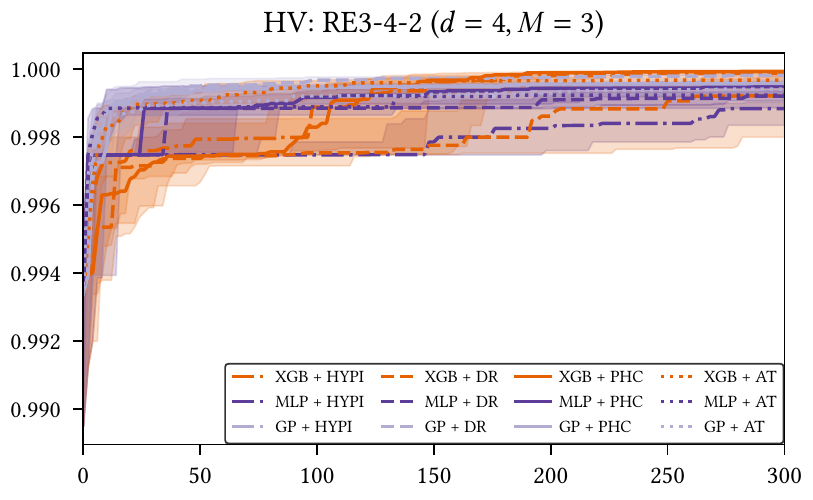}\\
\includegraphics[width=0.25\linewidth]{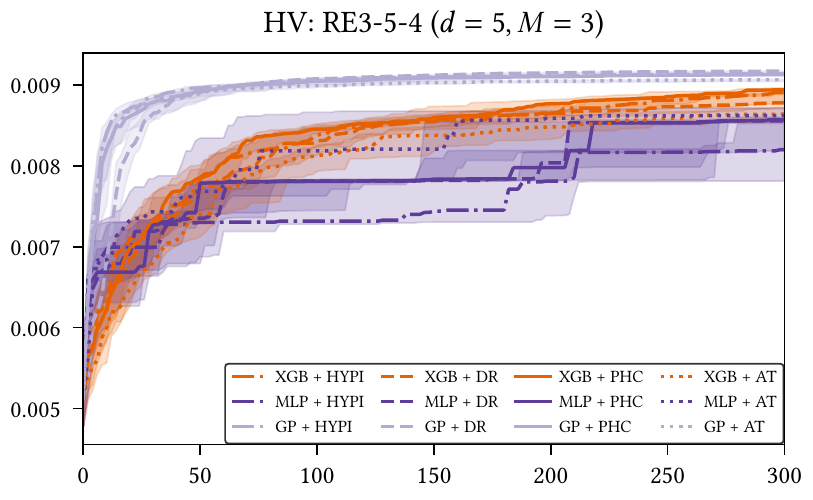}%
\includegraphics[width=0.25\linewidth]{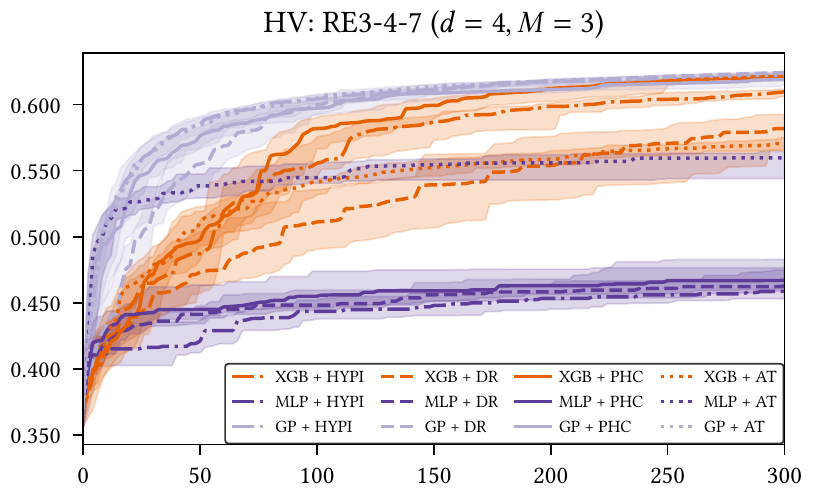}%
\includegraphics[width=0.25\linewidth]{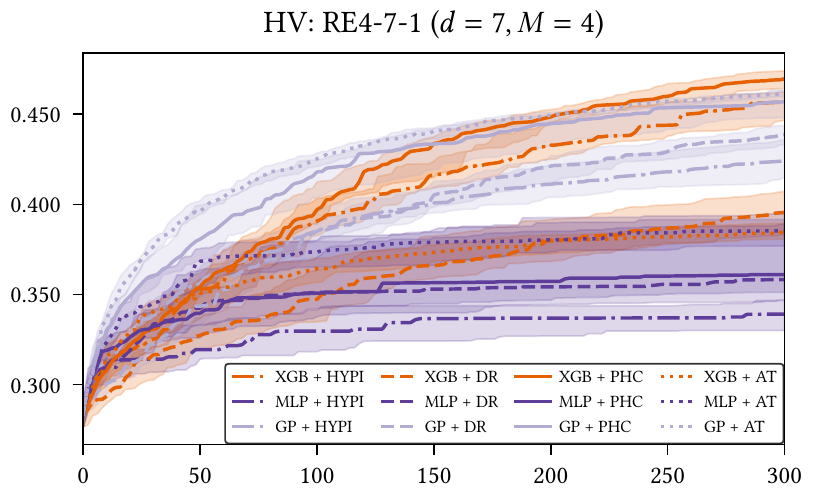}%
\includegraphics[width=0.25\linewidth]{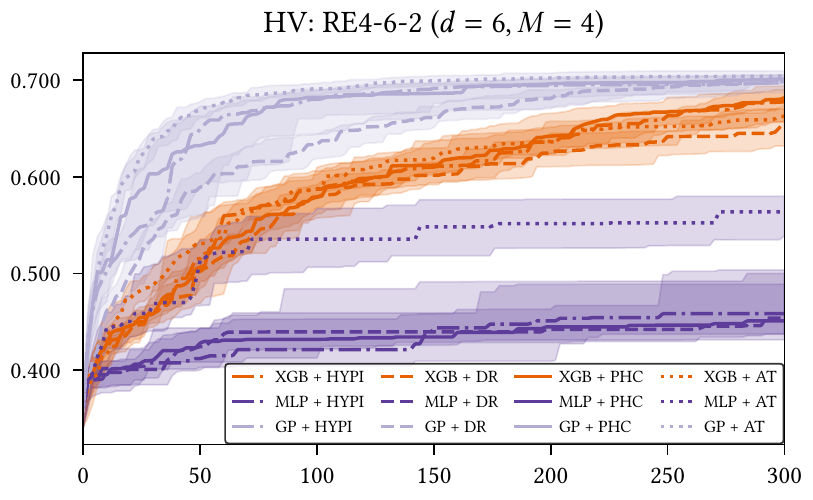}\\
\includegraphics[width=0.25\linewidth]{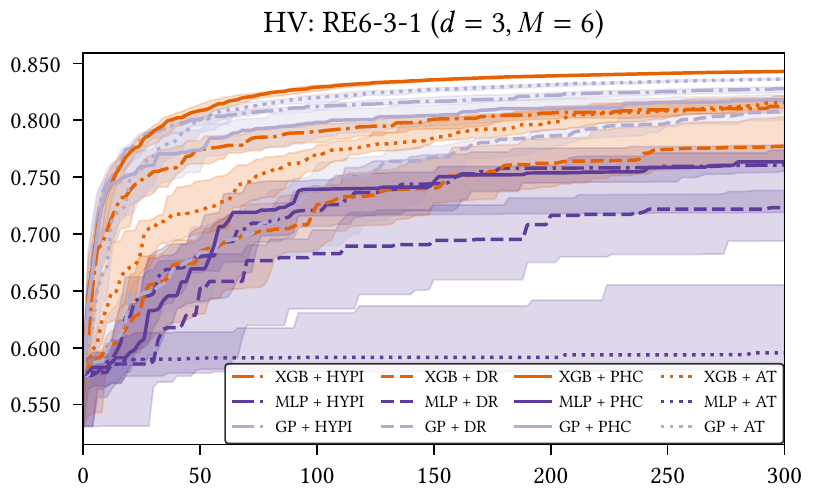}%
\includegraphics[width=0.25\linewidth]{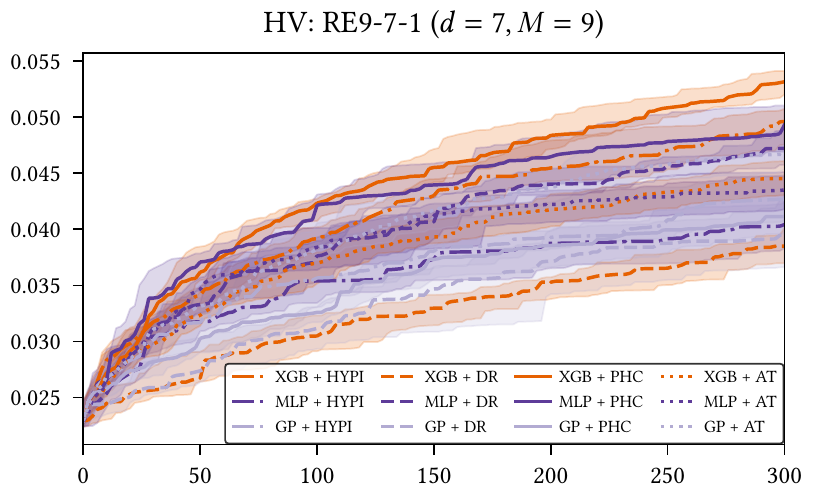}\\
\rule{\linewidth}{0.4pt}
\includegraphics[width=0.25\linewidth]{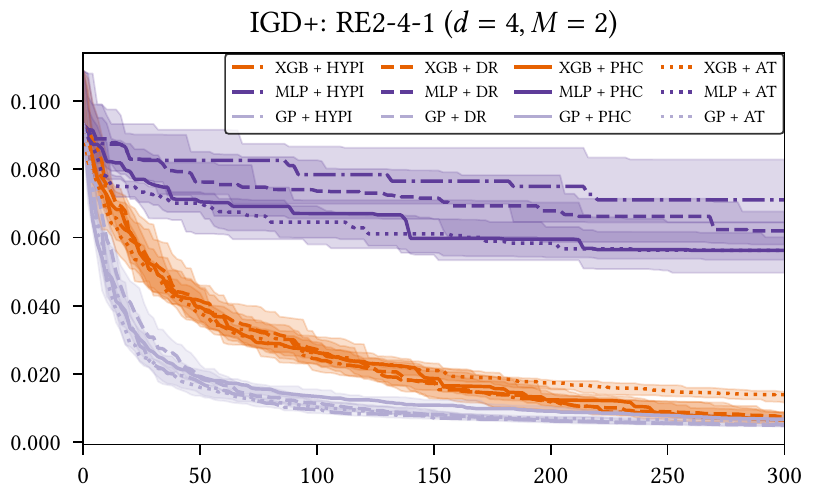}%
\includegraphics[width=0.25\linewidth]{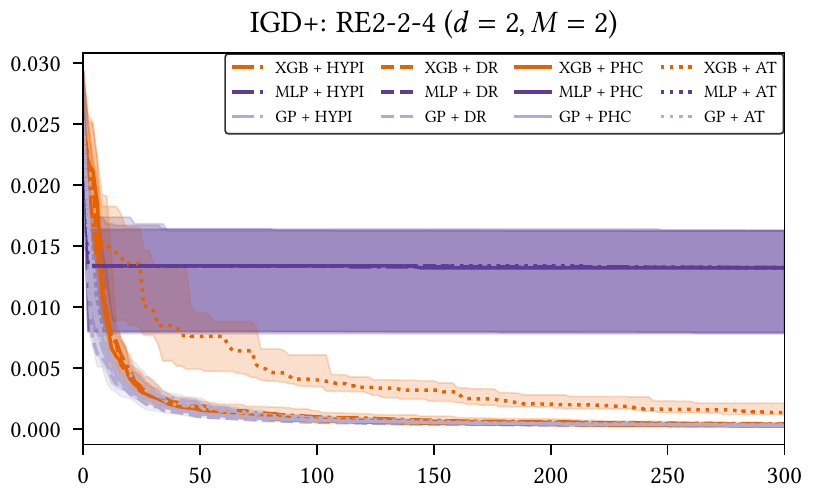}%
\includegraphics[width=0.25\linewidth]{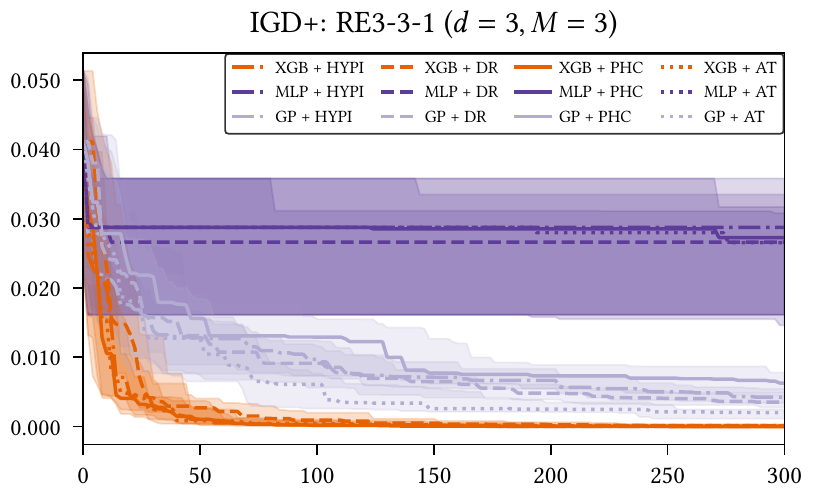}%
\includegraphics[width=0.25\linewidth]{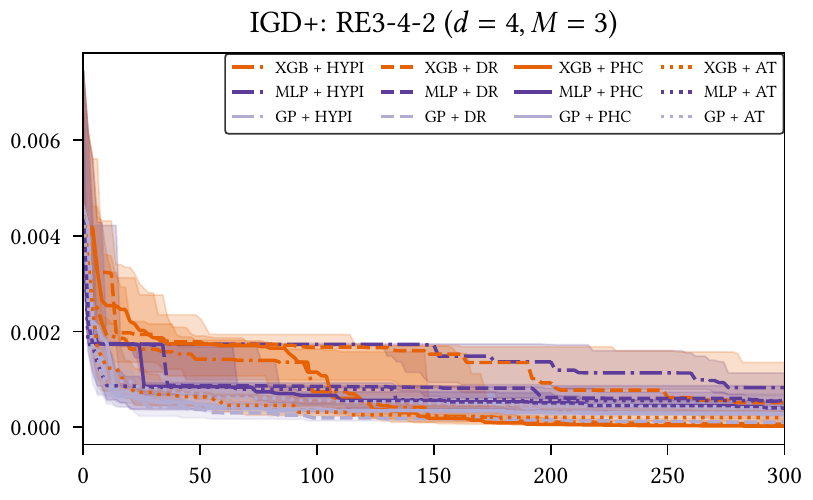}\\
\includegraphics[width=0.25\linewidth]{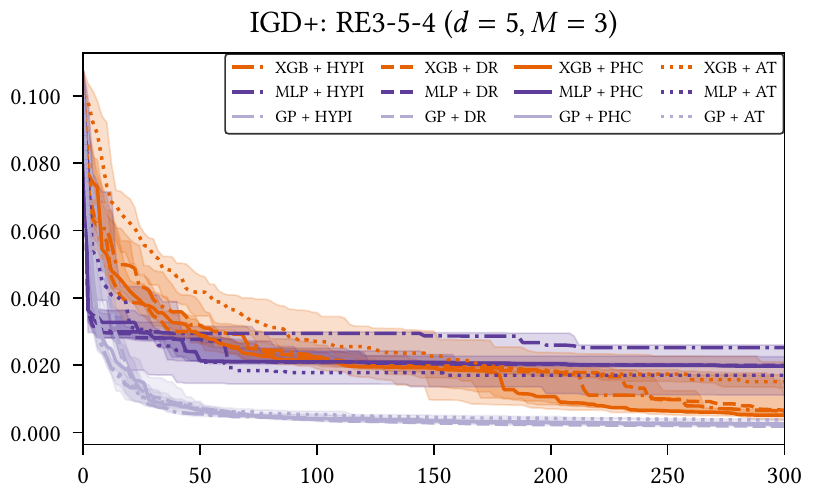}%
\includegraphics[width=0.25\linewidth]{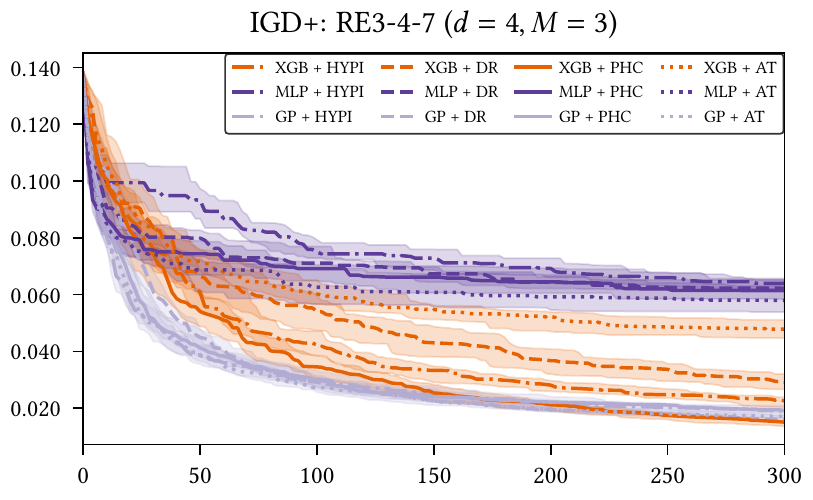}%
\includegraphics[width=0.25\linewidth]{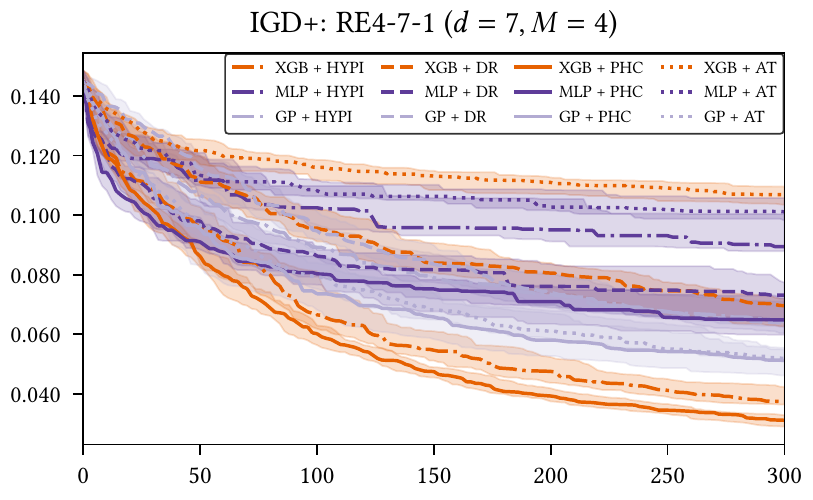}%
\includegraphics[width=0.25\linewidth]{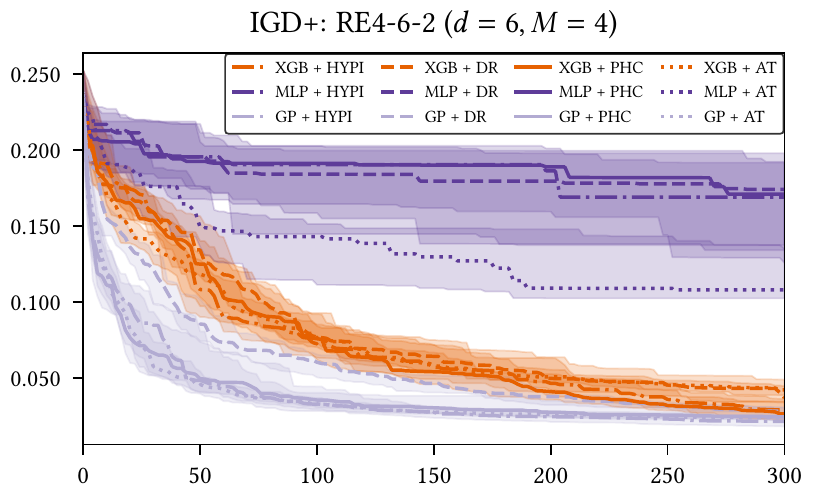}\\
\includegraphics[width=0.25\linewidth]{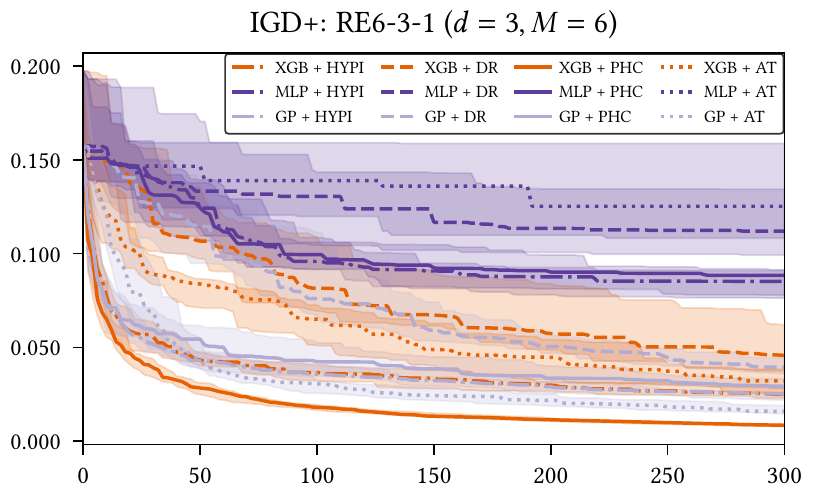}%
\includegraphics[width=0.25\linewidth]{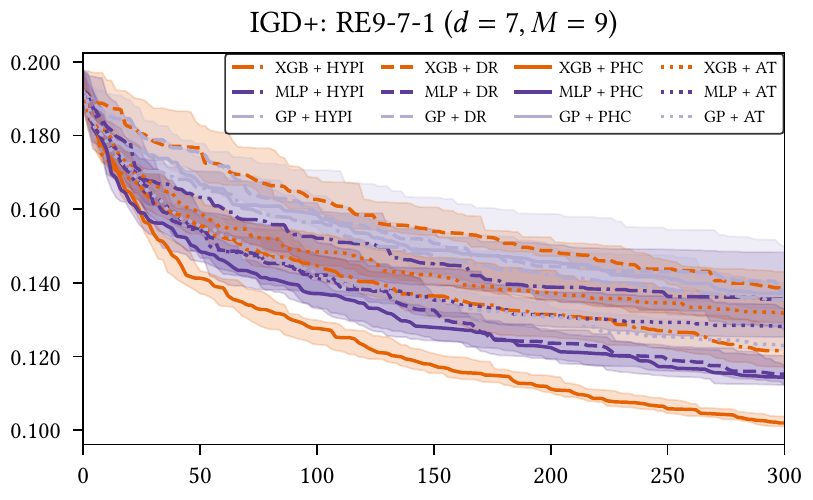}\\
\caption{%
    Hypervolume (\emph{upper}) and IGD+ (\emph{lower})
    convergence plots the Real-world benchmark.}
\label{fig:conv_RW}
\end{figure}

\newpage
\subsection{WFG (high-dimensional) Convergence Plots}
\label{sec:conv:WFG_HD}
    
\begin{figure}[H]
\includegraphics[width=0.25\linewidth]{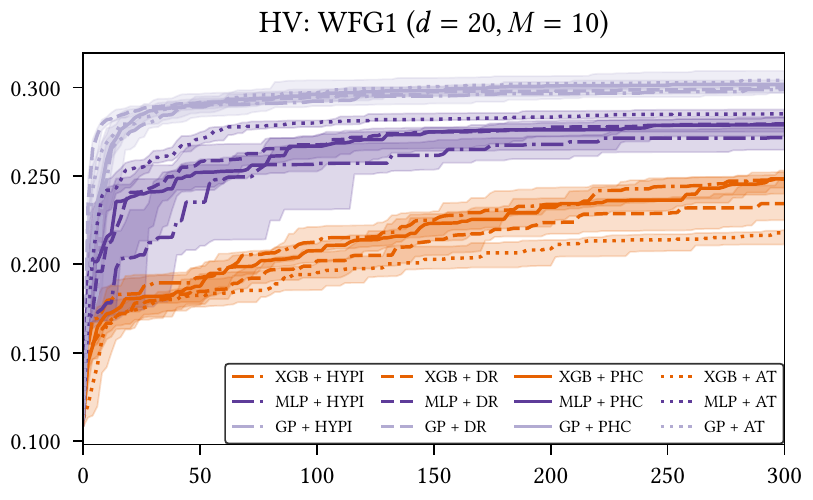}%
\includegraphics[width=0.25\linewidth]{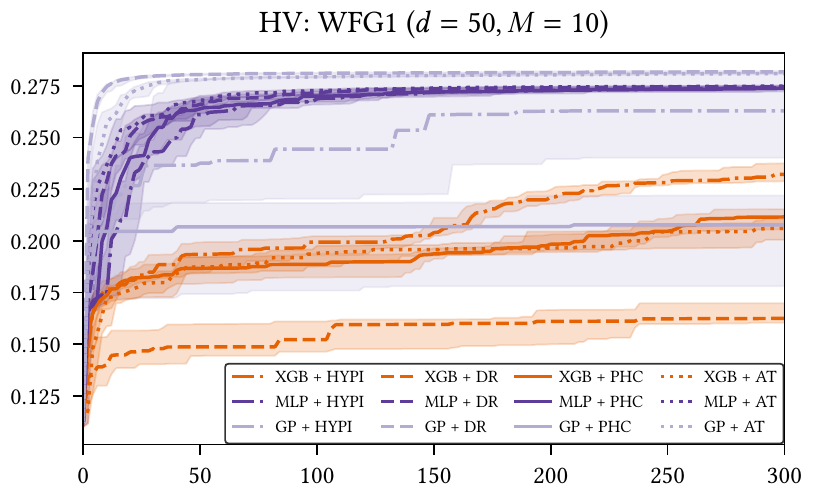}%
\includegraphics[width=0.25\linewidth]{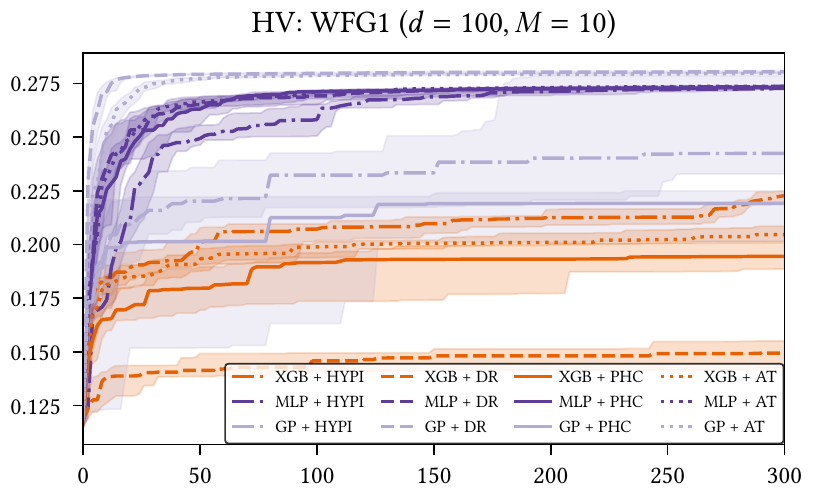}%
\includegraphics[width=0.25\linewidth]{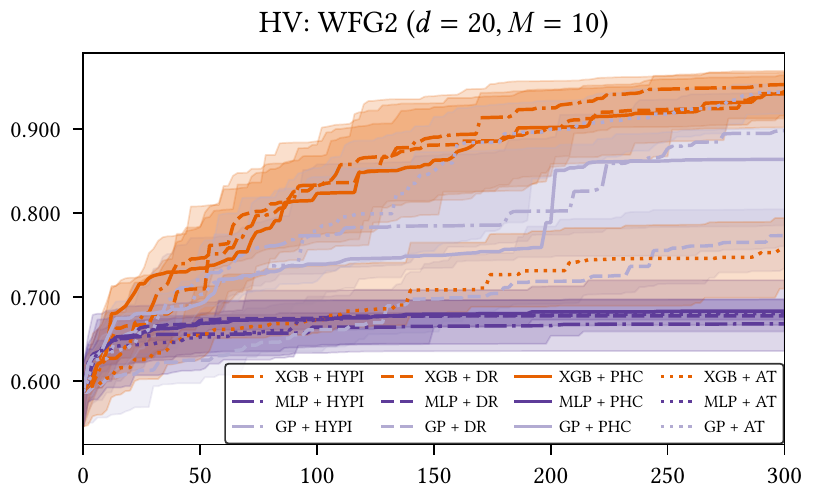}\\
\includegraphics[width=0.25\linewidth]{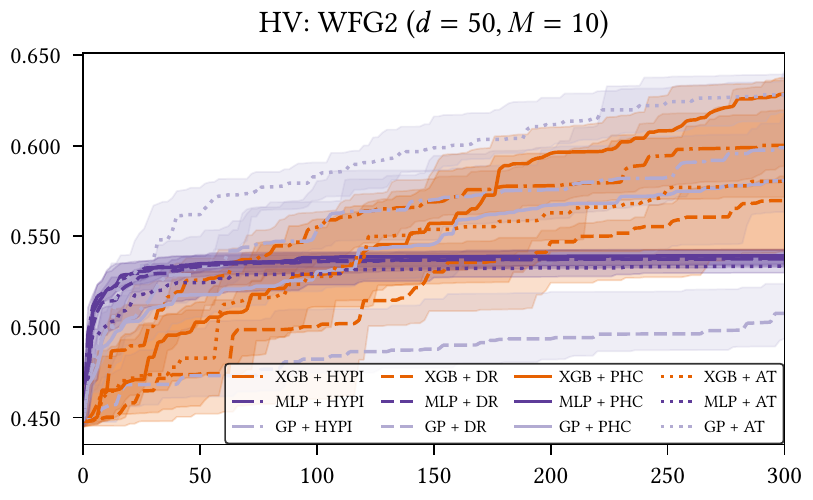}%
\includegraphics[width=0.25\linewidth]{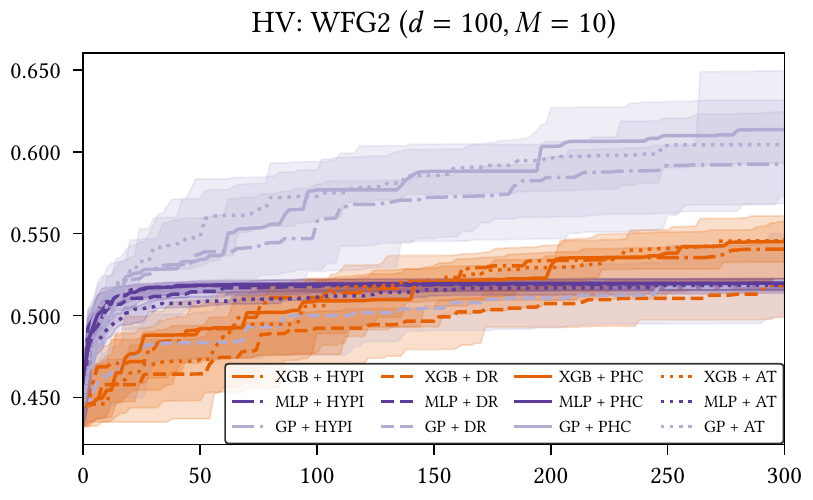}%
\includegraphics[width=0.25\linewidth]{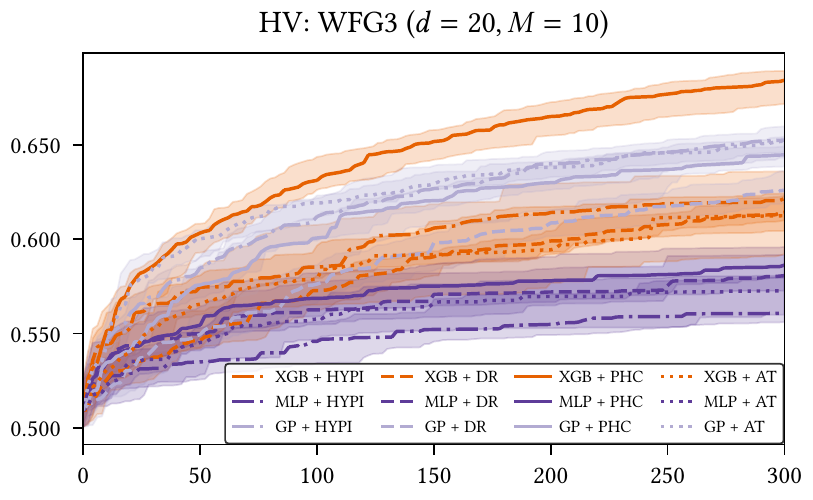}%
\includegraphics[width=0.25\linewidth]{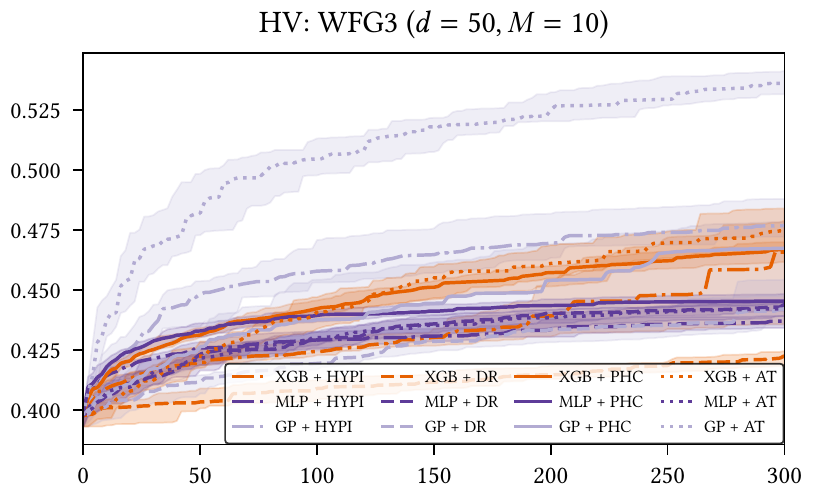}\\
\includegraphics[width=0.25\linewidth]{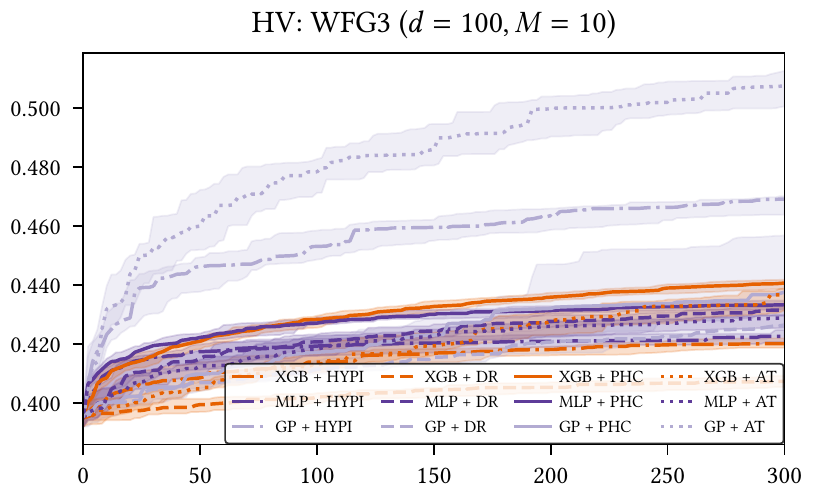}\\
\rule{\linewidth}{0.4pt}
\includegraphics[width=0.25\linewidth]{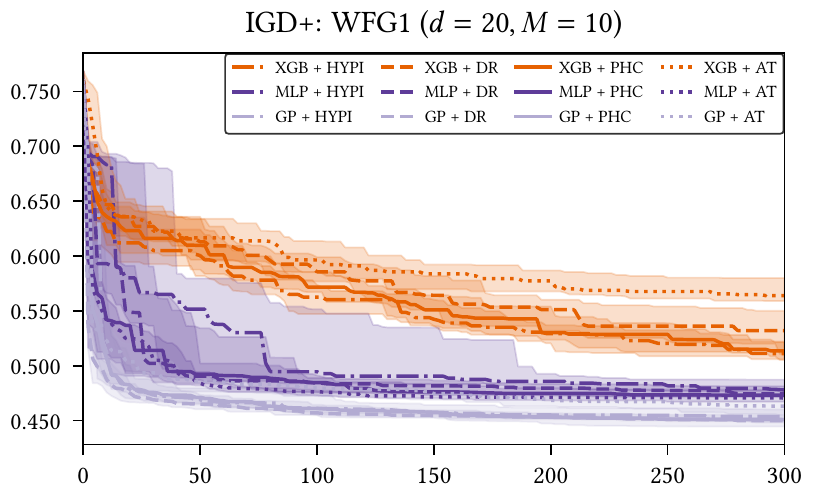}%
\includegraphics[width=0.25\linewidth]{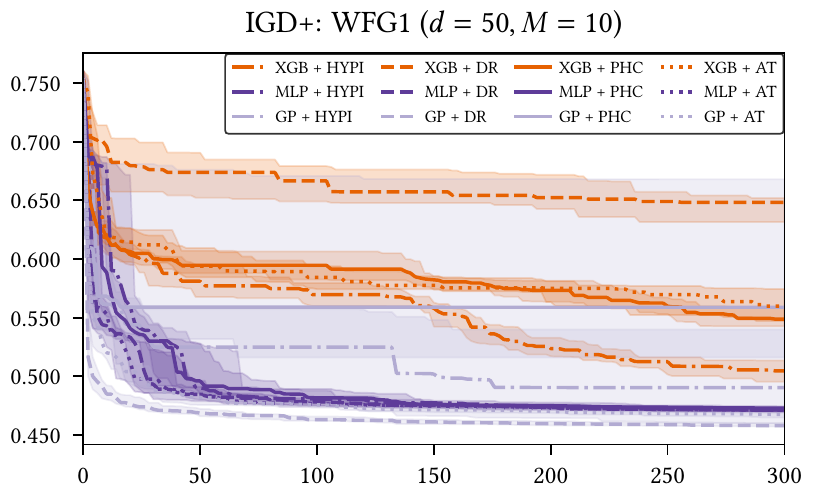}%
\includegraphics[width=0.25\linewidth]{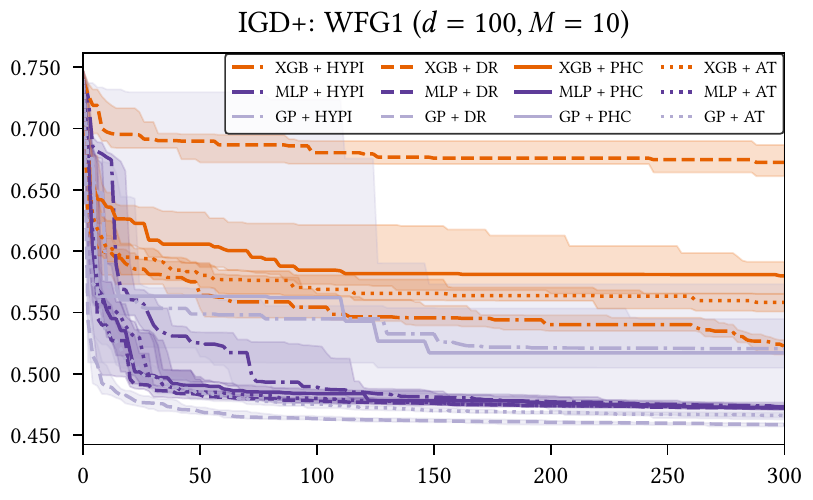}%
\includegraphics[width=0.25\linewidth]{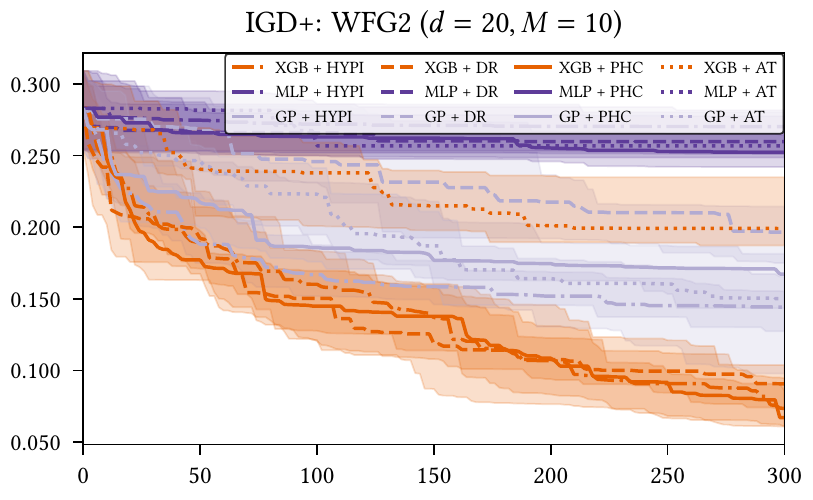}\\
\includegraphics[width=0.25\linewidth]{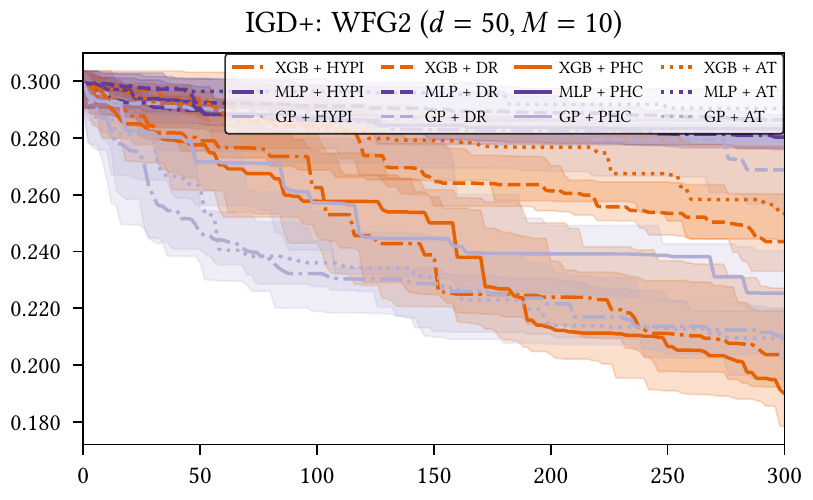}%
\includegraphics[width=0.25\linewidth]{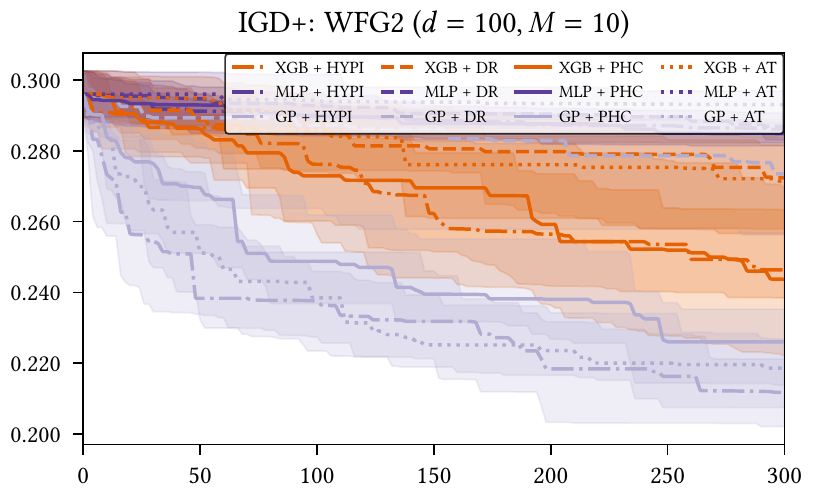}%
\includegraphics[width=0.25\linewidth]{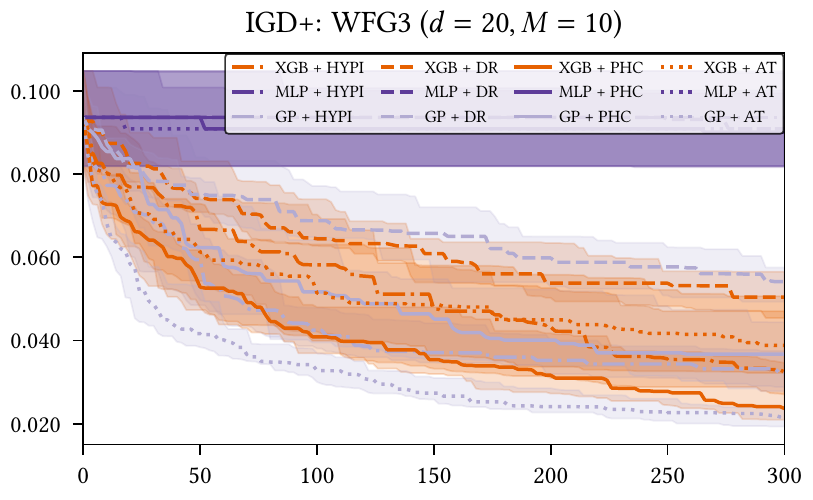}%
\includegraphics[width=0.25\linewidth]{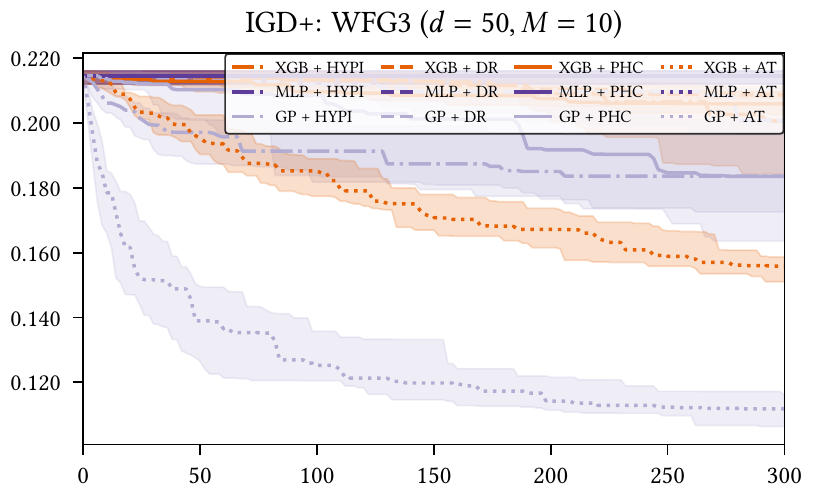}\\
\includegraphics[width=0.25\linewidth]{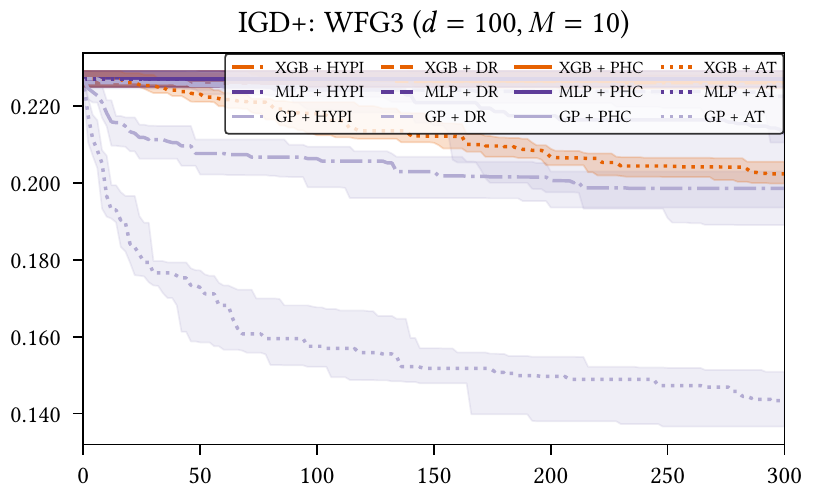}%
\caption{%
    Hypervolume (\emph{upper}) and IGD+ (\emph{lower})
    convergence plots for the high-dimensional WFG1, WFG2 and WFG3 problems.
}
\label{fig:conv_WFG_HD123}
\end{figure}

\begin{figure}[H]
\includegraphics[width=0.25\linewidth]{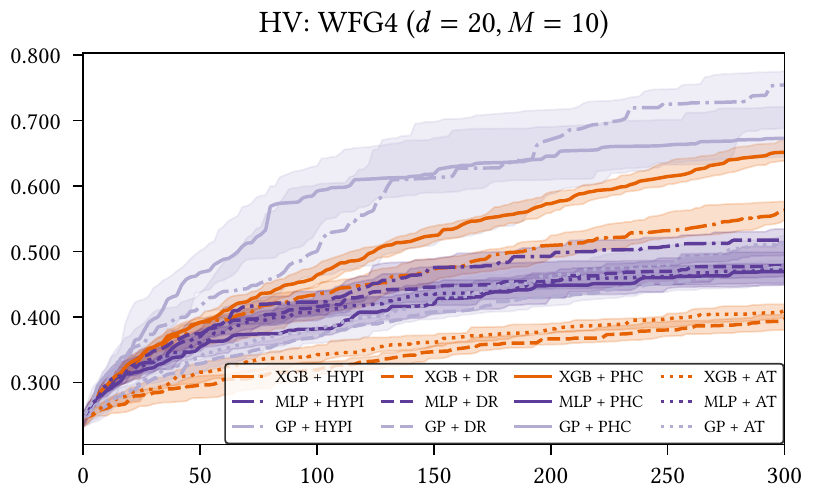}%
\includegraphics[width=0.25\linewidth]{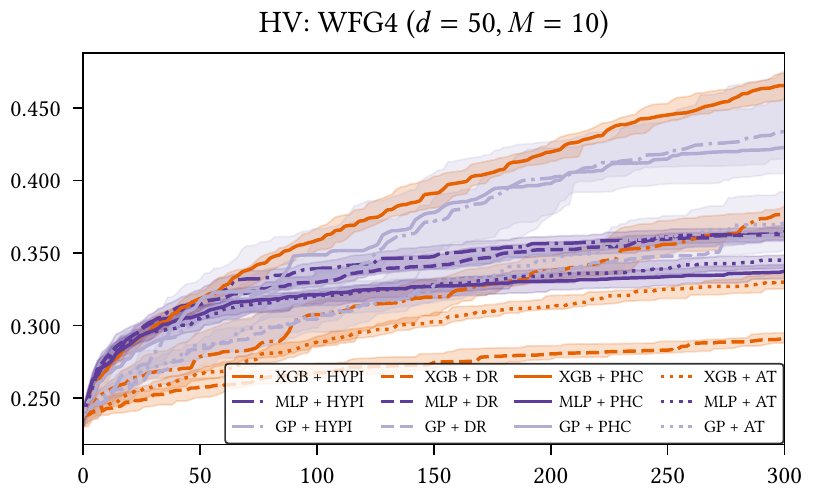}%
\includegraphics[width=0.25\linewidth]{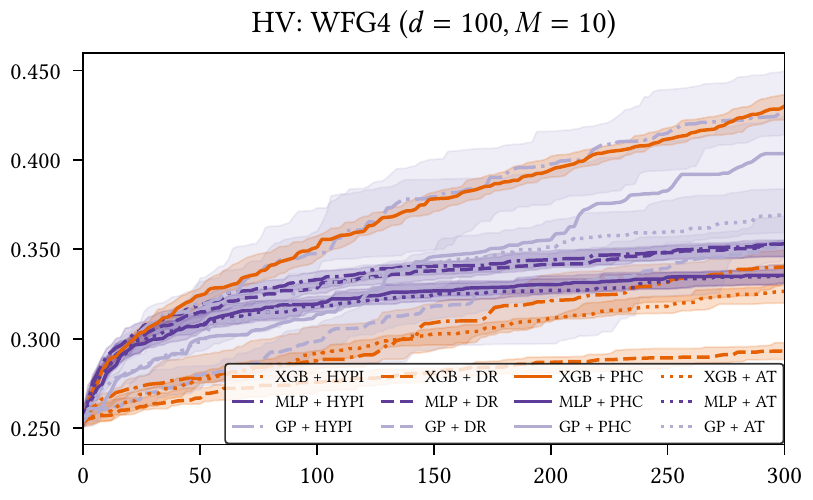}%
\includegraphics[width=0.25\linewidth]{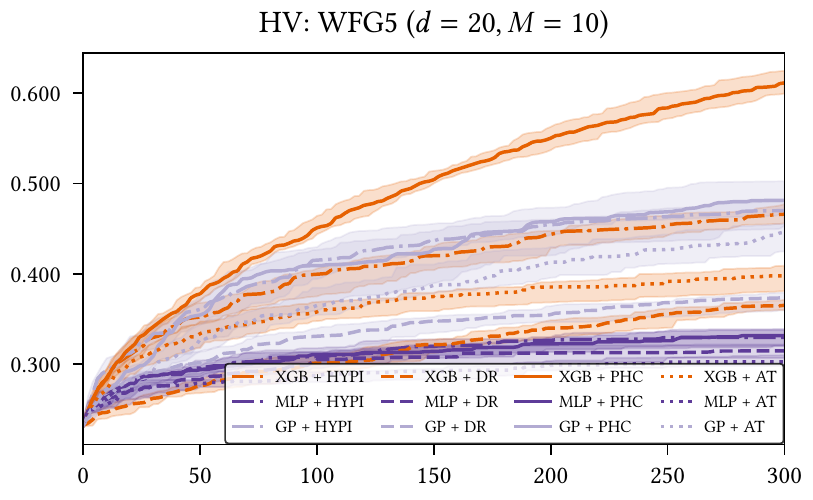}\\
\includegraphics[width=0.25\linewidth]{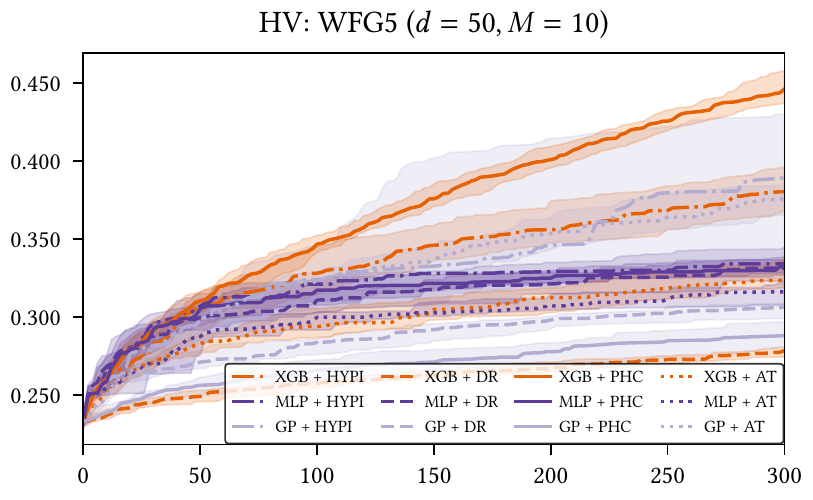}%
\includegraphics[width=0.25\linewidth]{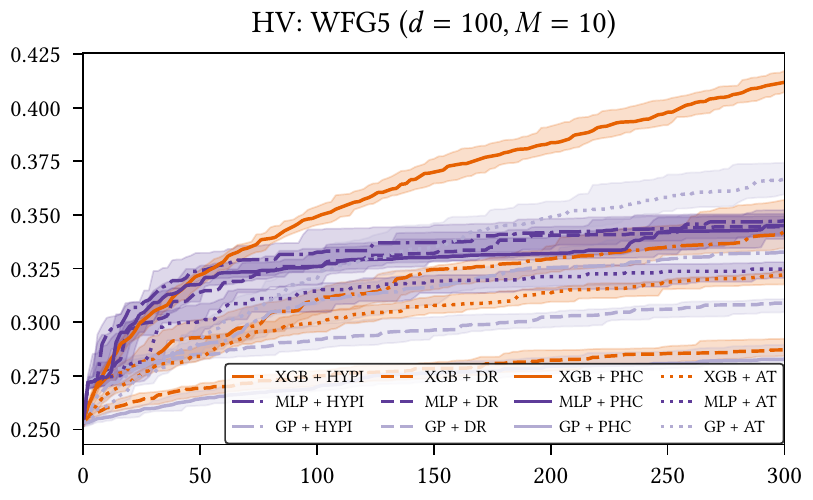}%
\includegraphics[width=0.25\linewidth]{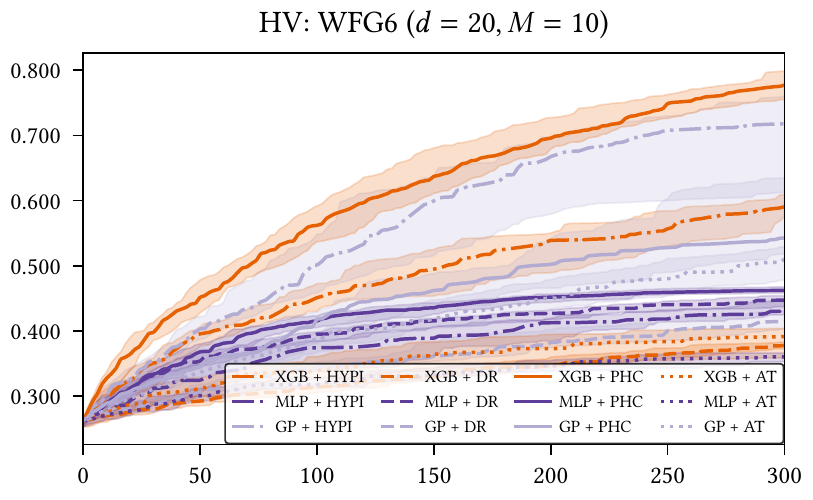}%
\includegraphics[width=0.25\linewidth]{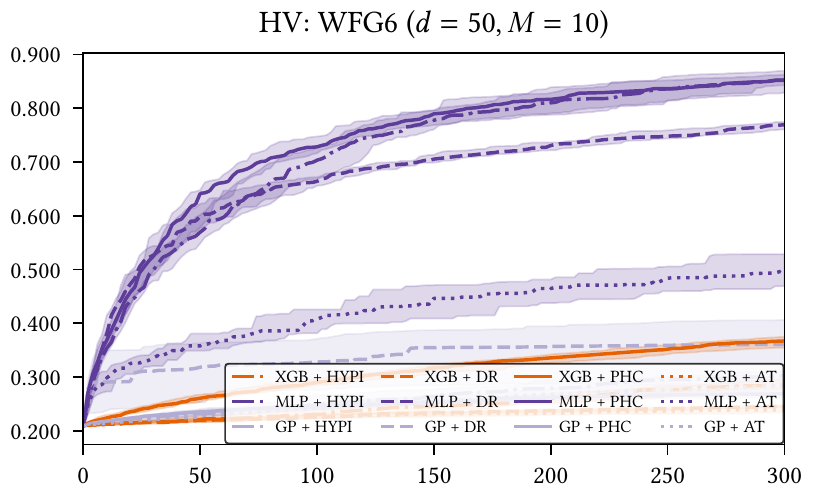}\\
\includegraphics[width=0.25\linewidth]{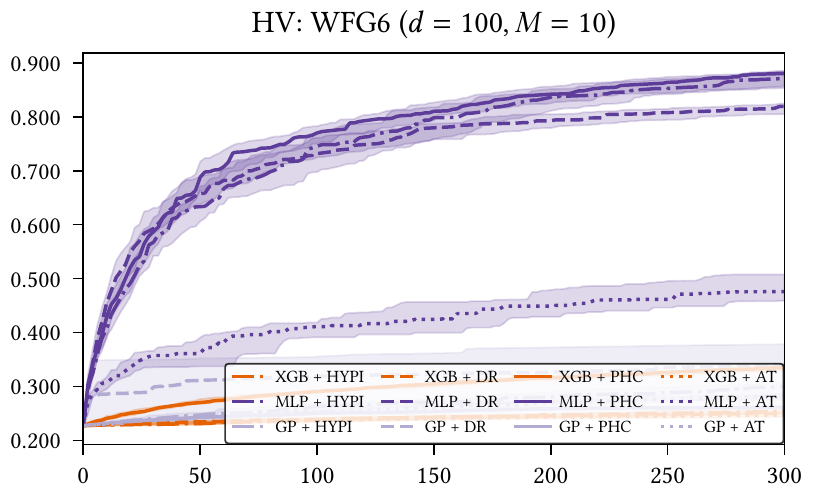}\\
\rule{\linewidth}{0.4pt}
\includegraphics[width=0.25\linewidth]{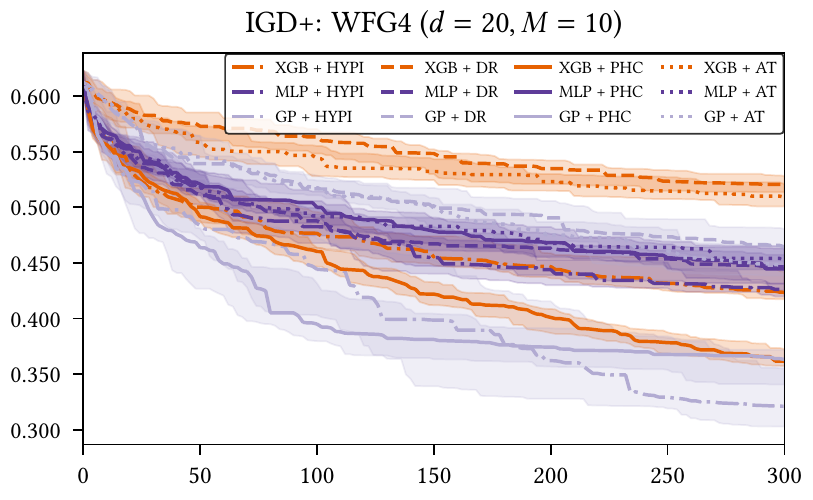}%
\includegraphics[width=0.25\linewidth]{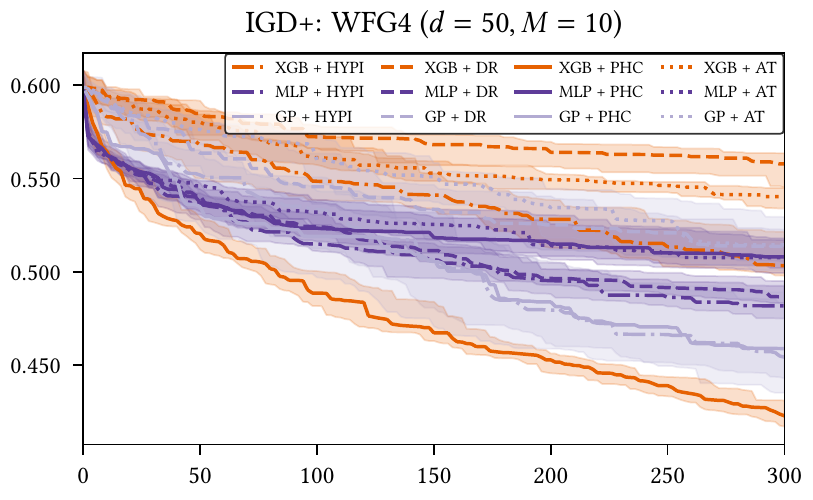}%
\includegraphics[width=0.25\linewidth]{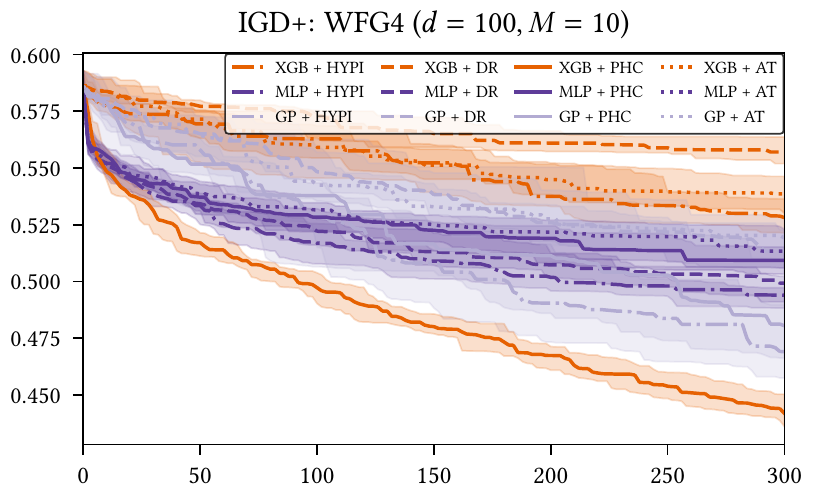}%
\includegraphics[width=0.25\linewidth]{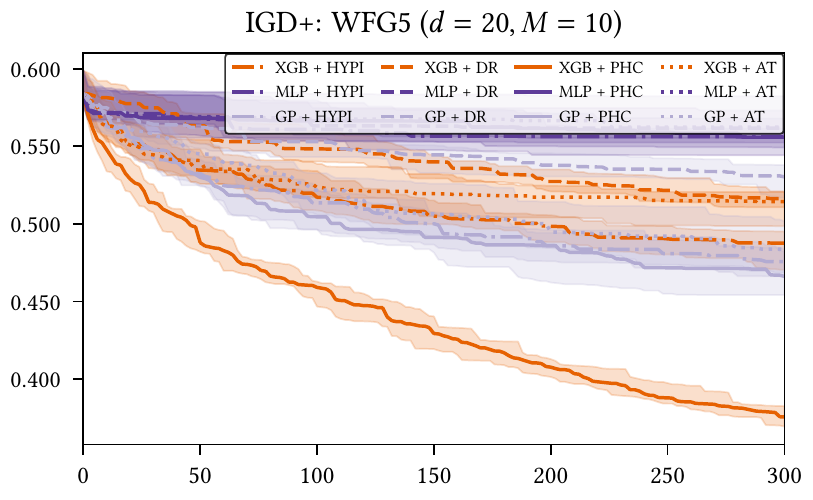}\\
\includegraphics[width=0.25\linewidth]{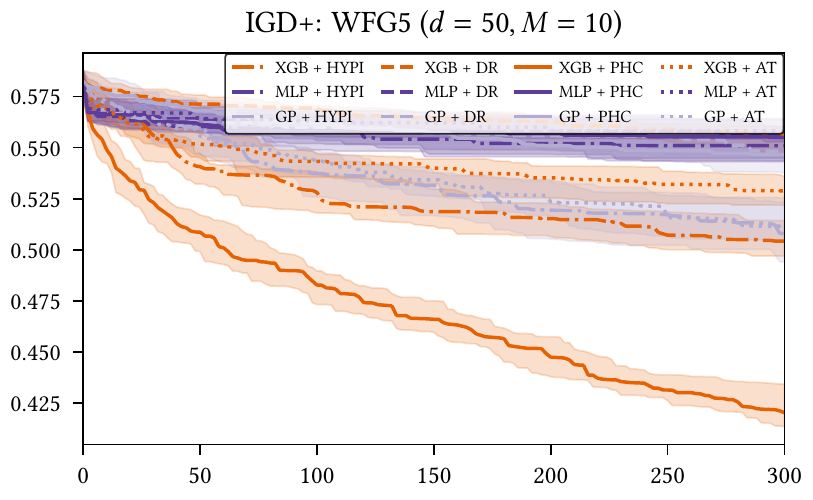}%
\includegraphics[width=0.25\linewidth]{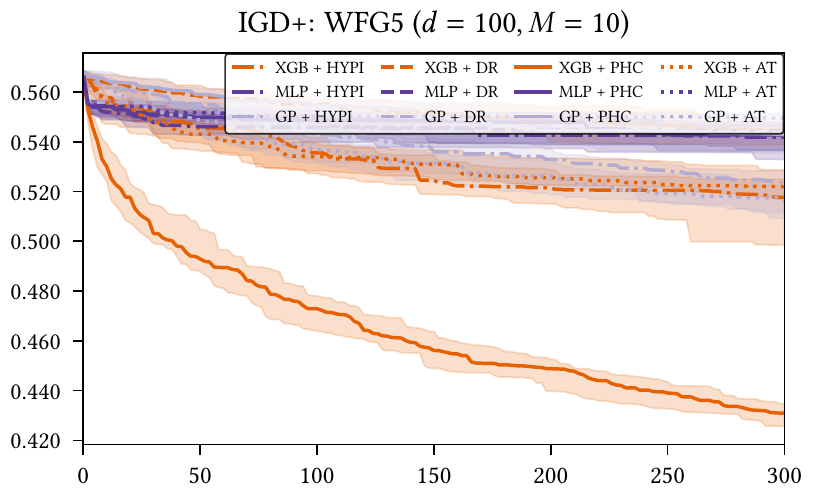}%
\includegraphics[width=0.25\linewidth]{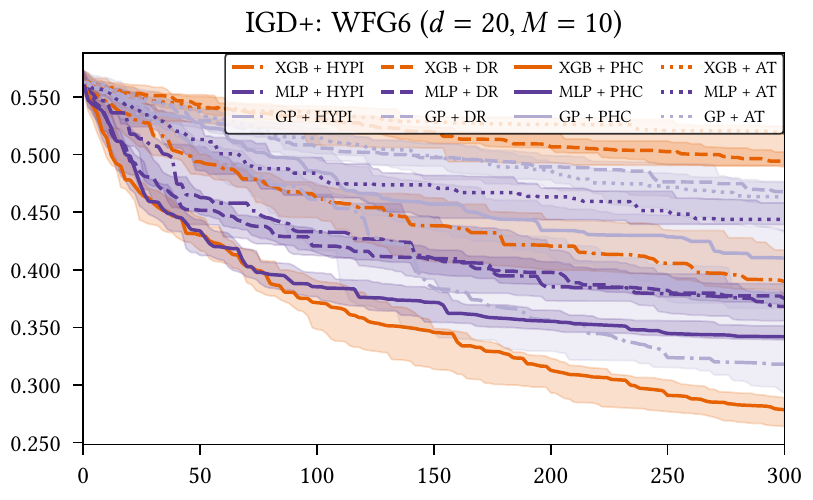}%
\includegraphics[width=0.25\linewidth]{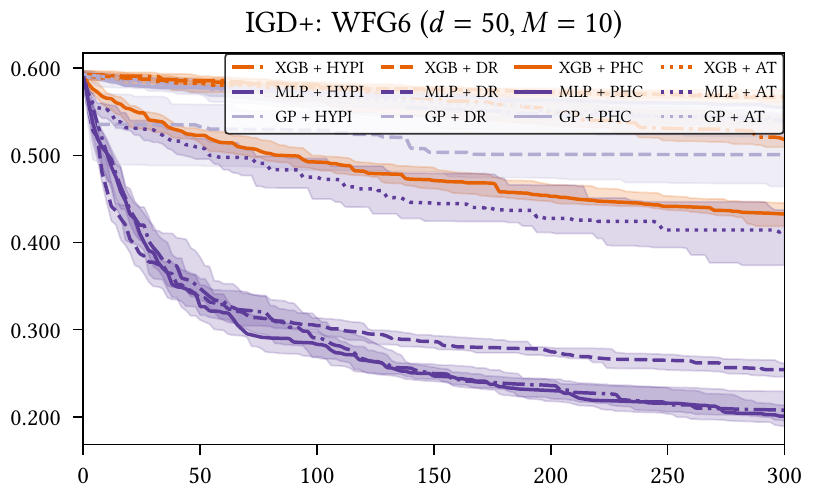}\\
\includegraphics[width=0.25\linewidth]{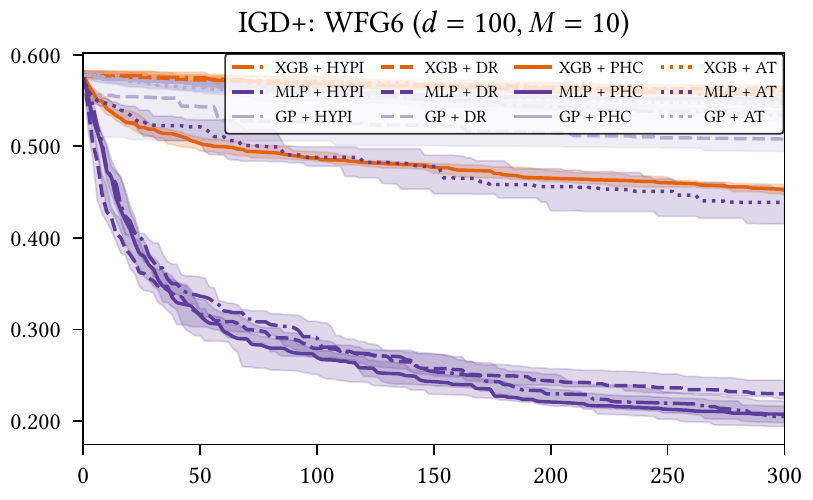}%
\caption{%
    Hypervolume (\emph{upper}) and IGD+ (\emph{lower})
    convergence plots for the high-dimensional WFG4, WFG5 and WFG6 problems.
}
\label{fig:conv_WFG_HD456}
\end{figure}

\begin{figure}[H]
\includegraphics[width=0.25\linewidth]{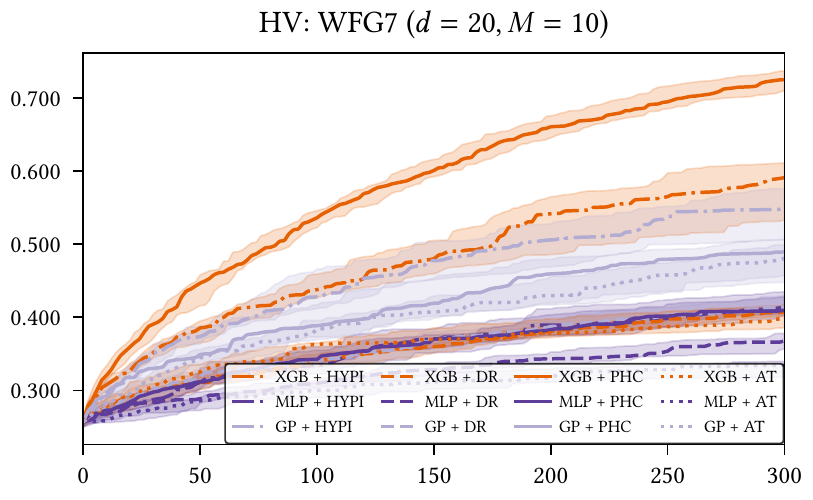}%
\includegraphics[width=0.25\linewidth]{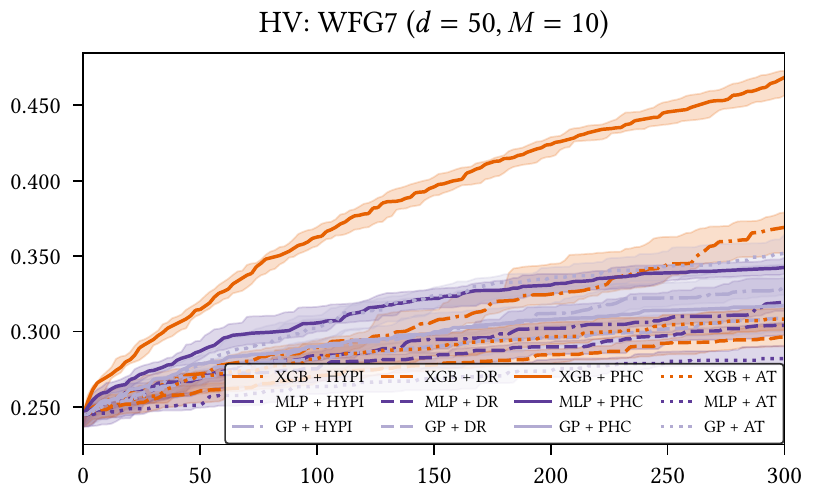}%
\includegraphics[width=0.25\linewidth]{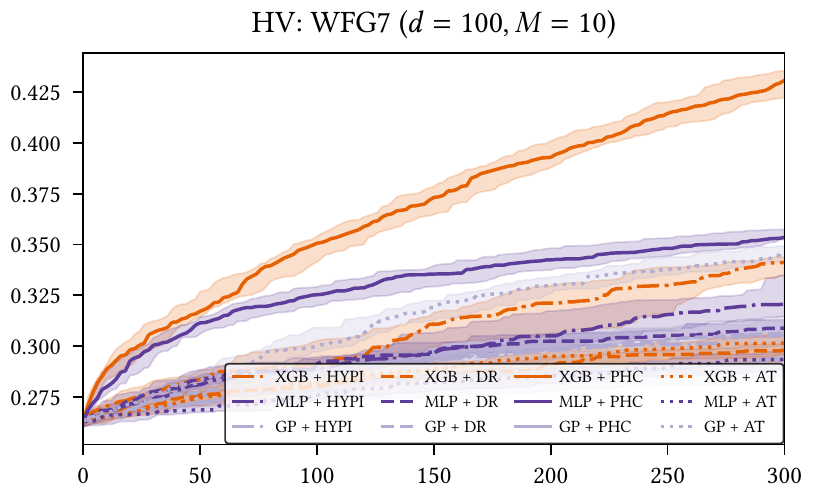}%
\includegraphics[width=0.25\linewidth]{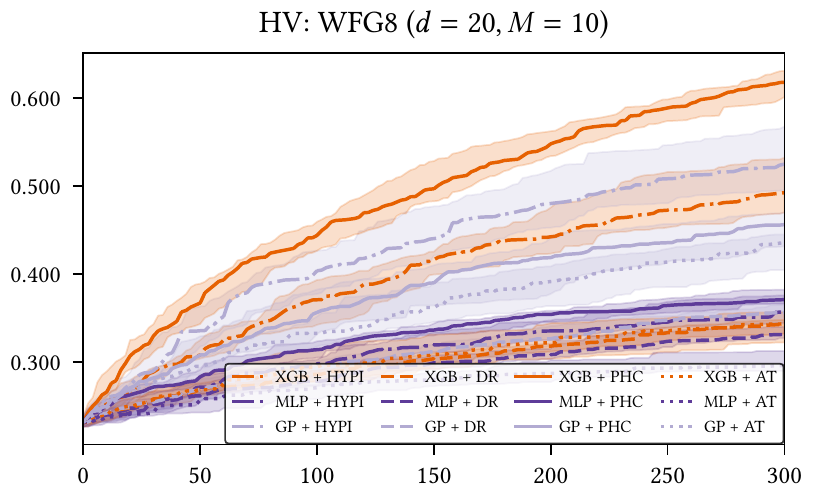}\\
\includegraphics[width=0.25\linewidth]{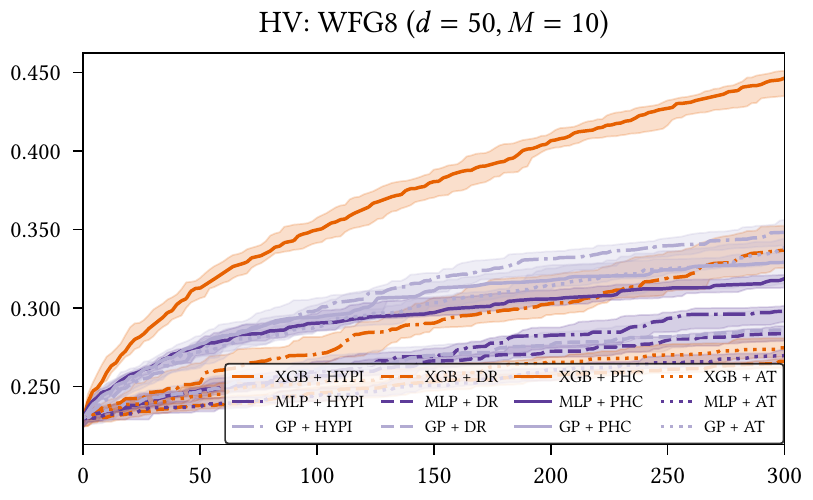}%
\includegraphics[width=0.25\linewidth]{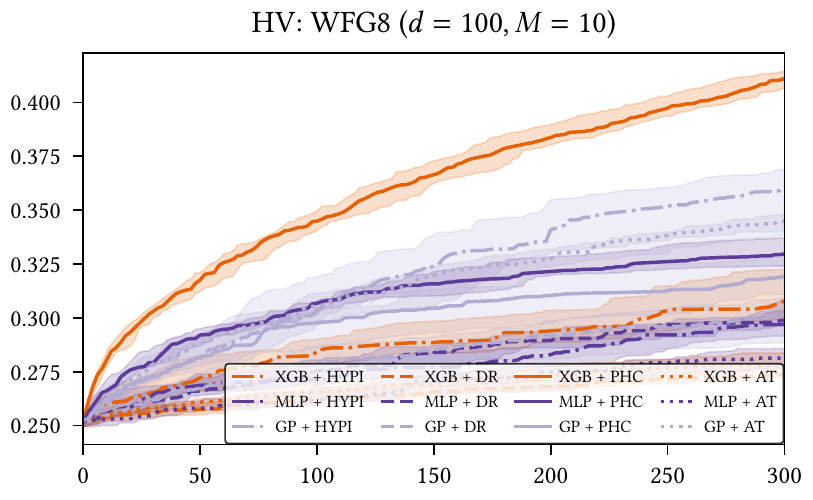}%
\includegraphics[width=0.25\linewidth]{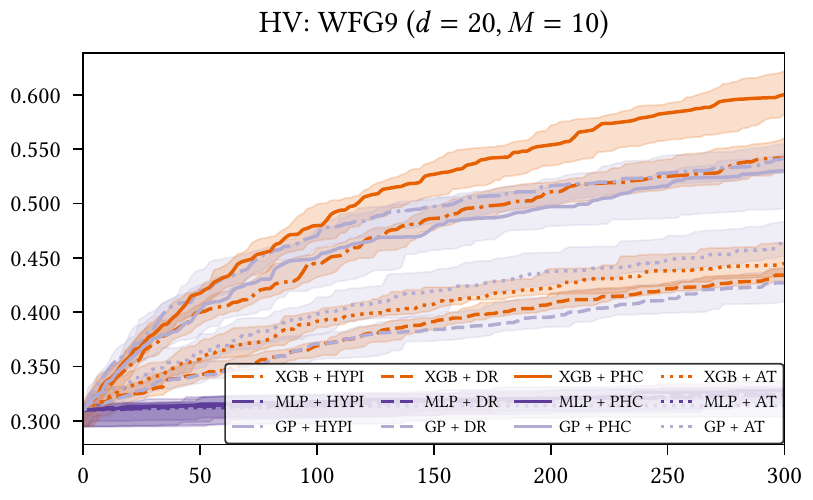}%
\includegraphics[width=0.25\linewidth]{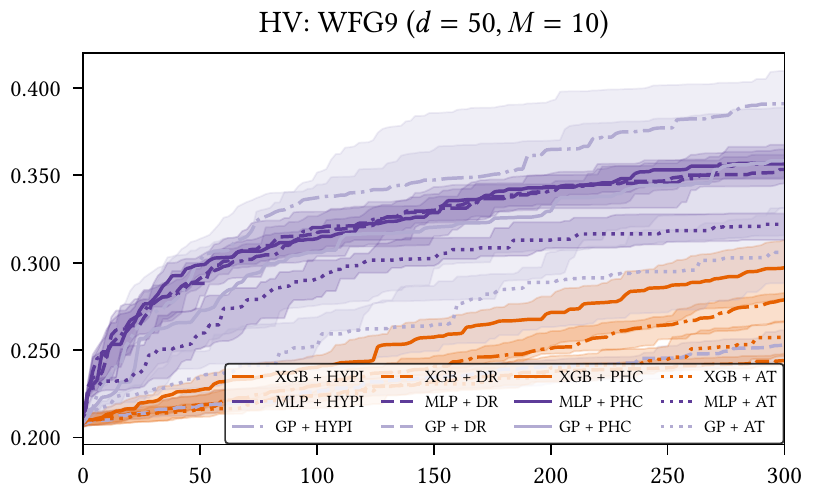}\\
\includegraphics[width=0.25\linewidth]{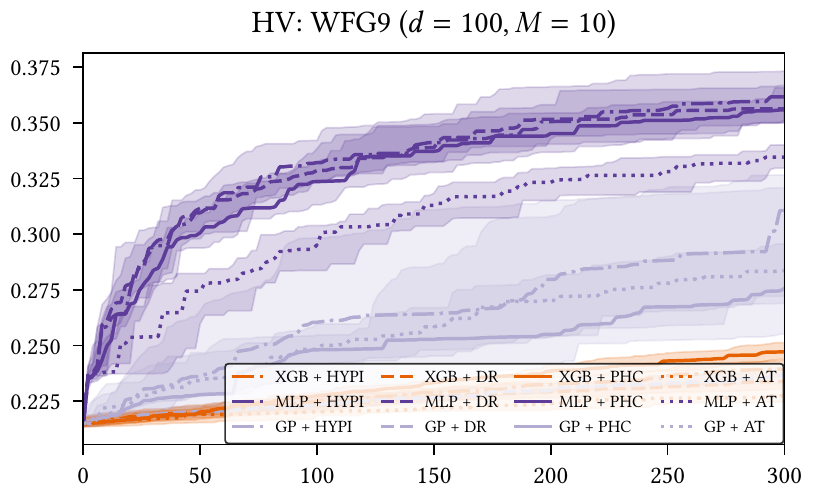}\\
\rule{\linewidth}{0.4pt}
\includegraphics[width=0.25\linewidth]{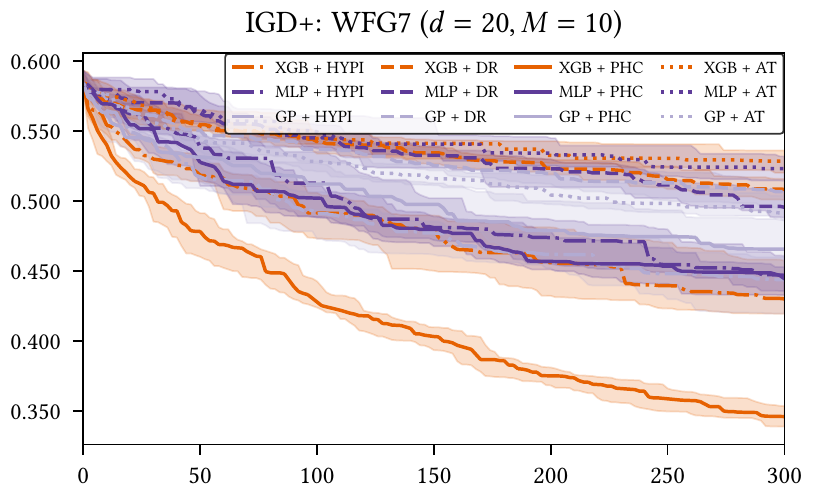}%
\includegraphics[width=0.25\linewidth]{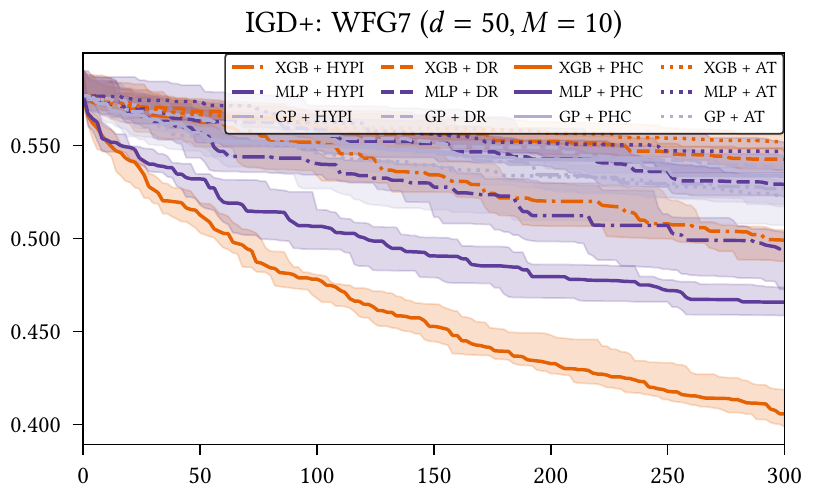}%
\includegraphics[width=0.25\linewidth]{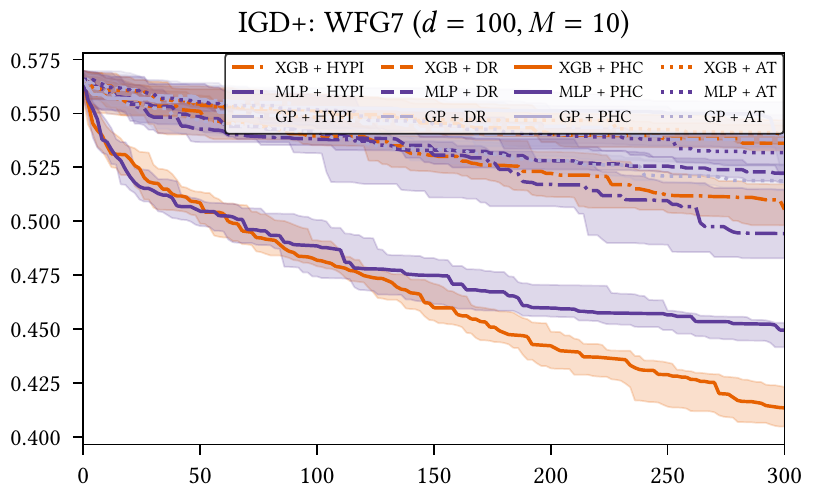}%
\includegraphics[width=0.25\linewidth]{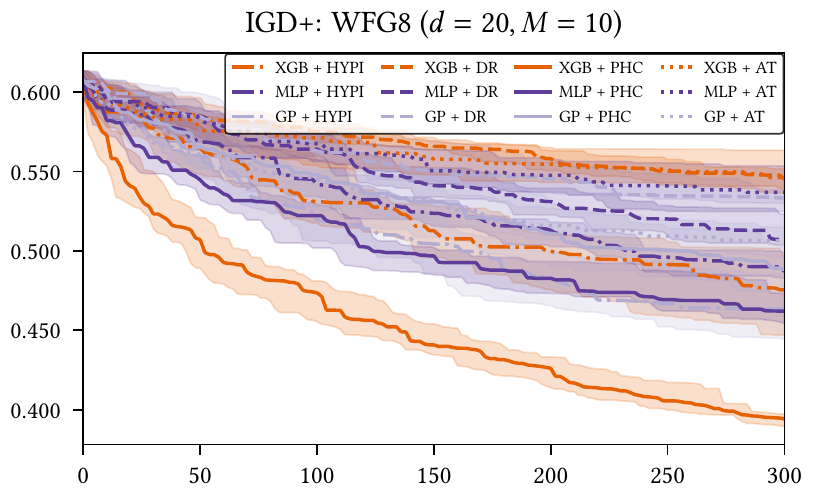}\\
\includegraphics[width=0.25\linewidth]{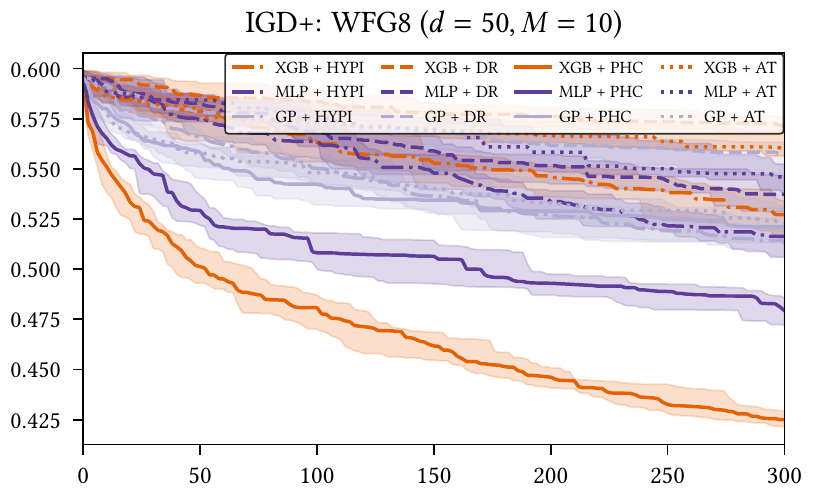}%
\includegraphics[width=0.25\linewidth]{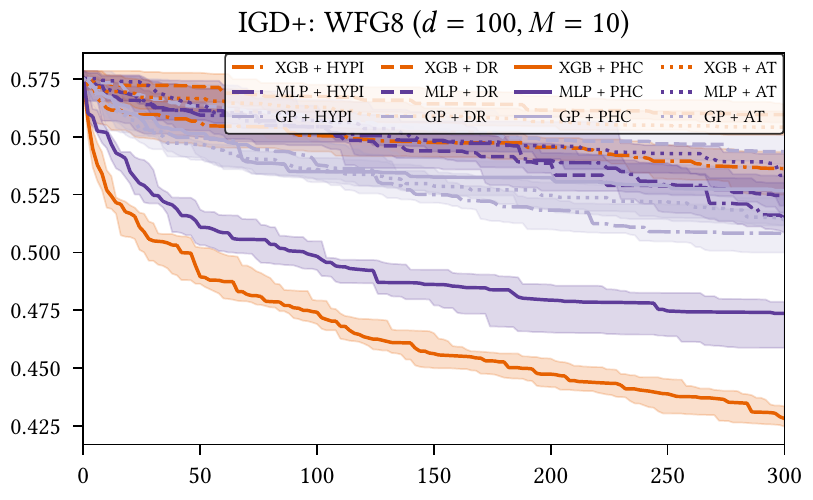}%
\includegraphics[width=0.25\linewidth]{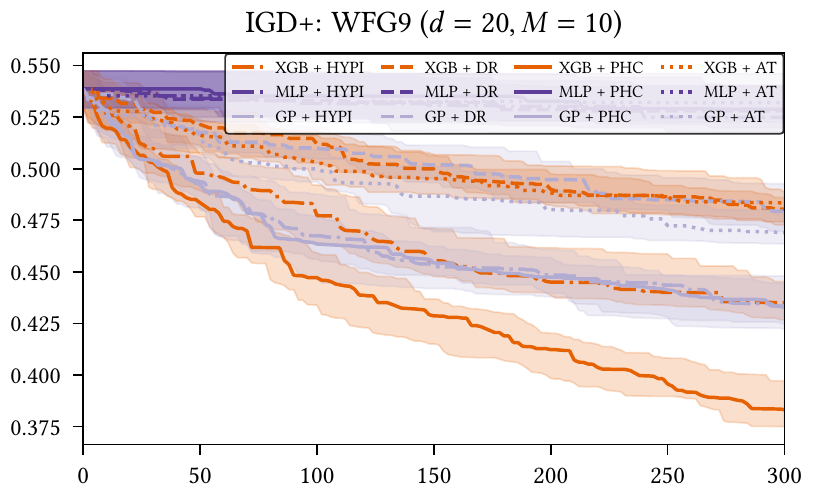}%
\includegraphics[width=0.25\linewidth]{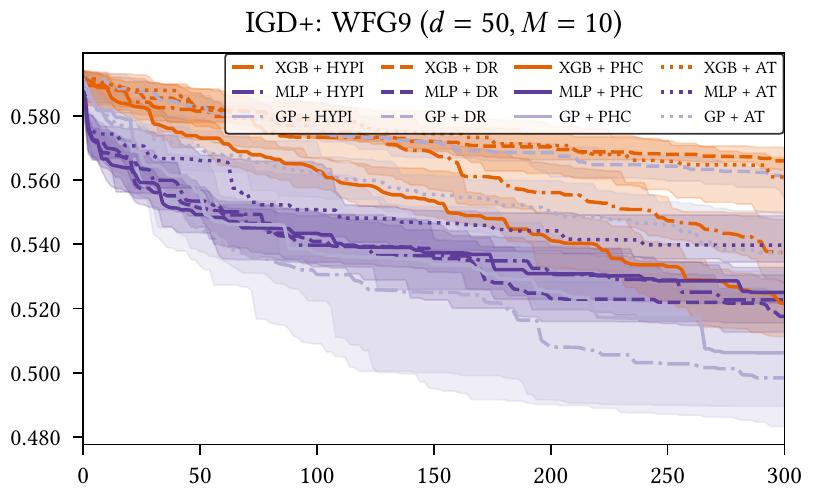}\\
\includegraphics[width=0.25\linewidth]{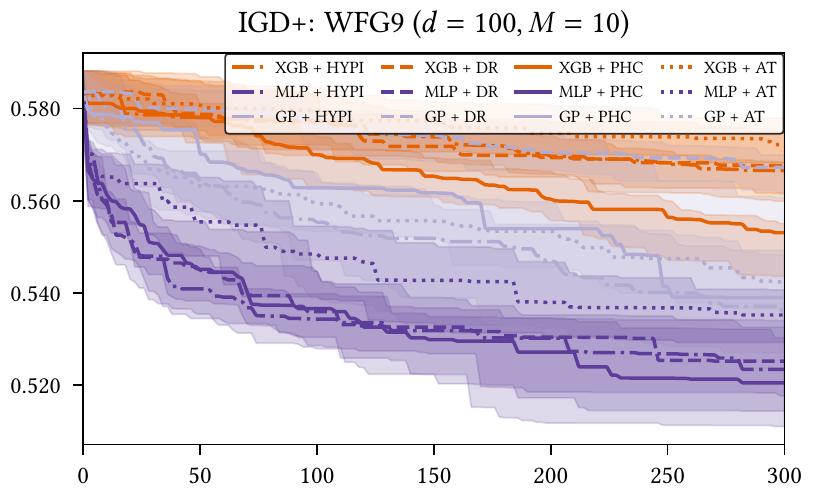}%
\caption{%
    Hypervolume (\emph{upper}) and IGD+ (\emph{lower})
    convergence plots for the high-dimensional WFG7, WFG8 and WFG9 problems.
}
\label{fig:conv_WFG_HD789}
\end{figure}

\bibliographystyle{ACM-Reference-Format}
\bibliography{ref}